\documentclass[10pt,twocolumn,letterpaper]{article}

\usepackage[pagenumbers]{cvpr} %

\usepackage{amsmath,amsfonts,bm}

\def\eqref#1{equation~\ref{#1}}

\def\1{\bm{1}}

\DeclareMathAlphabet{\mathsfit}{\encodingdefault}{\sfdefault}{m}{sl}
\SetMathAlphabet{\mathsfit}{bold}{\encodingdefault}{\sfdefault}{bx}{n}

\newcommand{\softmax}{\mathrm{softmax}}

\usepackage{multirow}
\usepackage{makecell}
\usepackage{array}
\usepackage{graphicx}
\usepackage{tabularx}
\usepackage[toc,page]{appendix}

\definecolor{cvprblue}{rgb}{0.21,0.49,0.74}
\usepackage[pagebackref,breaklinks,colorlinks,allcolors=cvprblue]{hyperref}

\def\paperID{5797} %
\def\confName{CVPR}
\def\confYear{2025}

\title{Nested Attention: Semantic-aware Attention Values for Concept Personalization }

\author{
 Or Patashnik$^{\dagger,\mathsection}$ \qquad Rinon Gal$^\dagger$ \qquad Daniil Ostashev$^\mathsection$ \qquad Sergey Tulyakov$^\mathsection$ \\ Kfir Aberman$^\mathsection$ \qquad Daniel Cohen-Or$^{\dagger,\mathsection}$ \\[5pt]
 $^\dagger$Tel Aviv University \qquad  $^\mathsection$Snap Research \\
}

\begin{document}
\twocolumn[{%
    \renewcommand\twocolumn[1][]{#1}%
    \maketitle
    \begin{center}
        \vspace{-14pt}
        \includegraphics[width=0.905\textwidth]{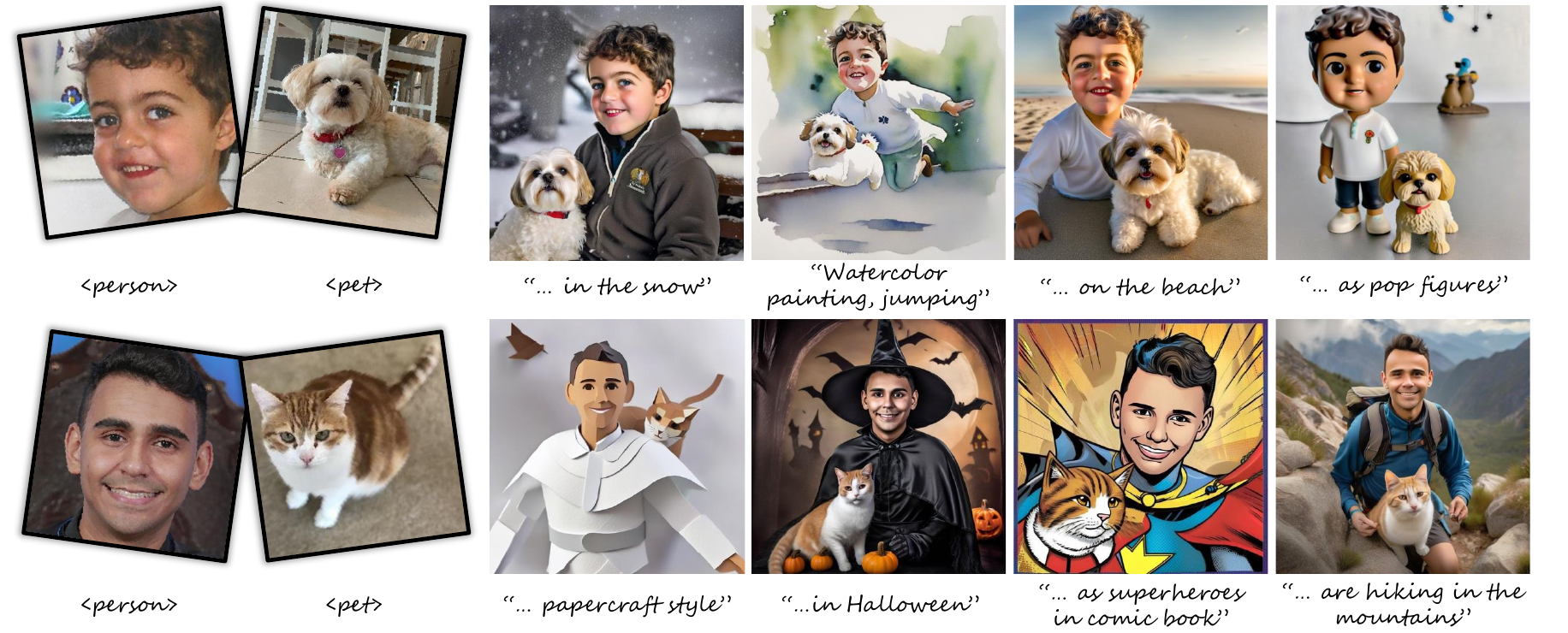}
        \vspace{-10pt}
        \captionof{figure}{
            Our nested attention mechanism attaches a localized, expressive representation of a subject to a single text token. This approach improves identity preservation while maintaining the model's prior, and can combine multiple personalized concepts in a single image.
        }
    \label{fig:teaser}
    \end{center}
}] 

\begin{abstract}
    Personalizing text-to-image models to generate images of specific subjects across diverse scenes and styles is a rapidly advancing field. Current approaches often face challenges in maintaining a balance between identity preservation and alignment with the input text prompt. Some methods rely on a single textual token to represent a subject, which limits expressiveness, while others employ richer representations but disrupt the model’s prior, diminishing prompt alignment. 
    In this work, we introduce Nested Attention, a novel mechanism that injects a rich and expressive image representation into the model’s existing cross-attention layers.
    Our key idea is to generate query-dependent subject values, 
    derived from nested attention layers that learn to select relevant subject features for each region in the generated image.
    We integrate these nested layers into an encoder-based personalization method, and show that they enable high identity preservation while adhering to input text prompts.
    Our approach is general and can be trained on various domains. Additionally, its prior preservation allows us to combine multiple personalized subjects from different domains in a single image.
\end{abstract}
    
\section{Introduction}
    Personalization of text-to-image models~\cite{nichol2021glide,ramesh2022hierarchical,ho2020denoising,dhariwal2021diffusion} enables users to generate captivating images featuring their own personal data. To introduce new subjects into the text-to-image model, initial approaches conduct per-subject optimization~\cite{gal2022textual,ruiz2022dreambooth,kumari2022customdiffusion}, achieving impressive results but requiring several minutes to capture each subject. To reduce this overhead, more recent approaches train image encoders~\cite{gal2023encoder, arar2023domain, ruiz2023hyperdreambooth, ye2023ipadapter, wang2024instantid, guo2024pulid, gal2024lcmlookahead, wang2024moa, xiao2023fastcomposer, valevski2023face0}. These encoders embed the subject into a latent representation, which is then used in conjunction with diverse text prompts to generate images of the subject in multiple contexts. 

    A key challenge in personalizing text-to-image models is balancing identity preservation and prompt alignment~\cite{ye2023ipadapter, guo2024pulid, gal2024lcmlookahead, arar2024palp}. Most encoder-based works~\cite{ye2023ipadapter,guo2024pulid,gal2024lcmlookahead,wang2024instantid,Wei_2023_ICCV} tackle personalization by encoding the subject into a large number of visual tokens which are injected into the diffusion model using new cross-attention layers. Such approaches are highly expressive and can achieve high fidelity to the subject, but they tend to overwhelm the model's prior, harming text-to-image alignment (see \Cref{sec:related-work}). A common alternative is to tie the encoded subject to a small set of word embeddings~\cite{gal2023encoder,li2023photomaker, xiao2023fastcomposer, wang2024moa}, introduced as part of the original cross-attention mechanism. This limits the impact on the model's learned prior, but greatly limits expressivity, reducing identity preservation.
    
    In this work, we propose a novel injection method that draws on the benefits of both approaches, employing a rich and expressive representation of the input image while still tying it to a single textual-token injected through the existing cross-attention layers. 
    Our key idea is to introduce query-dependent subject values using a \textit{Nested Attention} mechanism, comprised of two attention-layers. The external layer is the standard text-to-image cross-attention layer, where the novel subject is tied to a given text token. However, rather than assigning the same attention-value to this token across the entire image, we use an additional, ``nested'' attention layer to construct localized, query-dependent attention-values. In this nested layer, the generated image queries can attend to a rich, multi-vector representation of the novel subject, learning to select the most relevant subject features for each generated-image region. Intuitively, instead of having to encode the subject's entire appearance in a single token, the model can now encode smaller semantic visual elements (\eg, the mouth or the eyes), and distribute them as needed during generation.
    
    This nested mechanism thus has the advantages of both prior approaches -- a rich, multi-token representation, while bounding its influence to a single textual token which can be easily controlled. This not only leads to better trade-offs between prompt-alignment and subject-fidelity, but also to an increasingly disentangled representation, allowing us to combine several personalized concepts in a single image simply by using a different nested attention layer for each subject (\Cref{fig:teaser}). Importantly, while recent encoder-based methods focus on face-recognition based features and losses~\cite{guo2024pulid,gal2024lcmlookahead,wang2024instantid,li2023photomaker}, our approach is general and also enhances performance for non-human domains. Moreover, it does not require specialized datasets with repeated identities, and can be trained on small sets like FFHQ~\cite{karras2019style}.

    We show that our approach achieves high identity preservation while better preserving the model's prior, allowing diverse prompting capabilities. Importantly, our experiments reveal that under similar data and training-compute budgets, the nested attention approach outperforms common subject-injection methods like decoupled cross-attention~\cite{ye2023ipadapter}, on both identity similarity and editability.
    Finally, we analyze the behavior of the nested attention blocks, showing that our performance can be enhanced even further by supplying multiple subject-images at test time (without re-training), and show additional applications like identity-blending and semantic subject variations.

\section{Related Work} \label{sec:related-work}

\paragraph{Text-to-image personalization}
Text-to-image personalization aims to expand a pre-trained model's knowledge with new concepts, so that the model will be able to synthesize them in novel scenes following a user's prompt~\cite{gal2022textual,ruiz2022dreambooth}. Initial methods achieve this goal by learning a text-embedding~\cite{gal2022textual} to represent the concept, or by fine-tuning the generative network itself~\cite{ruiz2022dreambooth}. When learning text-embeddings, improved results can be achieved through careful expansions of the embedding space, for example by learning a different embedding for each denoising network layer~\cite{voynov2023p+}, for every time-step~\cite{alaluf2023neural,gal2023encoder} or by encoding information in negative prompts~\cite{dong2022dreamartist}. For fine-tuning based methods, a common approach is to restrict tuning to specific weights~\cite{simoLoRA2023,kumari2022customdiffusion,tewel2023key,avrahami2023chosen,han2023svdiff,Hu2021LoRALA,avrahami2023bas,frenkel2025implicit,shah2025ziplora, jones2024customizing}, with the aim of better preserving the pre-trained model's prior. 

While these approaches are largely successful, they require lengthy training for every subject, with training times and costs only increasing as models become larger and more complicated. A few recent methods~\cite{tewel2024trainingfree,rout2024rbmodulation} explore training-free personalization by mixing cross-image attention features. However, they struggle to preserve identities, and are largely limited to styles and simple objects. 

To overcome these challenges, considerable effort has gone into encoder-based solutions, which train a neural network to assist in the task of personalization. Our method improves on this encoder-based approach.

\vspace{-12pt}

\paragraph{Encoder-based personalization}
Initial efforts into text-to-image encoders focused on a two-step approach which first trains an encoder to provide an initial guess of a subject embedding~\cite{gal2023encoder,li2023blip} or a set of adjusted network weights~\cite{ruiz2023hyperdreambooth,arar2023domain}. These were then further tuned at inference time, to achieve high-quality personalization in as few as $5$ steps. 

More recently, a long line of works sought to avoid inference-time optimization, relying only on a pre-trained encoder to inject novel concepts into the network in a feed-forward manner~\cite{Wei_2023_ICCV,chen2023subjectdriven,shi2023instantbooth,ye2023ipadapter,jia2023taming,li2023photomaker}. Among these, particular effort has been directed at the personalization of human faces~\cite{valevski2023face0,wang2024instantid,yuan2023celebbasis,xiao2023fastcomposer}. This domain is of particular challenge, as humans are sensitive to minute details in human faces. Hence, common approaches seek to improve identity preservation by relying on features extracted from an identity recognition network~\cite{ye2023ipadapter,valevski2022unitune,wang2024instantid} or by leveraging an identity network as an auxiliary loss~\cite{gal2024lcmlookahead,guo2024pulid,peng2023portraitbooth}. 

A common thread among these methods is the use of an additional cross-attention layer as a means to inject the encoded subject's likeness. However, this approach commonly leads to degraded prompt adherence because the new layers draw the model away from its learned prior. Hence, such approaches commonly employ specialized datasets~\cite{gal2024lcmlookahead}, losses~\cite{guo2024pulid}, or significant test-time parameter tuning~\cite{ye2023ipadapter} to better enable a user to freely modify the encoded subject using text prompts. In contrast, methods that encode subjects into tokens in the existing cross-attention layers~\cite{gal2023encoder,li2023photomaker, xiao2023fastcomposer,wang2024moa,alaluf2023neural} can more easily preserve the prior by aligning the subject's attention masks to an existing word~\cite{tewel2023key}, but struggle to preserve subject identity.

Here, we propose to tackle this challenge through a novel nested attention mechanism. Instead of having to balance separate cross attention layers, we use the nested layer to compute a per-region attention-value vector, which can be injected using the existing cross-attention layers. Doing so allows us to enjoy an expressive multi-vector representation, while better preserving the model's prior.

\vspace{-3pt}
\section{Method}
\vspace{-1pt}

\begin{figure}
    \centering
    \includegraphics[width=\linewidth]{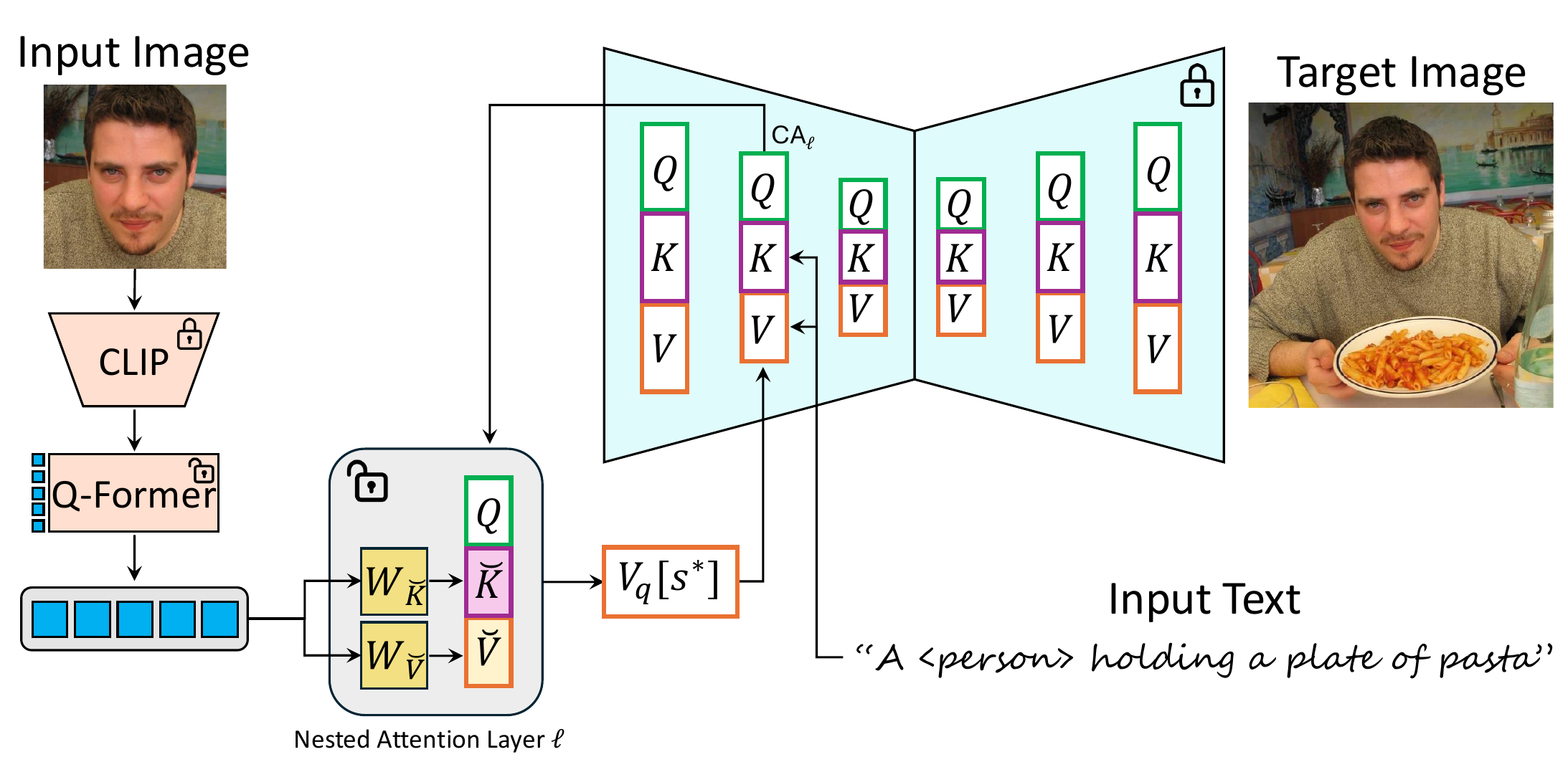}
    \vspace{-20pt}
    \caption{Method overview. The input image is passed through an encoder that produces multiple tokens to represent it. These tokens are  projected to form the keys and values of the nested attention layers. The result of each nested attention layer is a new set of per-query values, $V_q^*$, which then replace the cross-attention values of the token $s^*$ representing the subject. One nested attention layer is added to each of the cross-attention layers of the model. 
    }
    \vspace{-14pt}
    \label{fig:encoder-overview}
\end{figure}

Our method builds on a pretrained text-to-image diffusion model~\cite{podell2024sdxl}. 
Given an input image of a specific subject and a text prompt, we generate a novel image of this subject that aligns with the prompt. To achieve this, we employ an encoder-based approach that takes the input image and converts it into a set of tokens.
These tokens are then used to calculate per-query attention values using a novel nested attention layer (see overview in Figure~\ref{fig:encoder-overview}).
Specifically, the nested attention mechanism selectively overrides the cross-attention values associated with a target token (\eg, \textit{``person"}) to which we apply the personalization, enabling the model to incorporate the unique features of the subject while adhering to the given prompt.

In the following subsections, we first provide background on cross-attention layers in diffusion models. We then introduce our nested attention mechanism, which is central to our personalization approach. Finally, we describe the architecture and training process of the encoder used to generate personalized tokens from input images.

\begin{figure}
    \centering
    \includegraphics[width=0.93\linewidth]{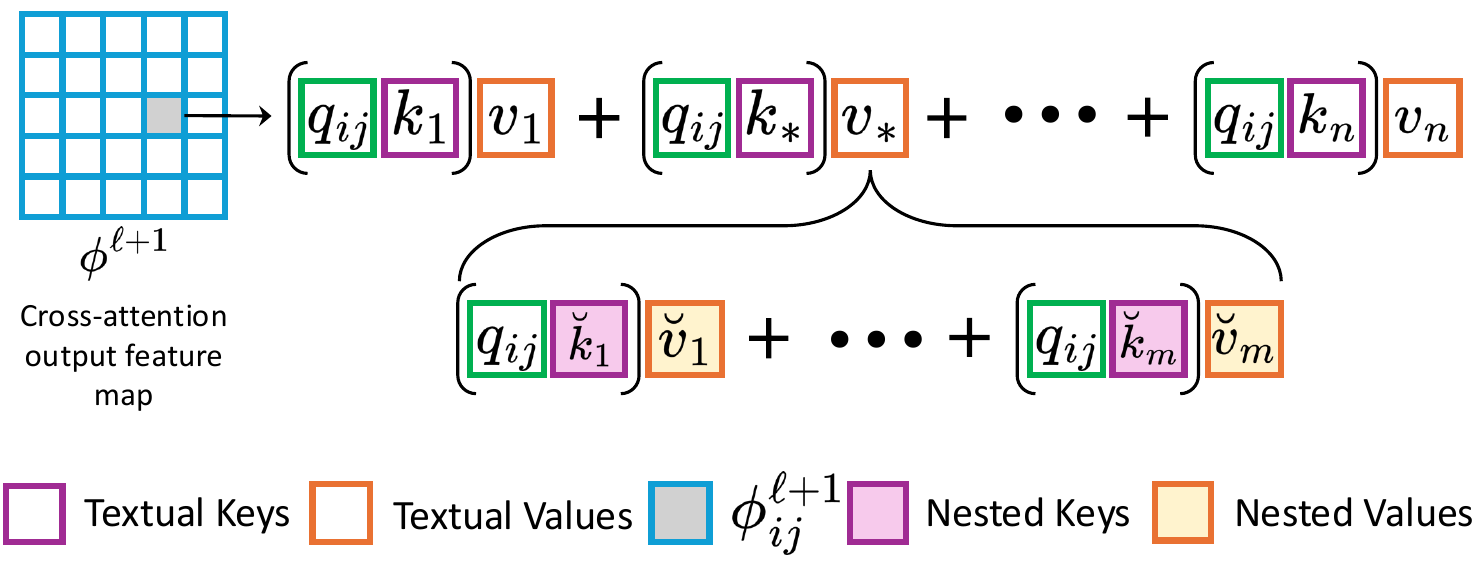}
    \vspace{-6pt}
    \caption{The nested attention mechanism. We replace the value of the token $s^*$ with the result of an attention operation between the query and the nested keys and values produced by the encoder, resulting in a query-dependent value.
    }
    \vspace{-12pt}
    \label{fig:nested-mechanism}
\end{figure}

\subsection{Preliminaries: Cross-Attention}
\vspace{-2pt}

Diffusion models typically incorporate text conditions into the generation process using cross-attention layers~\cite{rombach2021highresolution, podell2024sdxl}. Let $c$ denote the text encoding. In each cross-attention layer $\ell$, $c$ is projected into keys $K=f_K^{\ell}(c)$ and values $V=f_V^{\ell}(c)$, where $f_K^{\ell}$ and $f_V^{\ell}$ are learned linear layers parameterized by $W_K^\ell$ and $W_V^\ell$, respectively. The input feature map of the $\ell$-th layer of the diffusion model, denoted as $\phi_{\text{in}}^{\ell}(z_t)$, is projected into queries $Q = f_Q^{\ell}(\phi_{\text{in}}^{\ell}(z_t))$, where $f_Q^{\ell}$ is another learned linear layer parameterized by $W_Q^\ell$.
The output of the attention layer is formed using these queries, keys, and values. Each location in the output feature map, $\phi_{\text{out}}^{\ell}(z_t)_{ij}$, is a weighted sum of the values, as illustrated in Figure~\ref{fig:nested-mechanism}. Formally, the output feature map of the attention layer is given by:
\vspace{-8pt}
\begin{equation*}
    \vspace{-6pt}
    \phi_{\text{out}}^{\ell}(z_t) = \softmax \left( \frac{QK^T}{\sqrt{d}} \right)V.
\end{equation*}
Previous works~\cite{alaluf2023crossimage, cao_2023_masactrl, hertz2022prompt, tumanyan2023plug, Parmar_2023, tokenflow2023} have shown that each component of the attention mechanism in diffusion models serves a specific role. 
Consider $q_{ij}$, the query at spatial location $(i,j)$. Its dot product with each of the keys measures the semantic similarity between this spatial location and the concept represented by the key. These similarity scores are used to weight the concept values, which are then added to the existing features at the query's spatial location.
Hence, the query dictates which concepts should appear in each image region, while the values control the appearance. We will build on this insight for our nested attention layer. 

\subsection{Nested Attention}
\vspace{-2pt}

In standard diffusion models, the same value $V[s]$ corresponding to a specific textual token $s$ is used to form all the features $\phi_{\text{out}}^{\ell}(z_t)_{ij}$ corresponding to $s$ in the output feature map. For instance, when generating an image from the prompt ``a person on the beach'', the value corresponding to the token ``person'', $f_V^\ell(\text{``person''})$, influences all tokens representing the person, regardless of their diverse appearances (\eg, mouth and hair).
This means that the \textit{single} token value corresponding to the word ``person'' must represent all the high-dimensional information about the many different intricate details of the person being generated.

\begin{figure}
    \centering
    \setlength{\tabcolsep}{1pt}
    \scriptsize{
    \begin{tabular}{ccccc}
        \multirow{2}{*}{Input} & Generated & $V_q[s^*]$ w/o n- & \multicolumn{2}{c}{$V_q[s^*]$ w/} \\
         & image & ested attention & \multicolumn{2}{c}{nested attention} \\
        \includegraphics[width=0.19\linewidth]{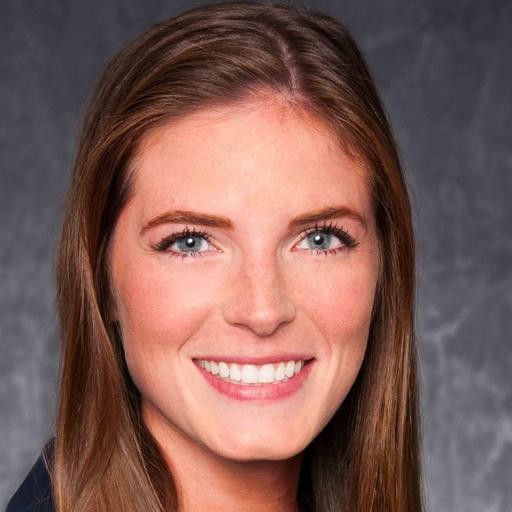} &
        \includegraphics[width=0.19\linewidth]{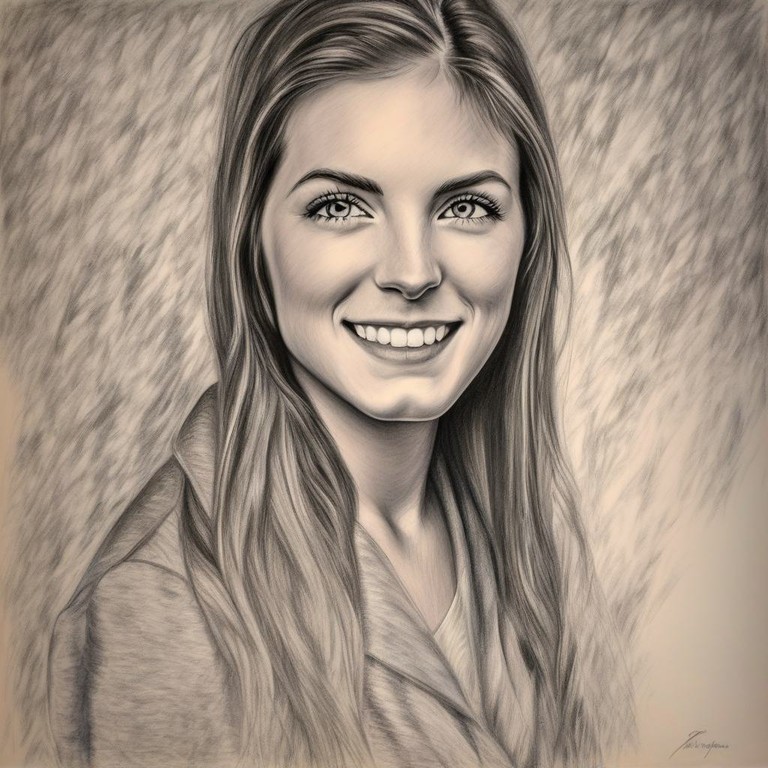} &
        \includegraphics[width=0.19\linewidth, height=0.19\linewidth]{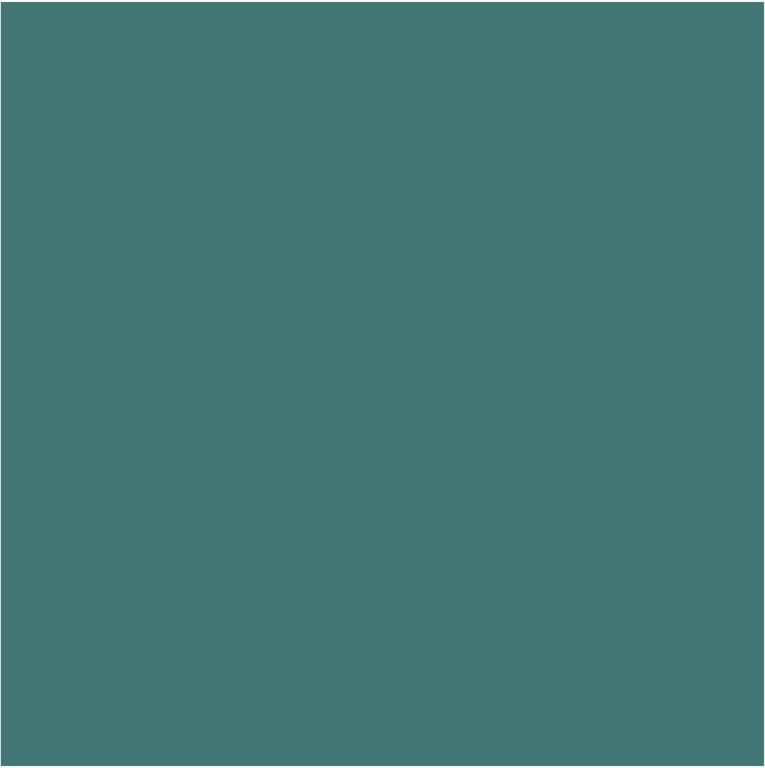} &
        \includegraphics[width=0.19\linewidth]{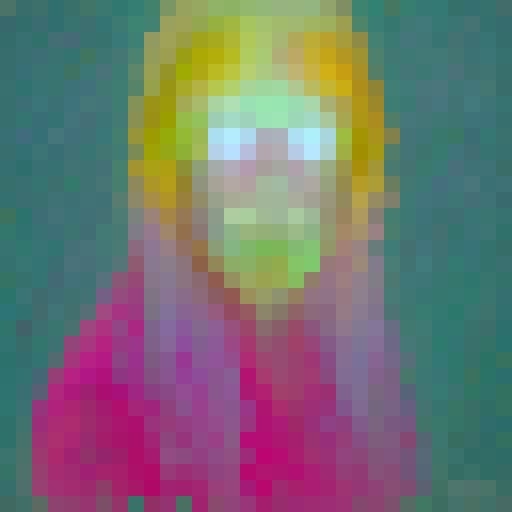} &
        \includegraphics[width=0.19\linewidth]{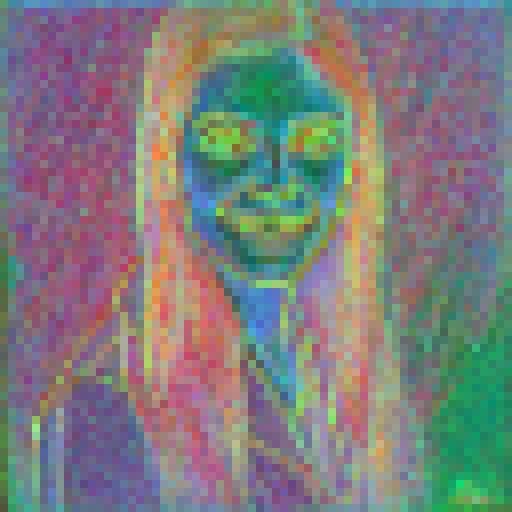} \\
        \multicolumn{5}{c}{``An abstract pencil drawing...''} \\
        \includegraphics[width=0.19\linewidth]{images/v_q/woman_input.jpg} &
        \includegraphics[width=0.19\linewidth]{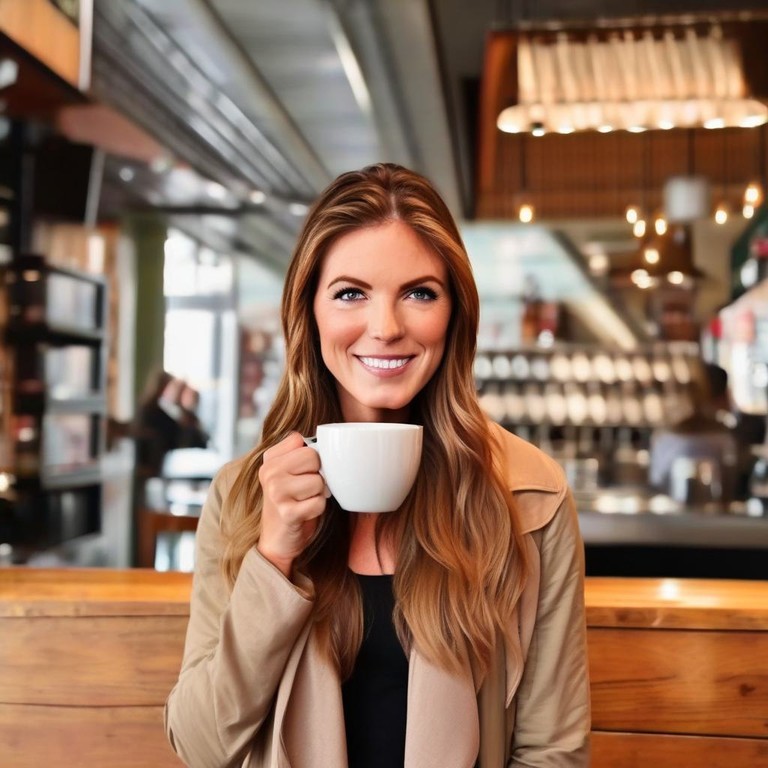} &
        \includegraphics[width=0.19\linewidth, height=0.19\linewidth]{images/v_q/v_q_ca.jpg} &
        \includegraphics[width=0.19\linewidth]{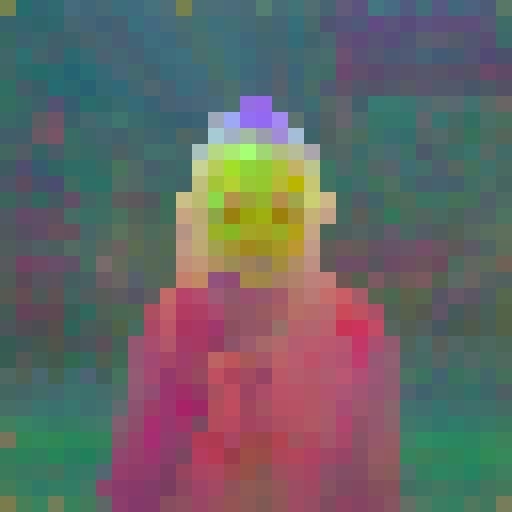} &
        \includegraphics[width=0.19\linewidth]{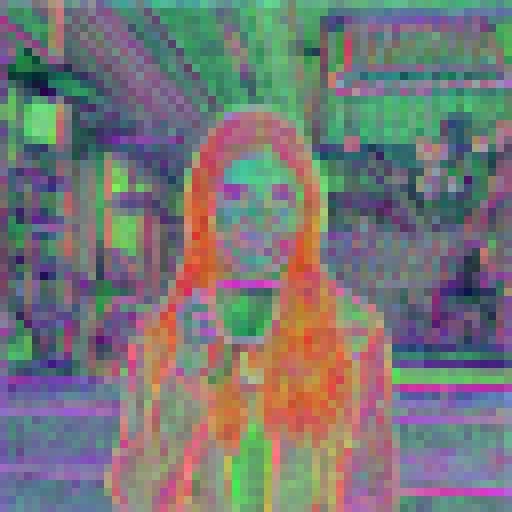} 
        \\
        \multicolumn{5}{c}{``... holding a coffee cup in a coffee shop''} 
        \vspace{-6pt}
    \end{tabular}
    }
    \caption{We visualize the values $V_q[s^*]$ generated for a subject in two different layers, with a vanilla cross-attention, and with our nested approach. Vanilla layers use the same value to represent the subject throughout the entire image (column 3). Nested attention assigns a different subject-value per query (columns 4 and 5), encoding fine-grained semantic information. 
    }
    \vspace{-12pt}
    \label{fig:vq}
\end{figure}

\begin{figure}
    \centering
    \setlength{\tabcolsep}{1pt}
    \footnotesize{
    \begin{tabular}{cc ccc}
        \includegraphics[width=0.23\linewidth]{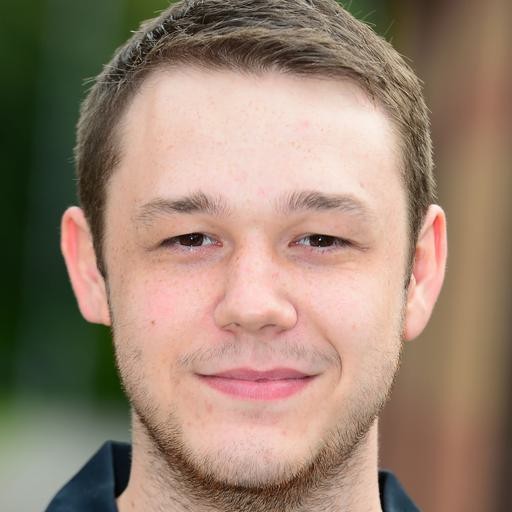} &
        { } &
        \multicolumn{3}{c}{\includegraphics[height=0.23\linewidth]{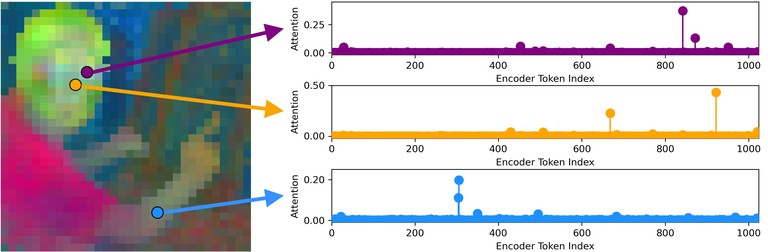}} \\
        Input image & { } & $V_q[s^*]$ (nested & \multicolumn{2}{c}{Nested attention map ($q\Breve{K}$)} \\
        & & attention output) \\
        \includegraphics[width=0.23\linewidth]{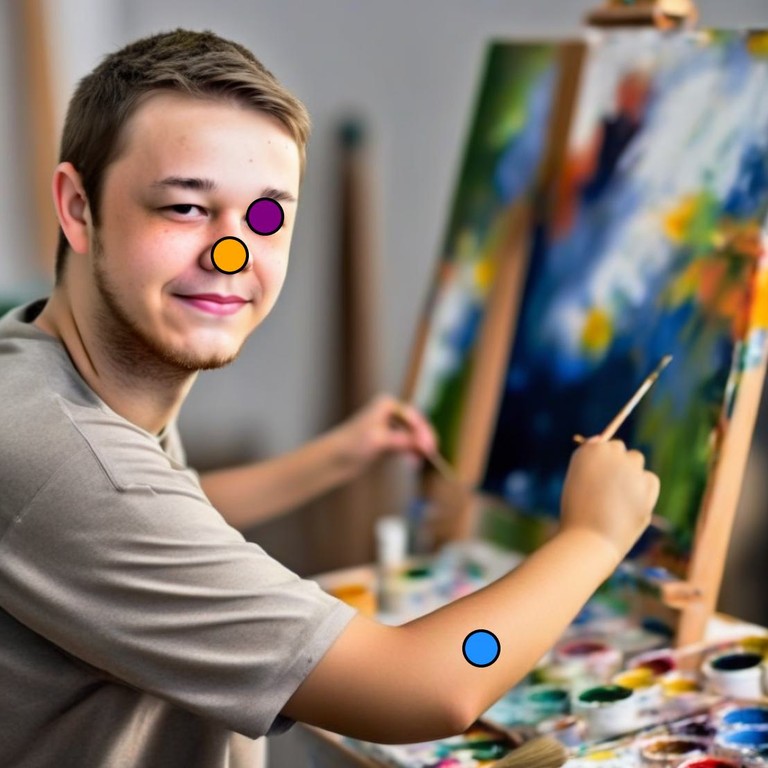} & 
        { } &
        \includegraphics[width=0.23\linewidth]{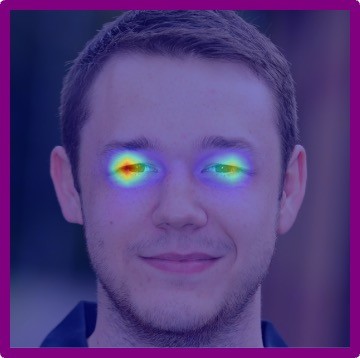} &
        \includegraphics[width=0.23\linewidth]{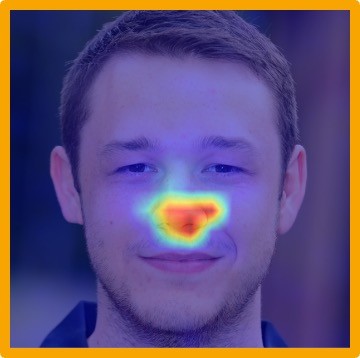} &
        \includegraphics[width=0.23\linewidth]{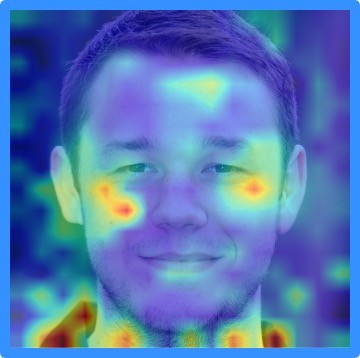} 
        \\
        Generated image 
        && \multicolumn{3}{c}{Q-Former attention map for the token with} \\
        && \multicolumn{3}{c}{highest attention in the nested attention map}
    \end{tabular}
    \vspace{-6pt}
    }
    \caption{
    Analyzing the query-dependent values ($V_q[s^*]$) from a nested attention layer. For three queries of the generated image (purple, orange, blue points), we first show their attention maps in a nested attention layer (graph). There, each point  corresponds to a token produced by the encoder. In each graph, 1-2 tokens dominate the attention. To analyze the information encoded in the most dominant token, we show the Q-Former attention map of its corresponding learned query. These show the semantic alignment between the probed query, and the source of values assigned to it.
    }
    \vspace{-12pt}
    \label{fig:vq-attn-maps}
\end{figure}

However, for personalization tasks, we require particularly high accuracy when generating a specific subject. Our key idea is to increase the expressiveness of the token corresponding to the subject, denoted by $s^*$, without overwhelming the rest of the prompt. We do so by introducing localized values that depend on the queries. These localized values can then be more specialized, representing for example the appearance of the individual's eyes or hair, without having to represent the individual's entire appearance in a single embedding (see \Cref{fig:vq}). 
To compute these per-region values, we propose to use another attention mechanism, which can itself link the semantic content of each region to a set of feature vectors extracted from the image (see \Cref{fig:vq-attn-maps}). We term this internal attention layer ``Nested Attention'', and its output is given by:
\vspace{-4pt}
\begin{equation*}
    \vspace{-3pt}
    v^*_{q_{ij}} =  \softmax \left( \frac{q_{ij}\Breve{K}^T}{\sqrt{d}} \right)\Breve{V} , 
\end{equation*}
where $q_{ij}$ is the query vector of spatial patch $(i,j)$ in the external cross-attention layer. $\Breve{K}$ and $\Breve{V}$ are the keys and values of the nested attention layer, given through linear projections parameterized by $W_{\Breve{K}}$ and $W_{\Breve{V}}$. Finally,  
$v^*_{q_{ij}}$ are the query-dependent values of the personalized token $s^*$ at spatial index $(i,j)$ (\ie, corresponding to $q_{ij}$).

These are then used in the external cross-attention layer through:
\vspace{-6pt}
\begin{equation*}
\vspace{-3pt}
\begin{aligned}
    \phi_{\text{out}}^{\ell}(z_t)_{ij} = \softmax \left( \frac{q_{ij}K^T}{\sqrt{d}} \right)V_{q_{ij}}, \\
    V_{q_{ij}}[s] =  \begin{cases}
      v^*_{q_{ij}}, & \text{if}\ s=s^* \\
      V[s], & \text{otherwise,}
    \end{cases}
\end{aligned}
\end{equation*}
where $s$ are the prompt's textual tokens. 
Through this two-stage attention mechanism, we allow the model to benefit from a rich-multi token representation of the image, while still tying all features to a single prompt token. The full nested attention mechanism is illustrated in Figure~\ref{fig:nested-mechanism}. 

\vspace{-10pt}
\paragraph{Regularizing $\pmb{V_q[s^*]}$} \label{sec:method-reg}
Prior work~\cite{tewel2023key,alaluf2023neural} has shown that personalization approaches that tie the novel subject to an embedding in the existing cross-attention layers can suffer from ``attention overfitting'', where the new token draws the attention from all image queries, leading the rest of the prompt to be ignored. Our approach aims to avoid this pitfall by predicting only attention values, while preserving the original keys assigned to the un-personalized word.

However, we note that this property can break if the norm of values generated by the nested attention, $V_q[s^*]$, is significantly higher than that of the original cross-attention values $V[s^*]$ obtained from the text embedding. Indeed, increasing the norm of $V_q[s^*]$ resembles the case where the attention given to $s^*$ is higher. To avoid this issue, we regularize the norm of each of the learned values $v^*_{q_{ij}}$ in $V_q[s^*]$ to be $\alpha |V[s^*]|$ where $\alpha$ is a fixed hyperparameter, and $|V[s^*]|$ is the norm of the cross-attention value of the un-personalized word. In our experiments, we set $\alpha=2$. Ablations on this choice are provided in Appendix~\ref{app:supp-ablate-normalization}.

\subsection{Encoder for Personalization}
\vspace{-2pt}

To personalize the text-to-image model, we incorporate nested attention layers into all of its cross-attention layers while keeping the original model's weights frozen during training. We train an encoder that produces tokens from which the keys and values of the nested layers are derived. An overview of this architecture is shown in Figure~\ref{fig:encoder-overview}.

The encoder's backbone is based on CLIP~\cite{radford2021learning}. Given an input image, we pass it through CLIP and extract tokens from its last layer before pooling. These tokens are then processed by a Q-Former~\cite{li2023blip}, where the number of learned queries of the Q-Former determines the number of nested keys and values. During training, CLIP remains frozen while the Q-Former is trained from scratch.

For training the encoder and nested attention layers, we utilize datasets consist of (input image, text prompt, target image) triplets. For human faces the target image is an in-the-wild image of the person and the input image is the cropped and aligned face. For pets, the input and target images are identical. In each triplet, the text prompt describes the target image, where we replace the word related to the subject (\eg, ``girl'') with the token $s^*$, which is set to ``person'' for human faces, and ``pet'' for pets.
The training procedure follows that of diffusion models: we add noise to the target image and then predict this noise using the diffusion model conditioned on input image. This approach allows our model to learn personalized representations while maintaining the prior of the original diffusion model.

\section{Analysis}

\begin{figure}
    \centering
    \setlength{\tabcolsep}{0pt}
    \small{
    \begin{tabular}{ccccc}
        \includegraphics[width=0.19\linewidth]{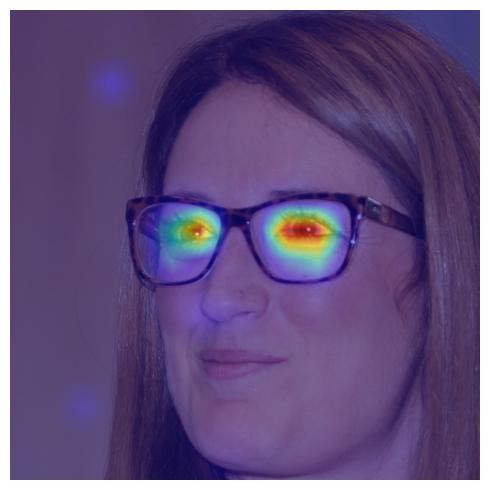} &
        \includegraphics[width=0.19\linewidth]{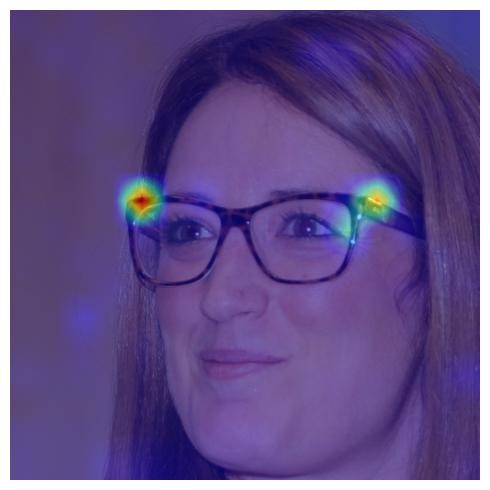} &
        \includegraphics[width=0.19\linewidth]{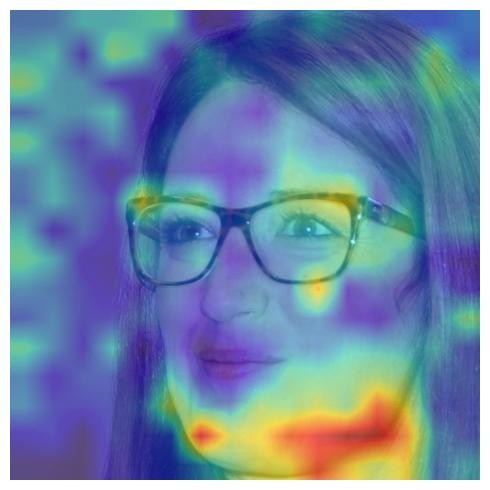} &
        \includegraphics[width=0.19\linewidth]{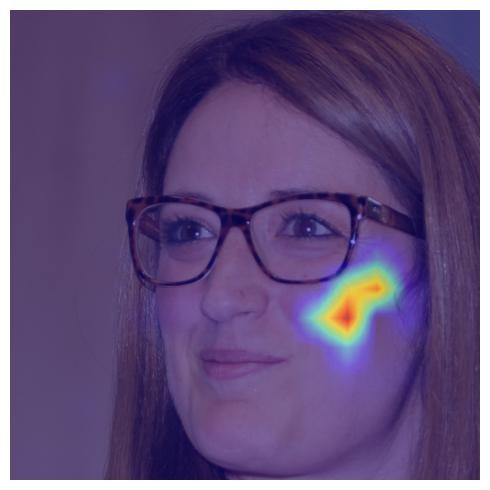} &
        \includegraphics[width=0.19\linewidth]{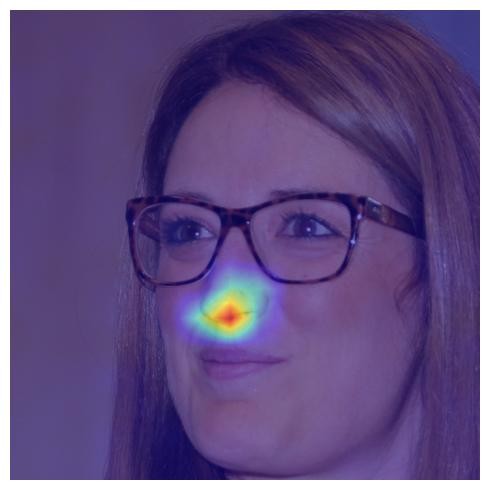} \\
        \includegraphics[width=0.19\linewidth]{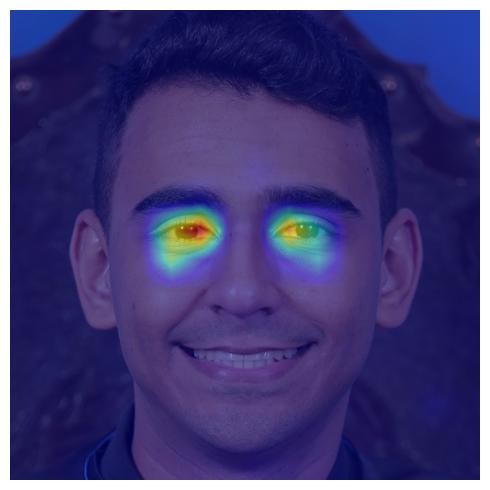} &
        \includegraphics[width=0.19\linewidth]{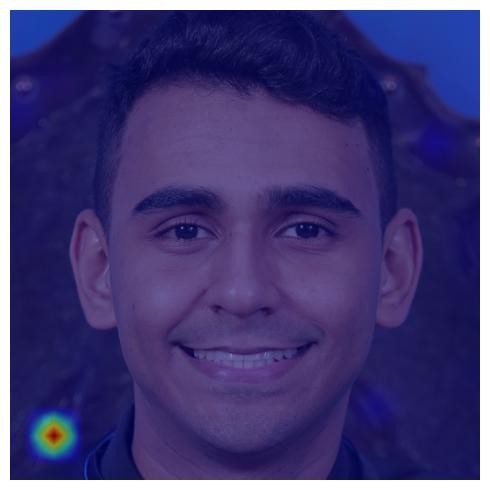} &
        \includegraphics[width=0.19\linewidth]{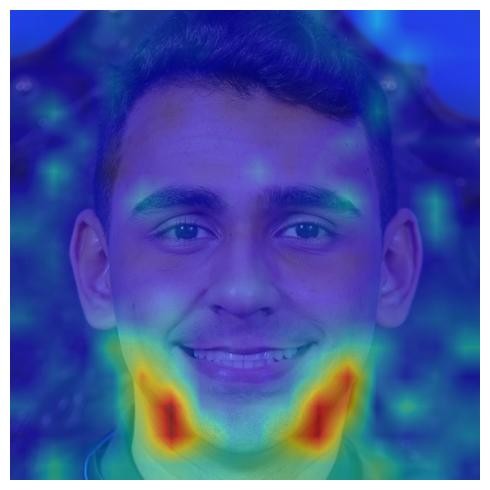} &
        \includegraphics[width=0.19\linewidth]{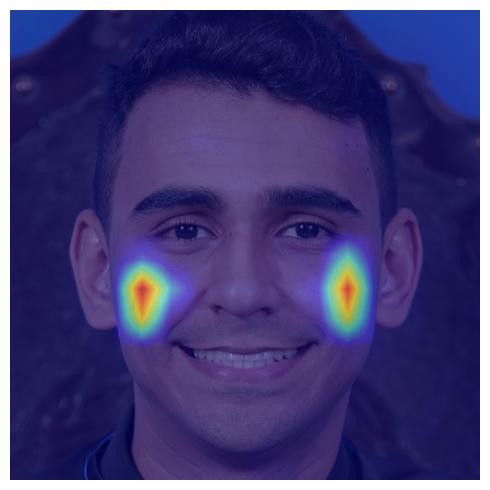} &
        \includegraphics[width=0.19\linewidth]{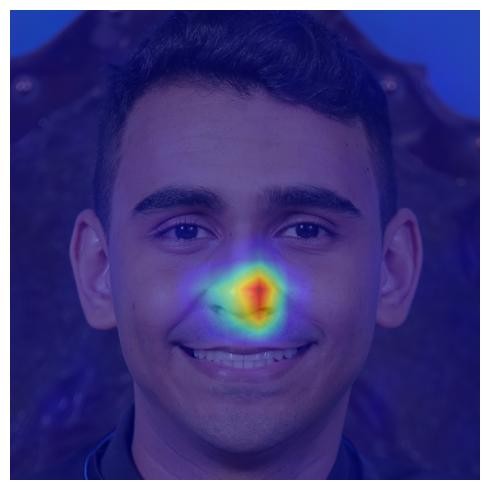} \\
    \end{tabular}
    \vspace{-12pt}
    }
    \caption{
        Attention maps between Q-Former learned queries and input image features. Each column shows a distinct query's attention map, illustrating how queries capture different facial features.
    }
    \vspace{-14pt}
    \label{fig:leanred-q-attn}
\end{figure}

\paragraph{What does the Q-Former learn?}
We begin by examining the features learned by the Q-Former component of the encoder. To do so, we visualize the attention maps between each learned query and the input image features. \Cref{fig:leanred-q-attn} displays attention maps for five sample learned queries across two different input images. The figure demonstrates that each learned query captures distinct semantic facial features. For instance, the leftmost column's query focuses on the eyes, while the rightmost column's query captures the nose. In the second column, the query attends to part of the glasses in the top image. Notably, the man in the bottom row does not wear glasses, resulting in a less meaningful attention map for this query with that particular input image.

\vspace{-13pt}
\paragraph{The importance of query-dependent values}
To achieve accurate identity preservation, nested attention layers generate query-dependent values, $V_q[s^*]$, for the personalized subject token, $s^*$. These query-dependent values, enhance identity preservation by encoding fine-grained details from the input image. \Cref{fig:vq} visualizes these values for two prompts, using the same input image. We show $V_q[s^*]$ from two different layers: at a $32\times32$ resolution, and at $64\times64$. These are captured at two-thirds of the way through the full denoising process.
For clarity, we show the values of $s^*$ from a standard cross-attention layer without nested attention, which remains constant across the image.
The visualizations show that our generated values are context-aware and capture the intricate visual details of the input image.

Finally, we show the semantic connection between the localized attention-values produced by the nested attention layer, and the input image.
In Figure~\ref{fig:vq-attn-maps}, we select three locations on the generated image: the eye, nose, and arm. For each of these queries, we show the attention map of a single nested attention layer. As can be seen, there are typically 1-2 dominant encoder tokens for the each query location. We can then follow these tokens to their source in the input image, using the approach of \Cref{fig:leanred-q-attn}. As can be seen, the information encoded in the query-dependent value corresponding to the generated eye and nose mostly comes from the eye and nose regions of the input image, respectively. This indicates that the query-dependent values produced by our nested attention layer contain relevant, localized information matching the semantics of the input image. When considering the arm region, the input image does not contain an area with matching semantics, but we can observe that the model focuses on the neck region (and the boy's shirt), and partially on skin areas on the boy's face.

\section{Experiments}\label{sec:experiments}

\paragraph{Implementation details}
We implement our method with SDXL~\cite{podell2024sdxl} which generates $1024\times1024$ images. The encoder is trained in two phases: $100$ epochs on resolution of $512$, and then additional $100$ epochs on resolution of $1024$. 
We train the human face model on FFHQ-Wild~\cite{karras2019style}, and the pets model on a combination of datasets~\citep{mokady2022selfdistilled,choi2020starganv2}. Additional implementation details are provided in Appendix~\ref{app:details}.

\subsection{Qualitative Results}
\begin{figure*}[t]
    \centering
    \begin{minipage}[t]{0.48\textwidth}
        \centering
        \setlength{\tabcolsep}{1pt}
        \scriptsize{
        \begin{tabular}{cccc}
            \includegraphics[width=0.23\linewidth]{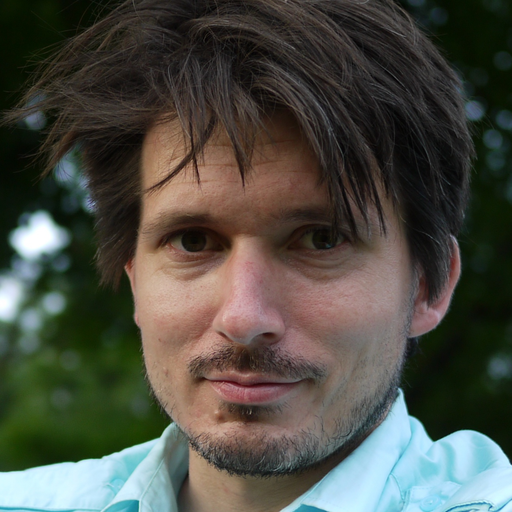} &
            \includegraphics[width=0.23\linewidth]{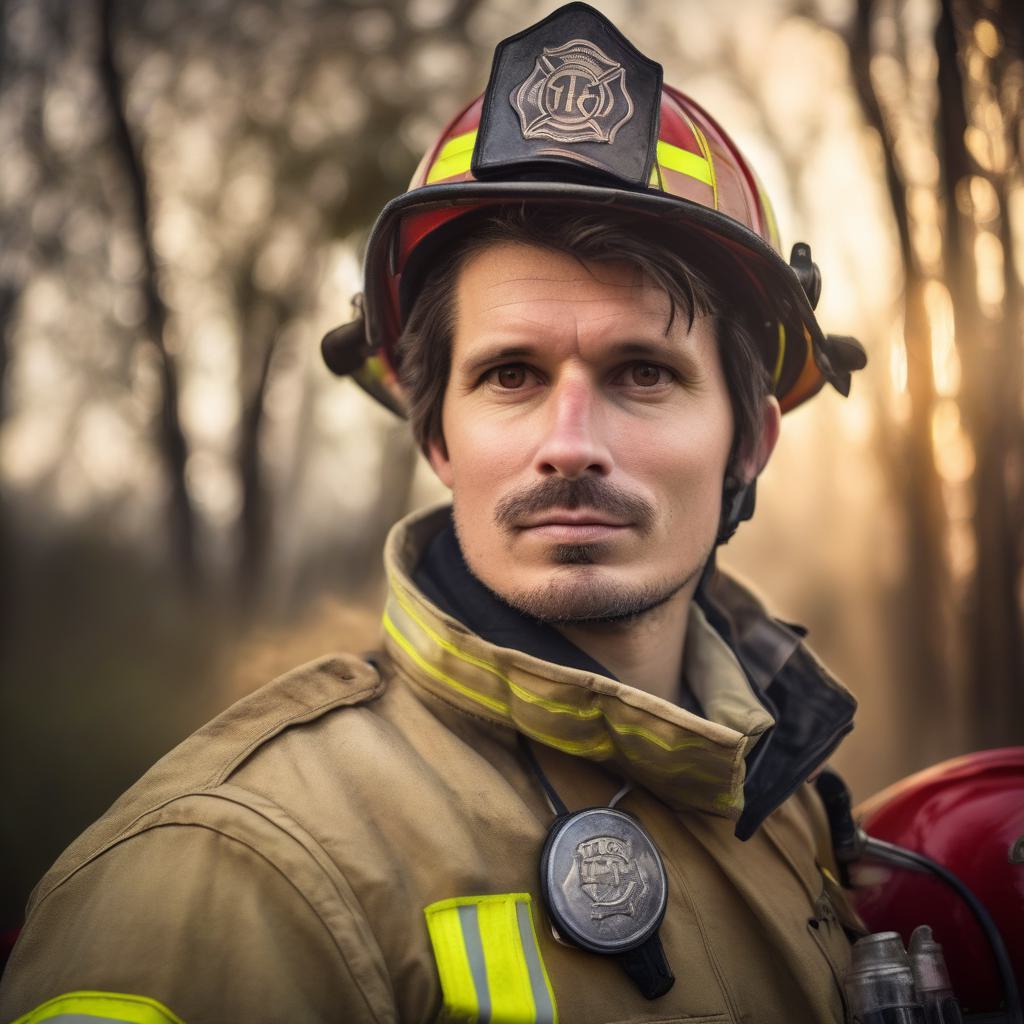} &
            \includegraphics[width=0.23\linewidth]{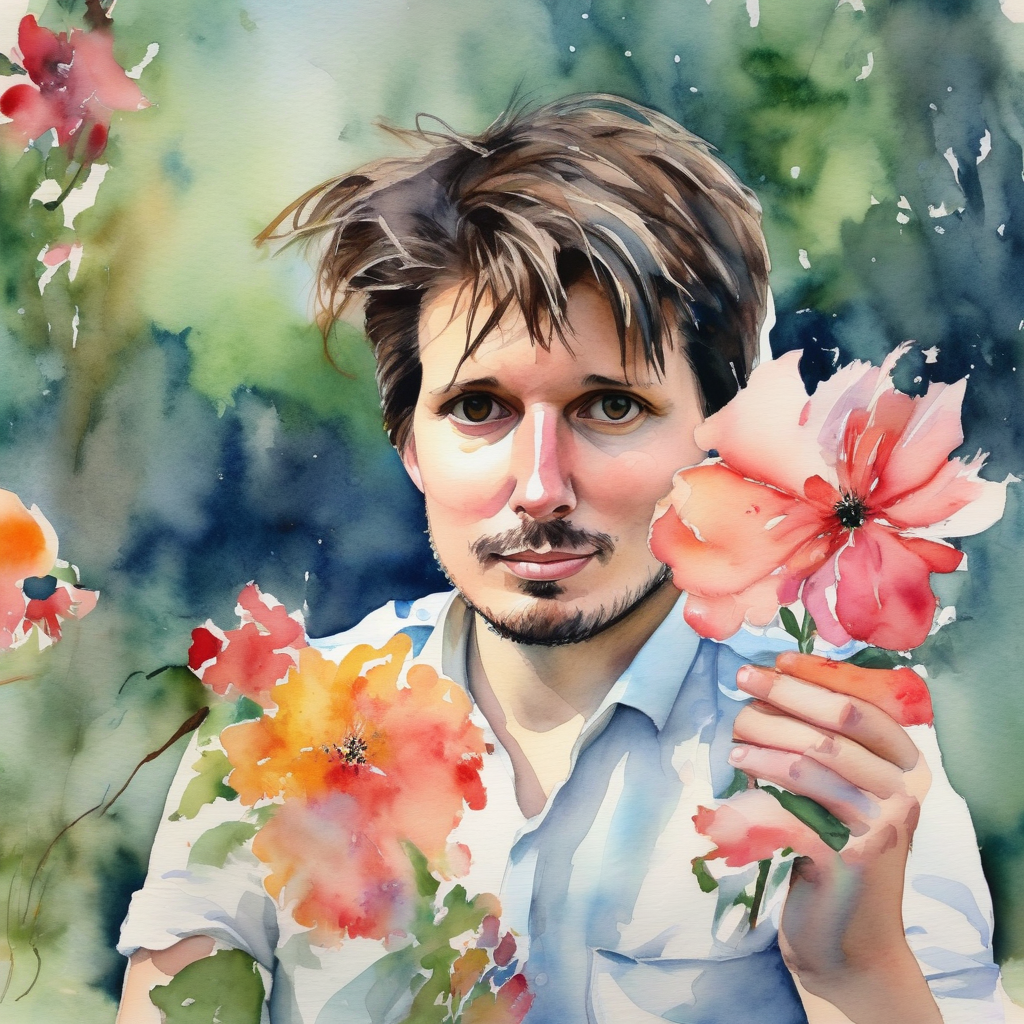} &
            \includegraphics[width=0.23\linewidth]{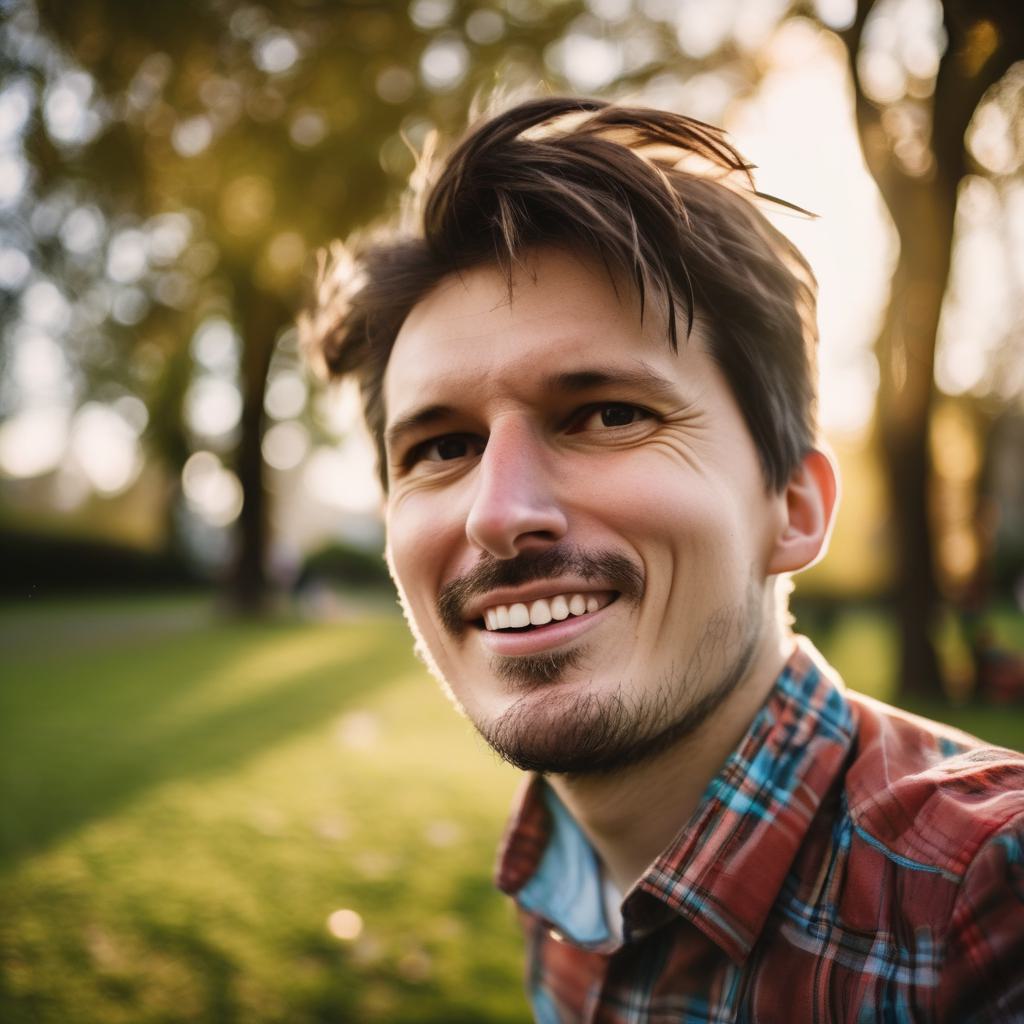} \\
            \includegraphics[width=0.23\linewidth]{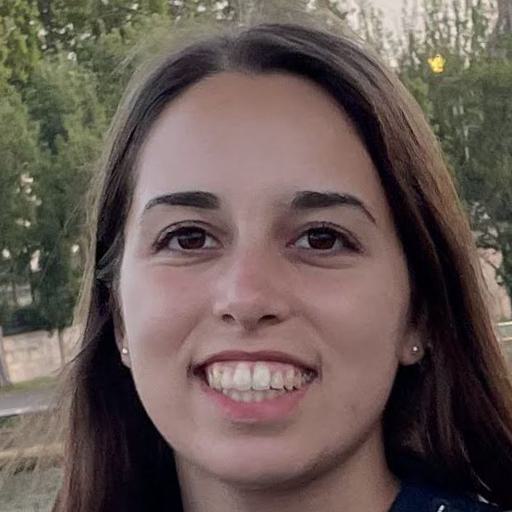} &
            \includegraphics[width=0.23\linewidth]{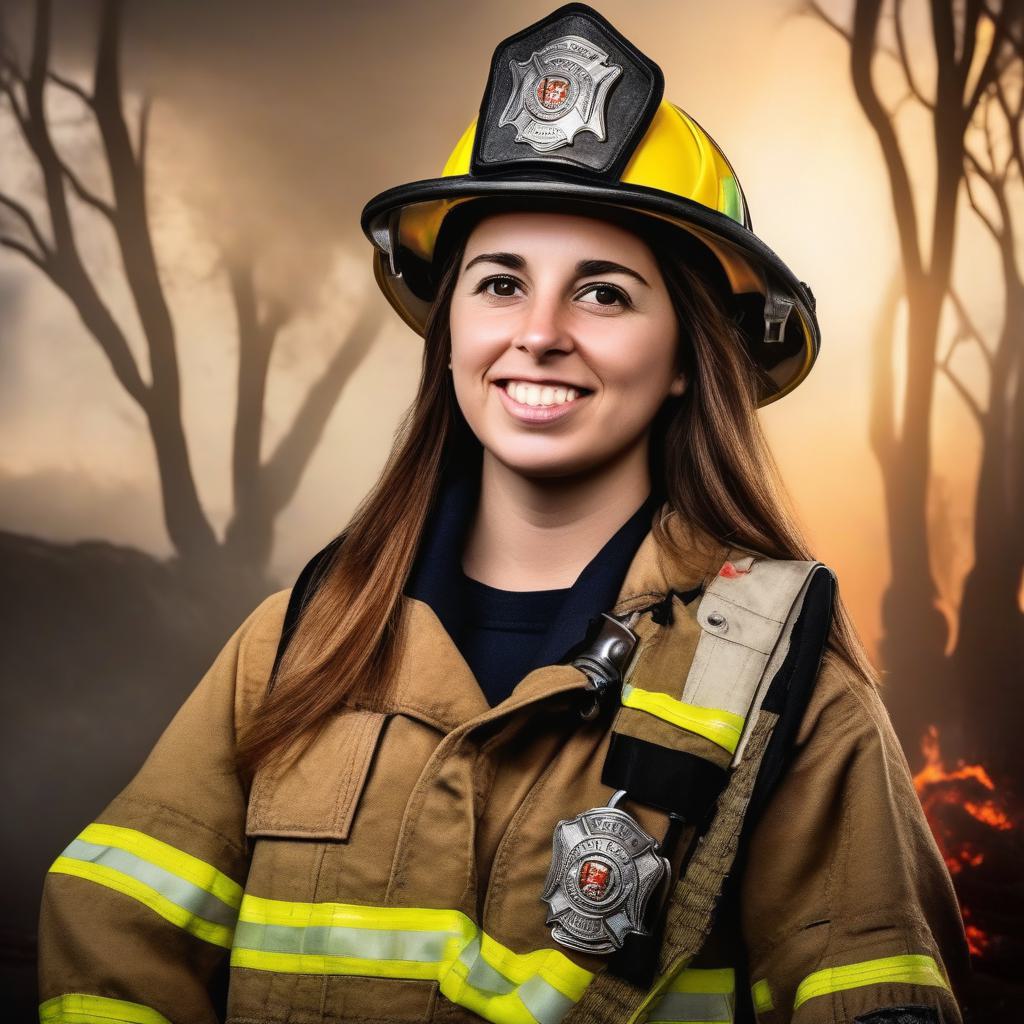} &
            \includegraphics[width=0.23\linewidth]{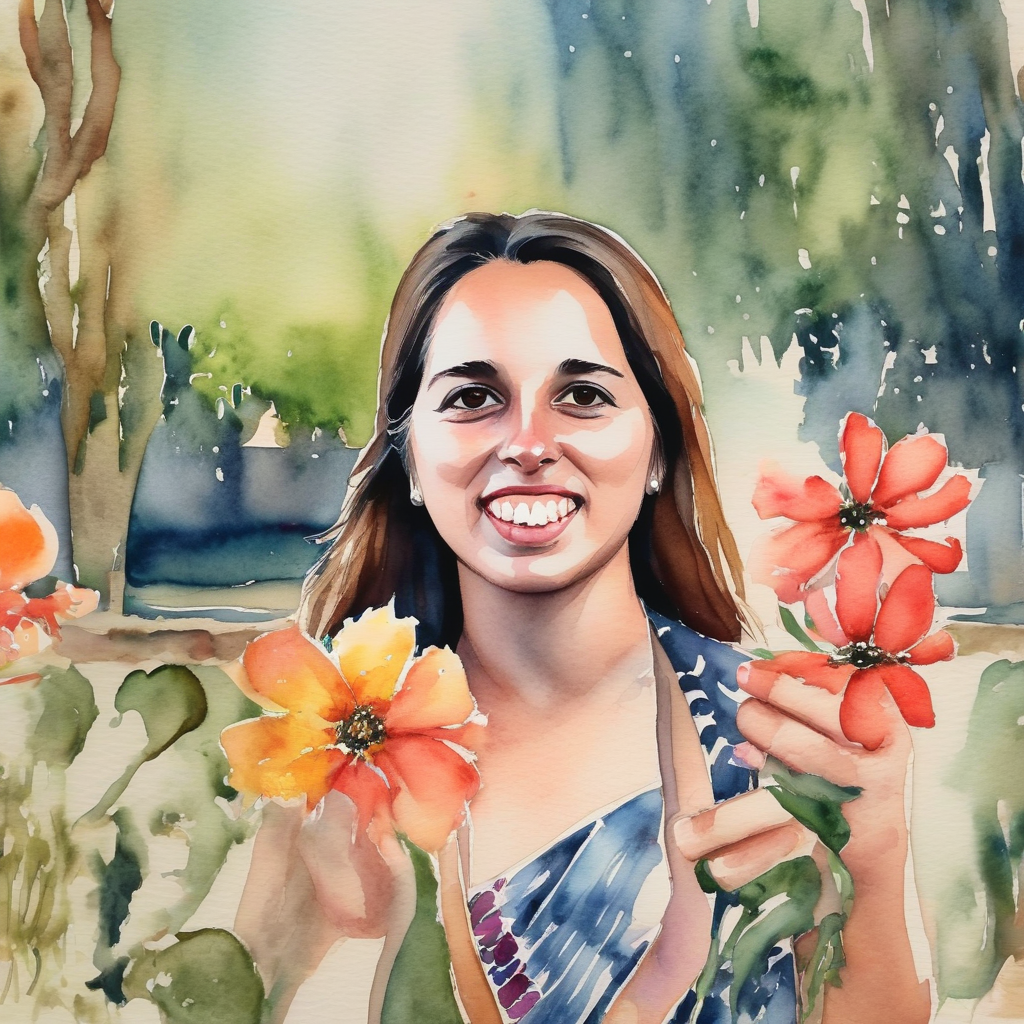} &
            \includegraphics[width=0.23\linewidth]{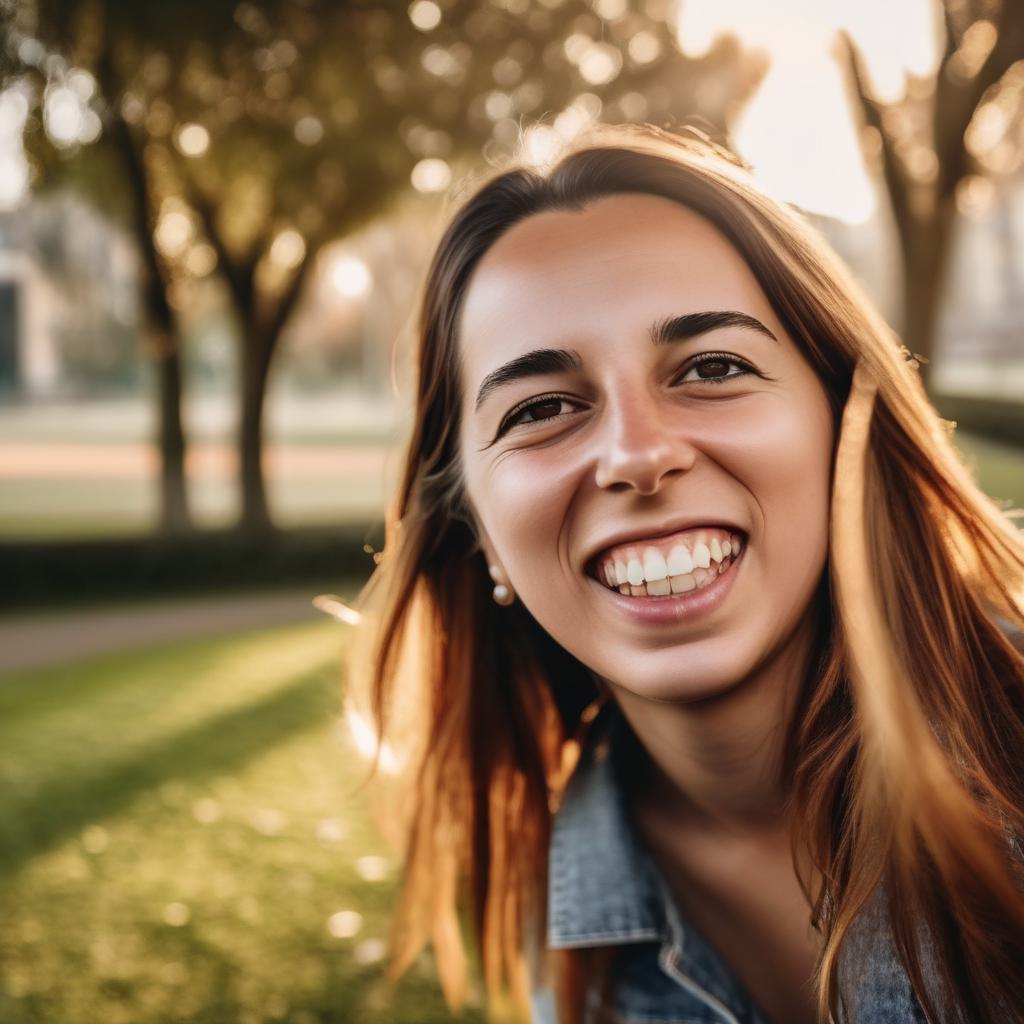} \\
            \includegraphics[width=0.23\linewidth]{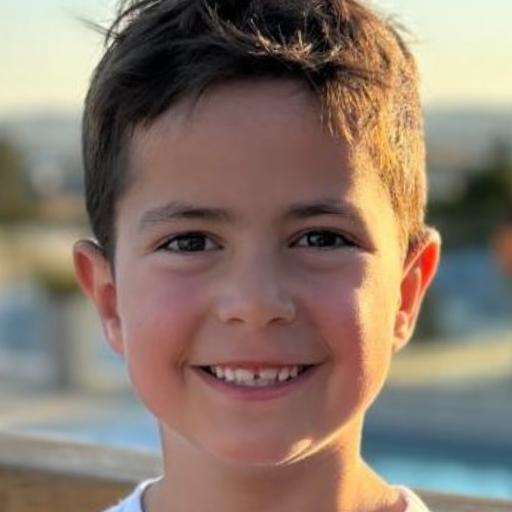} &
            \includegraphics[width=0.23\linewidth]{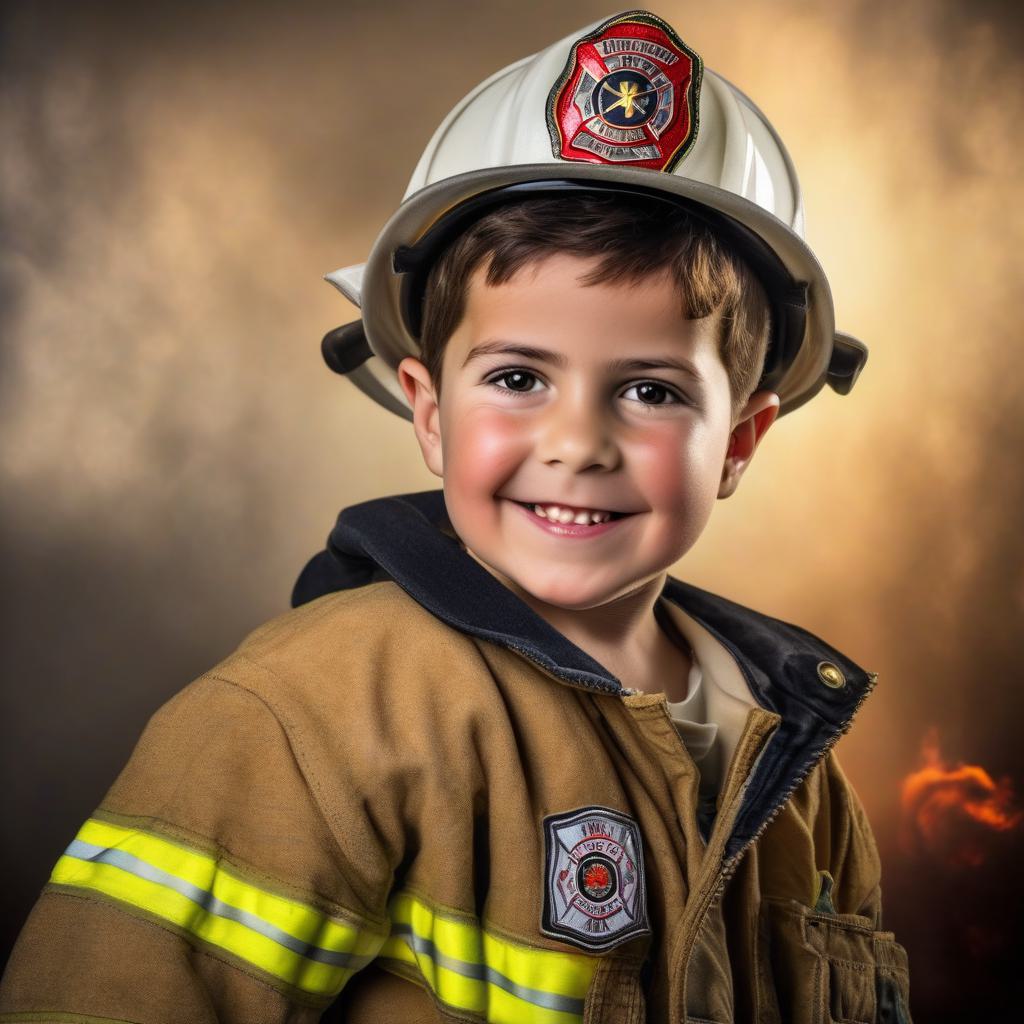} &
            \includegraphics[width=0.23\linewidth]{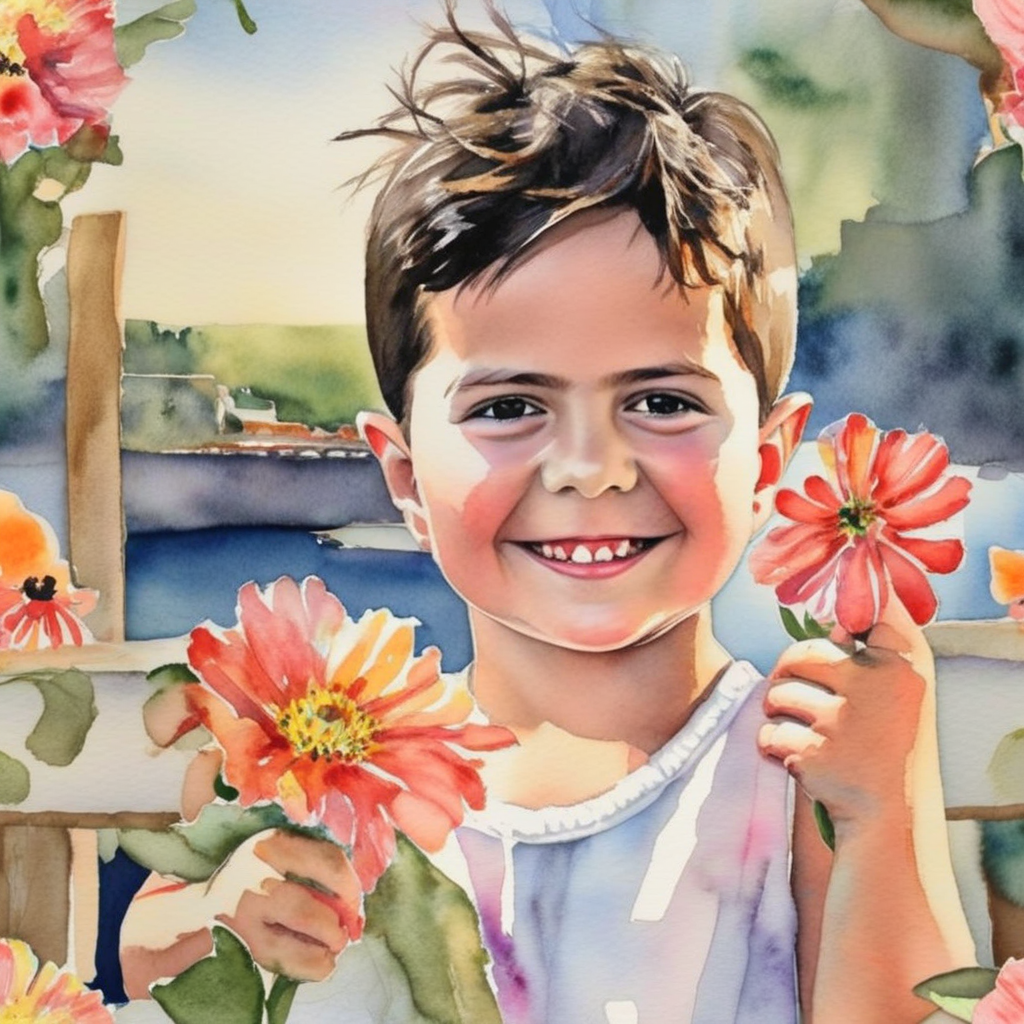} &
            \includegraphics[width=0.23\linewidth]{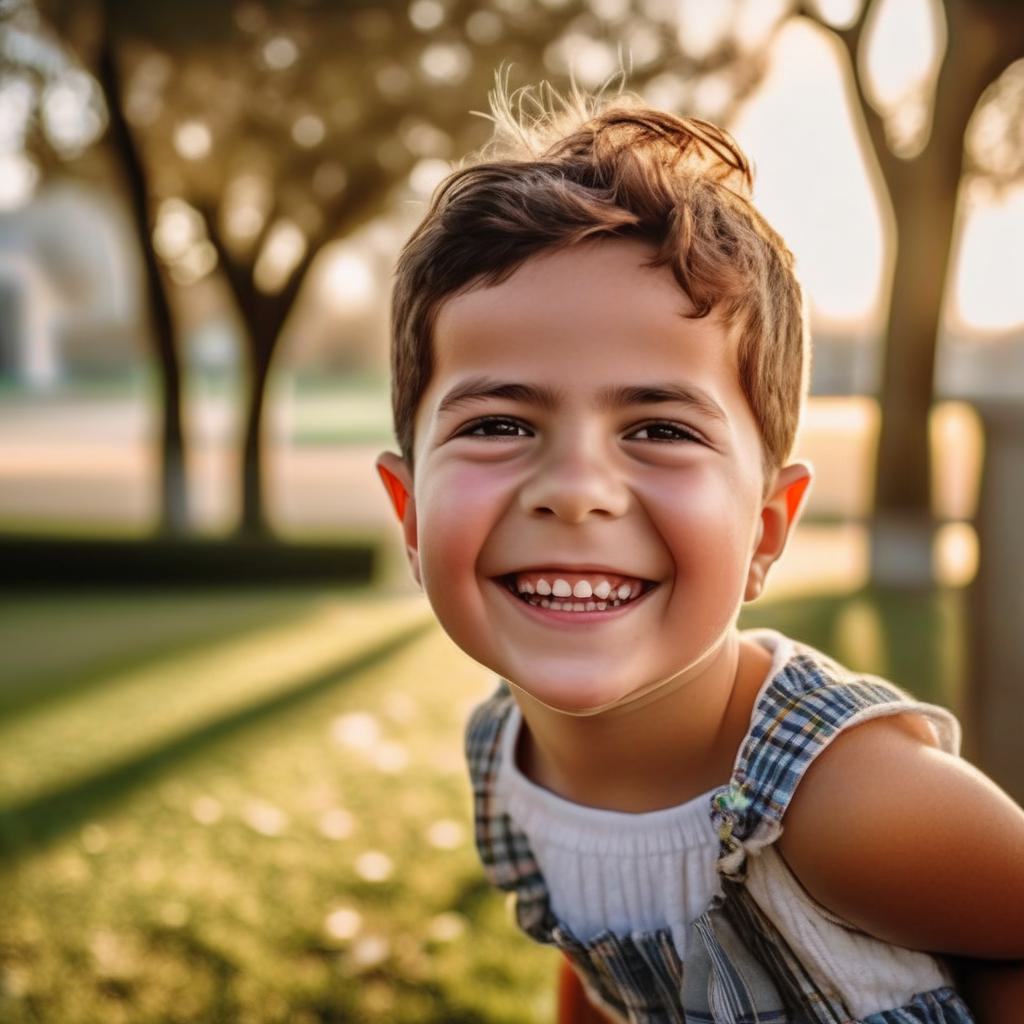} \\
            Input & ``Firefighter'' & ``Watercolor, & ``Laughing, in \\
            & & holding flower'' & the park''
        \end{tabular}
        }
    \end{minipage}%
    \hfill
    \begin{minipage}[t]{0.48\textwidth}
        \centering
        \setlength{\tabcolsep}{1pt}
        \scriptsize{
        \begin{tabular}{cccc}
            \includegraphics[width=0.23\linewidth]{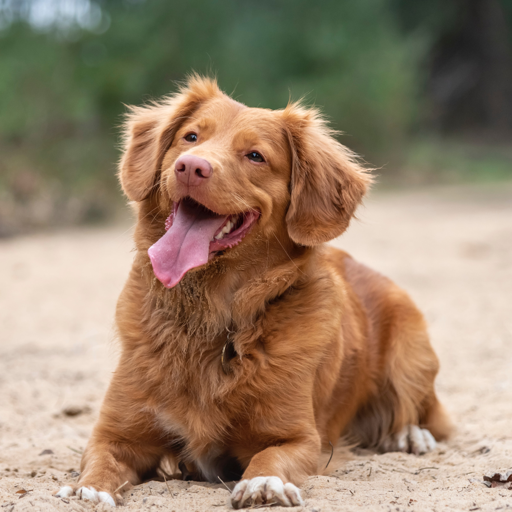} &
            \includegraphics[width=0.23\linewidth]{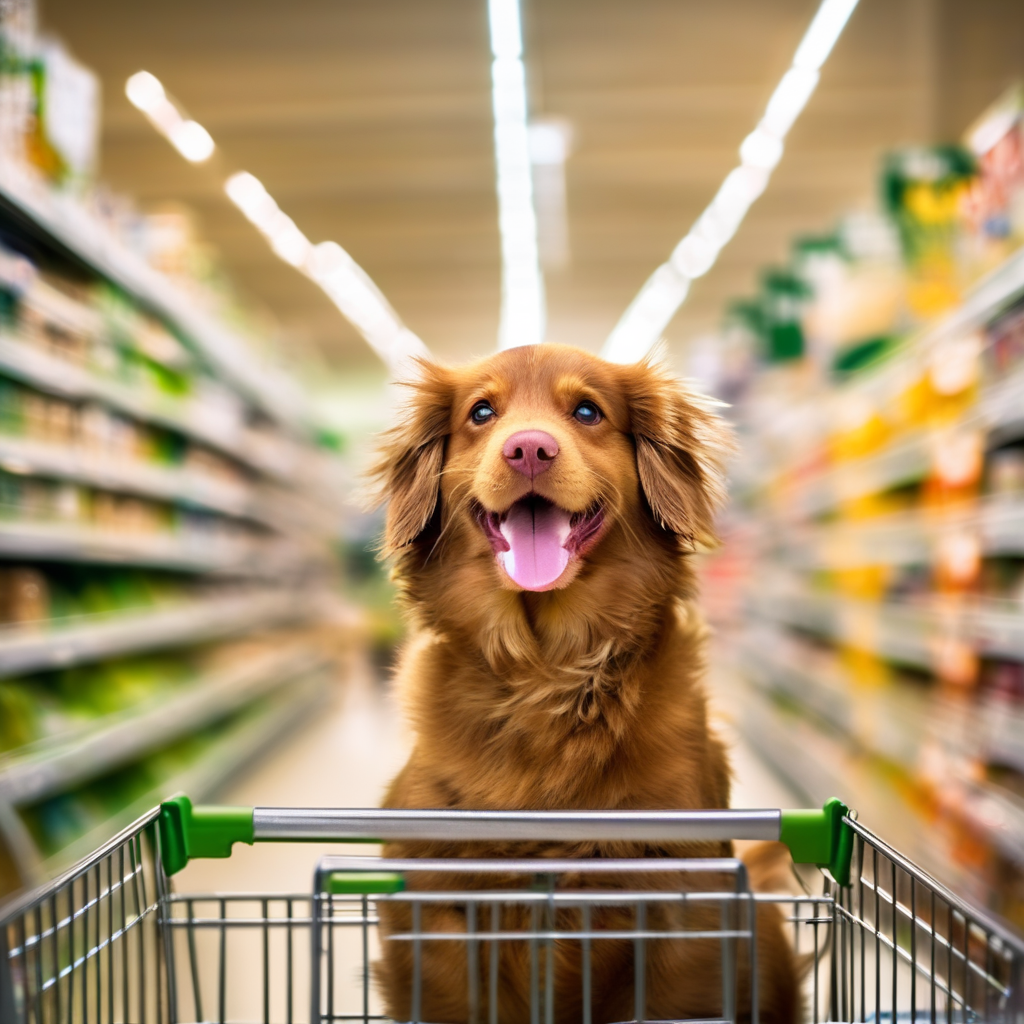} &
            \includegraphics[width=0.23\linewidth]{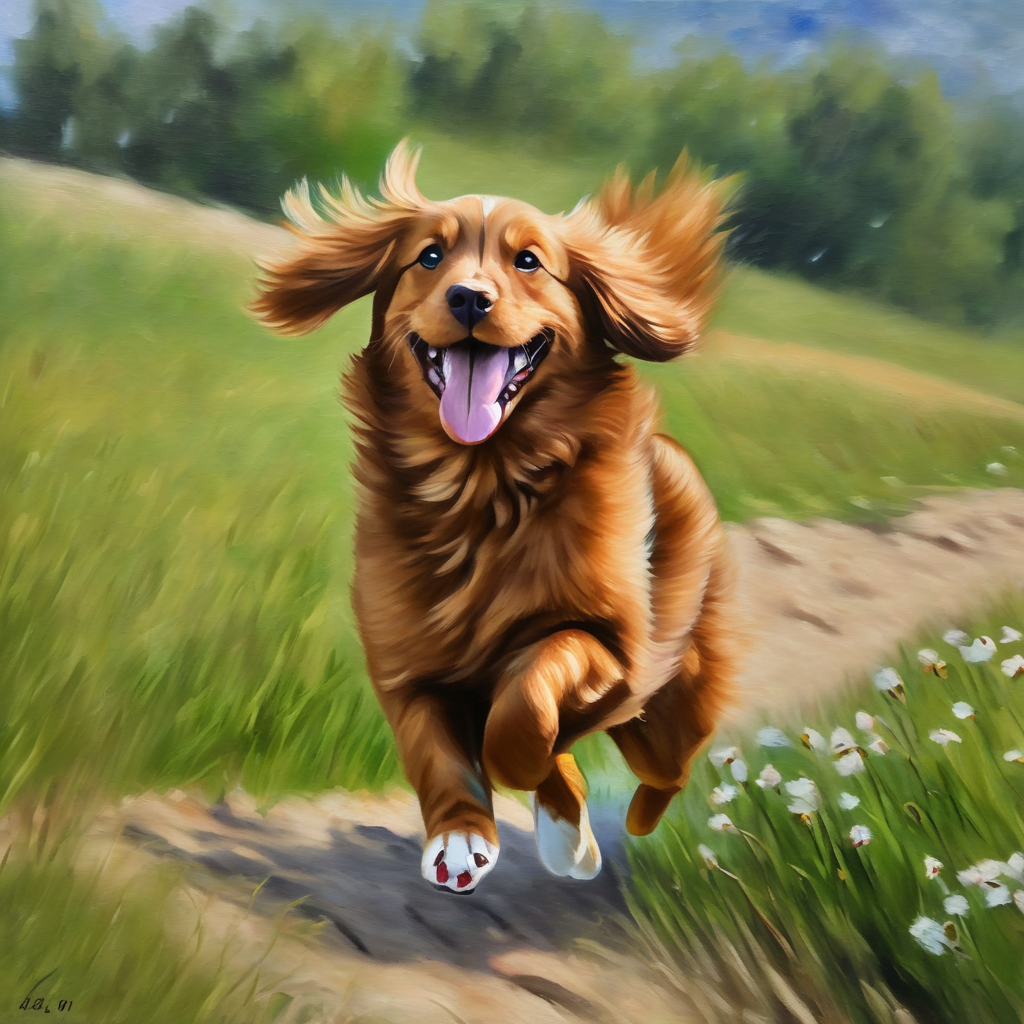} &
            \includegraphics[width=0.23\linewidth]{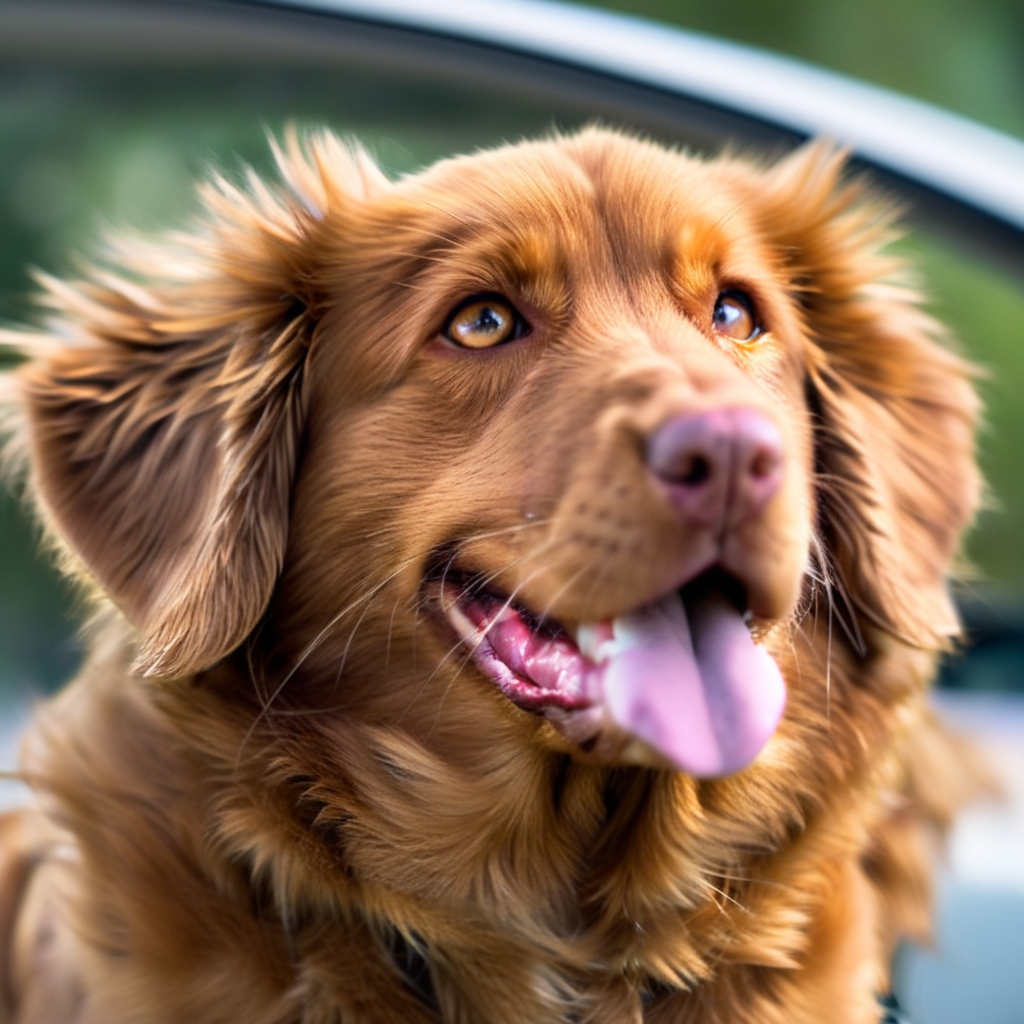} \\
            \includegraphics[width=0.23\linewidth]{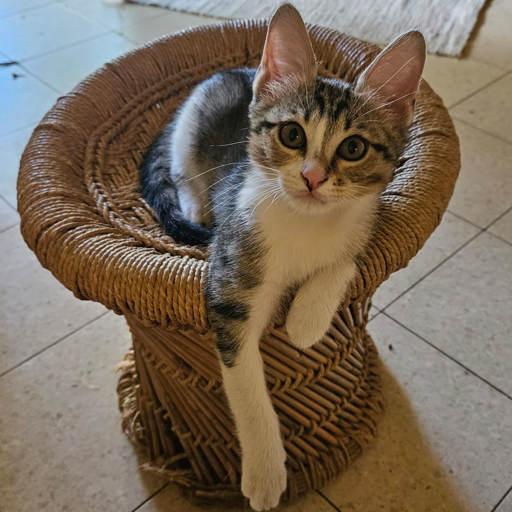} &
            \includegraphics[width=0.23\linewidth]{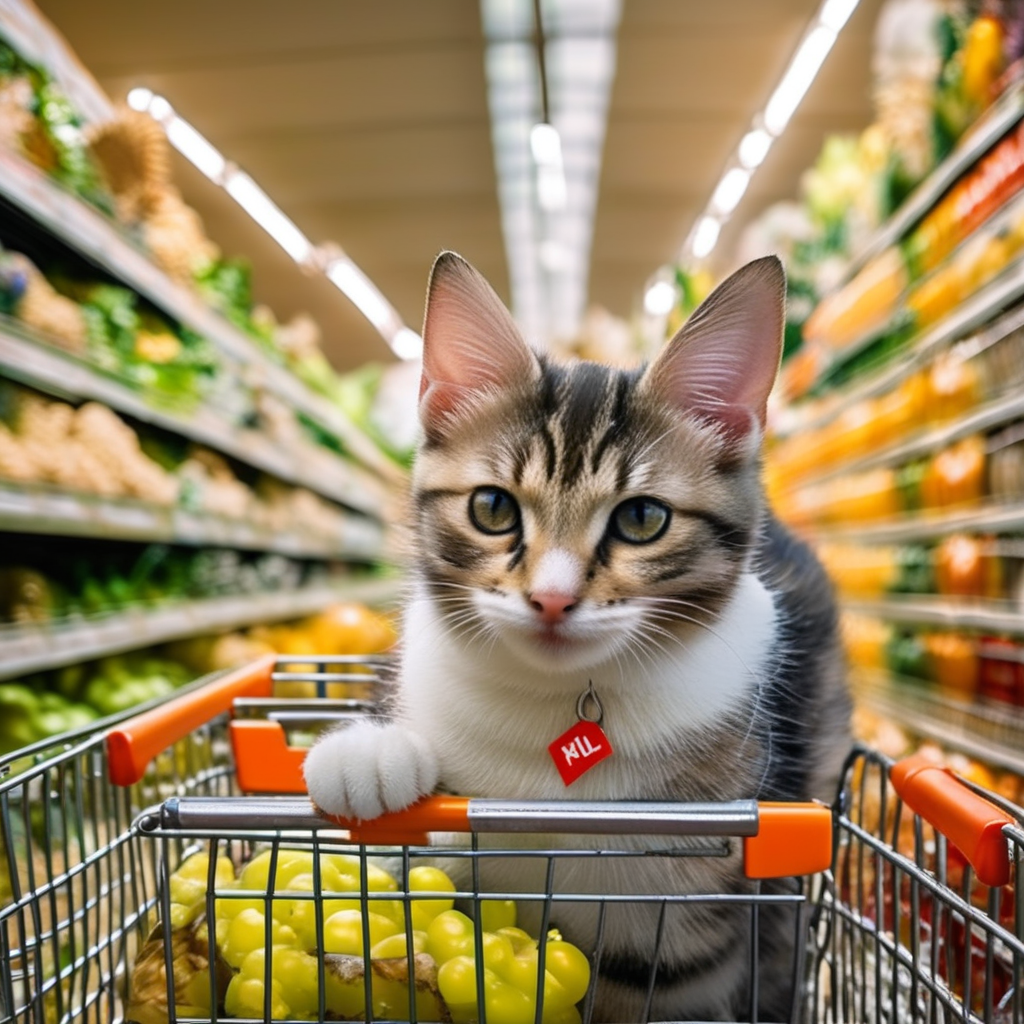} &
            \includegraphics[width=0.23\linewidth]{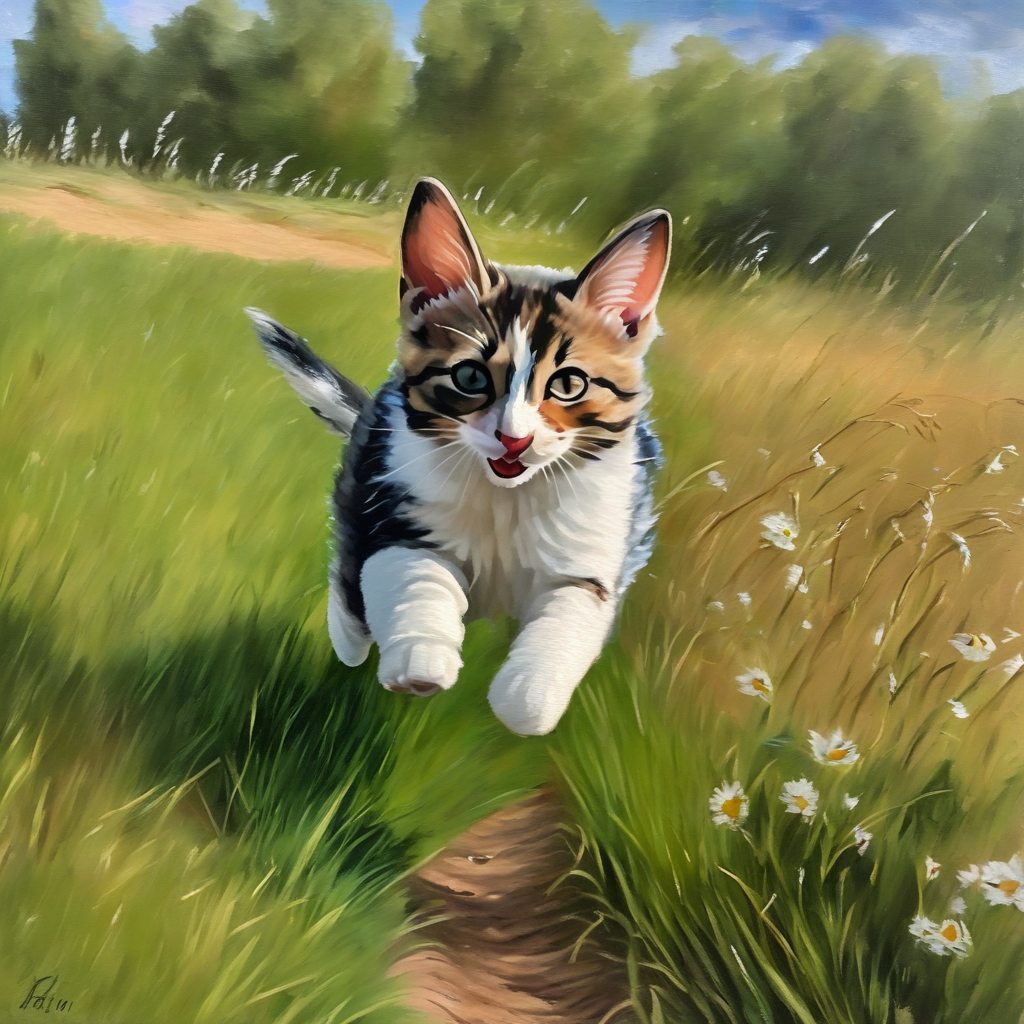} &
            \includegraphics[width=0.23\linewidth]{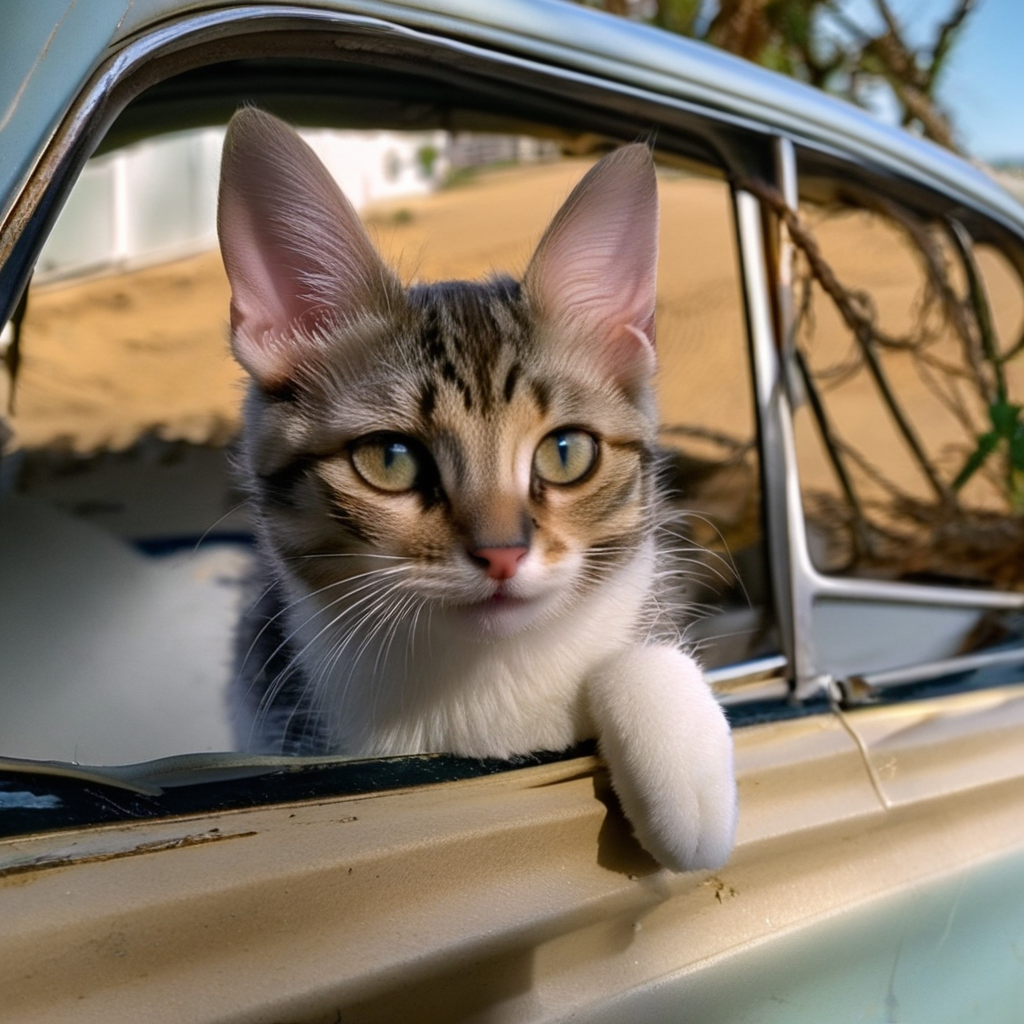} \\
            \includegraphics[width=0.23\linewidth]{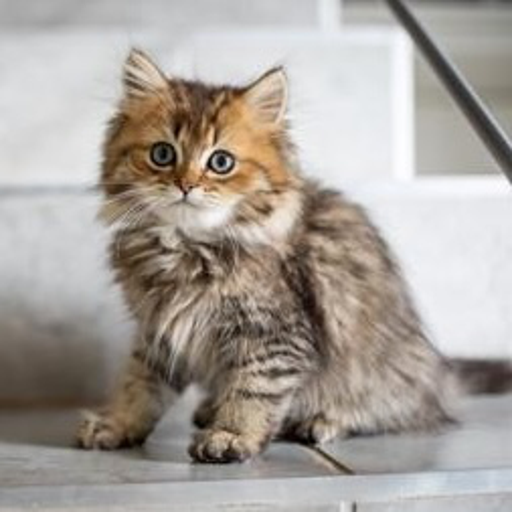} &
            \includegraphics[width=0.23\linewidth]{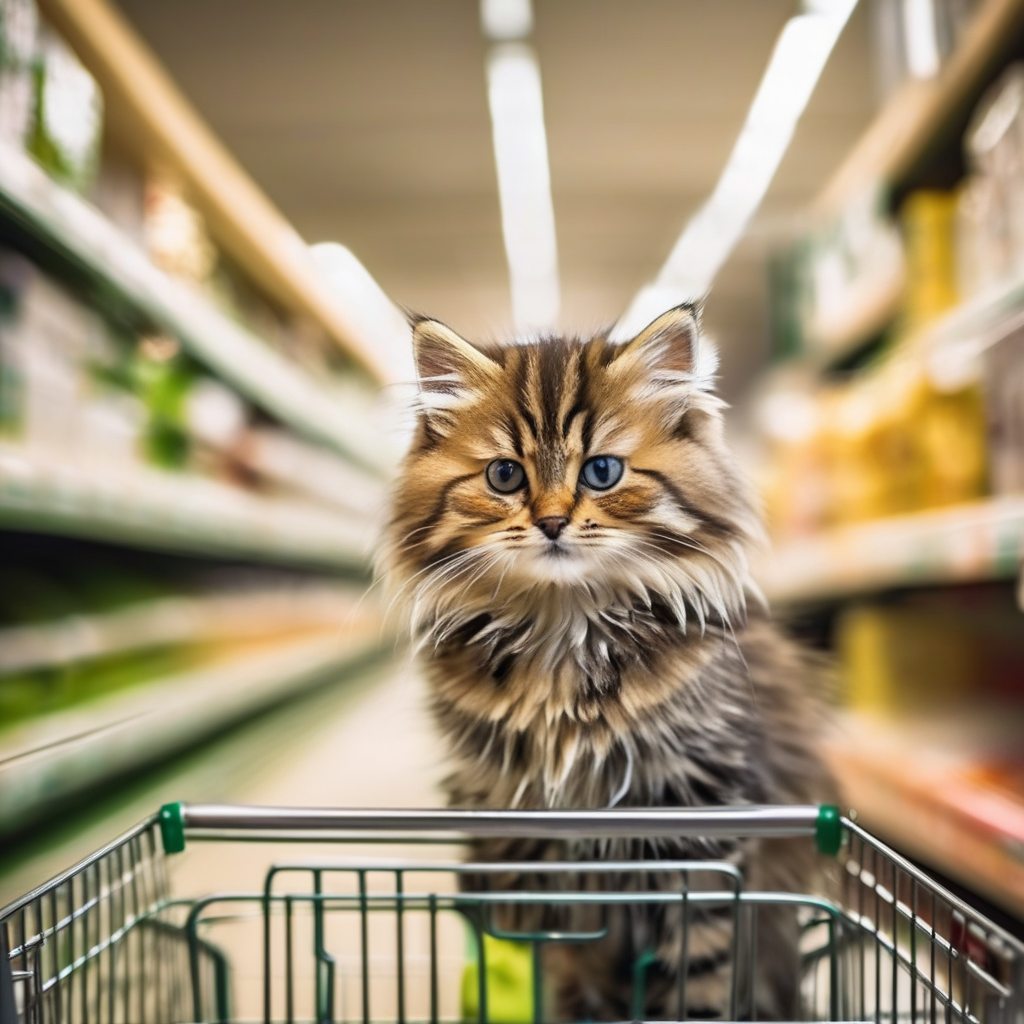} &
            \includegraphics[width=0.23\linewidth]{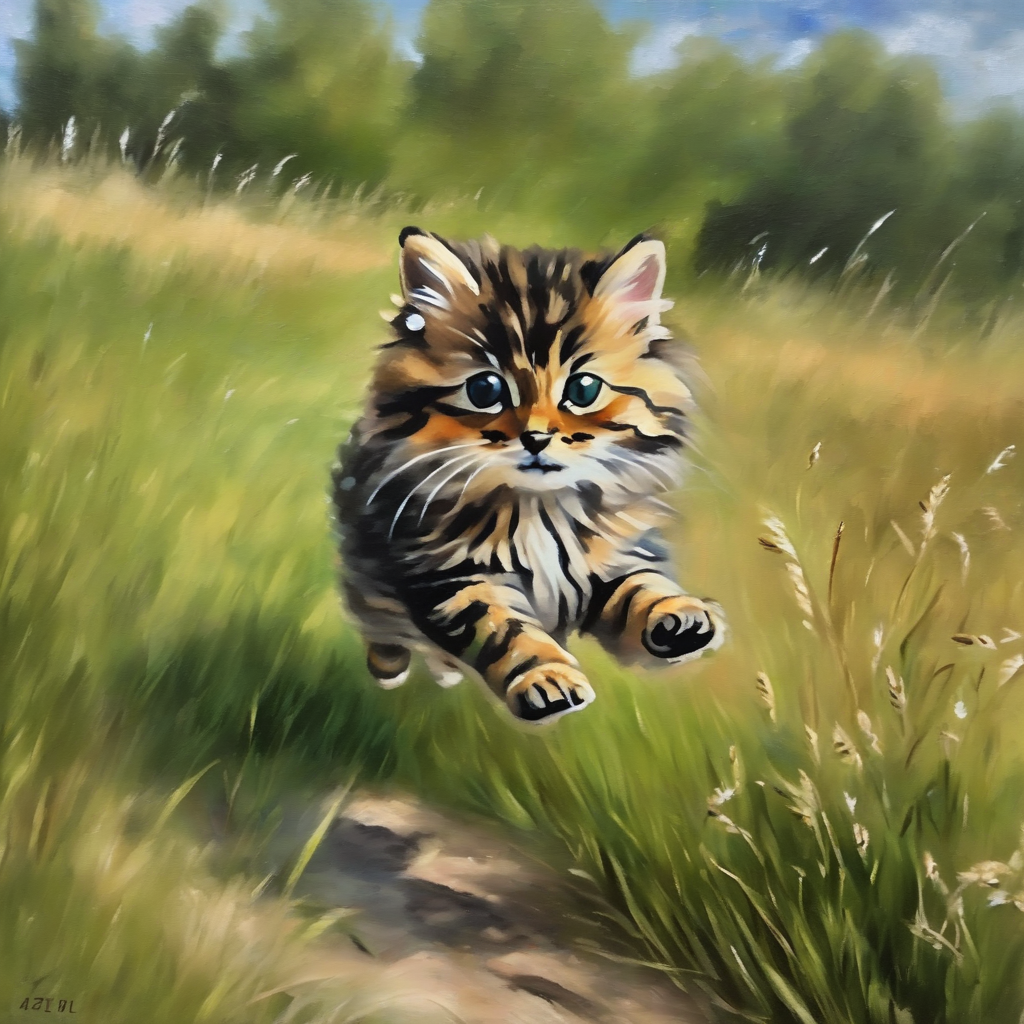} &
            \includegraphics[width=0.23\linewidth]{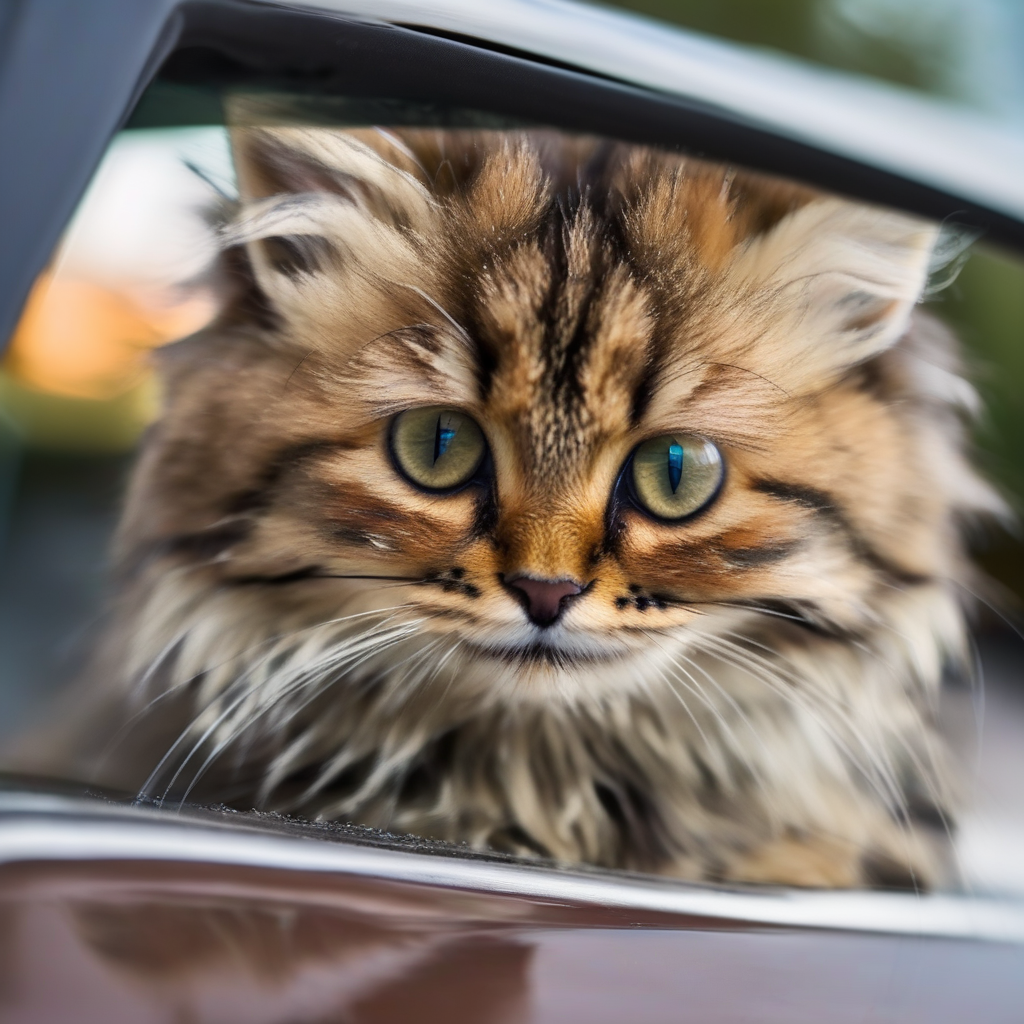} \\
            Input & ``In the super- & ``Oil Painting, & ``Looking outside \\
            & market with a cart'' & running, meadow'' &  a car window''
        \end{tabular}
        }
    \end{minipage}
    \vspace{-6pt}
    \caption{Qualitative results of our method trained on human faces (left) and pets (right). The sampled noise is fixed across each column.}
    \label{fig:qualitative-results}
    \vspace{-16pt}
\end{figure*}

We first show qualitative results generated by our model for both the humans and pet domains (\cref{fig:qualitative-results}). Our method accurately preserves the identity of the subject while adhering to prompts ranging from clothing and expression changes, to scenery modifications and new styles. The initial sampled noise is fixed across each column, leading to consistent composition, colors, and background that demonstrate our method's ability to preserve the model's prior.

\begin{figure}
    \centering
    \setlength{\tabcolsep}{1pt}
    \scriptsize{
    \begin{tabular}{ccccc}
    Input & $\lambda=1$ & $\lambda=2$ & $\lambda=3$ & $\lambda=4$ \\
        \includegraphics[width=0.19\linewidth]{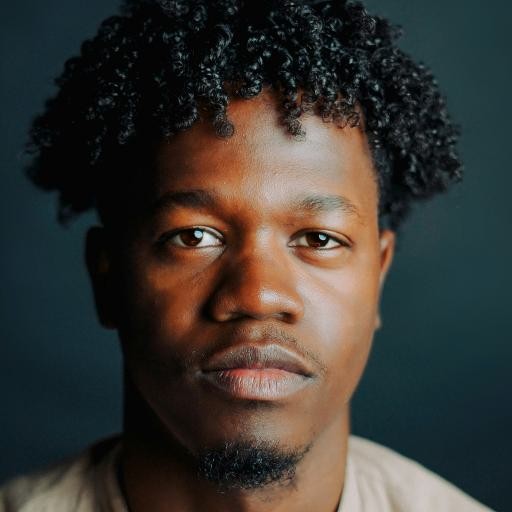} &
        \includegraphics[width=0.19\linewidth]{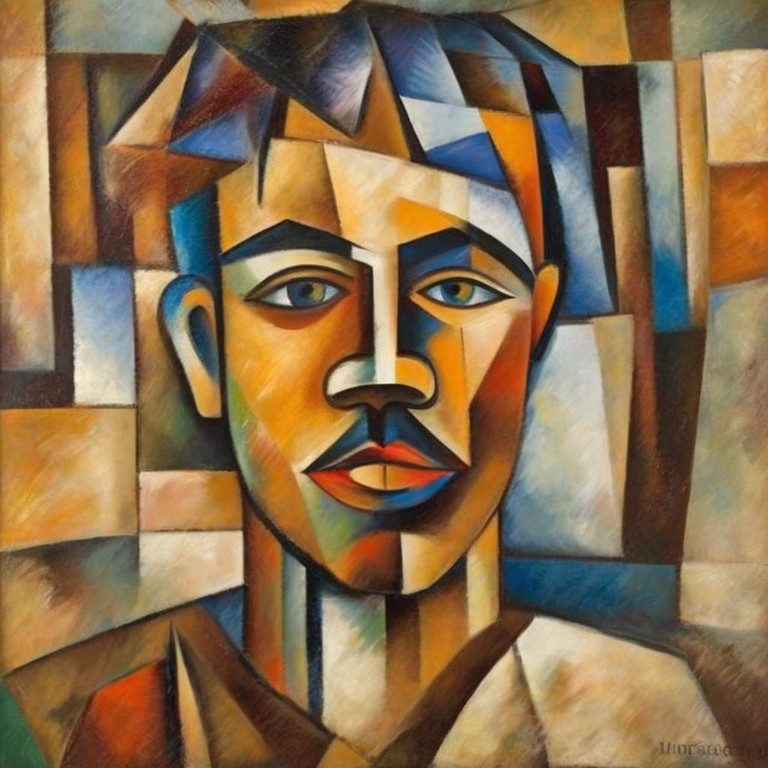} &
        \includegraphics[width=0.19\linewidth]{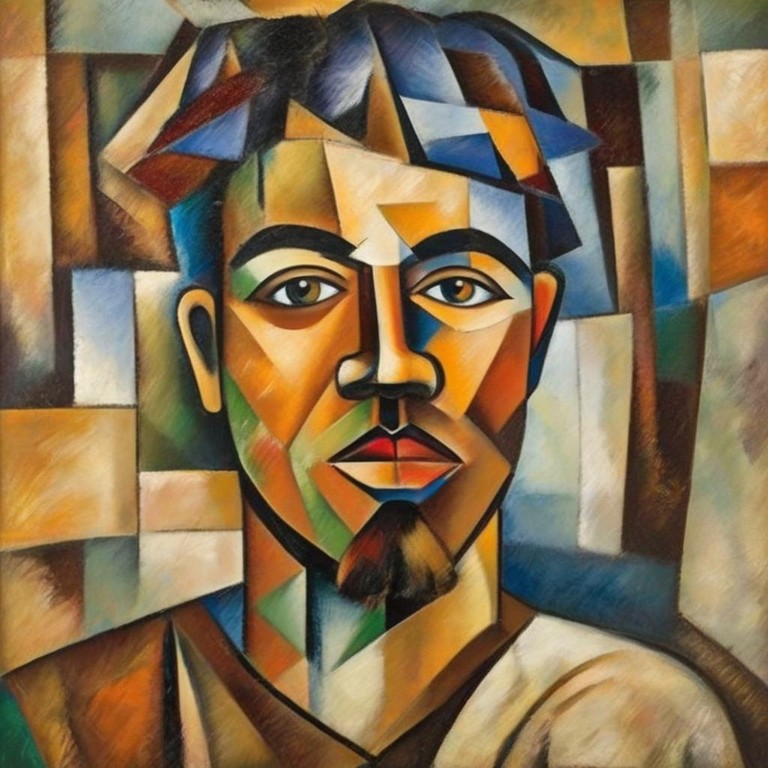} &
        \includegraphics[width=0.19\linewidth]{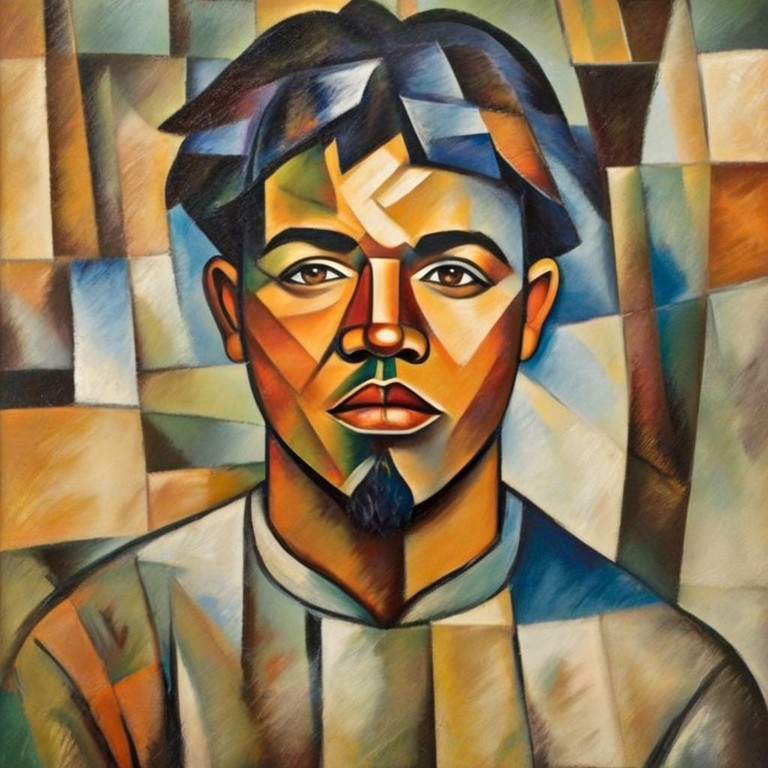} &
        \includegraphics[width=0.19\linewidth]{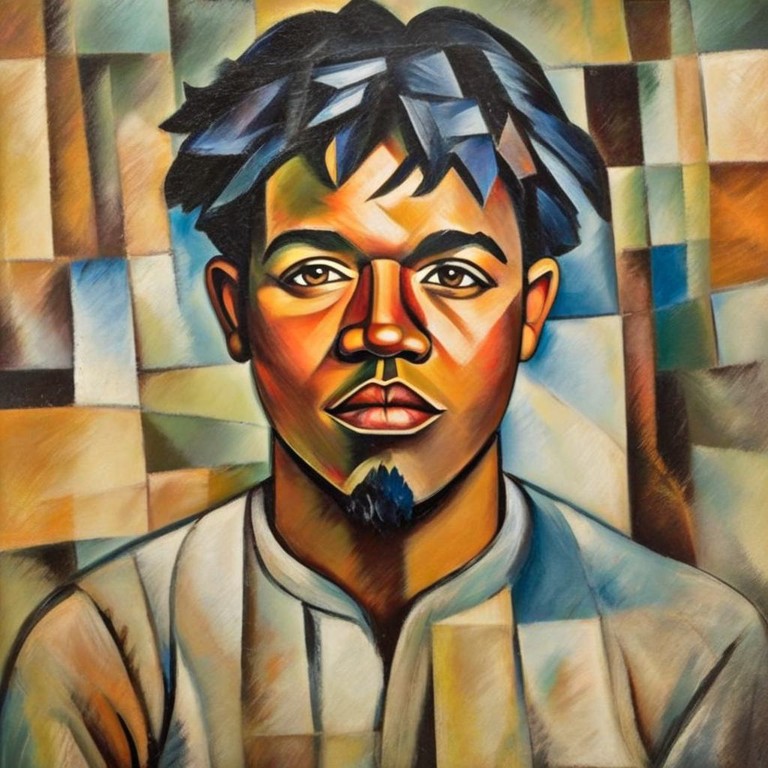} \\
        \multicolumn{5}{c}{``A cubism painting of a \textit{person}''} \\
        \includegraphics[width=0.19\linewidth]{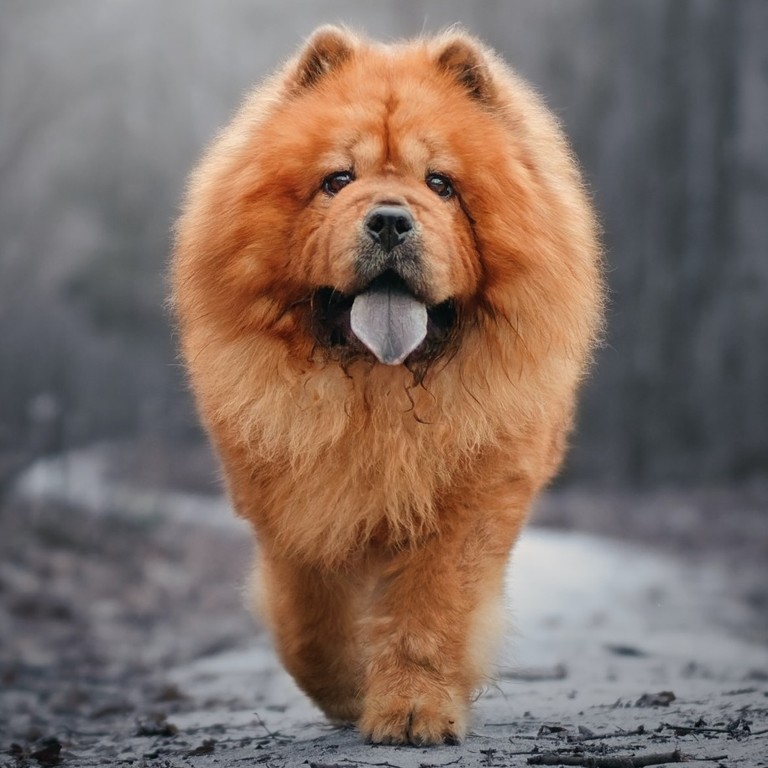} &
        \includegraphics[width=0.19\linewidth]{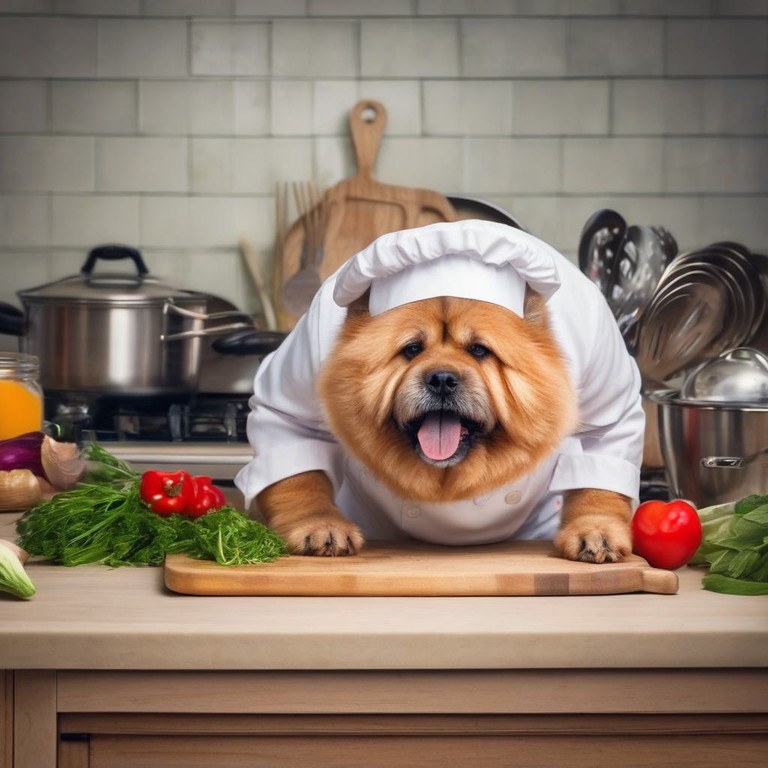} &
        \includegraphics[width=0.19\linewidth]{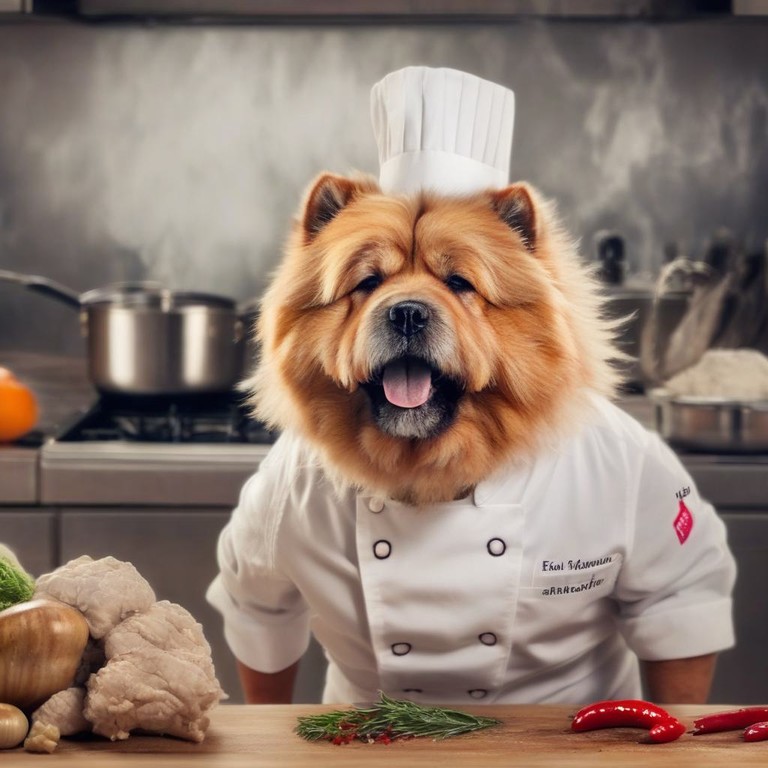} &
        \includegraphics[width=0.19\linewidth]{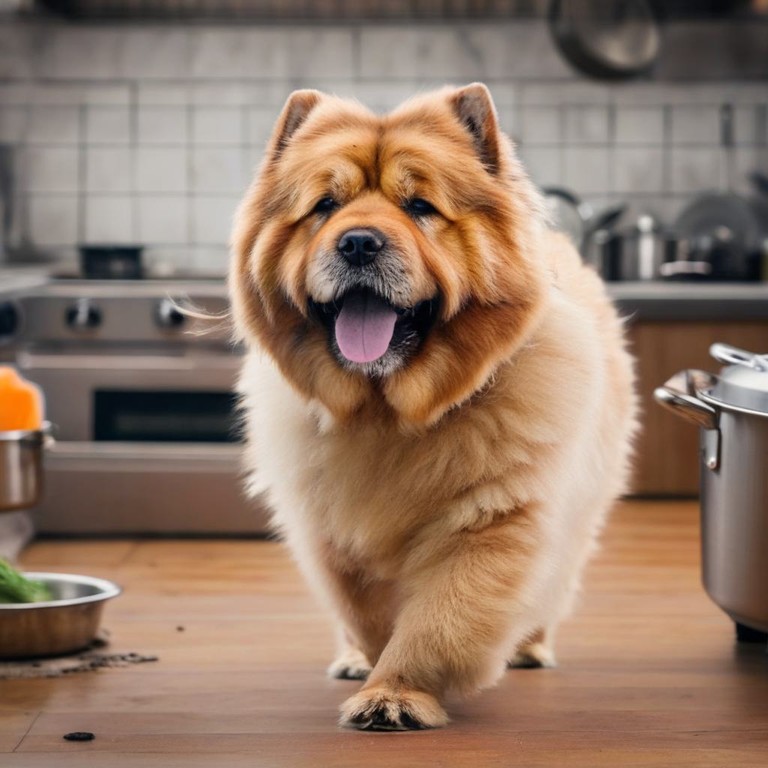} &
        \includegraphics[width=0.19\linewidth]{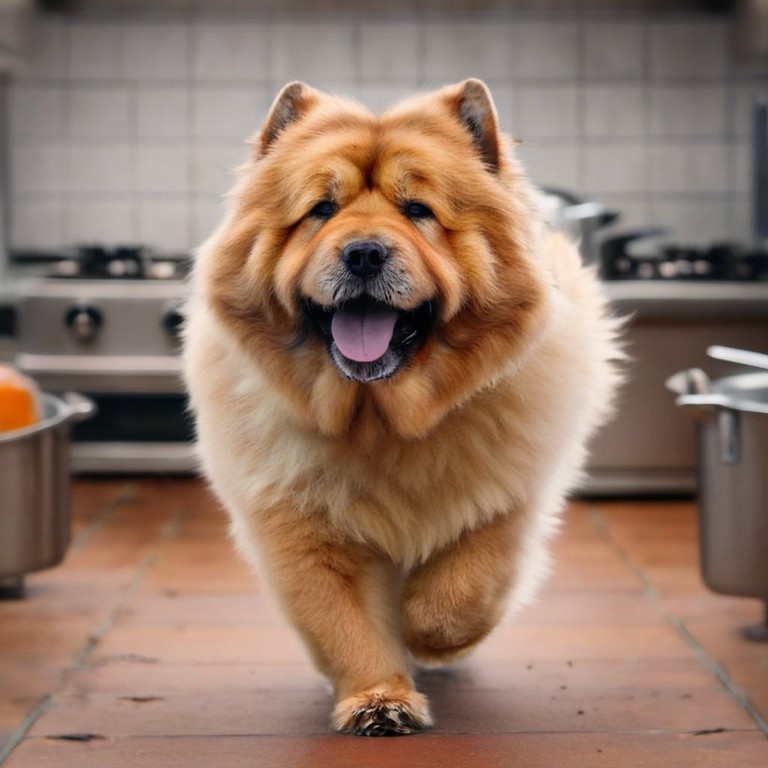} \\
        \multicolumn{5}{c}{``A high quality portrait photo of a \textit{pet} as a chef in the kitchen''} \\

    \end{tabular}
    \vspace{-6pt}
    }
    \caption{
        By manipulating the attention given to the personalized token, we control the identity-editability tradeoff. $\lambda$ denotes the factor in which we increase the attention to the special token.
    }
    \vspace{-16pt}
    \label{fig:adjust-attn}
\end{figure}

\vspace{-12pt}
\paragraph{Controlling identity-editability tradeoff} \label{sec:attn-factor}
Since our approach attaches the personalized concept to a single textual token, we can easily adjust the attention it gets during inference time. This can be used to control the tradeoff between identity preservation and prompt alignment, in a similar manner to the adapter-scale commonly used in decoupled-attention methods~\cite{ye2023ipadapter}. Specifically, we adjust the subject's attention as follows:
\vspace{-6pt}
\begin{equation*}
\vspace{-6pt}
    QK^T[s^*] = \max (QK^T[s^*], \lambda QK^T[s^*]),
\end{equation*}
where $K^T[s^*]$ is the special token's key, and $\lambda$ is the hyperparameter that controls the tradeoff. 
We use max operation because attention value before applying $\softmax$ can be negative, and we do not want to further reduce the subject's attention. \Cref{fig:adjust-attn} shows the effect of varying $\lambda$.

\subsection{Comparisons}
\vspace{-3pt}

\begin{figure*}
    \centering
    \setlength{\tabcolsep}{1pt}
    \scriptsize{
    \begin{tabular}{c c ccc c ccc}
        & Input & \multicolumn{3}{c}{$\longleftarrow$ Varying $\lambda$ $\longrightarrow$} & Input & \multicolumn{3}{c}{$\longleftarrow$ Varying $\lambda$ $\longrightarrow$} \\
        \raisebox{21pt}{\rotatebox[origin=t]{90}{Decoupled CA}} &
        \includegraphics[width=0.10\linewidth]{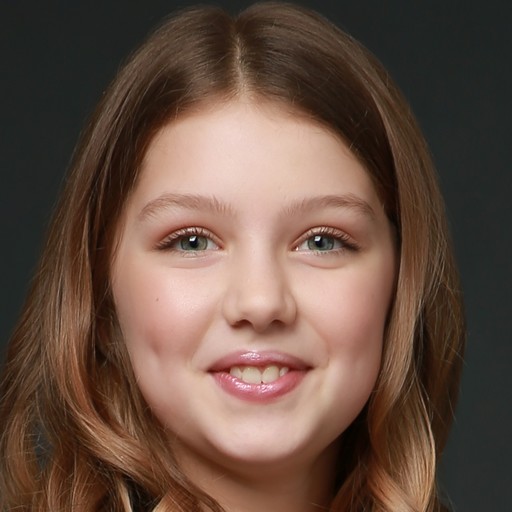} &
        \includegraphics[width=0.10\linewidth]{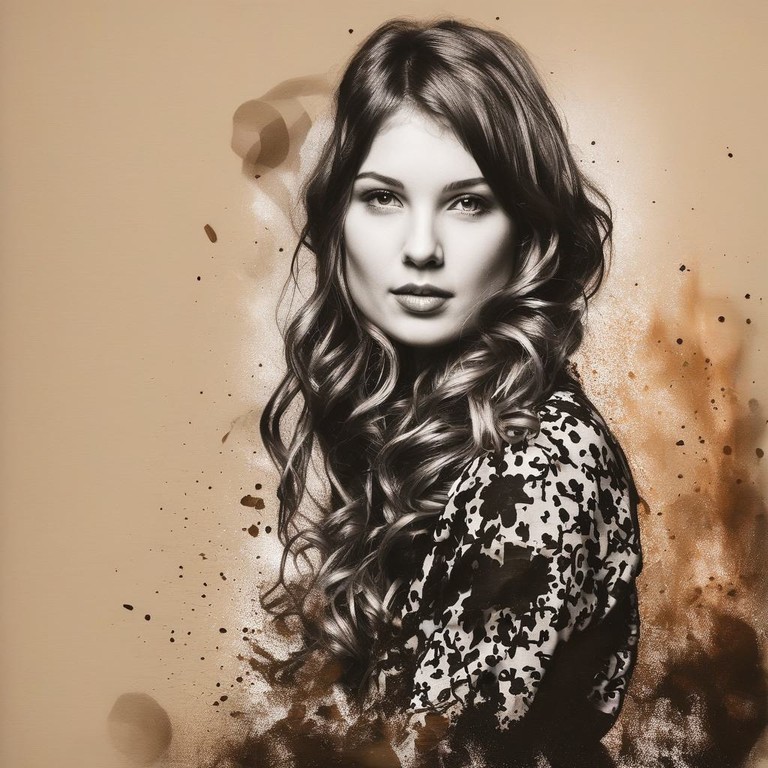} &
        \includegraphics[width=0.10\linewidth]{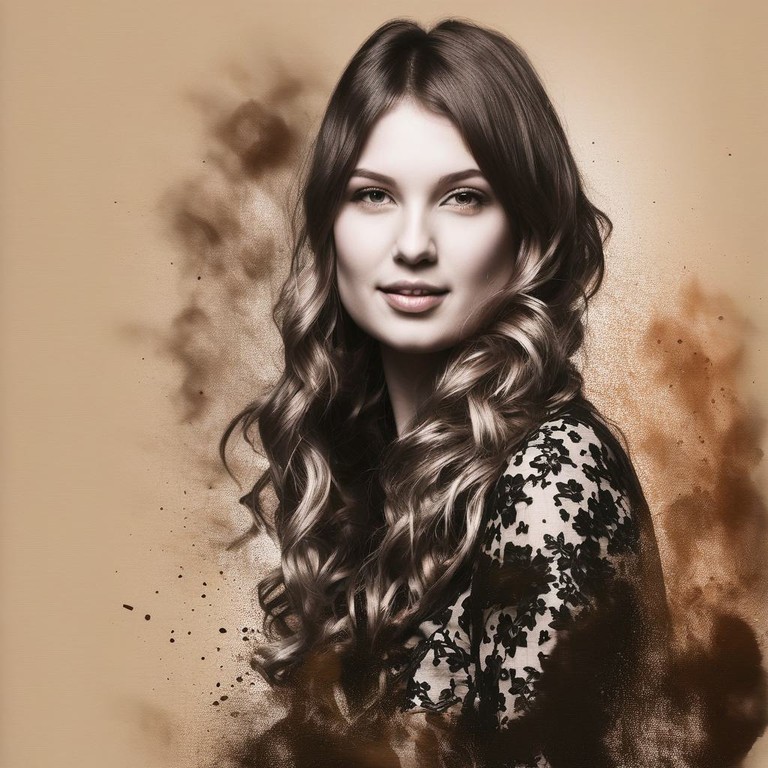} &
        \includegraphics[width=0.10\linewidth]{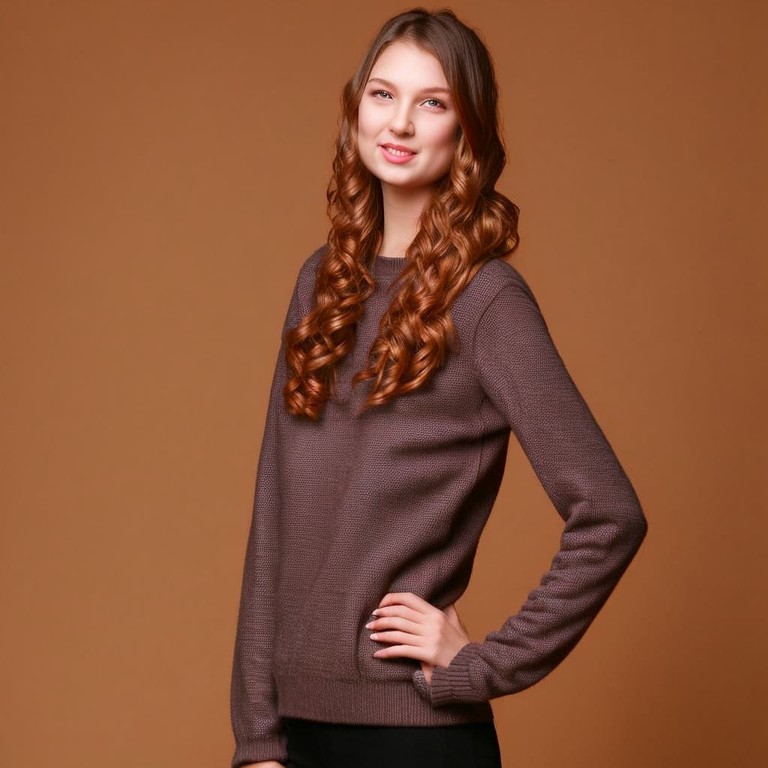} &
        \includegraphics[width=0.10\linewidth]{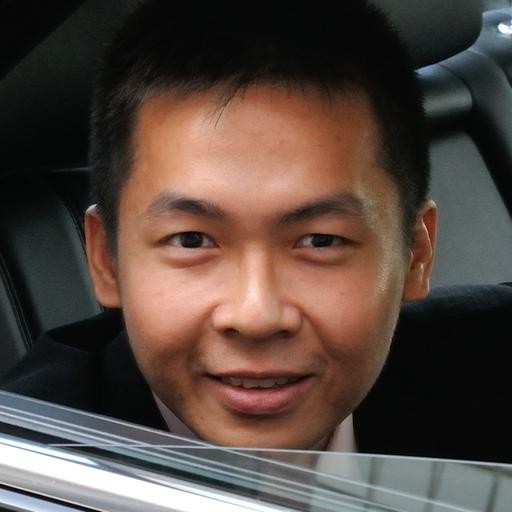} &
        \includegraphics[width=0.10\linewidth]{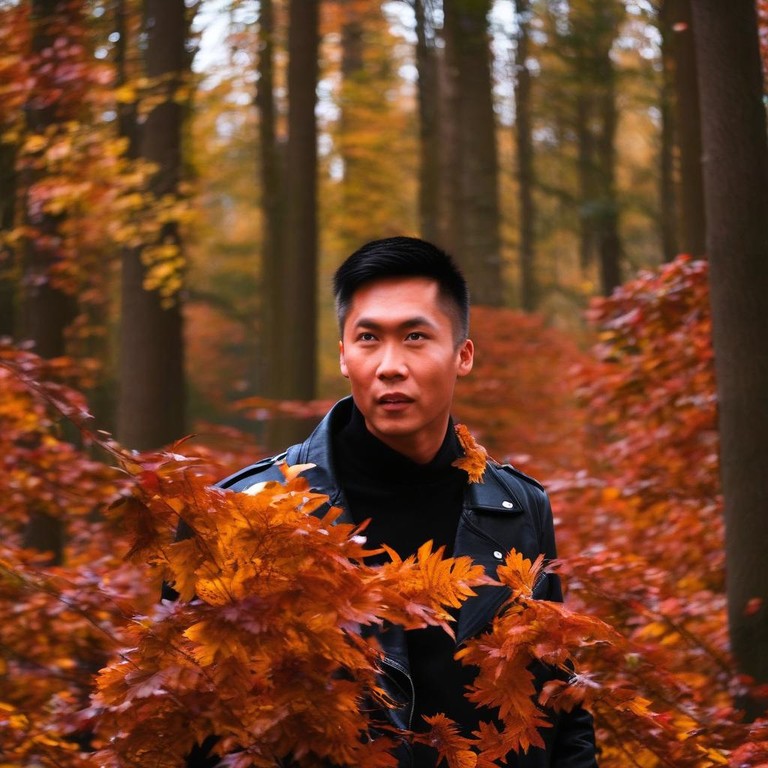} &
        \includegraphics[width=0.10\linewidth]{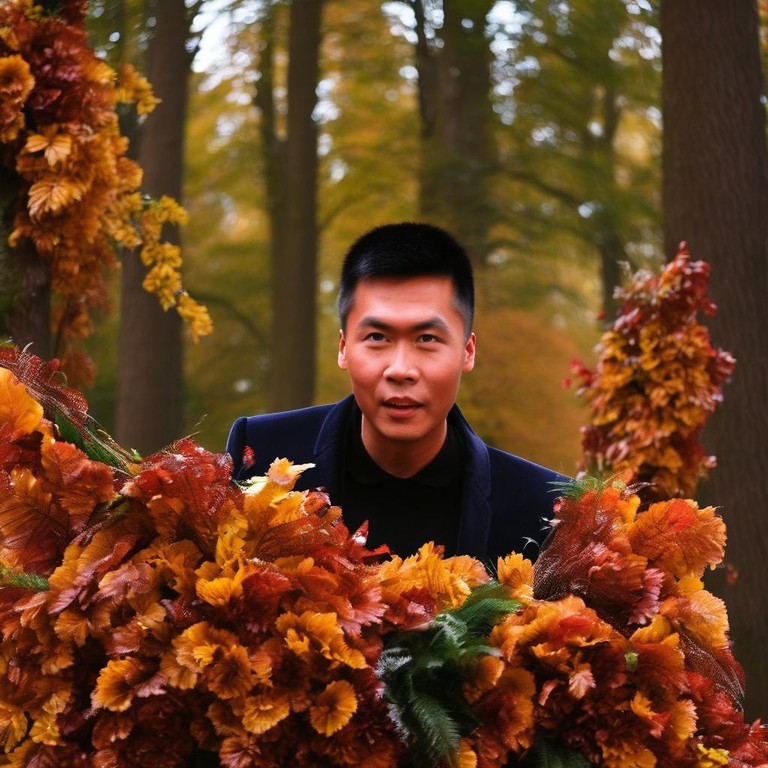} &
        \includegraphics[width=0.10\linewidth]{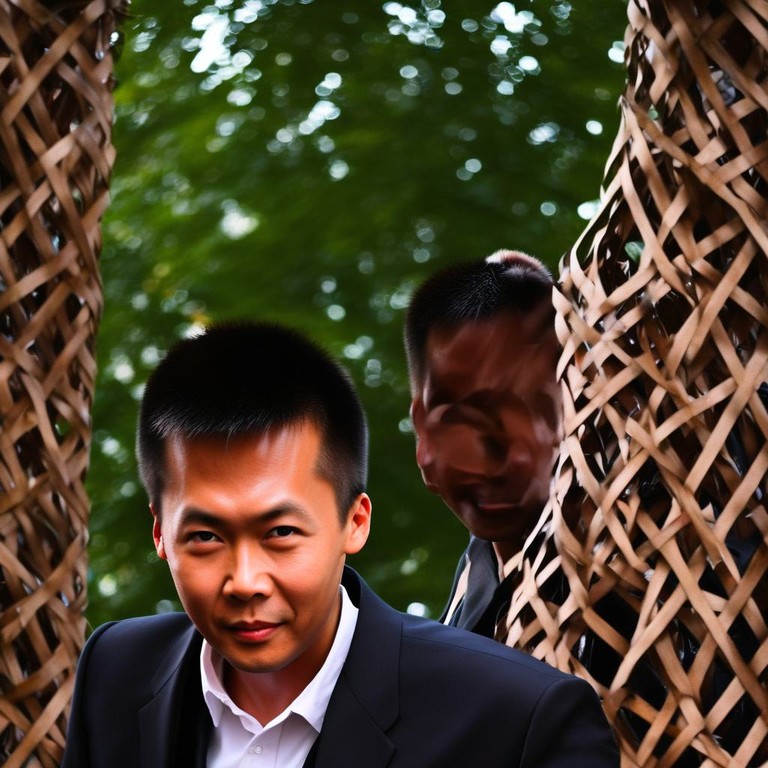} \\
        \raisebox{21pt}{\rotatebox[origin=t]{90}{Nested Attention}} &
        \includegraphics[width=0.10\linewidth]{images/mechanism_comparison/32007.jpg} &
        \includegraphics[width=0.10\linewidth]{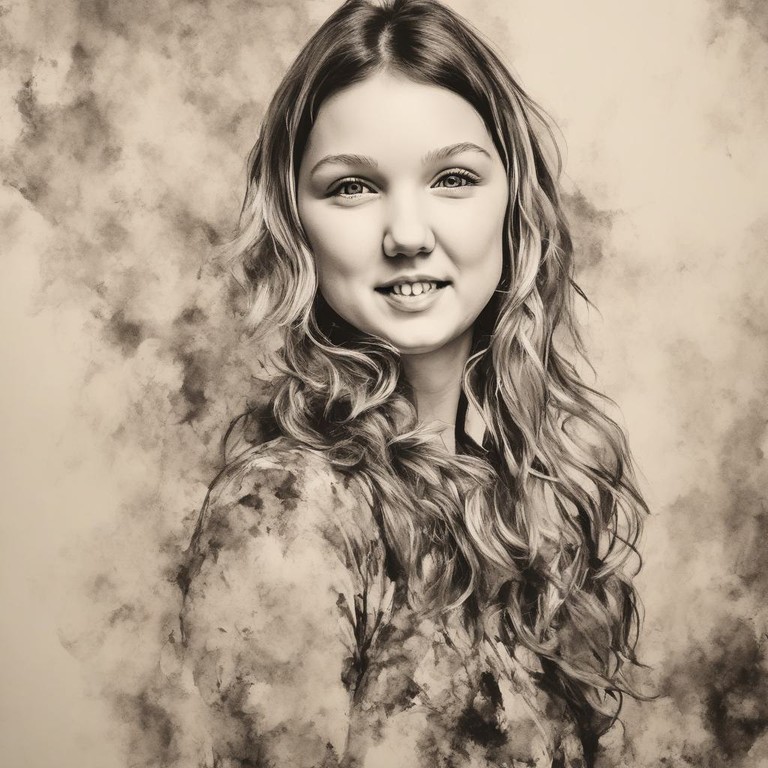} &
        \includegraphics[width=0.10\linewidth]{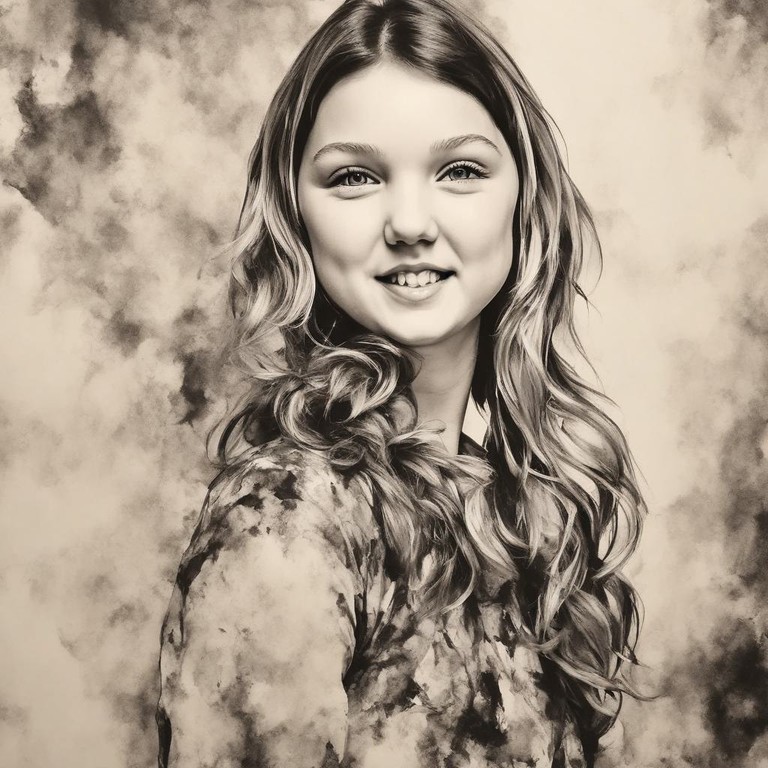} &
        \includegraphics[width=0.10\linewidth]{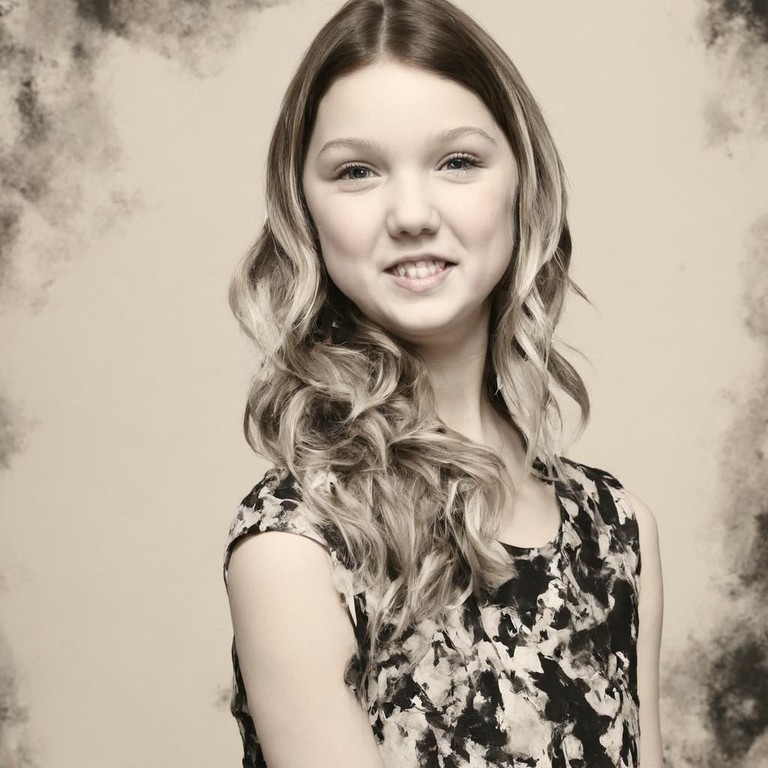} &
        \includegraphics[width=0.10\linewidth]{images/mechanism_comparison/man_input.jpg} &
        \includegraphics[width=0.10\linewidth]{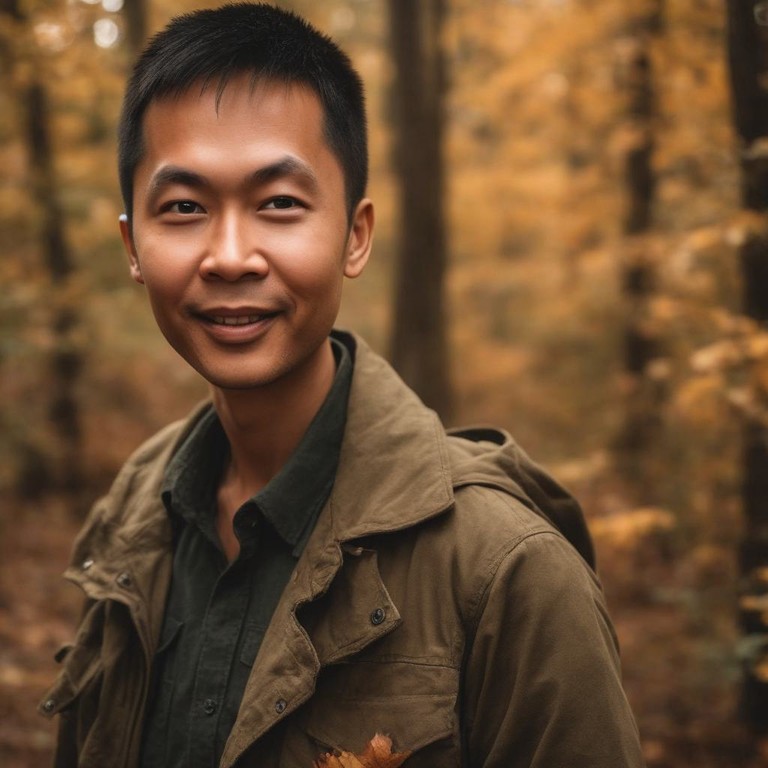} &
        \includegraphics[width=0.10\linewidth]{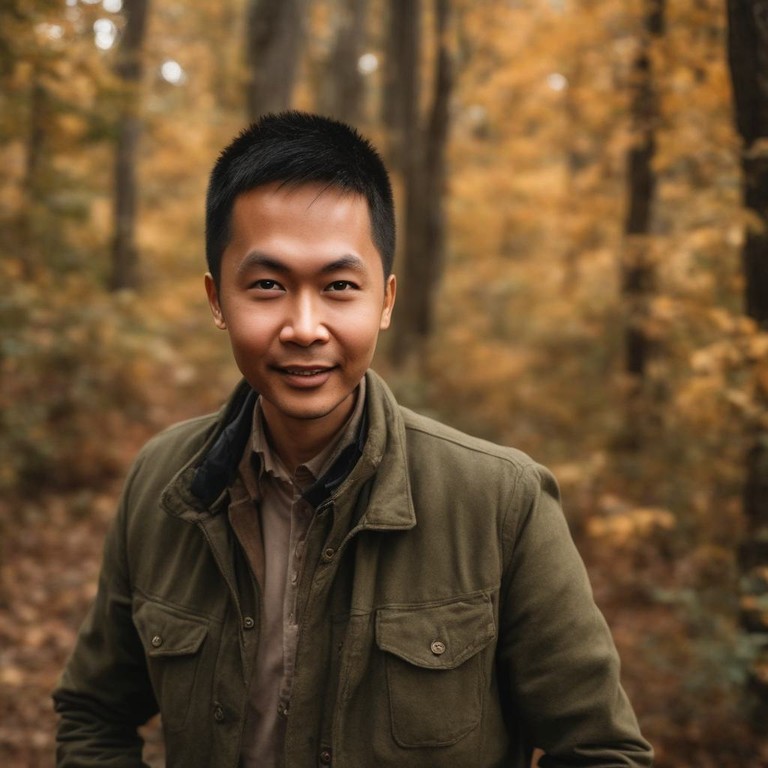} &
        \includegraphics[width=0.10\linewidth]{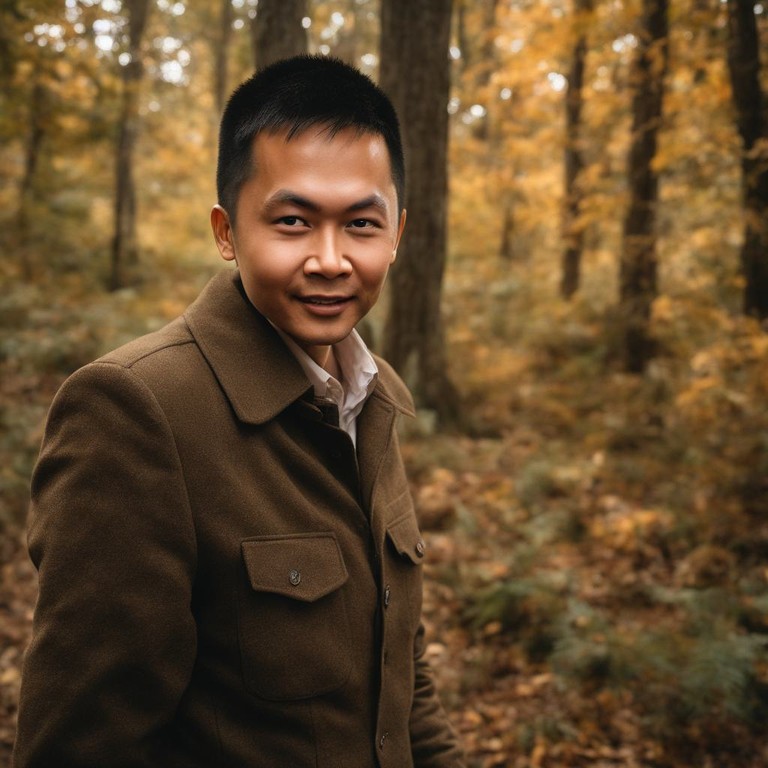} \\
        & \multicolumn{4}{c}{``An abstract ink drawing of a \emph{person}''} &
        \multicolumn{4}{c}{``A high quality portrait photo of a \emph{person} in the forest during fall''} \\
        
        \raisebox{21pt}{\rotatebox[origin=t]{90}{Decoupled CA}} &
        \includegraphics[width=0.10\linewidth]{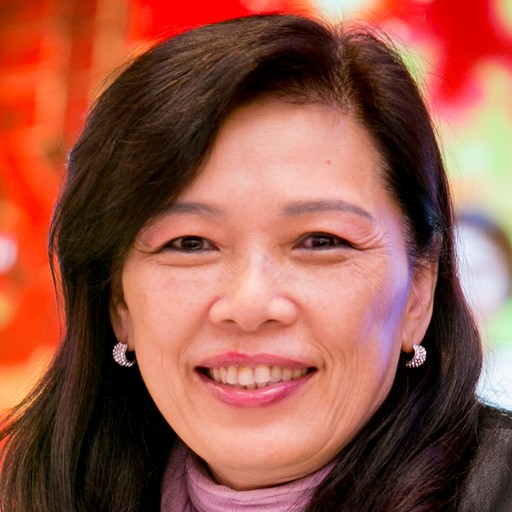} &
        \includegraphics[width=0.10\linewidth]{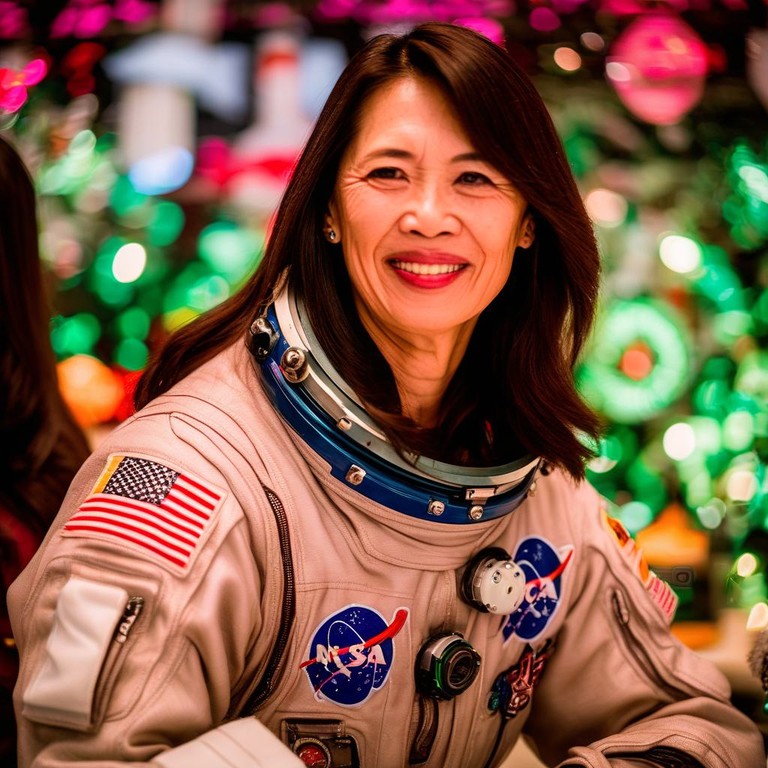} &
        \includegraphics[width=0.10\linewidth]{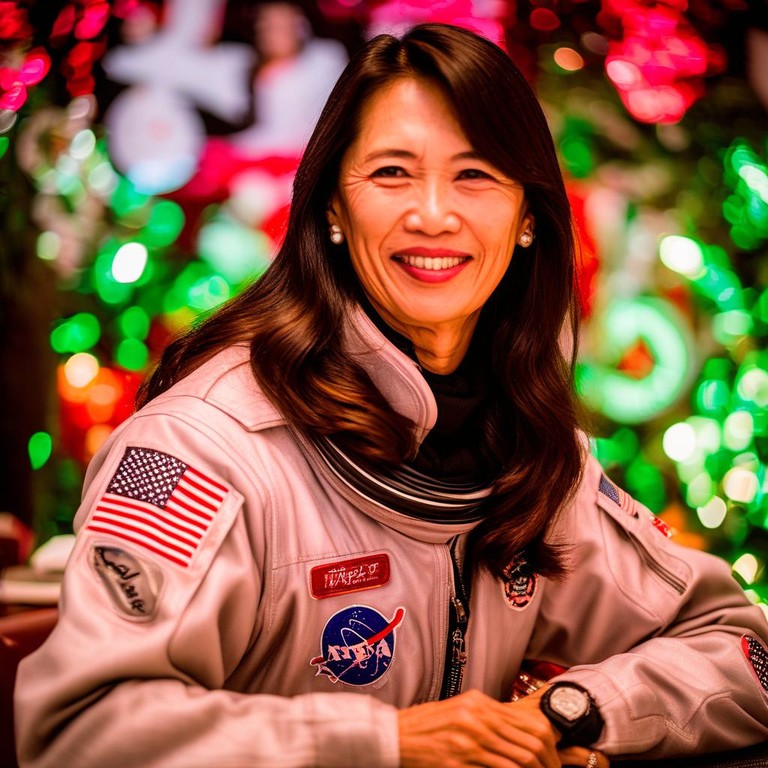} &
        \includegraphics[width=0.10\linewidth]{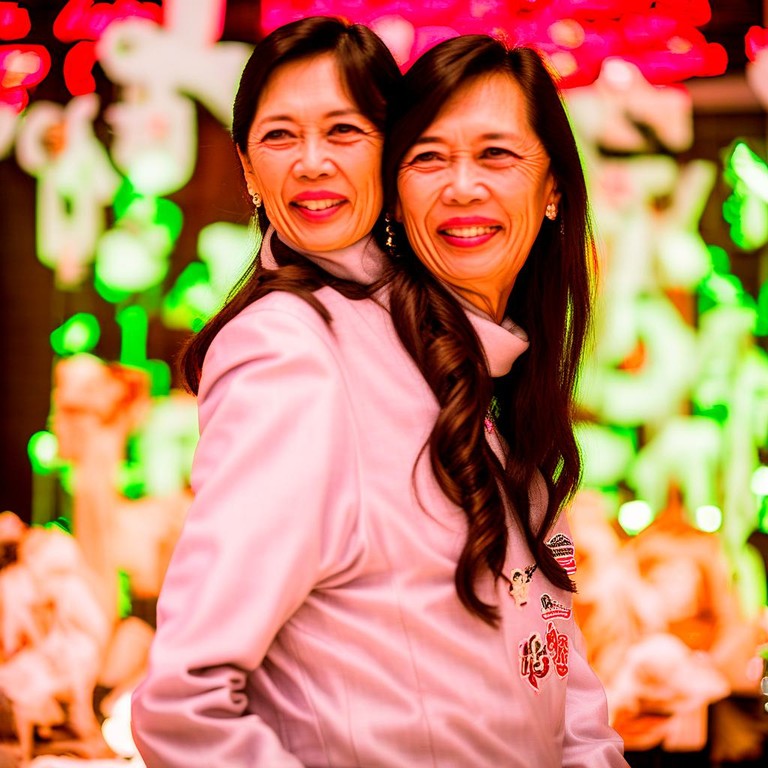} &
        \includegraphics[width=0.10\linewidth]{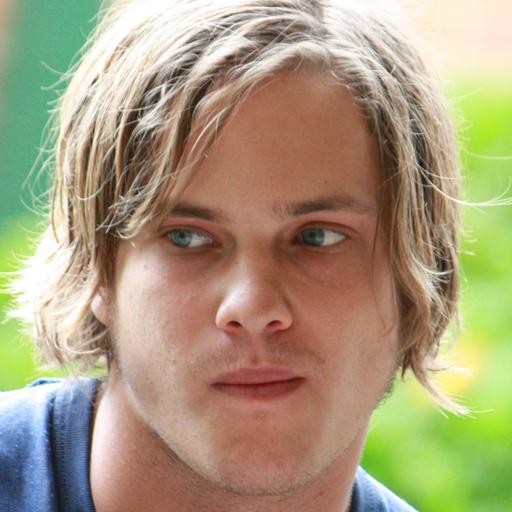} &
        \includegraphics[width=0.10\linewidth]{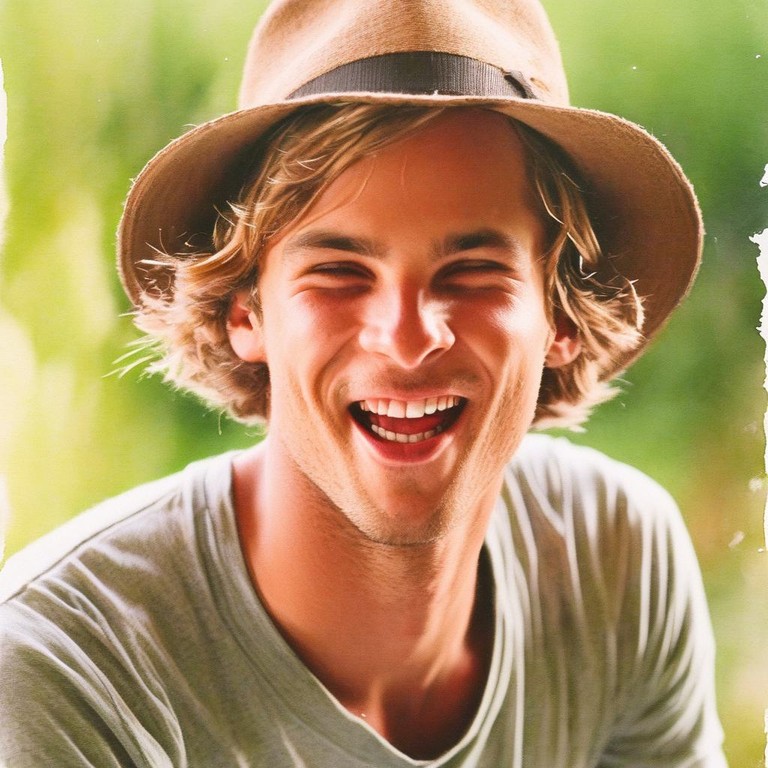} &
        \includegraphics[width=0.10\linewidth]{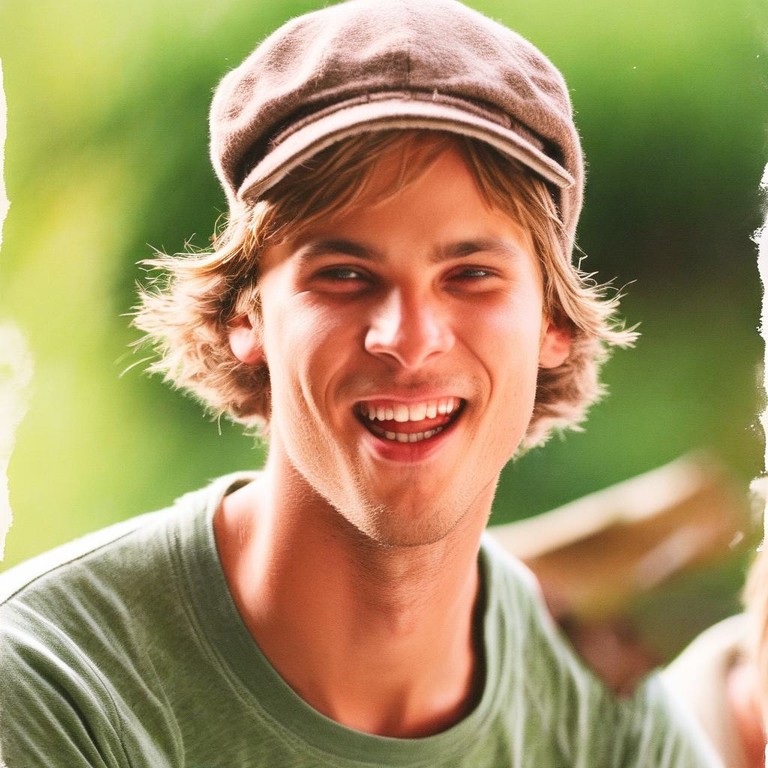} &
        \includegraphics[width=0.10\linewidth]{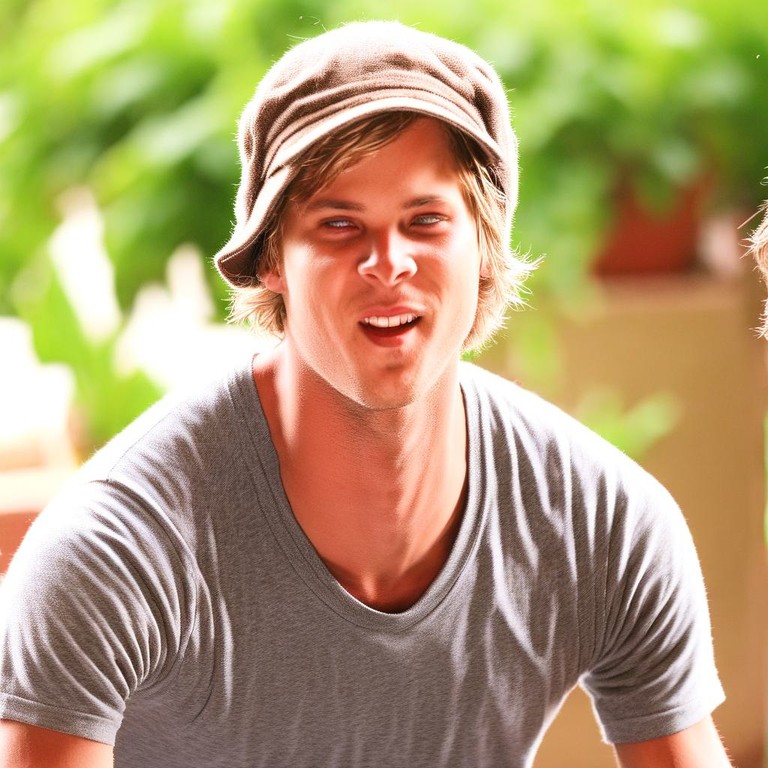} \\
        \raisebox{21pt}{\rotatebox[origin=t]{90}{Nested Attention}} &
        \includegraphics[width=0.10\linewidth]{images/mechanism_comparison/37512.jpg} &
        \includegraphics[width=0.10\linewidth]{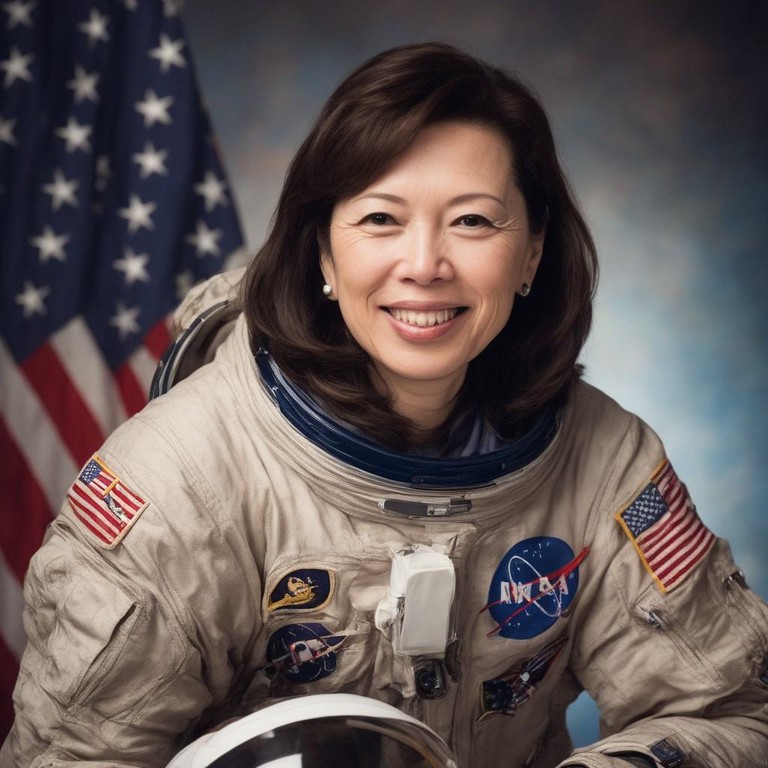} &
        \includegraphics[width=0.10\linewidth]{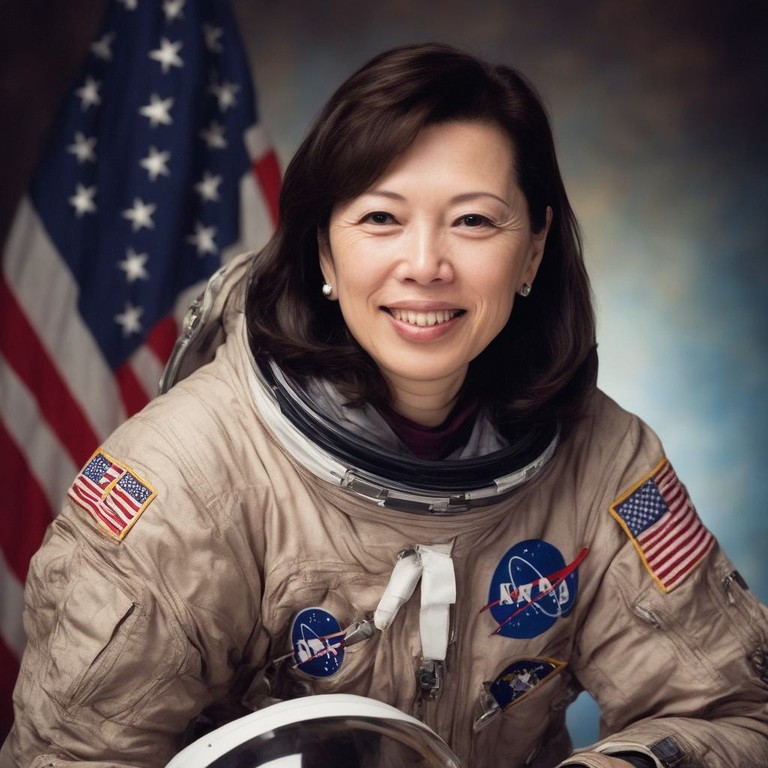} &
        \includegraphics[width=0.10\linewidth]{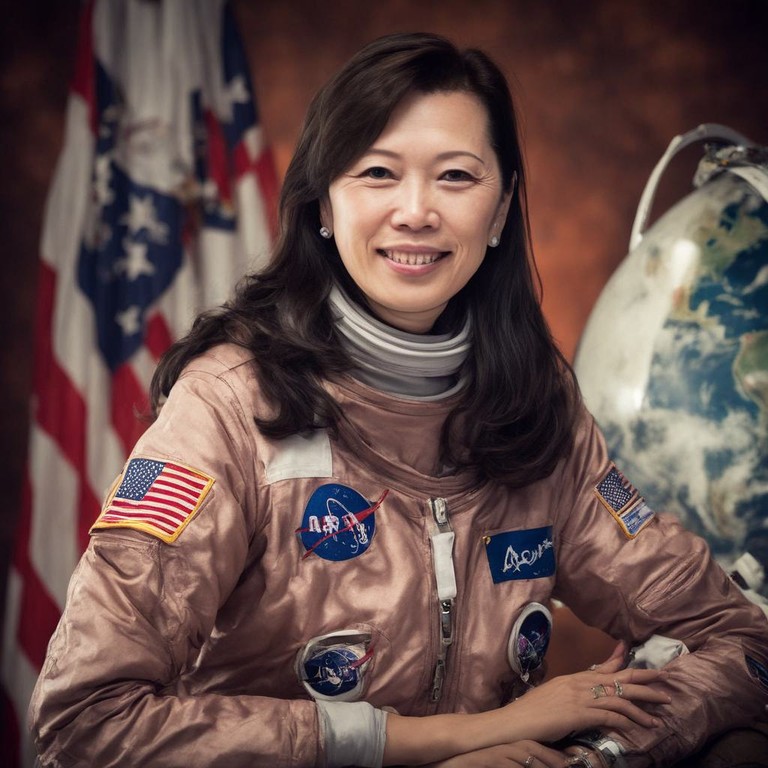} &
        \includegraphics[width=0.10\linewidth]{images/mechanism_comparison/man_watercolor_input.jpg} &
        \includegraphics[width=0.10\linewidth]{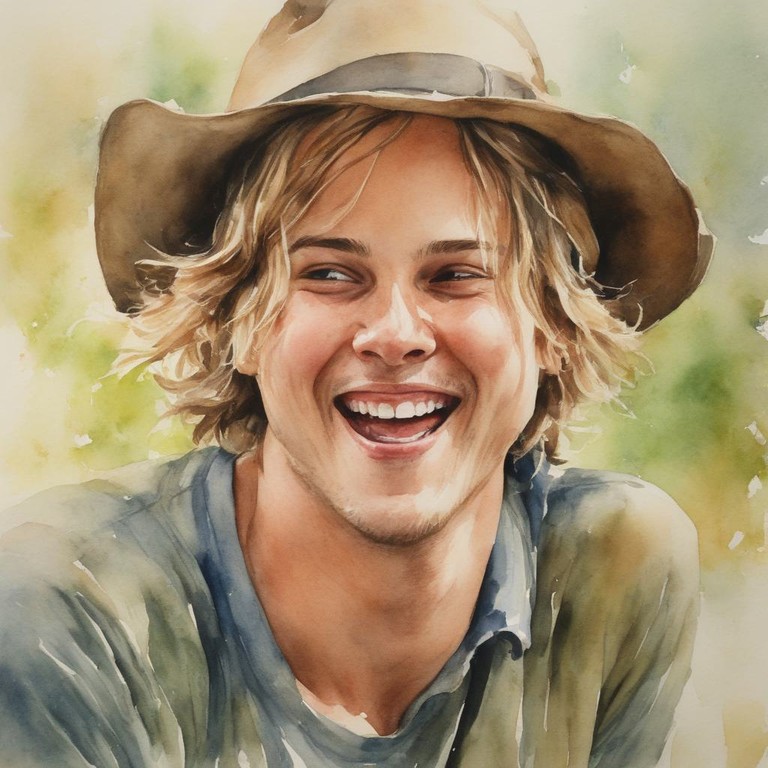} &
        \includegraphics[width=0.10\linewidth]{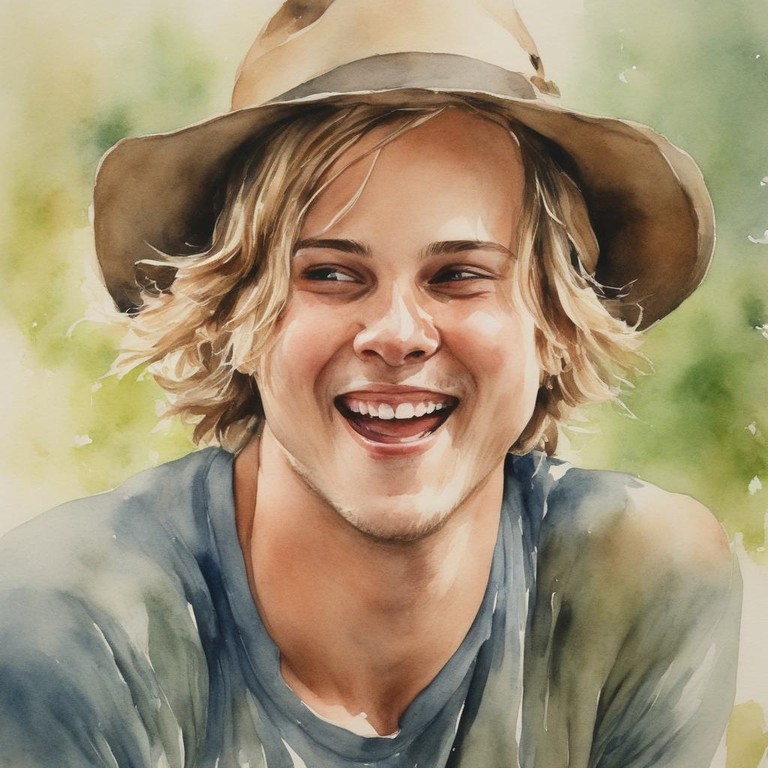} &
        \includegraphics[width=0.10\linewidth]{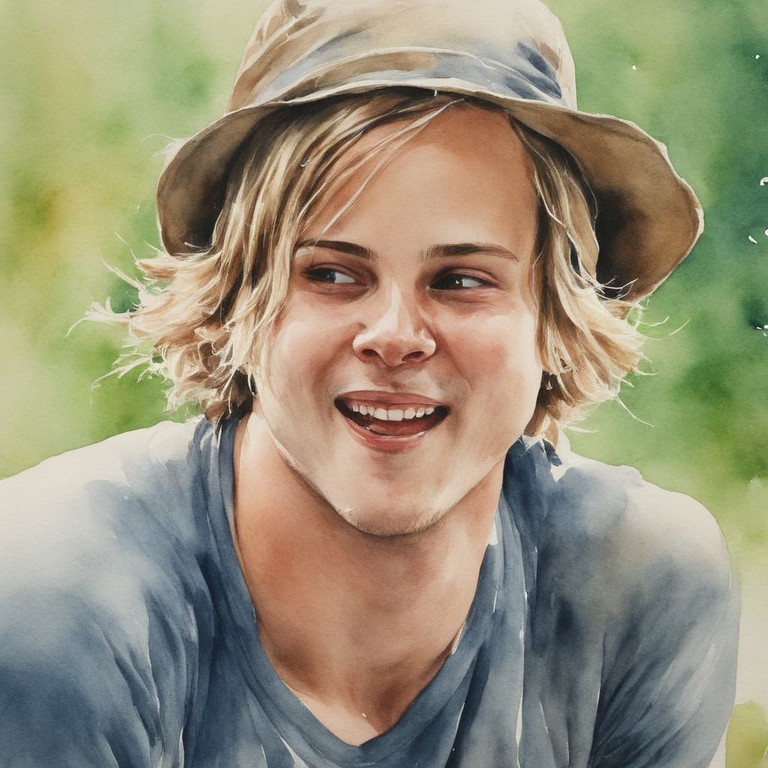} \\
        & \multicolumn{4}{c}{``A high quality photo of a \emph{person} as an astronaut''} &
        \multicolumn{4}{c}{``A watercolor painting of a \emph{person} laughing, he is wearing a hat''} \\

    \end{tabular}
    \vspace{-6pt}
    }
    \caption{
    Comparing nested attention with decoupled cross attention. 
    $\lambda$ balances between identity preservation and prompt alignment. 
    We use the following $\lambda$ values from left to right (top two rows). Decoupled CA: 0.5, 0.6, 1.0, nested attention: 1.0, 2.0, 4.0.
    Our method achieves better identity preservation while being aligned with the text prompt.
    }
    \vspace{-10pt}
    \label{fig:mechanism-comparison}
\end{figure*}

\begin{figure}
    \centering
    \includegraphics[width=0.85\linewidth]{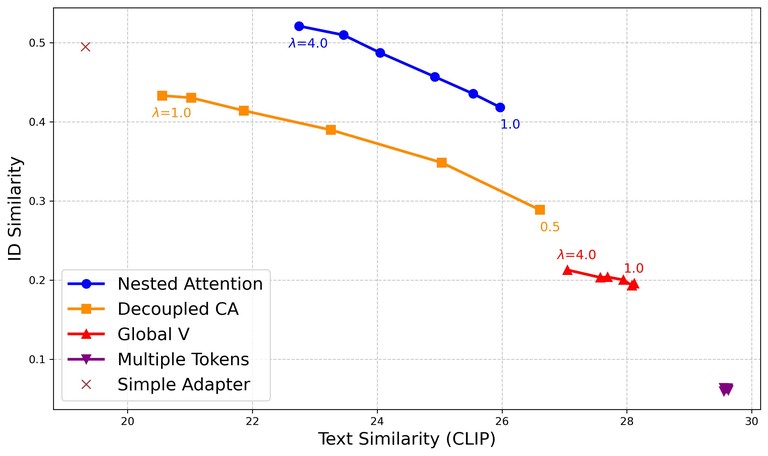}
    \vspace{-8pt}
    \caption{Quantitative comparison of various personalization injection mechanisms. All models were trained from scratch under the same setting, with a resolution of $512\times 512$.}
    \vspace{-16pt}
    \label{fig:comparison-quant-methods}
\end{figure}

\newcolumntype{C}[1]{>{\centering\let\newline\\\arraybackslash\hspace{0pt}}m{#1}}

\begin{figure*}
    \centering
    \setlength{\tabcolsep}{1pt}
    \scriptsize{
    \begin{tabular}{cc cccccccc }
        & & Input & IPA-Face & InstantID & PhotoMaker & Lookahead & PulID & Ours \\

        \raisebox{18pt}{\rotatebox[origin=t]{90}{Smiling at}} &
        \raisebox{18pt}{\rotatebox[origin=t]{90}{her birthday}} &
        \includegraphics[width=0.09\linewidth]{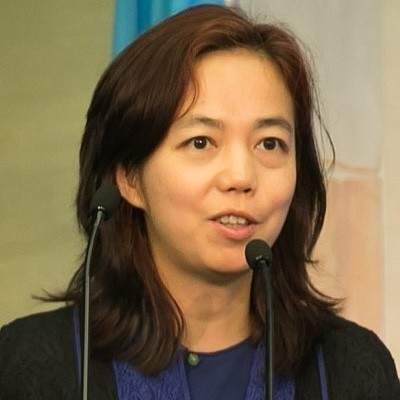} &
        \includegraphics[width=0.09\linewidth]{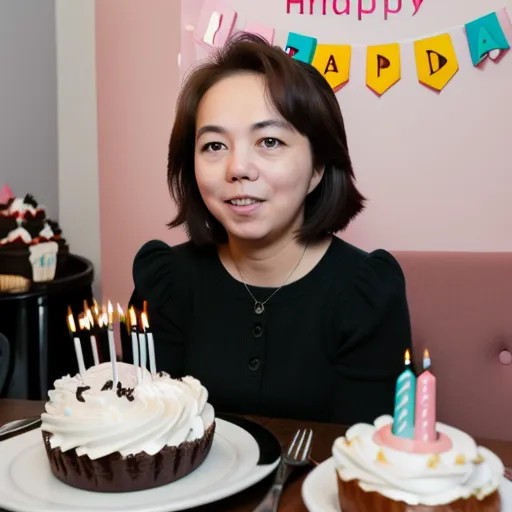} &
        \includegraphics[width=0.09\linewidth]{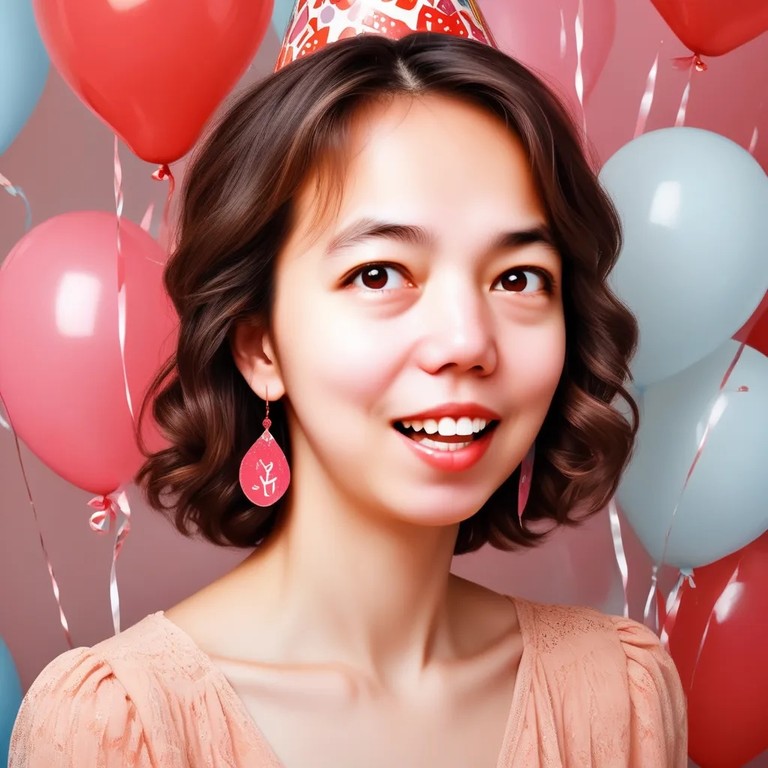} &
        \includegraphics[width=0.09\linewidth]{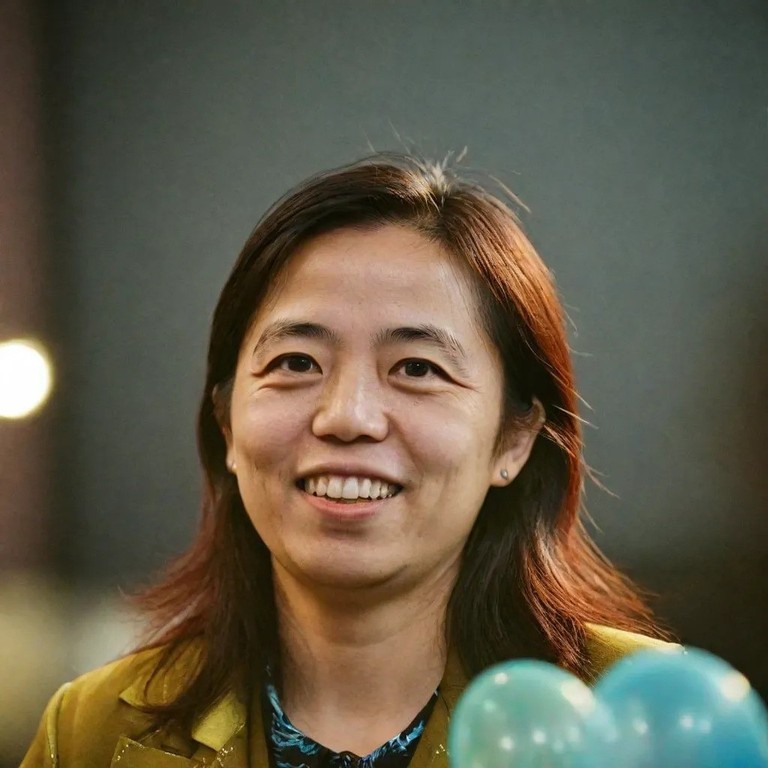} &
        \includegraphics[width=0.09\linewidth]{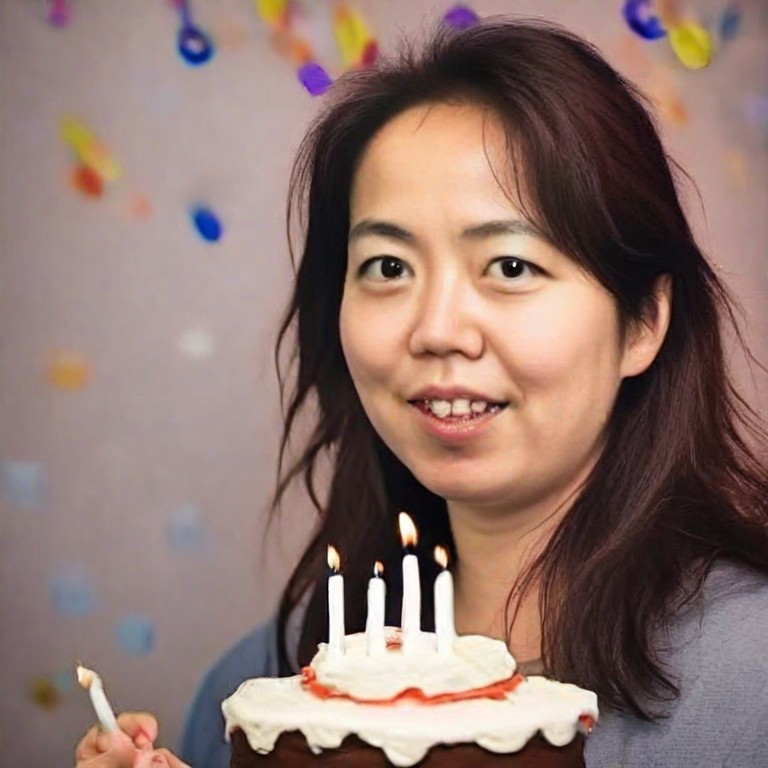} &
        \includegraphics[width=0.09\linewidth]{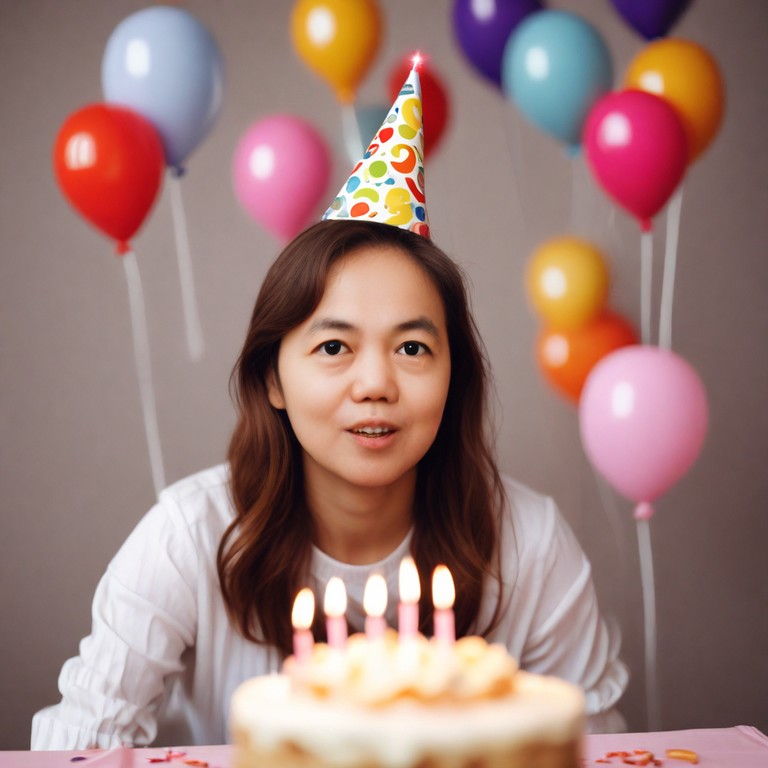} &
        \includegraphics[width=0.09\linewidth]{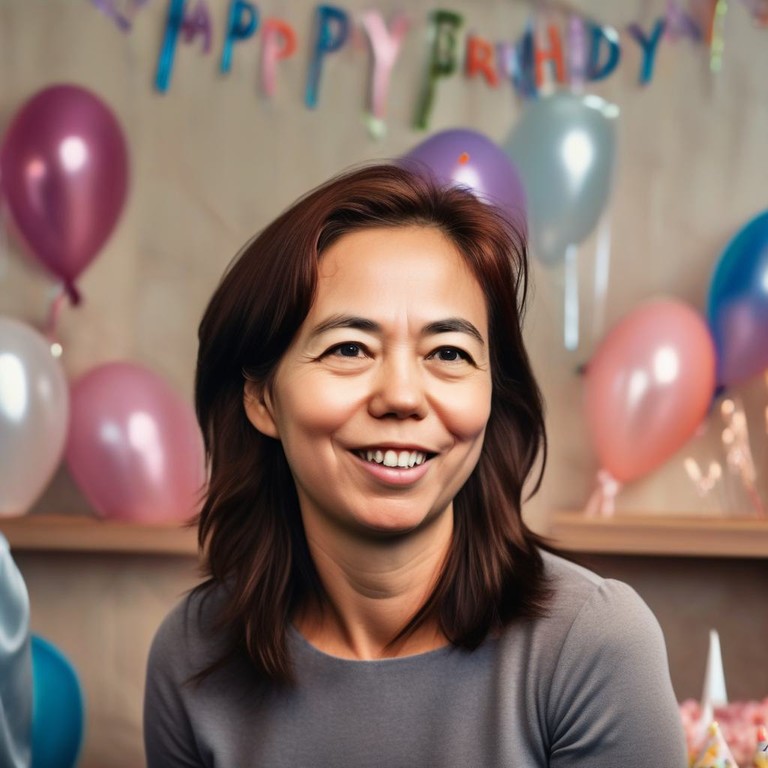} &
        \raisebox{40pt}{\multirow{4}{*}{\includegraphics[height=0.38\linewidth]{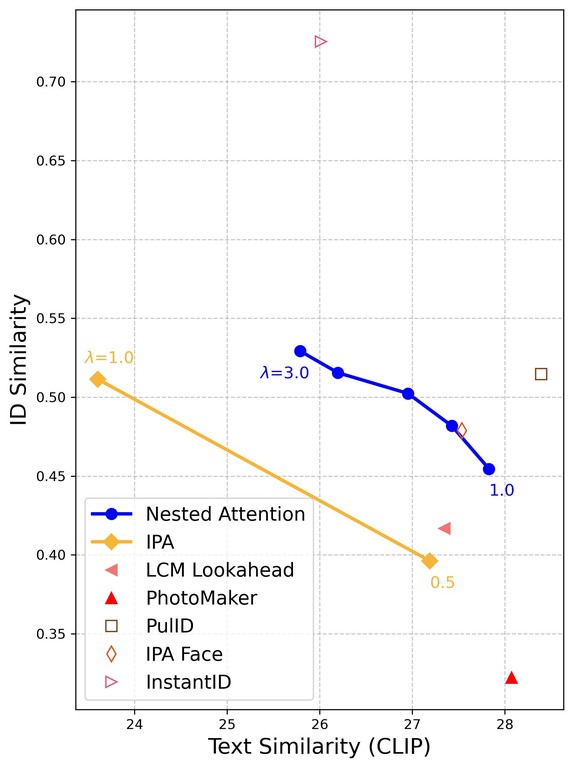}}}
        \\

        \raisebox{18pt}{\rotatebox[origin=t]{90}{Watercolor pai-}} &
        \raisebox{18pt}{\rotatebox[origin=t]{90}{nting, sideview}} &
        \includegraphics[width=0.09\linewidth]{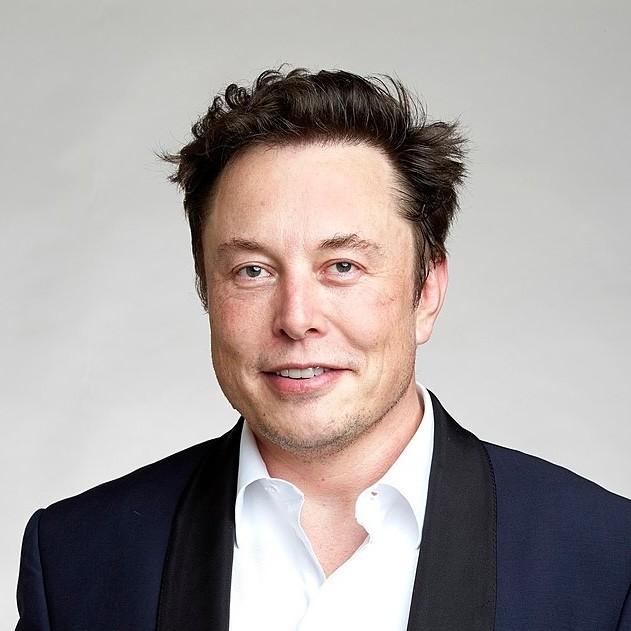} &
        \includegraphics[width=0.09\linewidth]{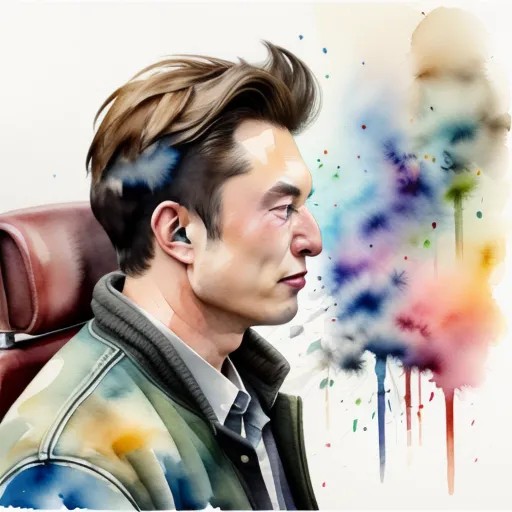} &
        \includegraphics[width=0.09\linewidth]{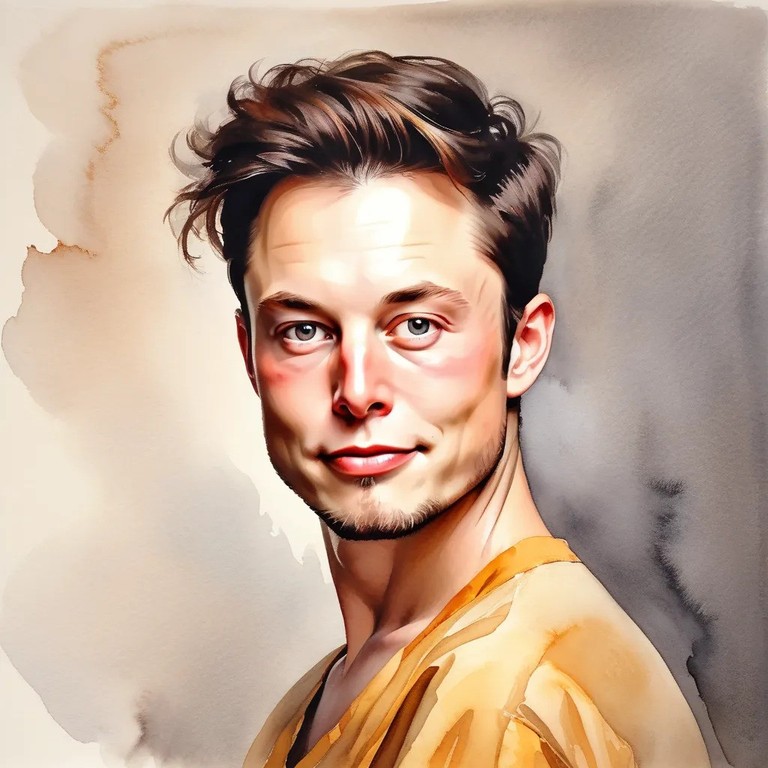} &
        \includegraphics[width=0.09\linewidth]{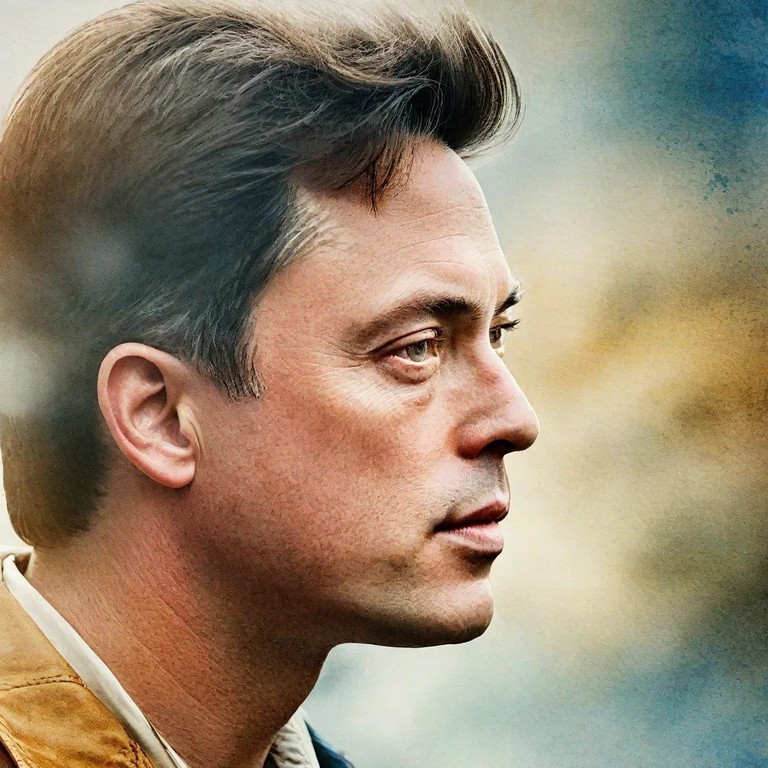} &
        \includegraphics[width=0.09\linewidth]{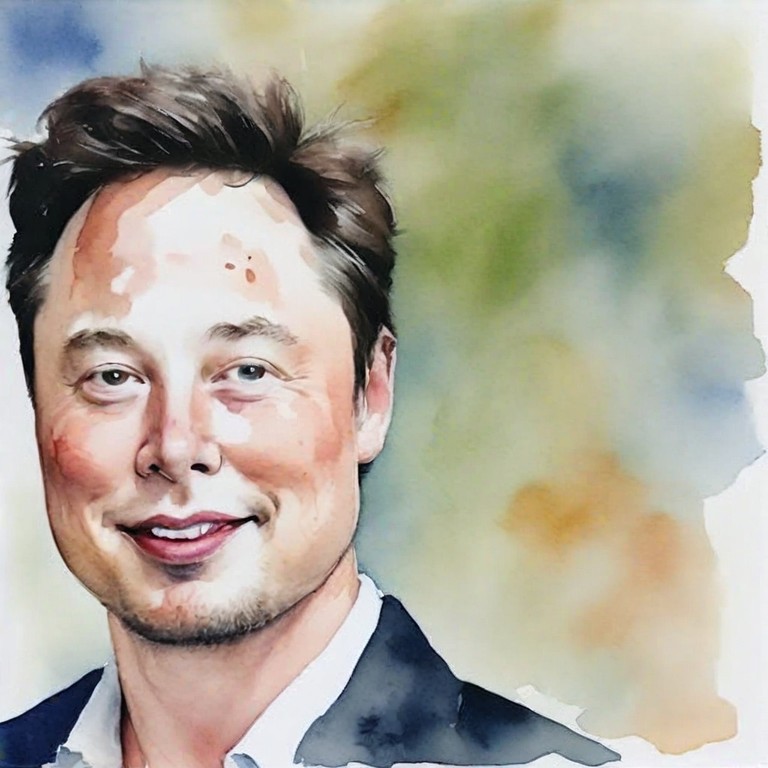} &
        \includegraphics[width=0.09\linewidth]{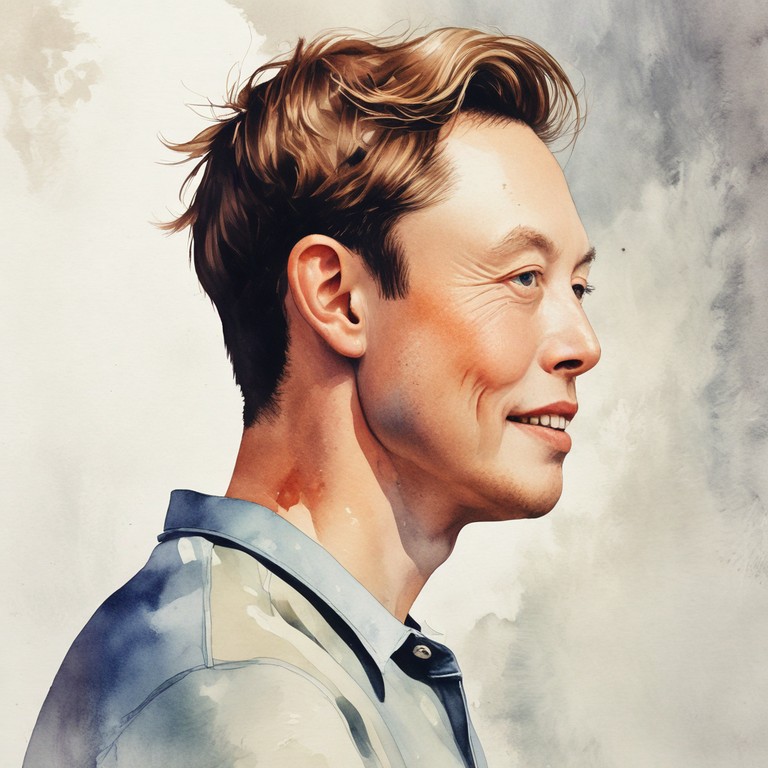} &
        \includegraphics[width=0.09\linewidth]{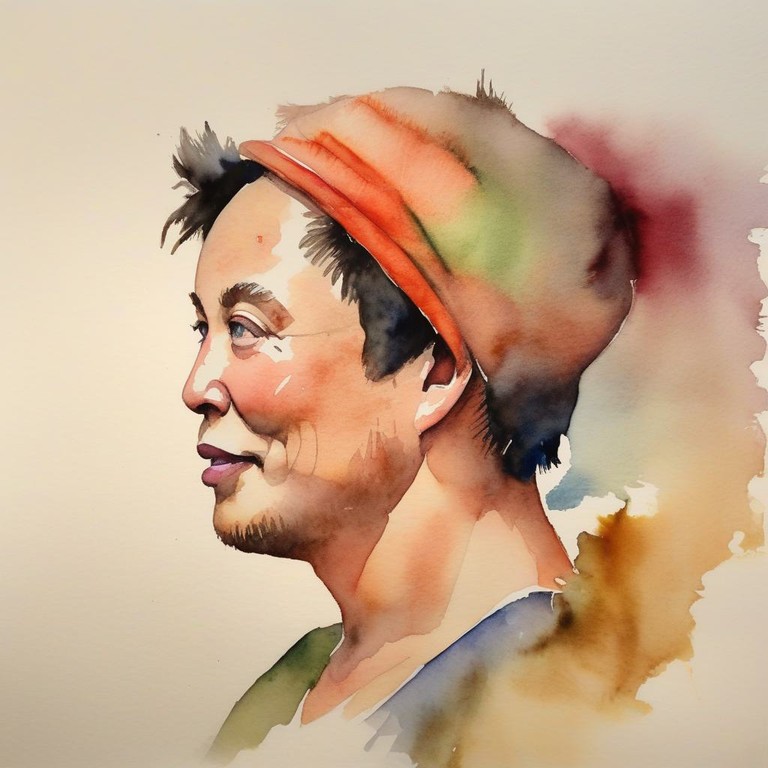} \\

        \multicolumn{2}{c}{\raisebox{18pt}{\rotatebox[origin=t]{90}{Pointillism}}} &
        \includegraphics[width=0.09\linewidth]{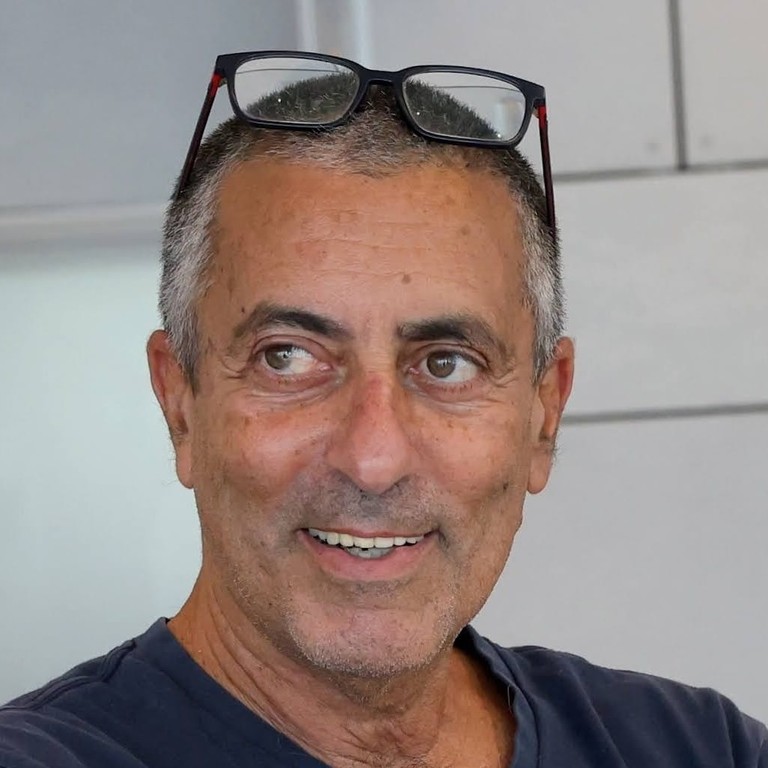} &
        \includegraphics[width=0.09\linewidth]{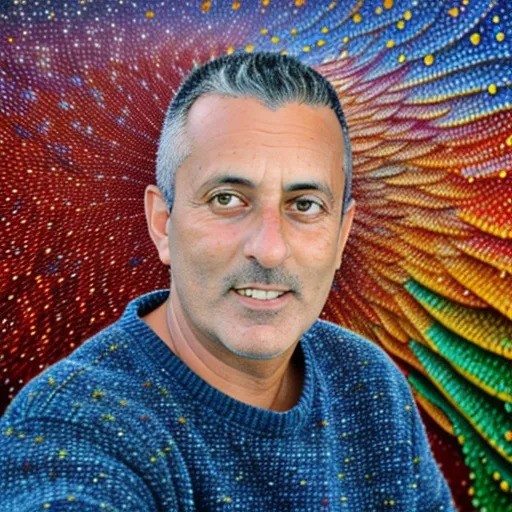} &
        \includegraphics[width=0.09\linewidth]{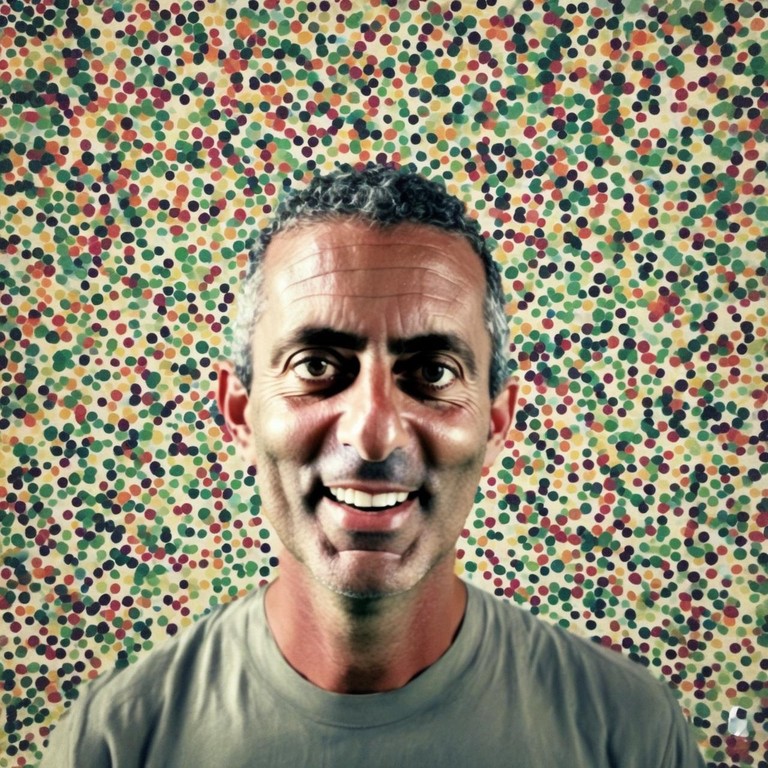} &
        \includegraphics[width=0.09\linewidth]{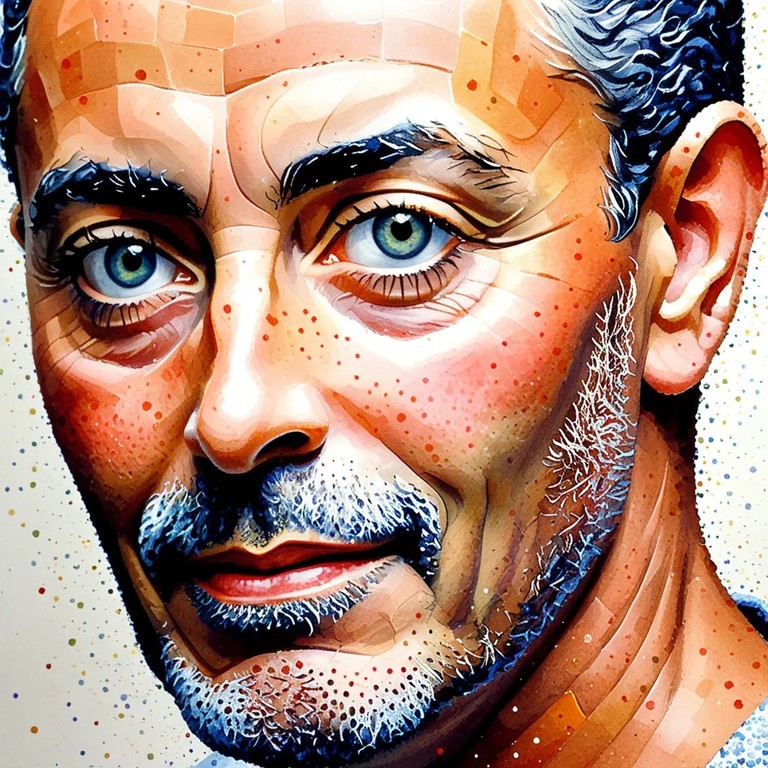} &
        \includegraphics[width=0.09\linewidth]{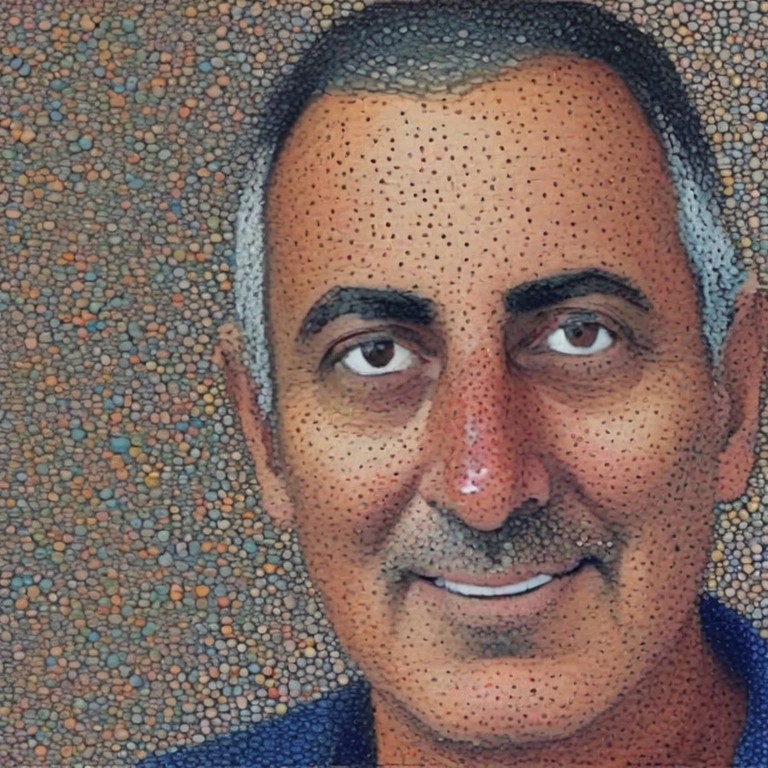} &
        \includegraphics[width=0.09\linewidth]{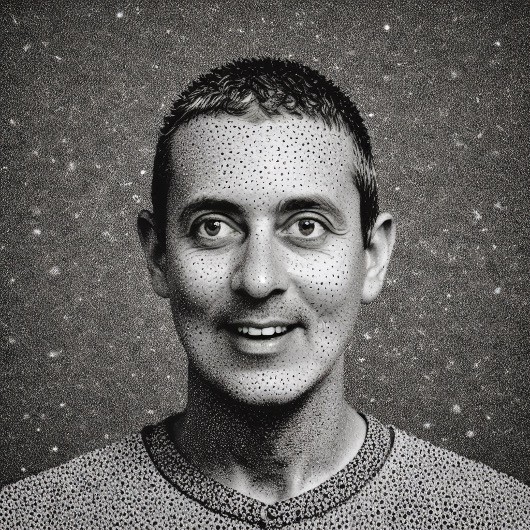} &
        \includegraphics[width=0.09\linewidth]{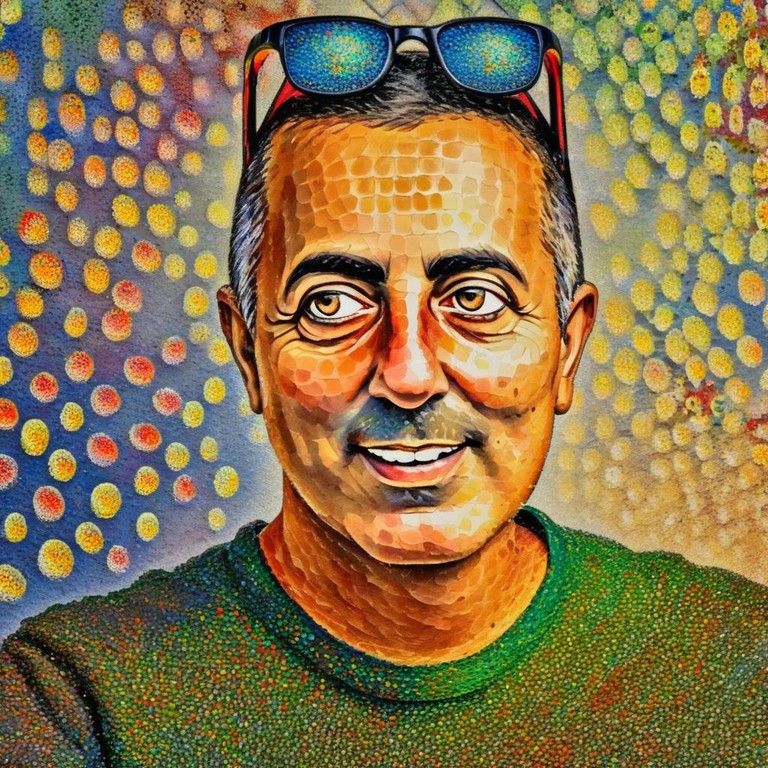} \\

        \raisebox{18pt}{\rotatebox[origin=t]{90}{Sticker, stick-}} &
        \raisebox{18pt}{\rotatebox[origin=t]{90}{ing tongue out}} &
        \includegraphics[width=0.09\linewidth]{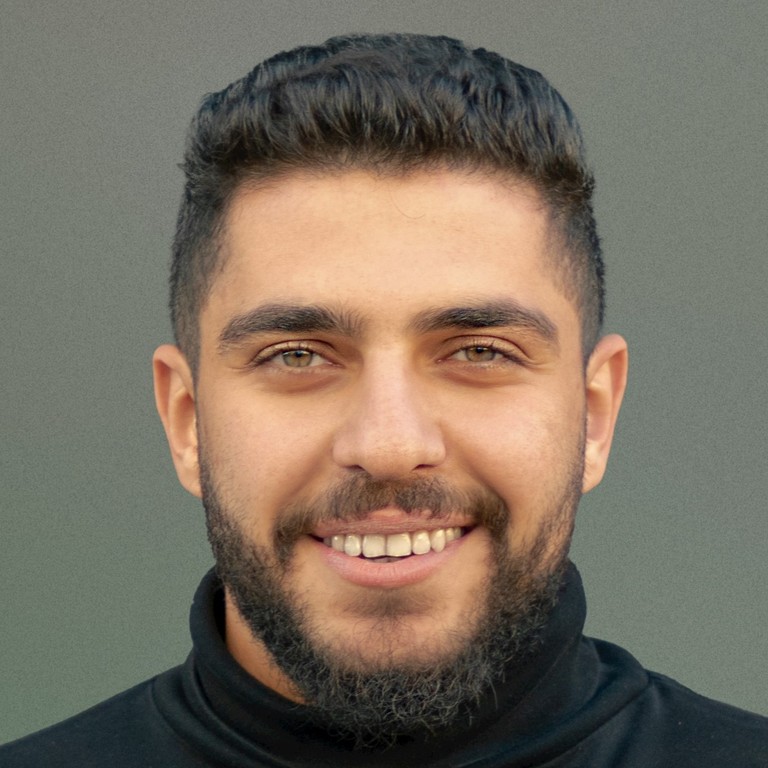} &
        \includegraphics[width=0.09\linewidth]{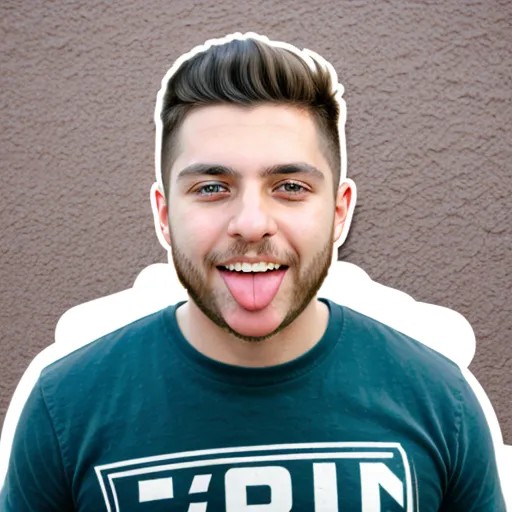} &
        \includegraphics[width=0.09\linewidth]{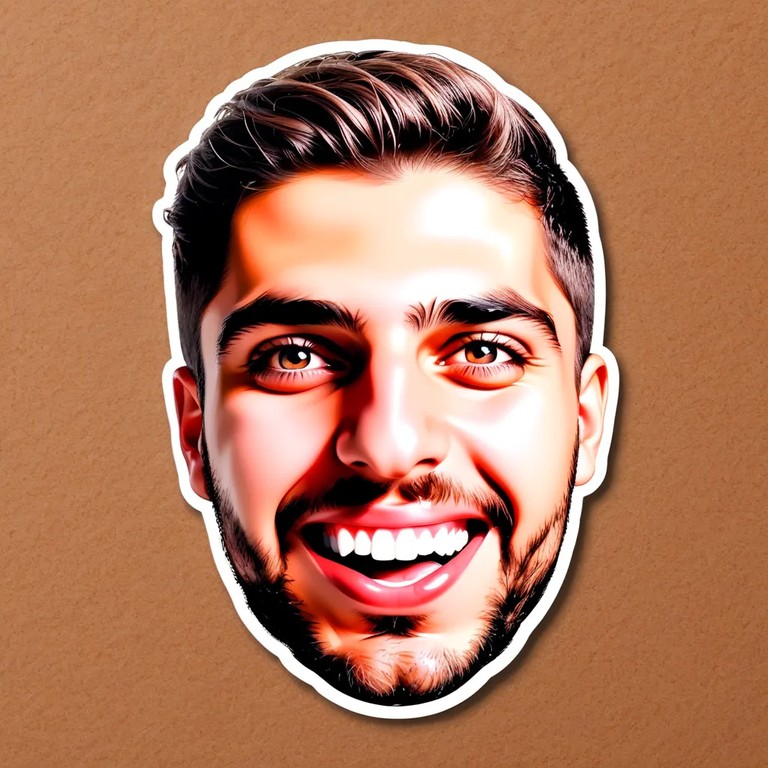} &
        \includegraphics[width=0.09\linewidth]{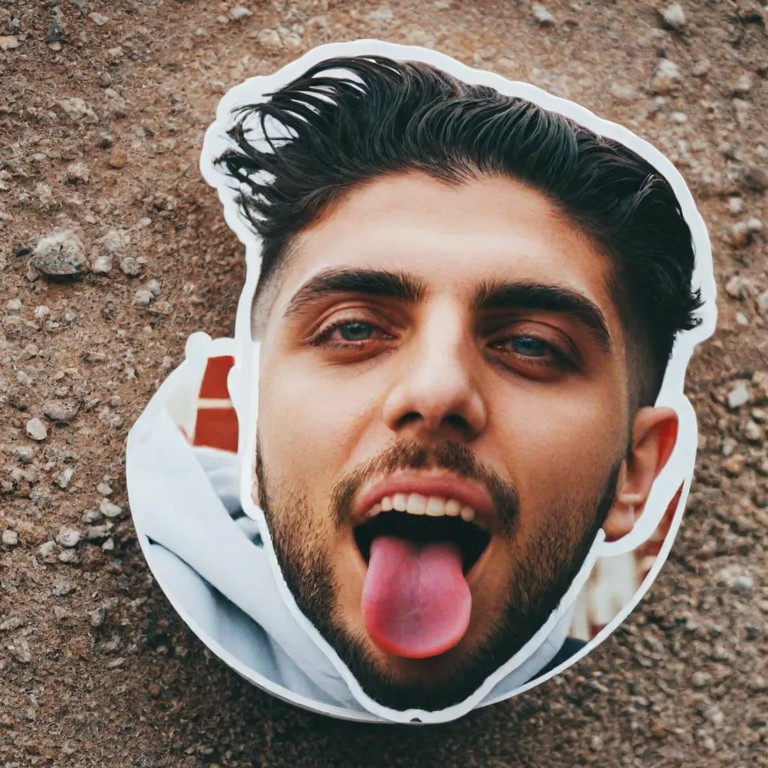} &
        \includegraphics[width=0.09\linewidth]{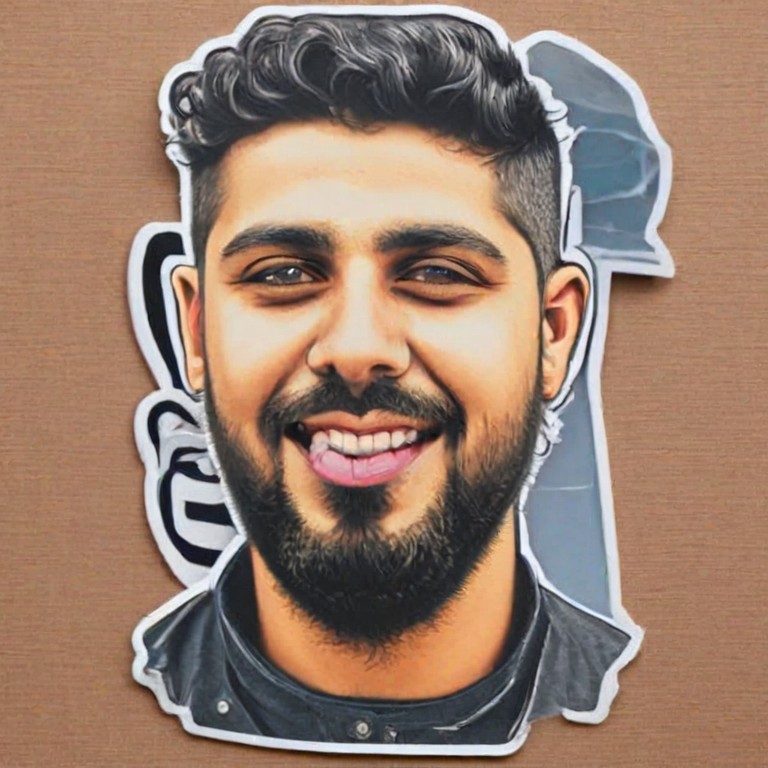} &
        \includegraphics[width=0.09\linewidth]{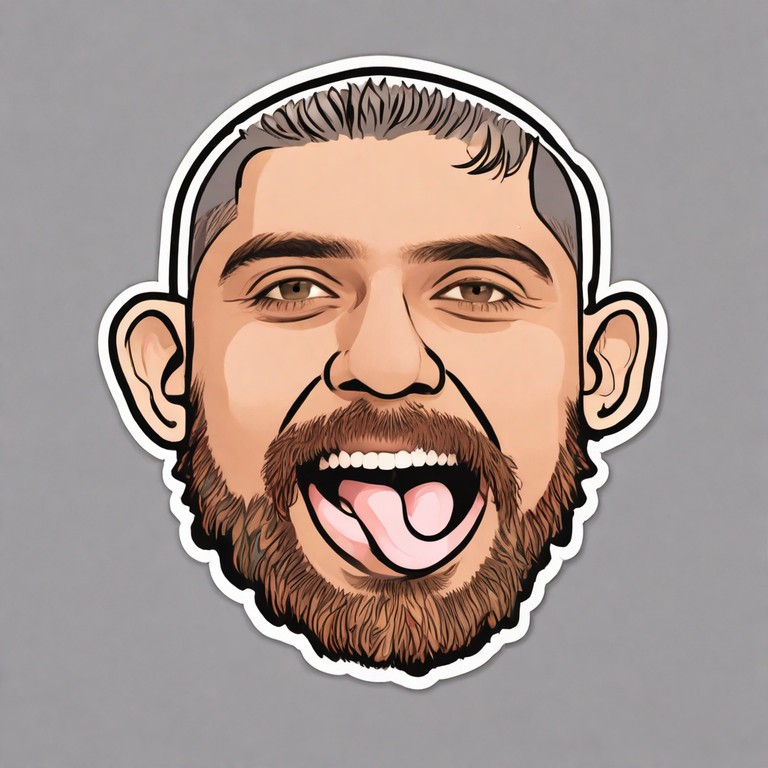} &
        \includegraphics[width=0.09\linewidth]{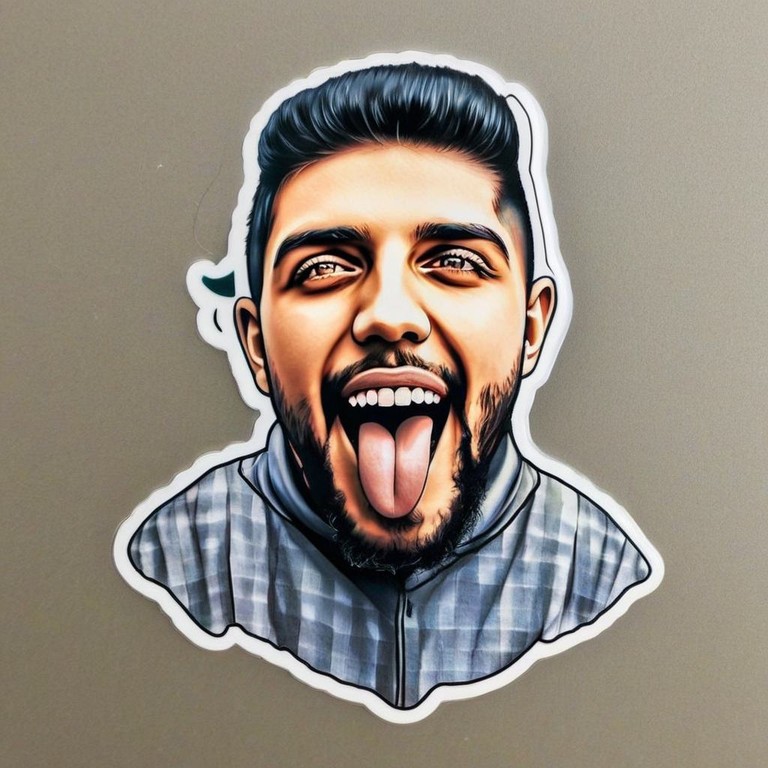} \\
    \end{tabular}
    \vspace{-8pt}
    }
    \caption{
        Left: qualitative comparison of human faces personalization methods. Our method successfully changes expressions and pose while preserving the identity. Right: quantitative comparison. Methods that use CLIP as their backbone encoder are marked with filled markers, while methods that build on face recognition models as their backbone encoder are marked with unfilled markers.
    }
    \vspace{-12pt}
    \label{fig:face-comparison}
\end{figure*}

\paragraph{Image injection mechanism}
At its core, nested attention is a method for injecting reference image features into a pretrained text-to-image model through its cross-attention layers. To show the benefit of this approach, we first compare it with other feature injection methods. 

We consider four alternative mechanisms. 
First, IP-Adapter's \cite{ye2023ipadapter} decoupled cross-attention (CA) mechanism, where text and image features are processed through parallel cross-attention layers that share queries and their outputs are summed. 
The second mechanism, known as `Simple Adapter'~\cite{ye2023ipadapter}, concatenates image features with text tokens in existing cross-attention layers, requiring no additional parameters beyond the encoder. 
The third alternative, which we call `Global $V$', explores the importance of query-dependent values. There, the special token's value is set to the mean of the encoder-produced tokens, projected through a per-layer projection matrix. This approach results in an identical value being used for all subject-queries. The final mechanism, termed `Multiple Tokens', demonstrates the significance of the nested mechanism by replacing the subject's token in the prompt with one token for each of the encoder's outputs (\ie the number of Q-Former tokens). To avoid attention-overfitting~\cite{tewel2023key}, we fix the keys of these tokens to the key of the original subject, but maintain a different encoder-produced value for each.

Importantly, we conduct these comparisons while maintaining consistent experimental conditions across all relevant parameters. Specifically, each method uses an identical Q-Former encoder architecture with the same number of learned queries, trained from scratch using the same dataset and number of training epochs. For all methods, we conduct only the first training stage ($512\times 512$ training resolution).

In \Cref{fig:mechanism-comparison} we show a qualitative comparison of the two most performant approaches -- `Nested Attention' and `Decoupled CA'. For the results of other methods, see Appendix~\ref{app:comparison-supp}. For each method, we display results using three inference-time hyperparameters that balance identity preservation and prompt alignment. For nested attention, $\lambda$ is the attention factor detailed in Section~\ref{sec:attn-factor}. For decoupled CA, $\lambda$ is the scale parameter introduced in IP-Adapter~\cite{ye2023ipadapter}. Our results demonstrate that nested attention achieves better balance, providing superior identity preservation while maintaining better alignment with the text prompt.

In \Cref{fig:comparison-quant-methods} we show a quantitative comparison of the four different methods. We measure text similarity using CLIP~\cite{radford2021learning}, and ID similarity with a face recognition model~\cite{ghostfacenet, leondgarse}. For nested attention, global $V$ and multiple tokens, we use $\lambda$ values of 1.0, 1.5, 2.0, 2.5, 3.0, 4.0. For decoupled CA we use $\lambda$ values of 0.5, 0.6, 0.7, 0.8, 0.9, 1.0. As illustrated in the graph, nested attention provides the best trade-off between identity preservation and text alignment.

\vspace{-14pt}
\paragraph{Comparison to prior work}
Next, we compare our model to other recent face personalization models, including two versions of IP-Adapter~\cite{ye2023ipadapter} (IPA-Face and IPA-CLIP), InstantID~\cite{wang2024instantid}, PulID~\cite{guo2024pulid}, PhotoMaker~\cite{li2023photomaker}, and LCM-Lookahead~\cite{gal2024lcmlookahead}. Qualitative and quantitative results are presented in Figure~\ref{fig:face-comparison}, and user study results are in Table~\ref{tab:user-study}.

Note that while our method was trained solely on FFHQ, all other methods were trained on larger datasets, some consisting of tens of millions of images (compared with FFHQ's $70,000$). Additionally, most of these baselines are specifically designed for human faces, using face-identity detection networks for feature extraction or as a loss. Our approach is more general, and can be applied to different domains. To better differentiate the results, we mark methods that utilize a CLIP-encoder with a full marker, and those that use a face-detection network as a feature extractor are shown with an unfilled marker in the graph of \Cref{fig:face-comparison}.

Our method outperforms all other CLIP-based encoder methods in both automatic metrics and throughout the user study. Notably, it does so when training on the comparatively small FFHQ dataset, without specialized data or losses. When considering the identity-network based approaches, we note that those that preserve the input face landmarks (\eg, InstantID) show artificially inflated identity scores~\cite{gal2024lcmlookahead} and a user study finds our identity preservation comparable, but with better prompt alignment and higher overall preference. This is particularly noticeable in prompts that require changes to pose and expression (\Cref{fig:face-comparison}, rows 1 \& 4). Similarly, our approach significantly outperforms IPA-Face in user evaluations.
PulID outperforms our approach across both identity similarity and prompt-alignment, but we note that it was trained on roughly a million curated images, uses both an identity network backbone and an identity loss which limit its extension to other domains, and proposes ideas which are largely orthogonal to our own, and could likely be merged with them.

\vspace{-14pt}
\paragraph{Multiple subjects comparison}
Our method can generate images with multiple personalized subjects (\Cref{fig:teaser}). For each domain, we run its own encoder and use its own nested attention layers to calculate the localized attention-values associated with its matching subject word (\eg, ``person'' and ``pet''). The subject-specific values are injected into the original cross-attention layers. 
As demonstrated, our approach effectively handles the generation of images with both a person and a pet subject, without requiring additional training, specialized components or adjustments. However, generating multiple subjects from the same domain remains challenging due to overlapping attention maps and self-attention leakage between subjects~\cite{dahary2024yourself}.

\begin{figure}
    \centering
    \setlength{\tabcolsep}{1pt}
    \scriptsize{
    \begin{tabular}{ccccc}
        \raisebox{21pt}{\rotatebox[origin=t]{90}{``... pointillism''}} &
        \includegraphics[width=0.22\linewidth]{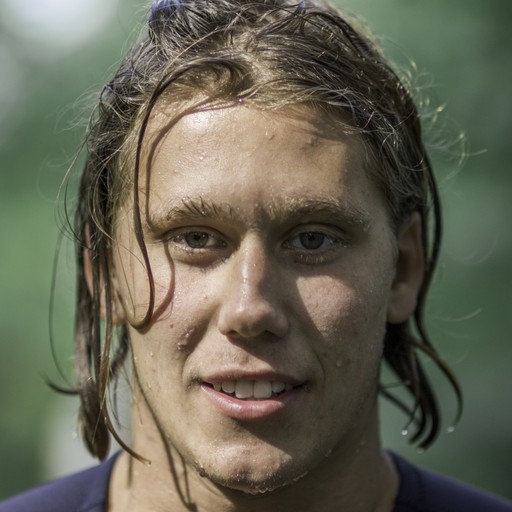} &
        \includegraphics[width=0.22\linewidth]{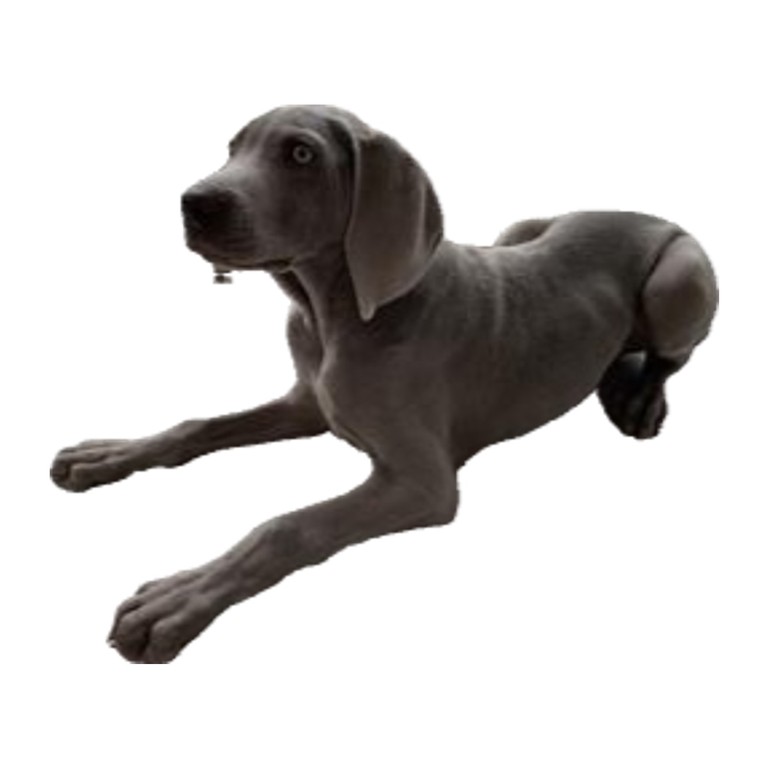} &
        \includegraphics[width=0.22\linewidth]{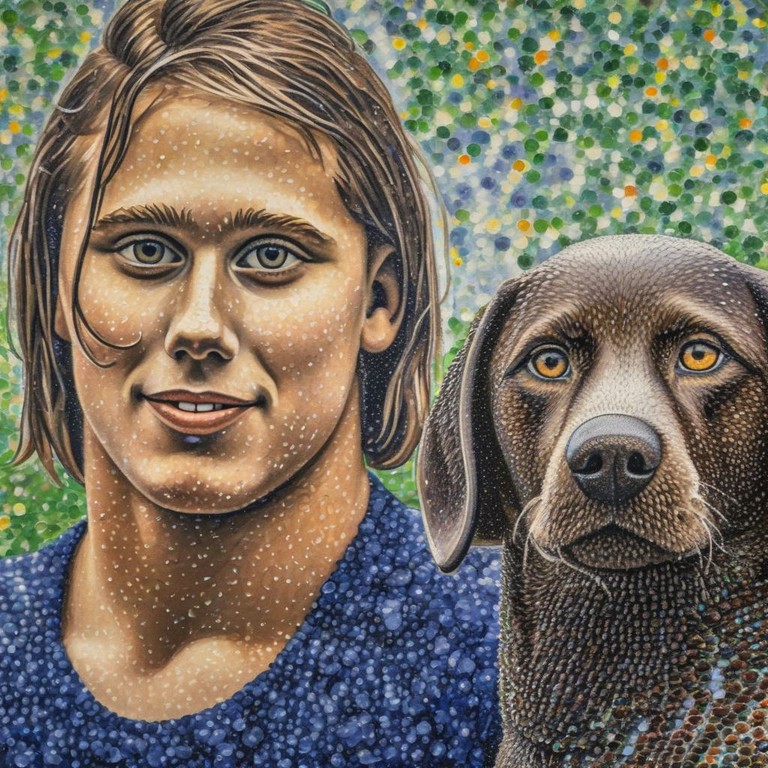} &
        \includegraphics[width=0.22\linewidth]{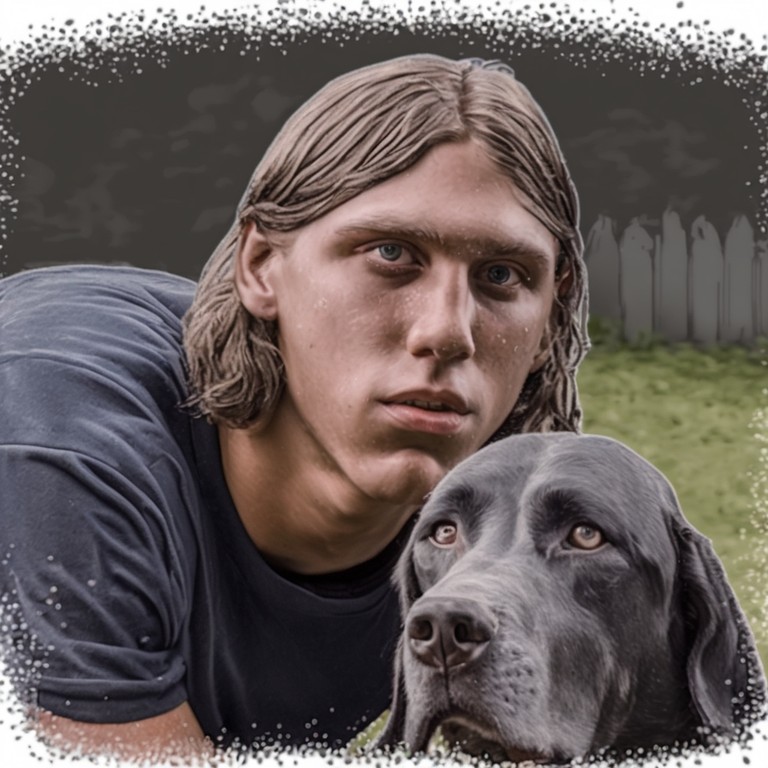} \\
        \raisebox{21pt}{\rotatebox[origin=t]{90}{``... digital art''}} &
        \includegraphics[width=0.22\linewidth]{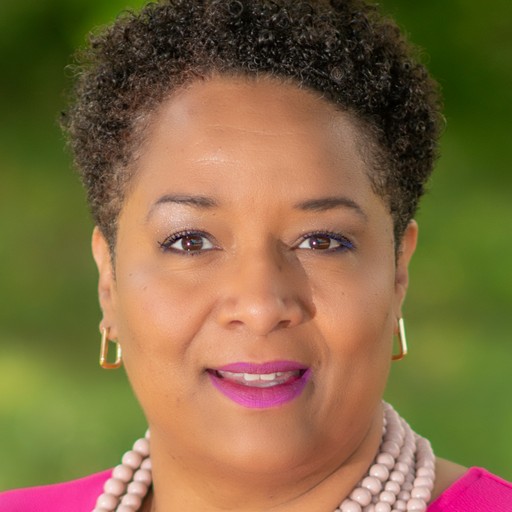} &
        \includegraphics[width=0.22\linewidth]{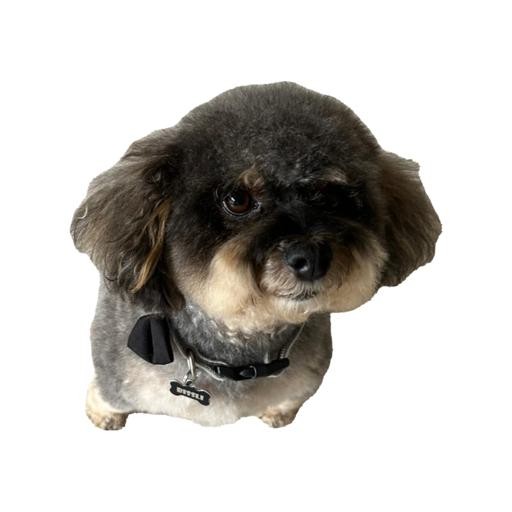} &
        \includegraphics[width=0.22\linewidth]{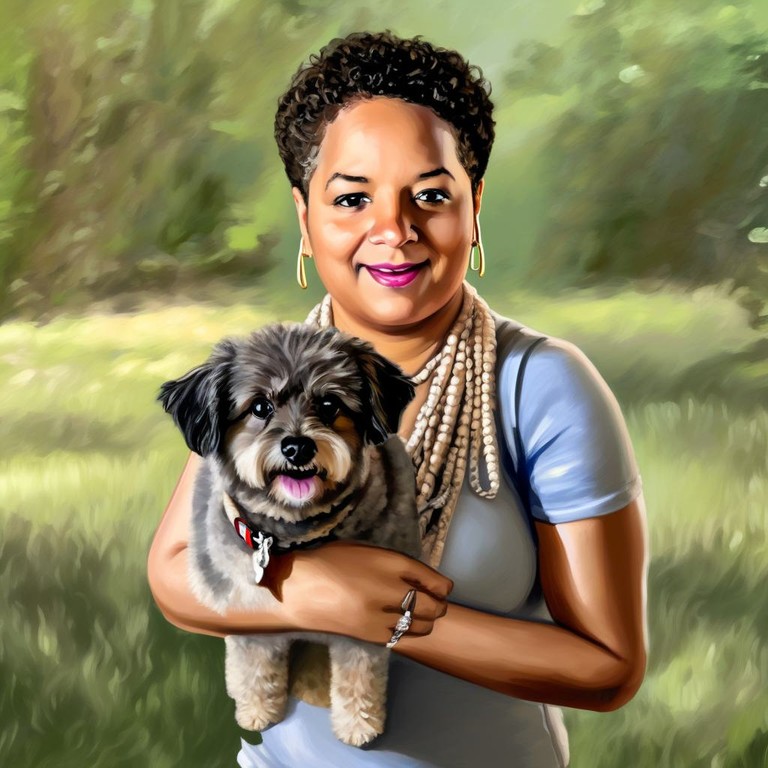} &
        \includegraphics[width=0.22\linewidth]{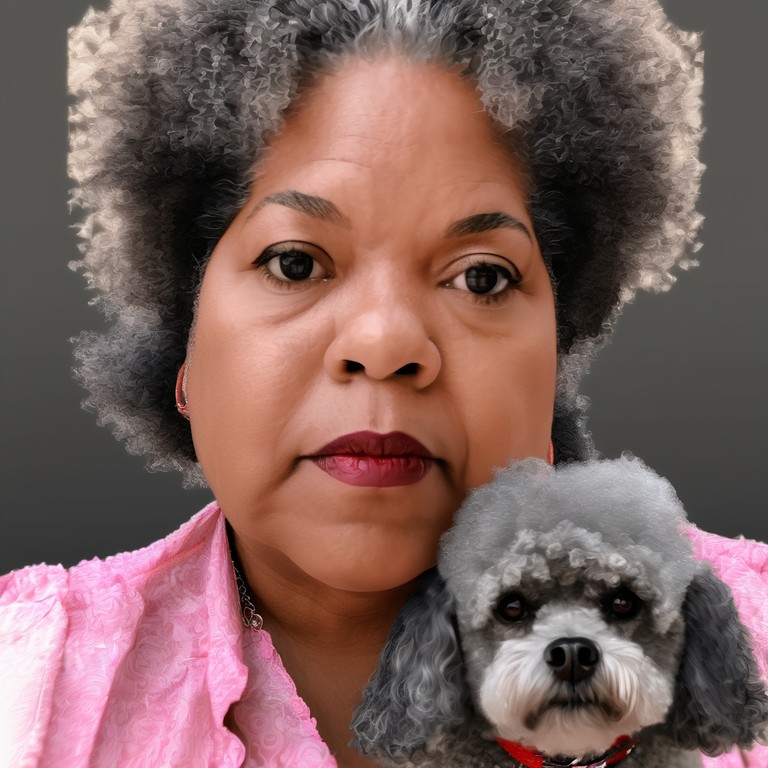} \\
        & Person input & Pet input & Ours & IPA
    \end{tabular}
    \vspace{-8pt}
    }
    \caption{
        Multi-subject generation comparison.
        }
    \vspace{-16pt}
    \label{fig:person_pet_comparison}
\end{figure}

Comparing multi-subject generation with IP-Adapter \cite{ye2023ipadapter}, the only strong baseline supporting non-face images, our approach shows superior identity preservation and prompt adherence when combining people and pets (Figure~\ref{fig:person_pet_comparison}). This is in part because the decoupled cross-attention approach is global, and its injected features can influence the entire image rather than the subject's regions.

\vspace{-2pt}
\subsection{Additional Results}
Here we show additional results. Additional results and applications are shown in the Appendix.

\begin{figure}
    \centering
    \setlength{\tabcolsep}{1pt}
    \scriptsize{
    \begin{tabular}{ccccc}
        \raisebox{22pt}{\rotatebox[origin=t]{90}{Input}} &
        \includegraphics[width=0.22\linewidth]{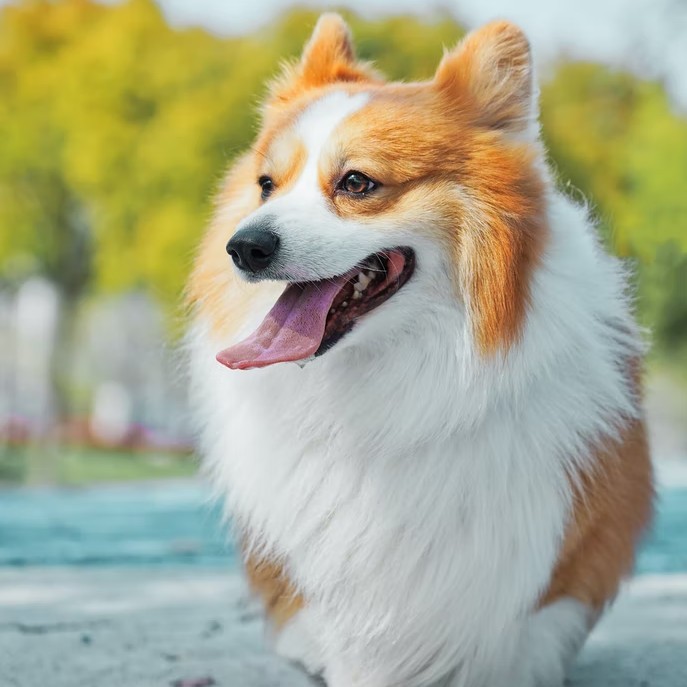} &
        \includegraphics[width=0.22\linewidth]{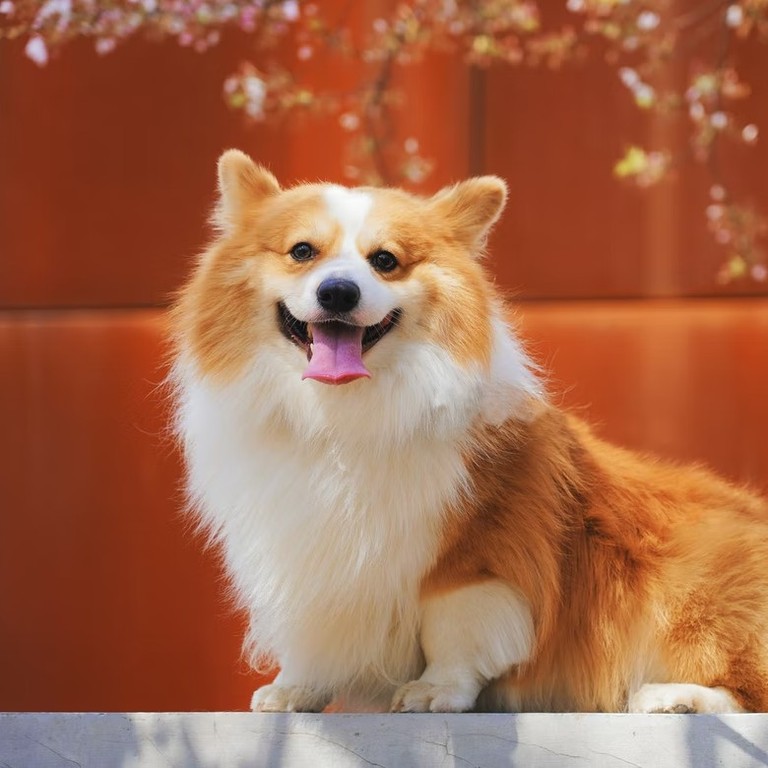} &
        \includegraphics[width=0.22\linewidth]{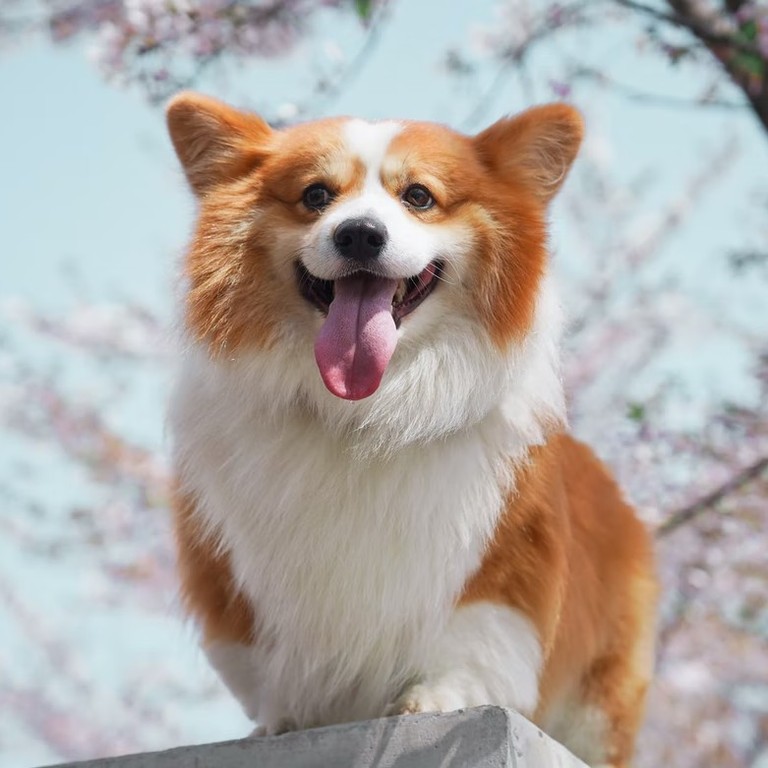} &
        \includegraphics[width=0.22\linewidth]{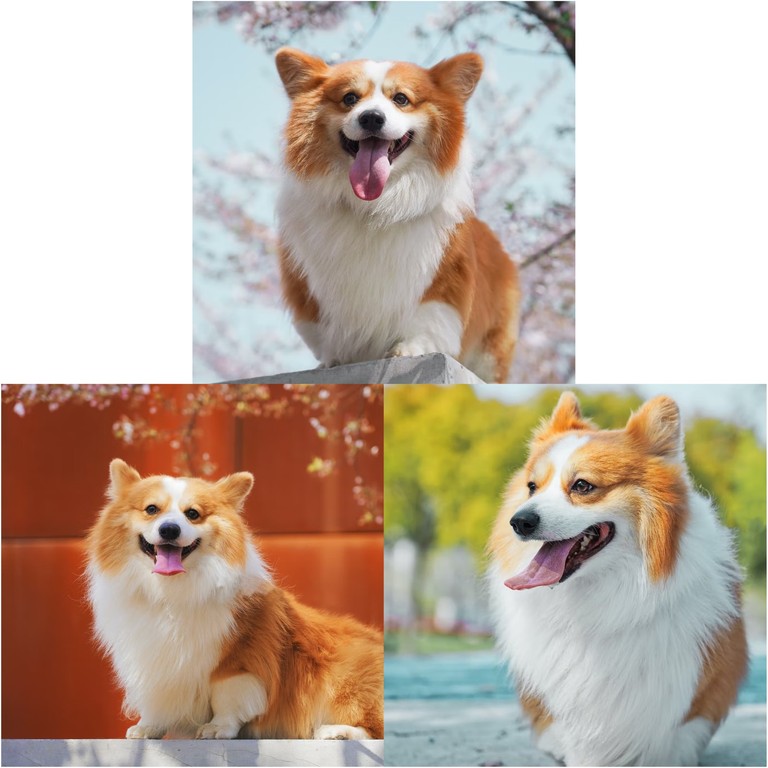} \\
        \raisebox{22pt}{\rotatebox[origin=t]{90}{``in a living room''}} &
        \includegraphics[width=0.22\linewidth]{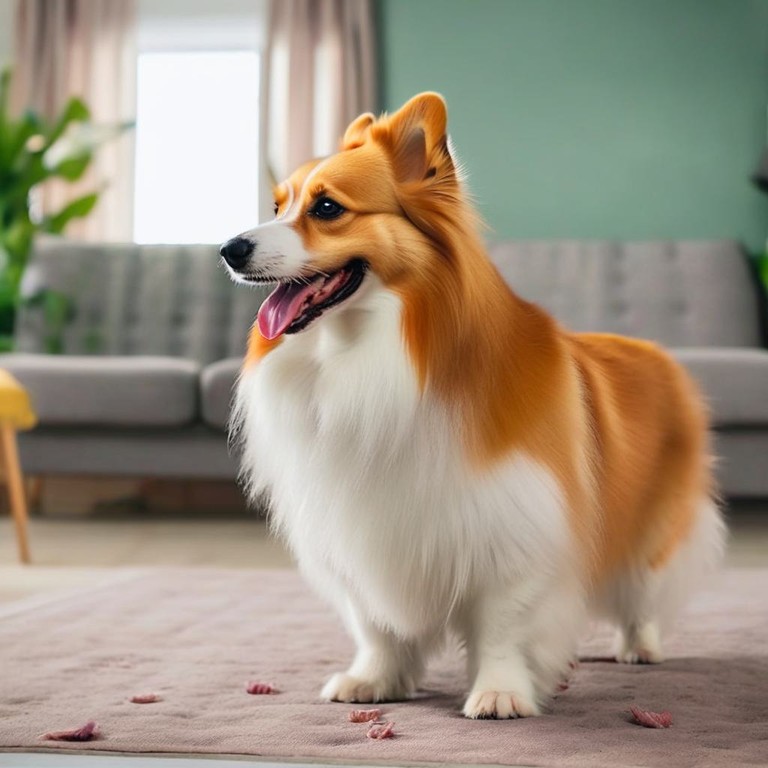} &
        \includegraphics[width=0.22\linewidth]{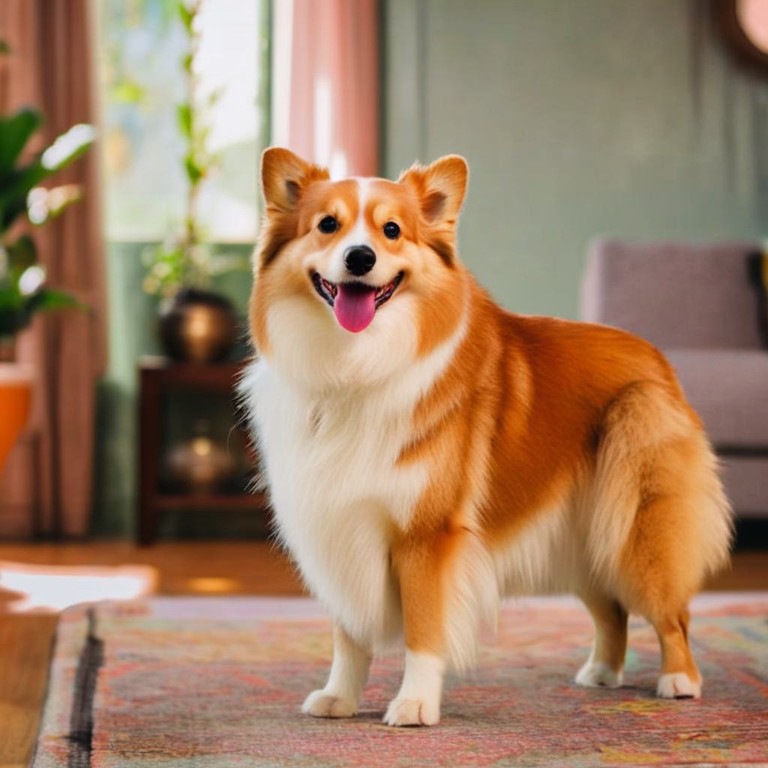} &
        \includegraphics[width=0.22\linewidth]{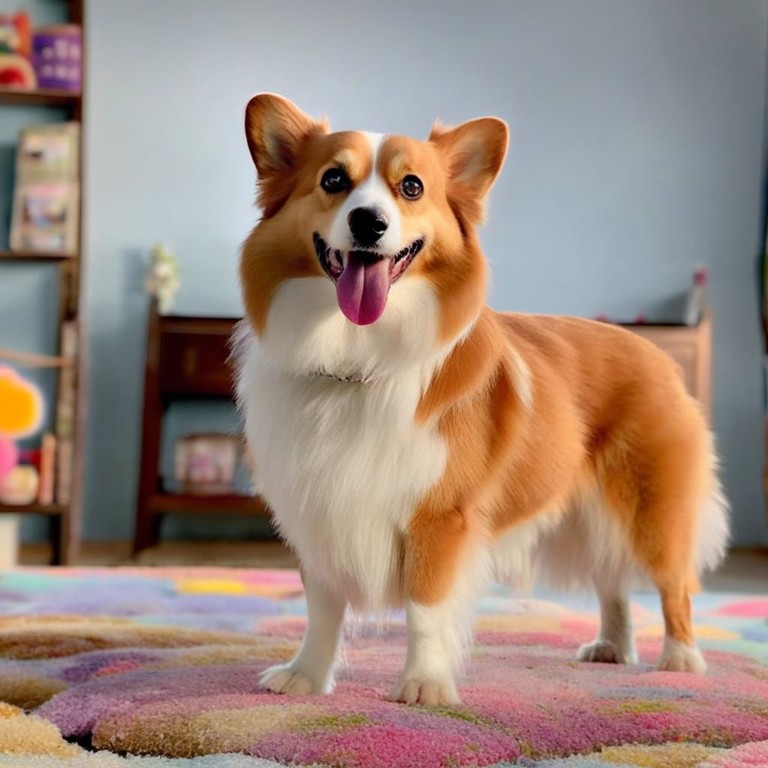} &
        \includegraphics[width=0.22\linewidth]{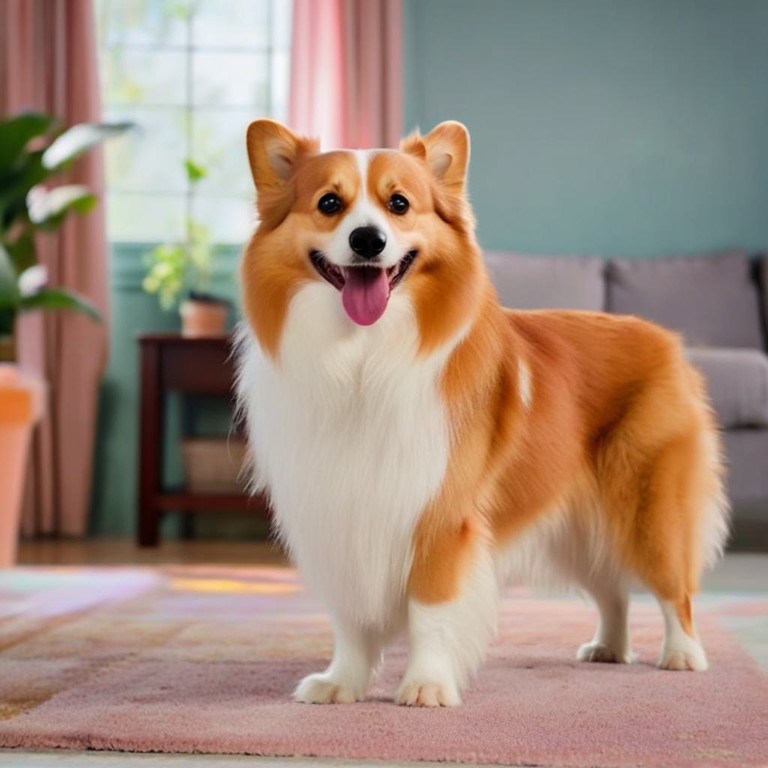} \\
    \end{tabular}
    \vspace{-8pt}
    }
    \caption{
        Using multiple images of the same concept increases the subject's fidelity in the generated image. 
    }
    \vspace{-16pt}
    \label{fig:multiple_images}
\end{figure}

\vspace{-12pt}
\paragraph{Multiple input images}
When multiple images of a subject are available, our approach can be improved even further, without any re-training or architecture changes. This can be particularly useful when capturing a subject's identity from a single image is challenging due to occlusions or ambiguity.
To leverage multiple input images, we encode each image separately and concatenate the resulting tokens. These tokens are then used as the input to the nested attention layers.
\Cref{fig:multiple_images} shows an example of combining multiple input images. Consider for example the leftmost input image, where it is ambiguous whether the orange part of the dog is part of the leg or the main body. Similarly, the dog's eyes appear smaller in the second column, and in the third column its fur appears shorter, with larger ears. By providing all input images to the encoder at the same time, the model is able to better capture the full identity of the dog, resulting in a higher-quality final output compared to using any single input image alone.

\begin{table}
    \centering
    \caption{User study results. We show winrate of our method in user preference against each method.
    }
    \vspace{-6pt}
    \label{tab:user-study}
    \footnotesize{
    \begin{tabular}{lcccc}
        \toprule
         Metric & IPA-Face & InstantID & Lookahead & PulID \\
        \midrule
        Prompt adherence & 65\% & 68\% & 55\% & 42\% \\
        ID similarity & 86\% & 47\% & 52\% & 50\% \\
        Overall preference & 71\% & 66\% & 56\% & 39\% \\
        \bottomrule
    \end{tabular}
    \vspace{-5pt}
    }
\end{table}

\section{Conclusions}
We introduced nested attention, a novel identity injection technique that provides a rich subject representation within the existing cross-attention layers of the model. It is based on two key principles: (i) modifying only the attention value of the subject token while keeping keys and other values unchanged, and (ii) making the subject token's attention value dependent on the query, \ie, assigning the subject a different value for each image region. In this sense, nested attention can be interpreted as an IP-Adapter that anchors the subject's encoding to a single textual token. This design better preserves the model's prior, while enabling a detailed and accurate representation of the subject.

Future work could explore adaptations of nested attention to other tasks, such as subject-conditioned inpainting or style transfer. Another promising direction involves extending the encoder to a domain-agnostic approach, which could tackle subjects from unseen classes. 
Finally, since IP-Adapter's decoupled-attention mechanism is a core component of many recent personalization encoders, we hope replacing it with our approach could boost their performance.

\section*{Acknowledgment}
We thank Ron Mokady, Amir Hertz, Yuval Alaluf, Amit Bermano, Yoad Tewel, and Maxwell Jones for their early feedback and helpful suggestions.

{
    \small
    \bibliographystyle{ieeenat_fullname}
    \bibliography{main}
}

\clearpage

\begin{appendices}

\section{Implementation Details} \label{app:details}

\paragraph{Method}
We train the human face model on FFHQ-Wild~\cite{karras2019style}, where the encoder's input image is aligned and cropped, and the target image is the full in-the-wild image. For the pets model we use SDXL to generate a synthetic data of $50,000$ pet images, where the background in these images is white. Additionally, we use $15,000$ images from AFHQ~\cite{choi2020starganv2} and $1,000$ dog images from \cite{mokady2022selfdistilled}.
The human face model has $1024$ learned queries in the Q-Former, while the pets model has $256$.

We find that better results for combining a person and a pet in the same image are obtained by segmenting the pet, and setting a white background.

All models are trained on NVIDIA A100 80GB GPUs. The face model uses 4 GPUs and a total batch size of $32$ in the first phase, and 8 GPUs with a total batch size of 16 in the second phase. For the pets model, both phases are trained on 8 GPUs with total batch sizes of 128 and 32 respectively.

\paragraph{User Study}
The results of our user study are provided in the main paper. A total of 22 participants took part, each evaluating 12 tuples consisting of an input image, an input prompt, our method's result, and a result from one of the other methods. For each tuple, participants were asked three questions: (1) which output image is better aligned with the prompt, (2) which output image better preserves the identity of the input image, and (3) which result is better overall. This setup yielded 264 responses for each question type.

\section{Additional Results}

\begin{figure*}
    \centering
    \setlength{\tabcolsep}{1pt}
    \scriptsize{
    \begin{tabular}{ccccccc}
        \includegraphics[width=0.142\linewidth]{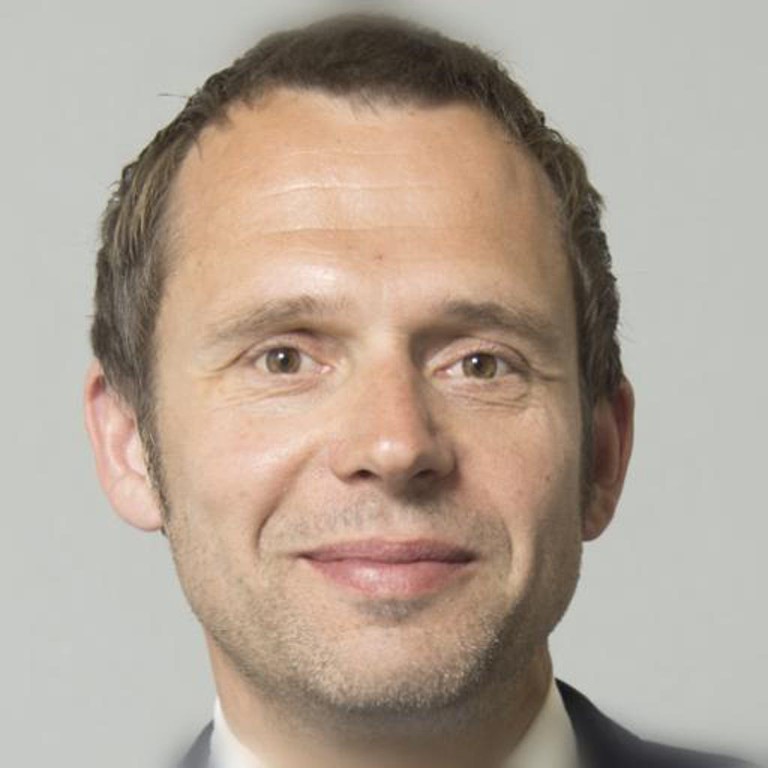} &
        \includegraphics[width=0.142\linewidth]{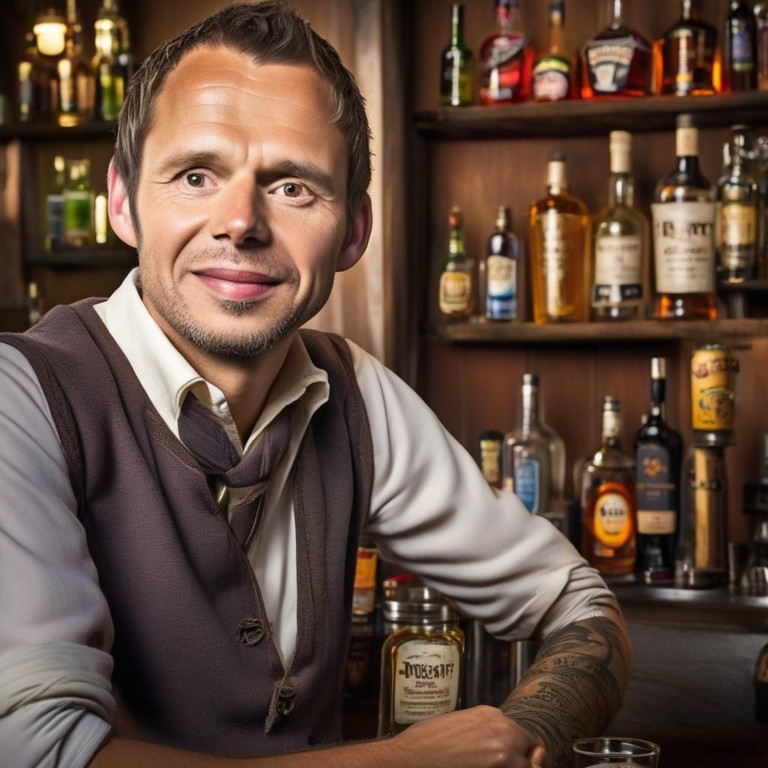} &
        \includegraphics[width=0.142\linewidth]{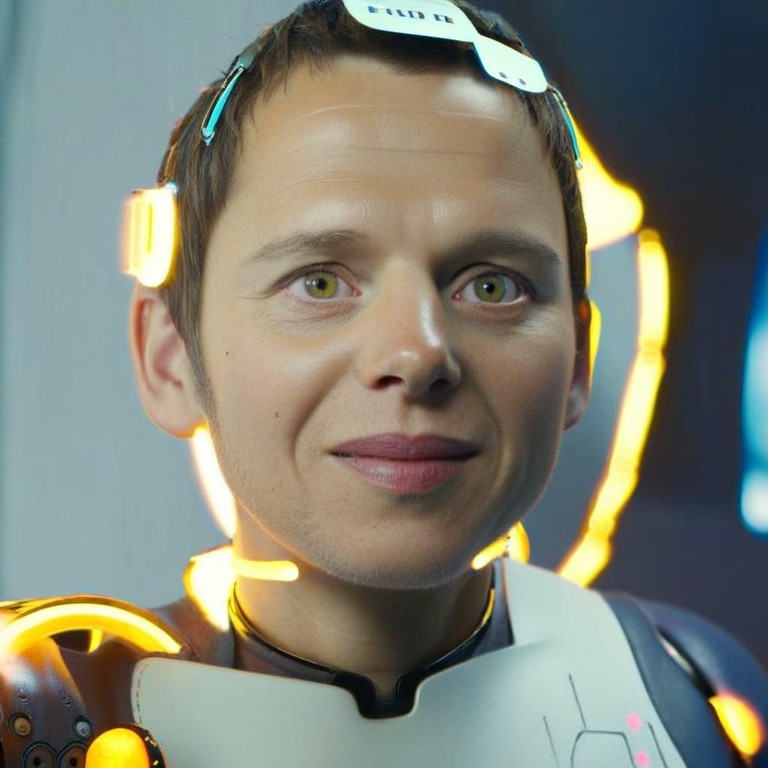} &
        \includegraphics[width=0.142\linewidth]{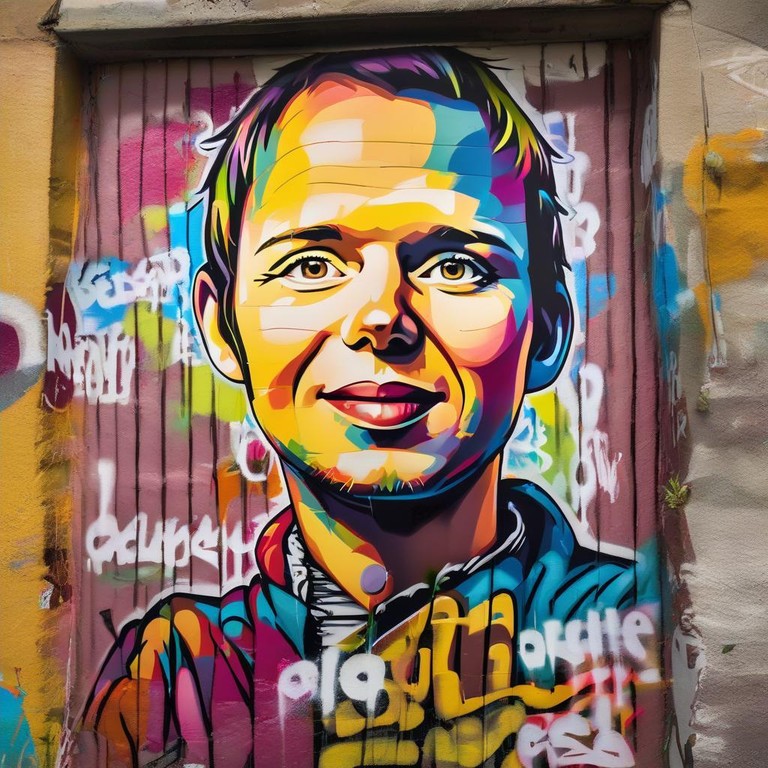} &
        \includegraphics[width=0.142\linewidth]{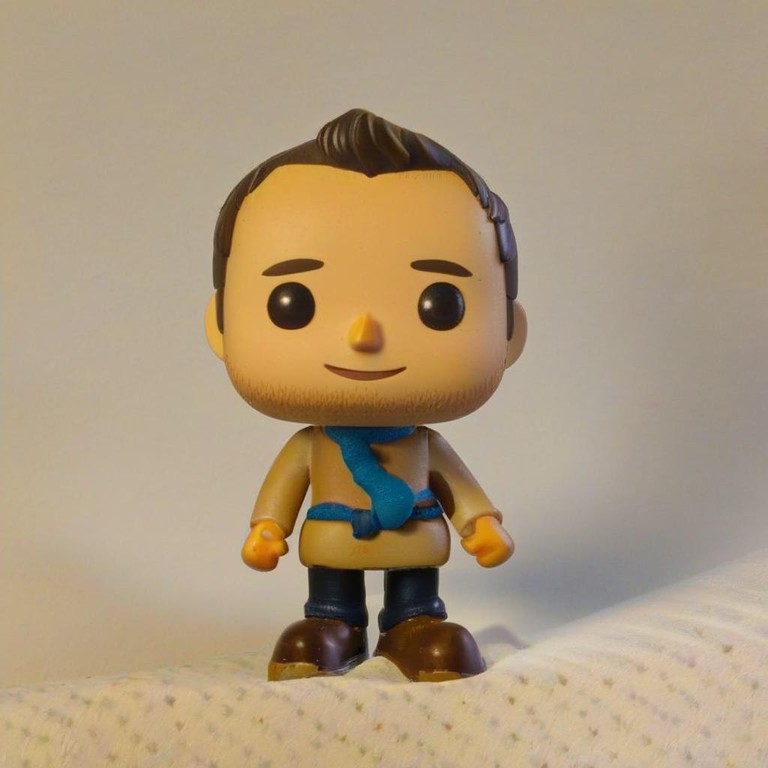} &
        \includegraphics[width=0.142\linewidth]{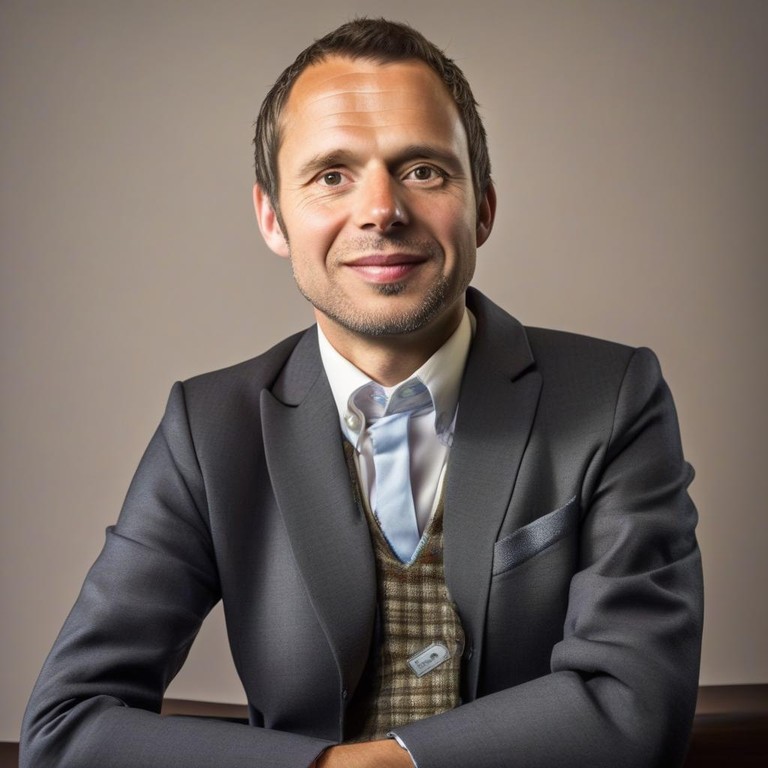} &
        \includegraphics[width=0.142\linewidth]{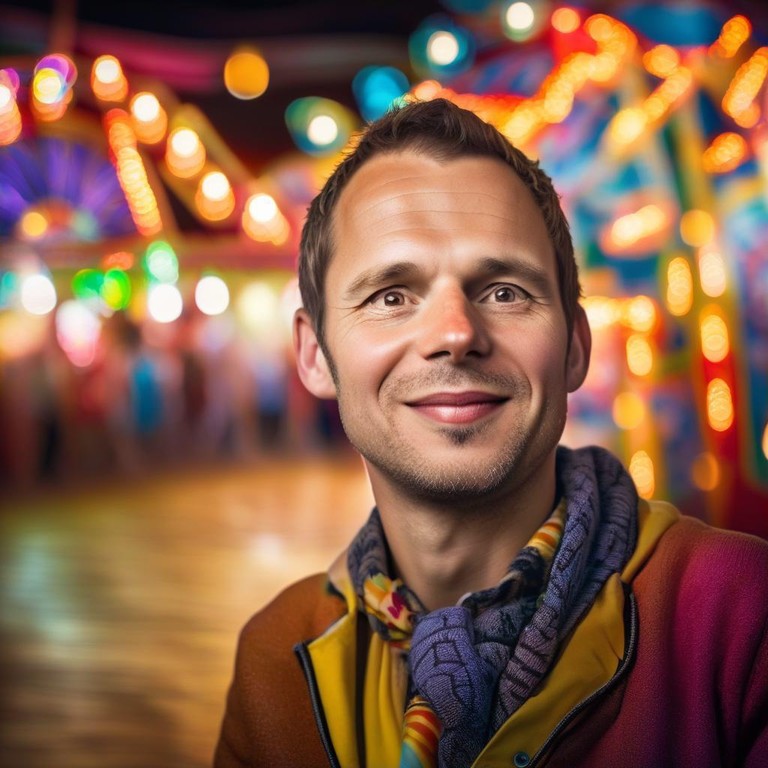} \\
        \includegraphics[width=0.142\linewidth]{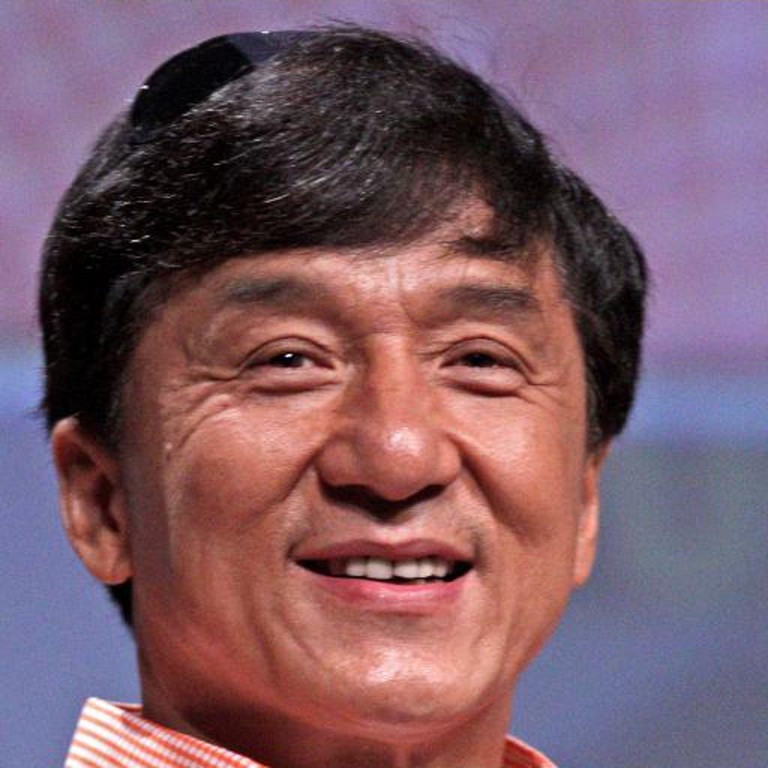} &
        \includegraphics[width=0.142\linewidth]{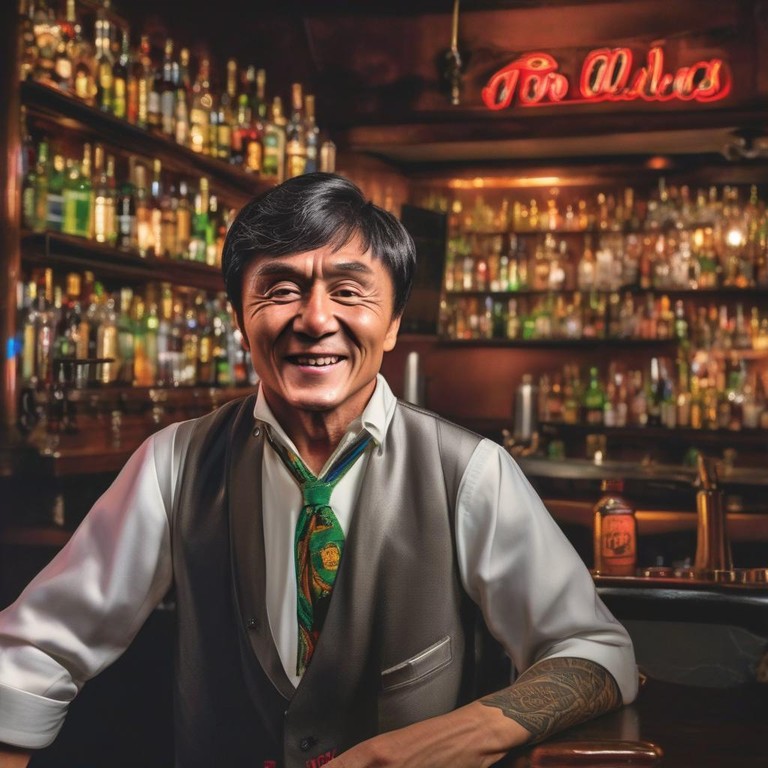} &
        \includegraphics[width=0.142\linewidth]{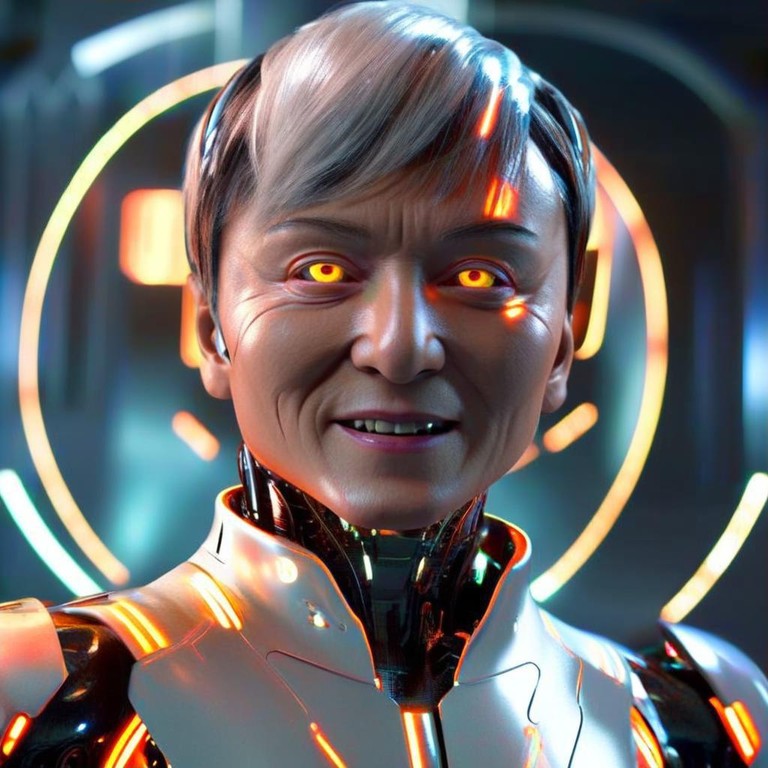} &
        \includegraphics[width=0.142\linewidth]{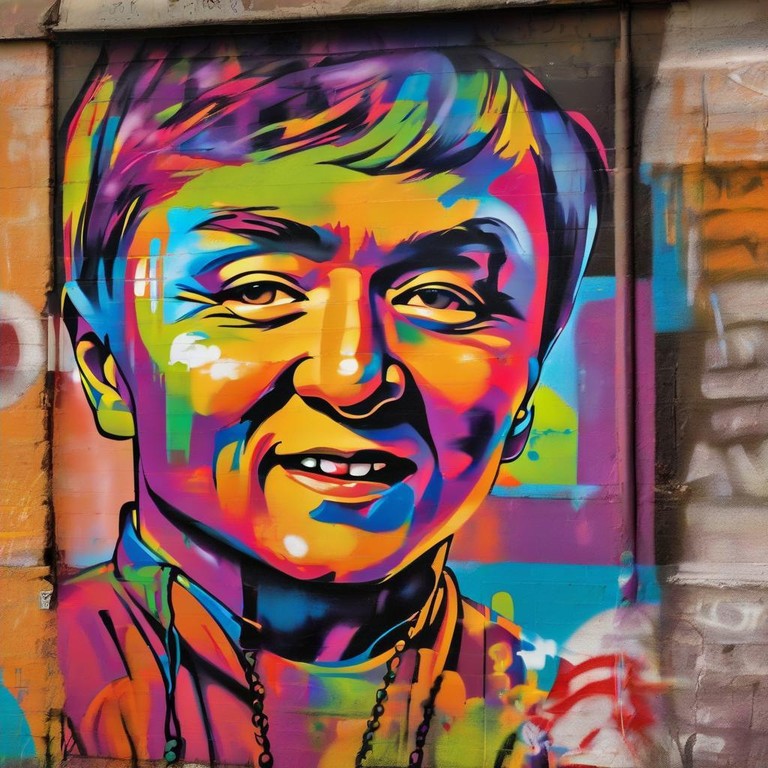} &
        \includegraphics[width=0.142\linewidth]{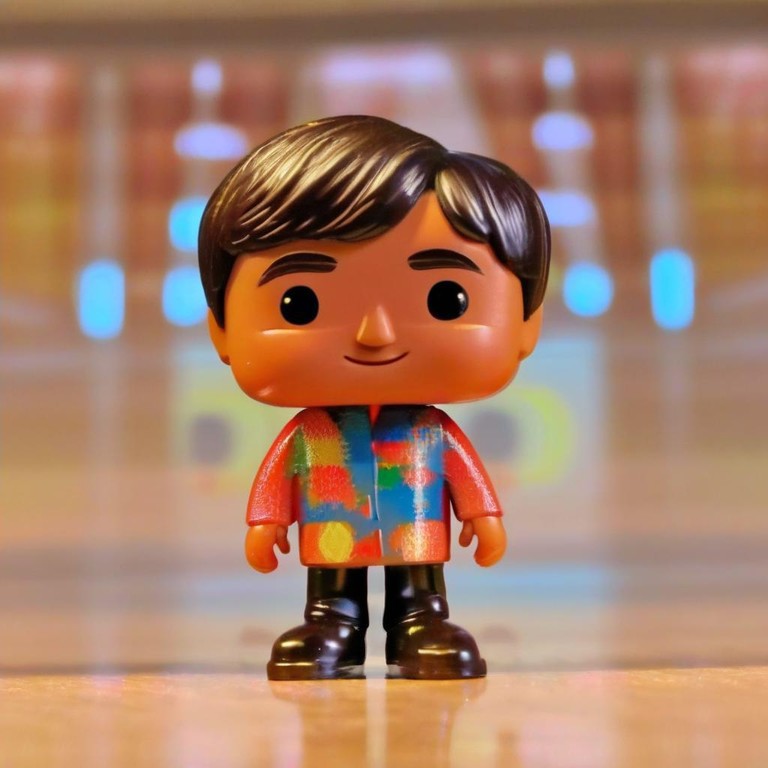} &
        \includegraphics[width=0.142\linewidth]{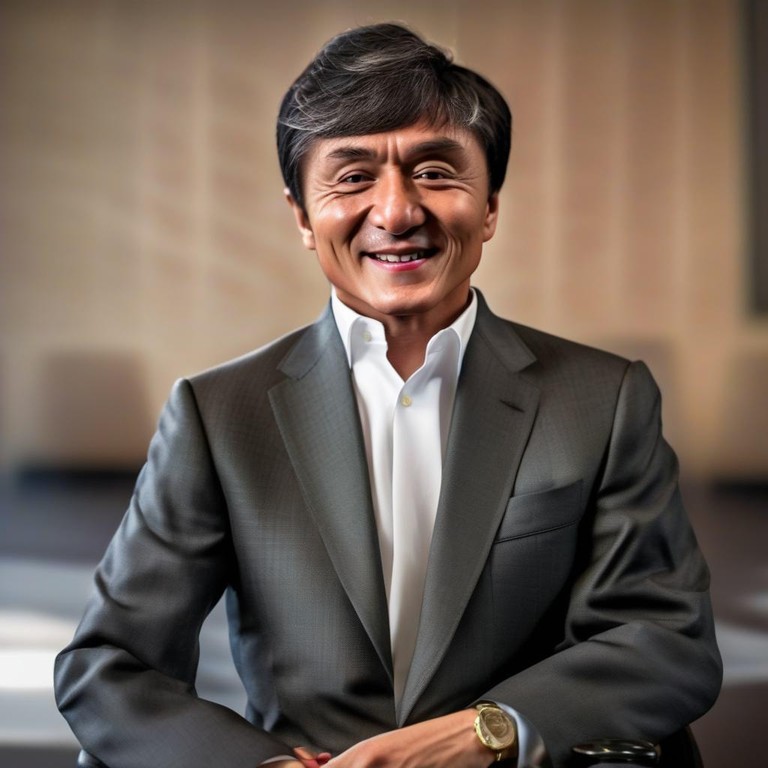} &
        \includegraphics[width=0.142\linewidth]{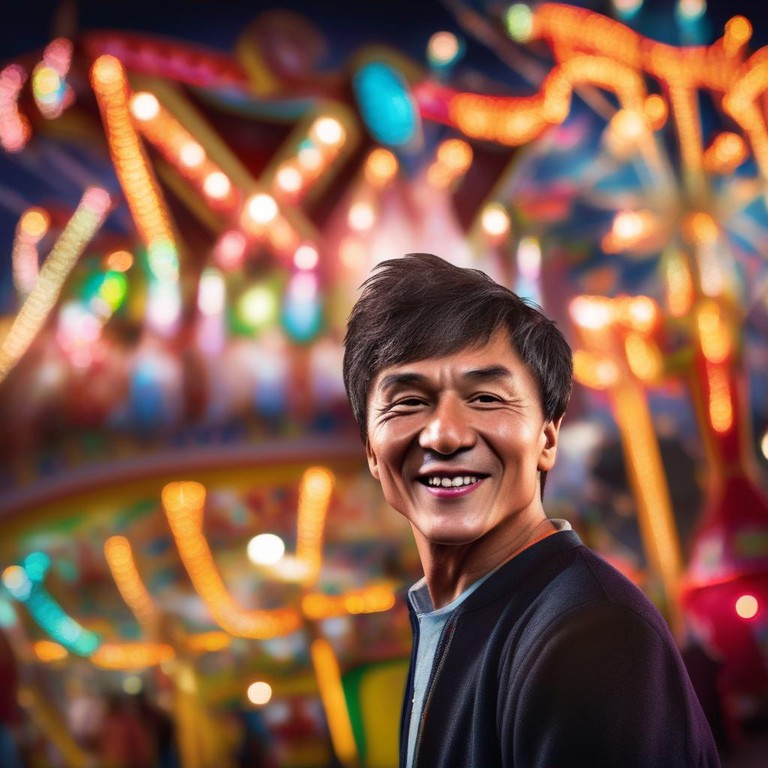} \\
        \includegraphics[width=0.142\linewidth]{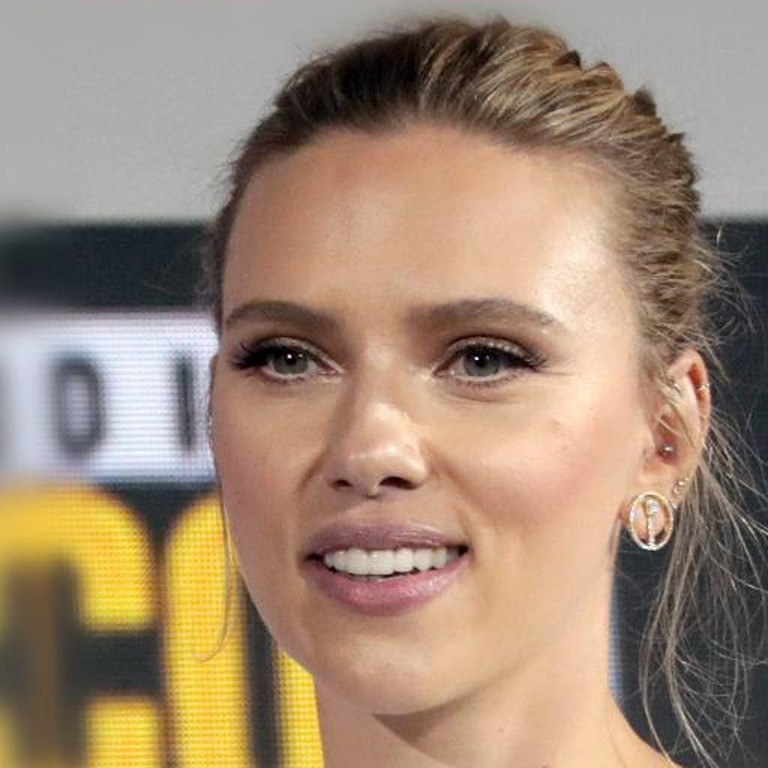} &
        \includegraphics[width=0.142\linewidth]{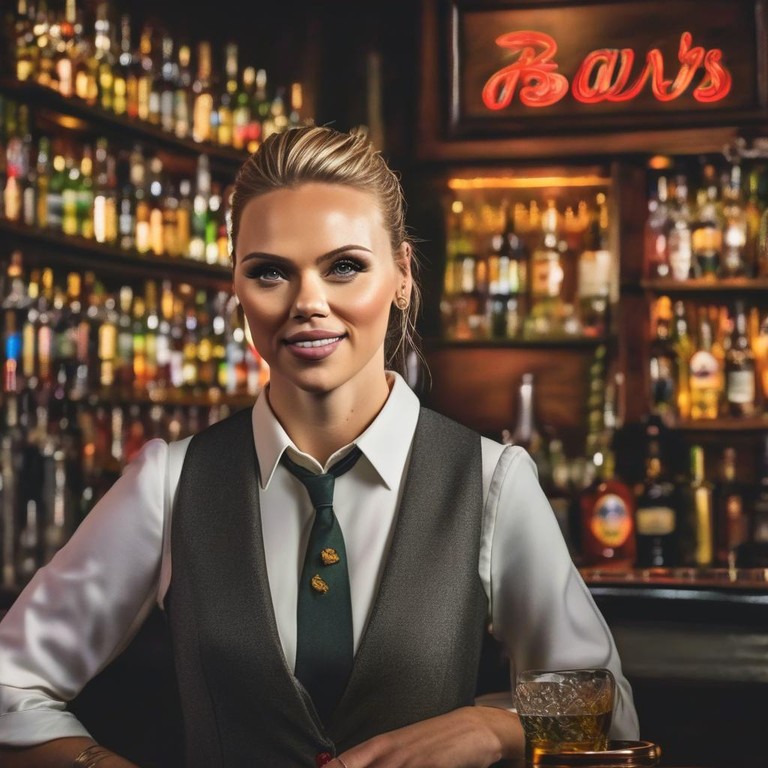} &
        \includegraphics[width=0.142\linewidth]{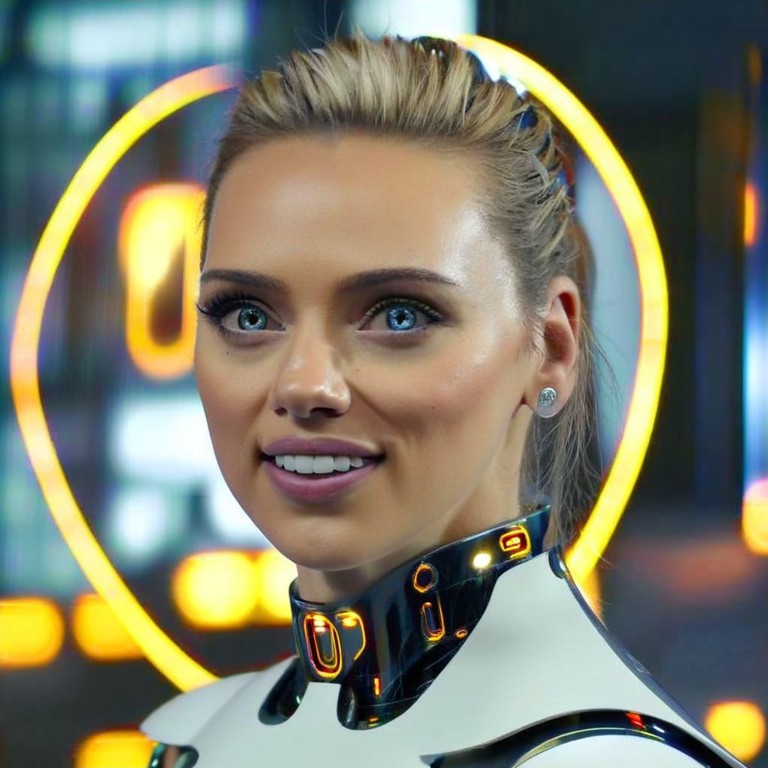} &
        \includegraphics[width=0.142\linewidth]{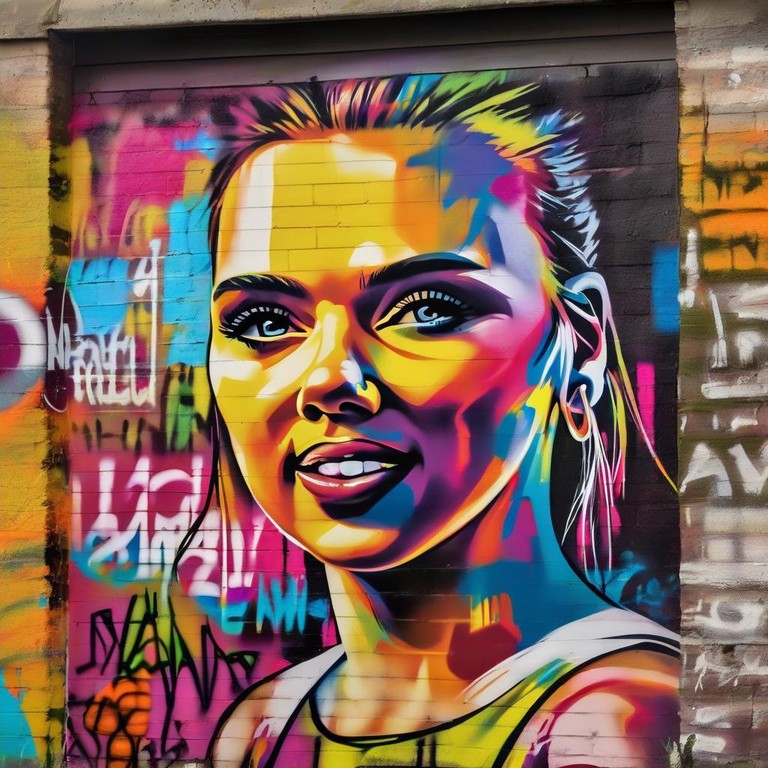} &
        \includegraphics[width=0.142\linewidth]{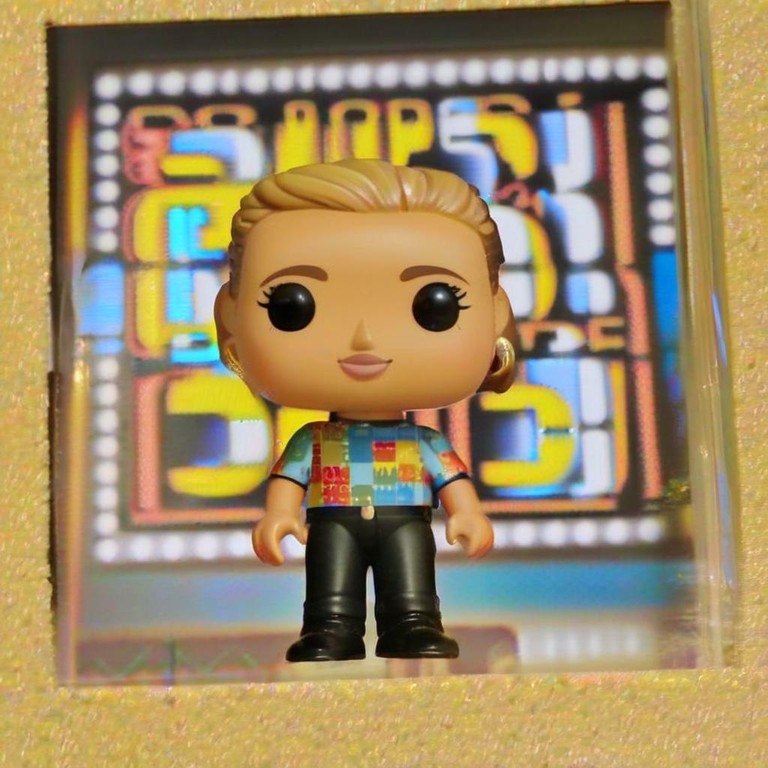} &
        \includegraphics[width=0.142\linewidth]{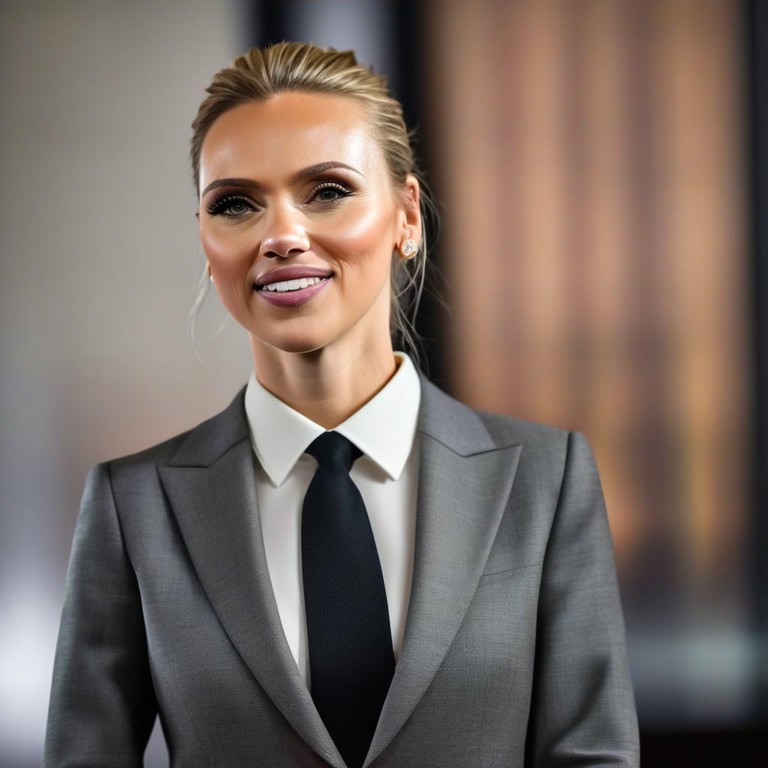} &
        \includegraphics[width=0.142\linewidth]{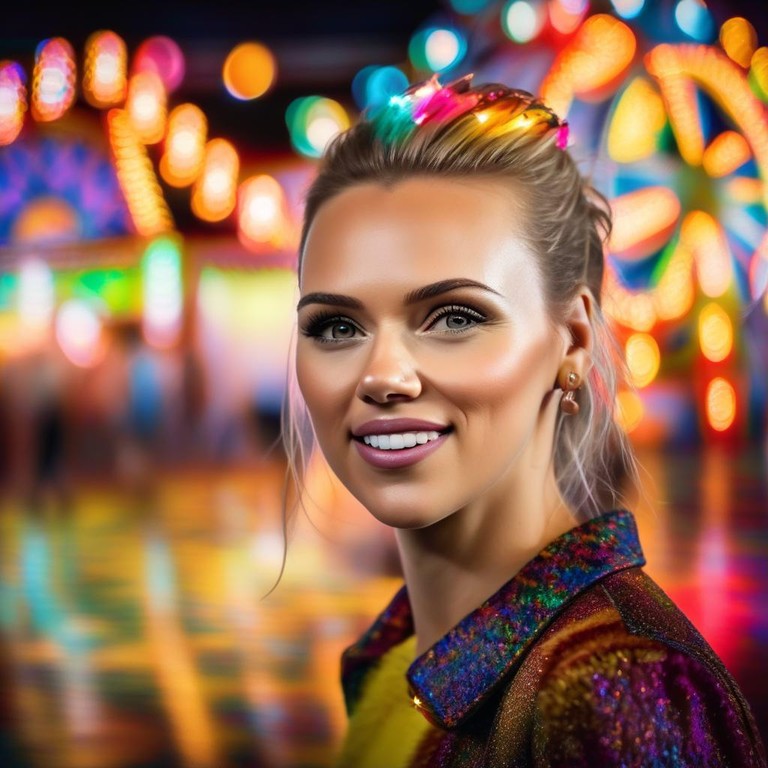} \\
        \includegraphics[width=0.142\linewidth]{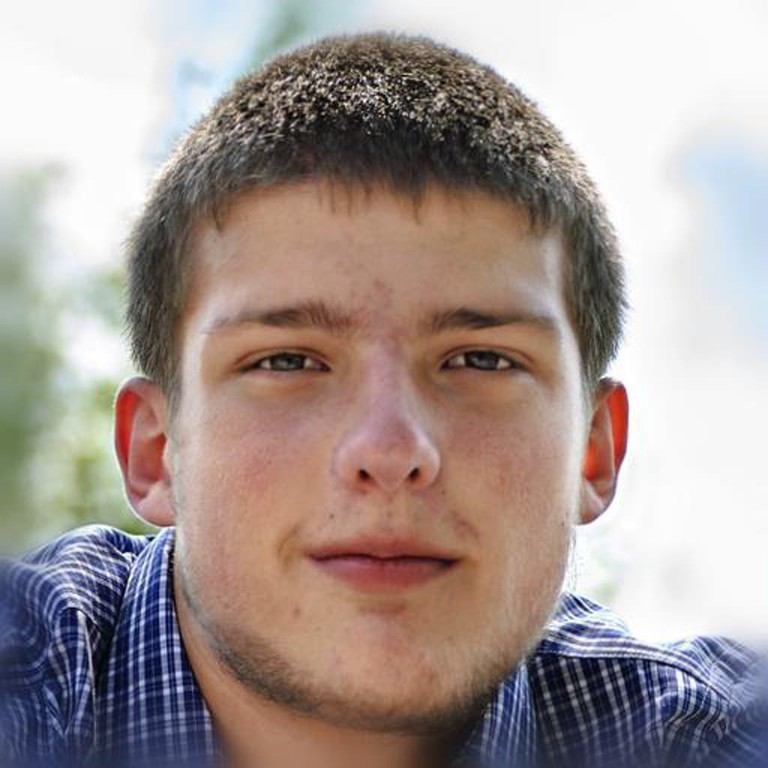} &
        \includegraphics[width=0.142\linewidth]{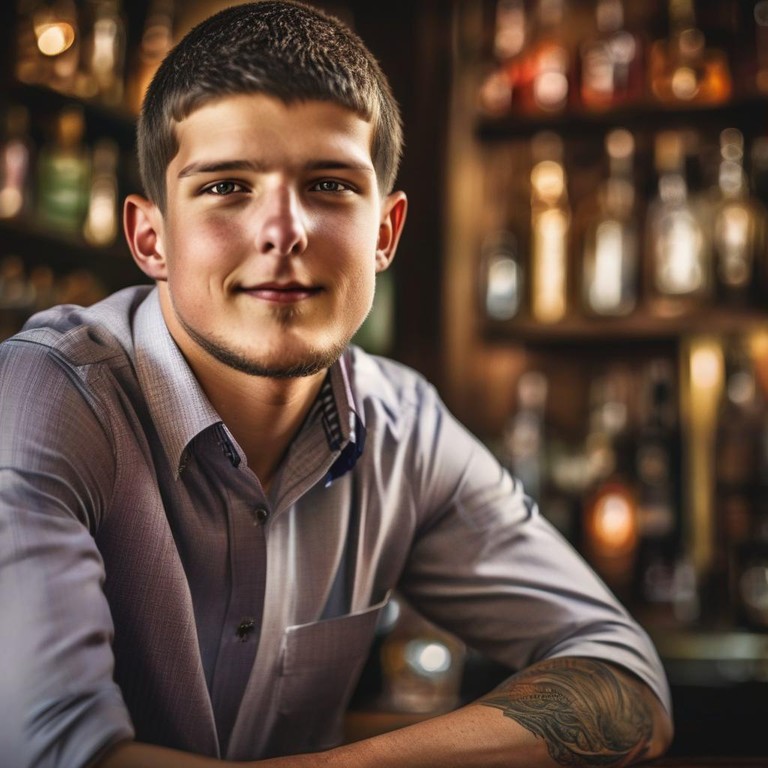} &
        \includegraphics[width=0.142\linewidth]{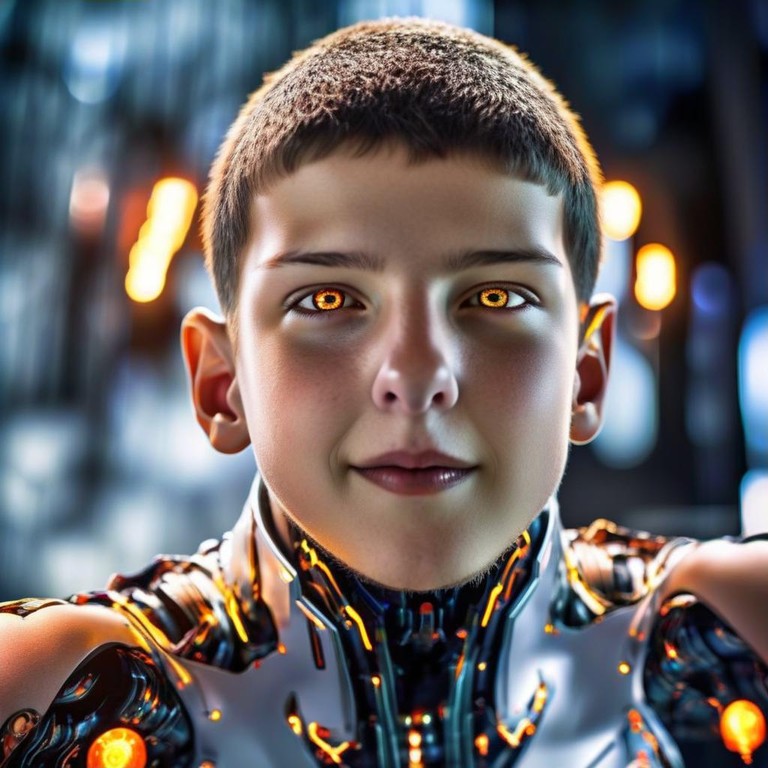} &
        \includegraphics[width=0.142\linewidth]{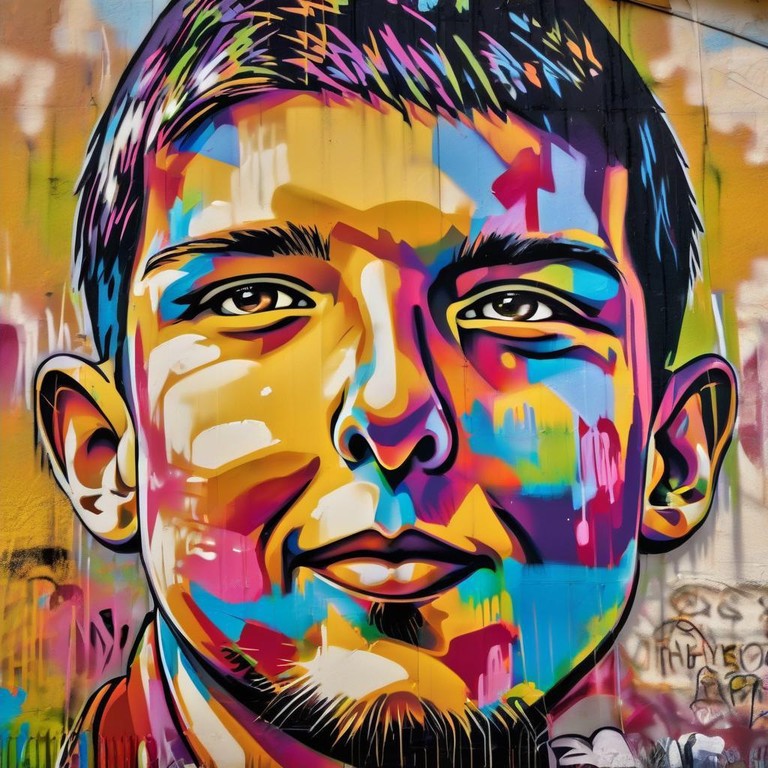} &
        \includegraphics[width=0.142\linewidth]{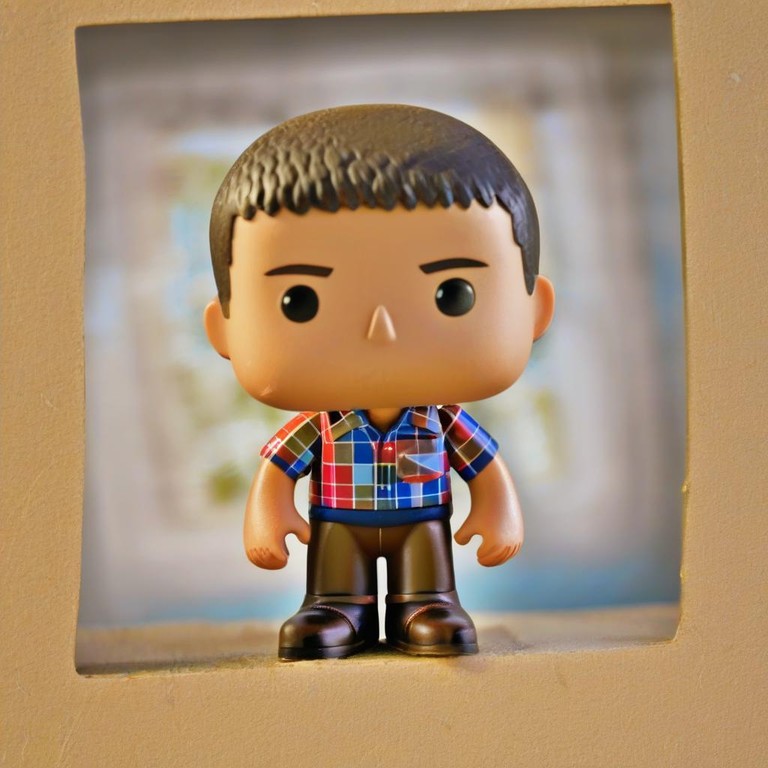} &
        \includegraphics[width=0.142\linewidth]{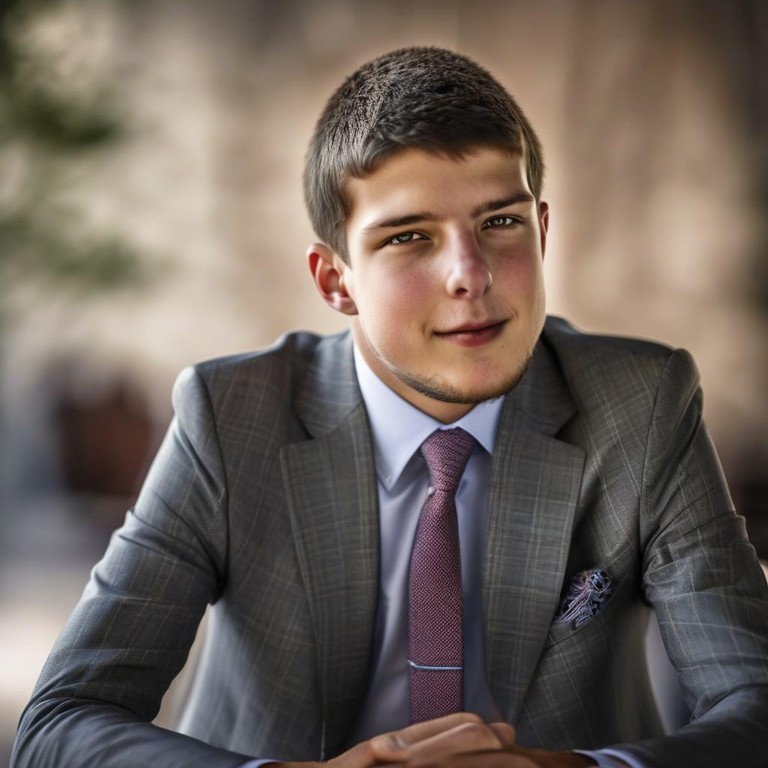} &
        \includegraphics[width=0.142\linewidth]{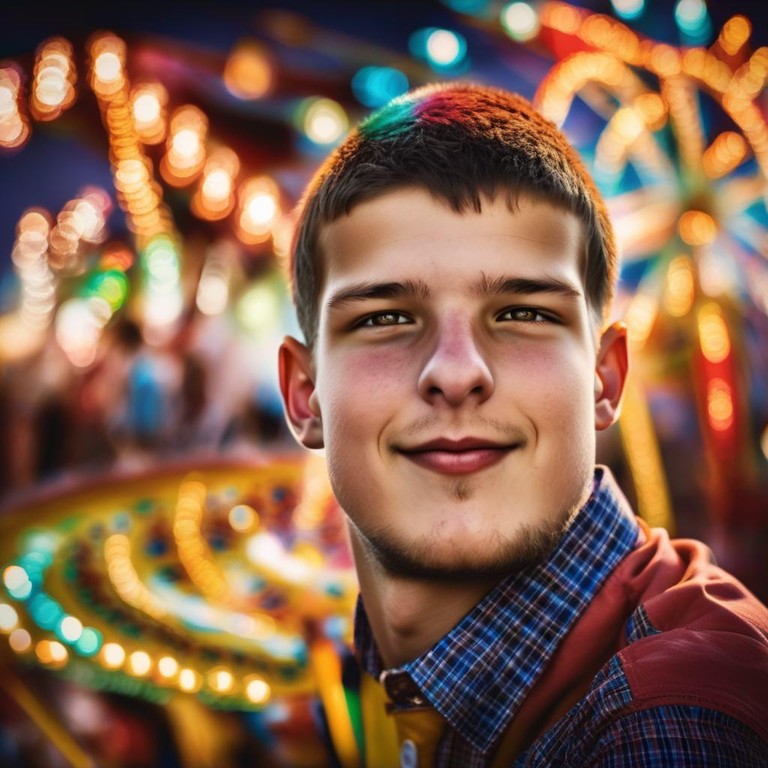} \\
        \includegraphics[width=0.142\linewidth]{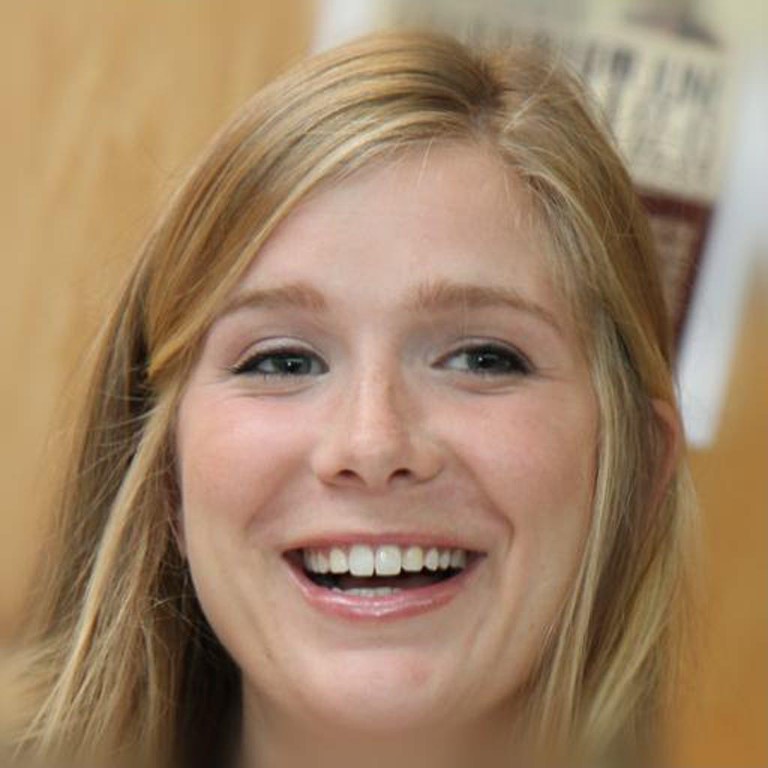} &
        \includegraphics[width=0.142\linewidth]{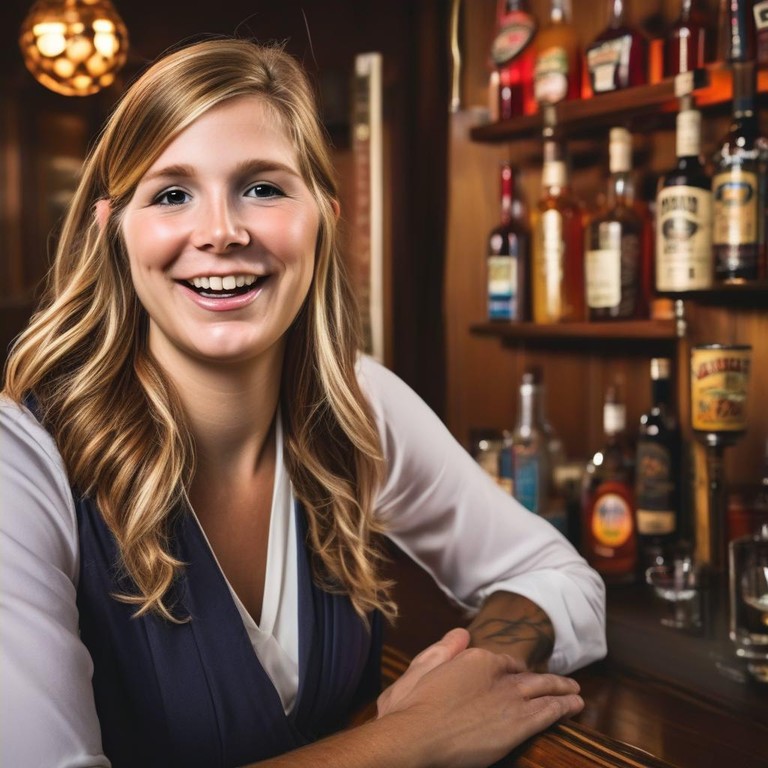} &
        \includegraphics[width=0.142\linewidth]{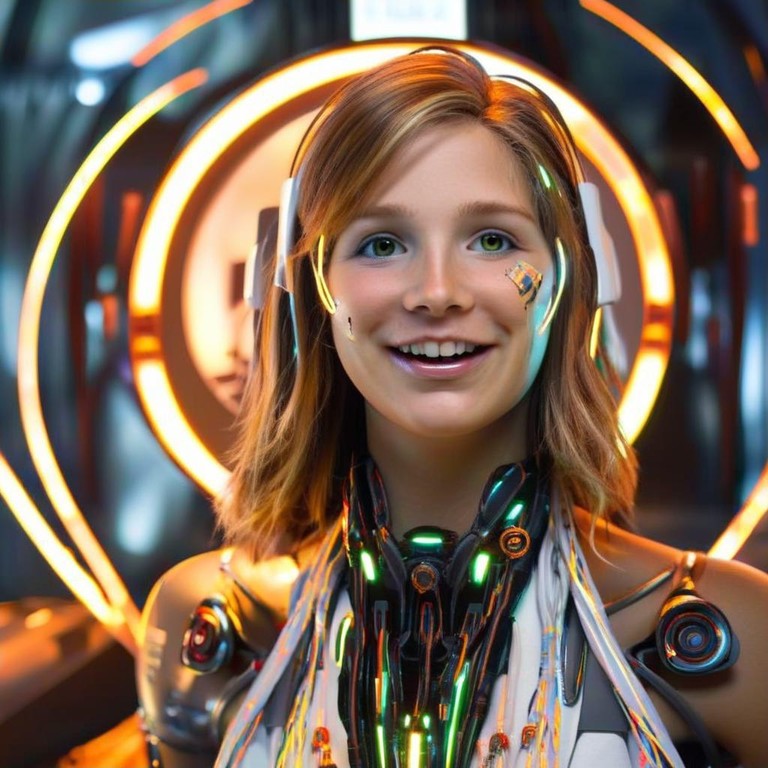} &
        \includegraphics[width=0.142\linewidth]{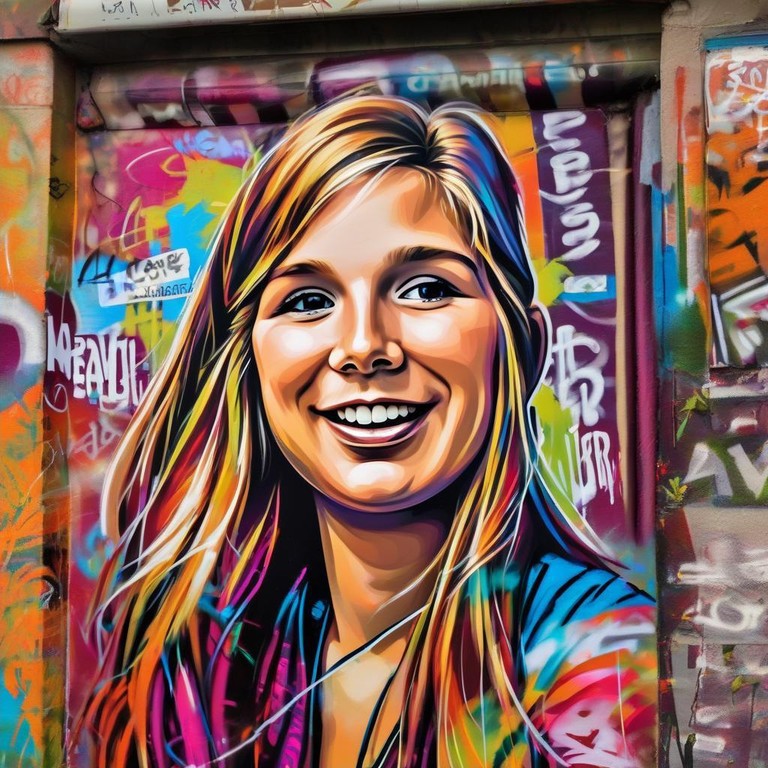} &
        \includegraphics[width=0.142\linewidth]{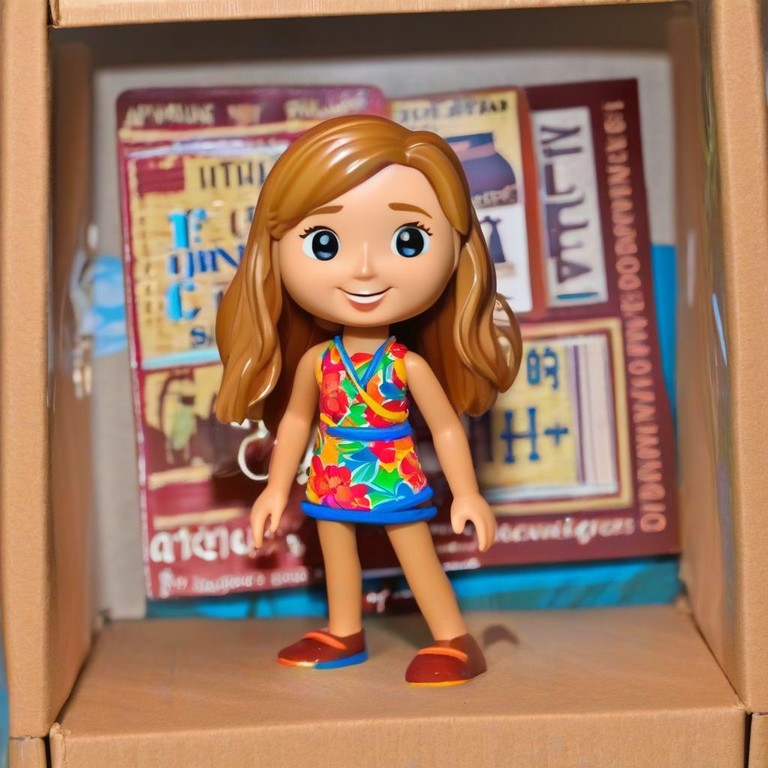} &
        \includegraphics[width=0.142\linewidth]{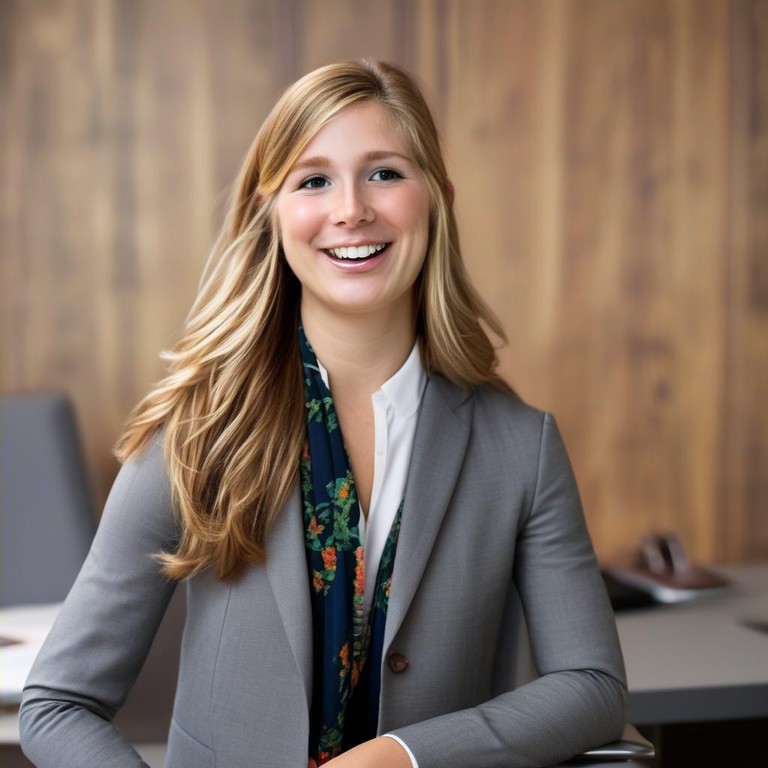} &
        \includegraphics[width=0.142\linewidth]{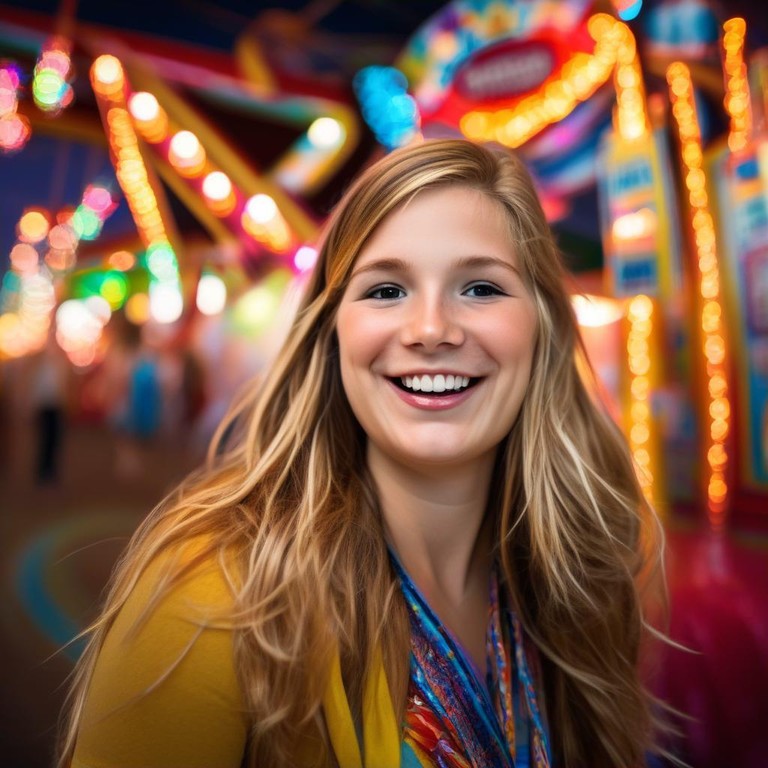} \\
        \includegraphics[width=0.142\linewidth]{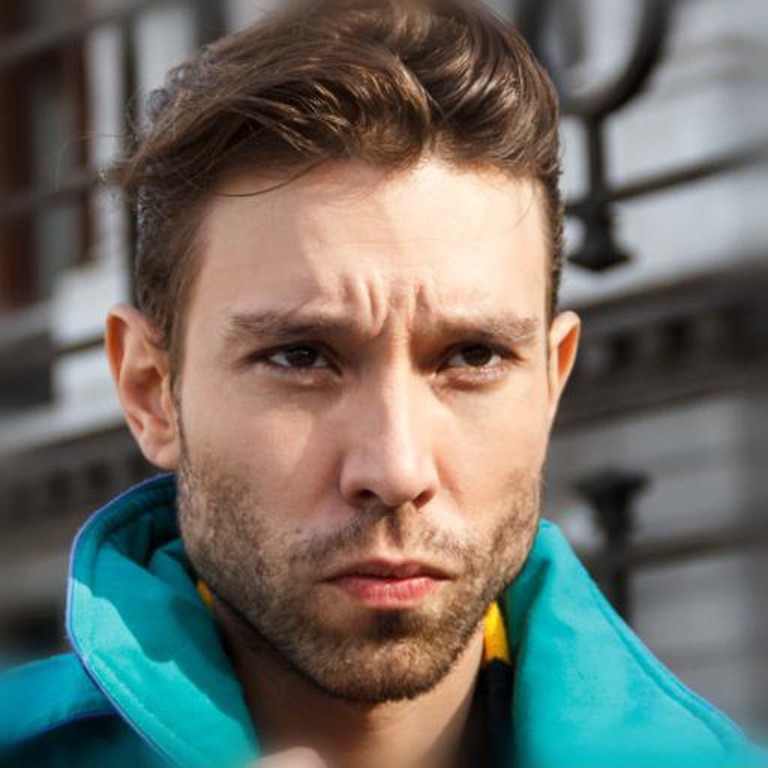} &
        \includegraphics[width=0.142\linewidth]{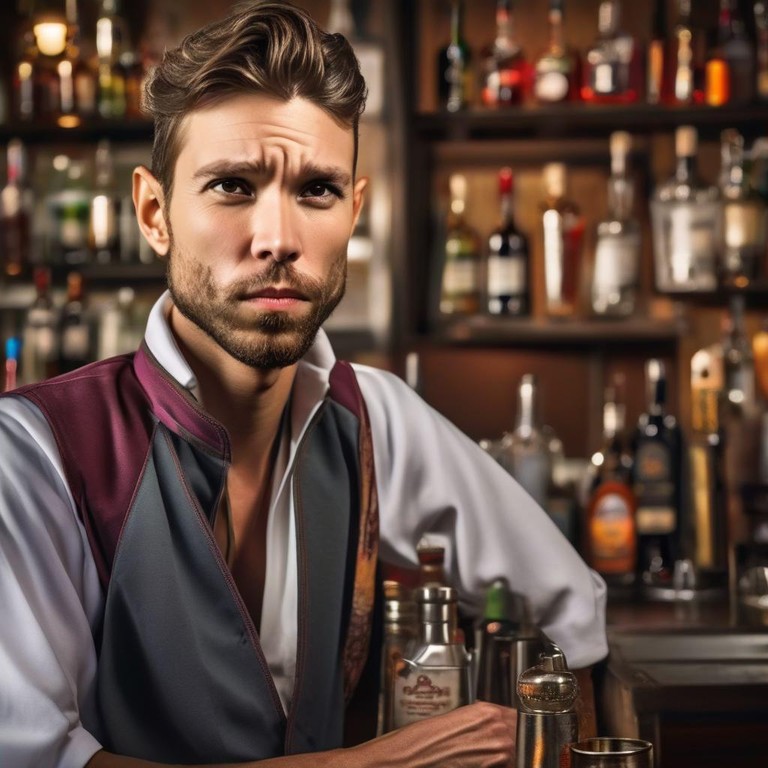} &
        \includegraphics[width=0.142\linewidth]{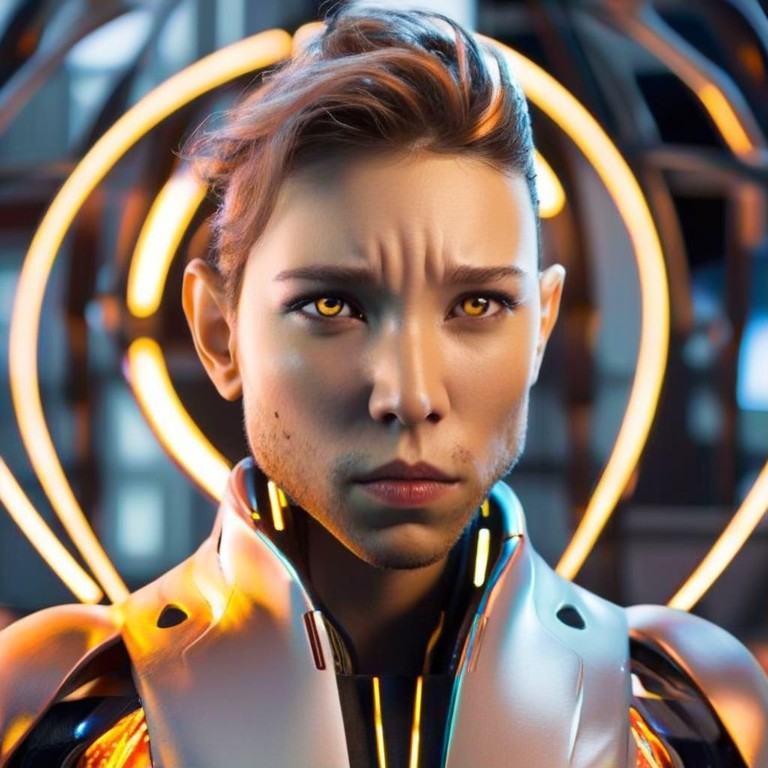} &
        \includegraphics[width=0.142\linewidth]{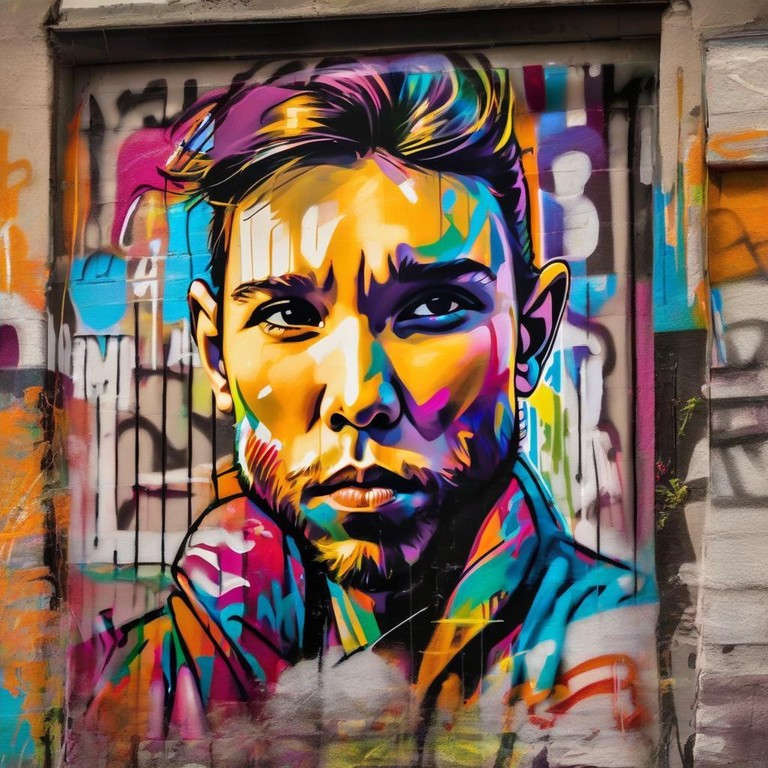} &
        \includegraphics[width=0.142\linewidth]{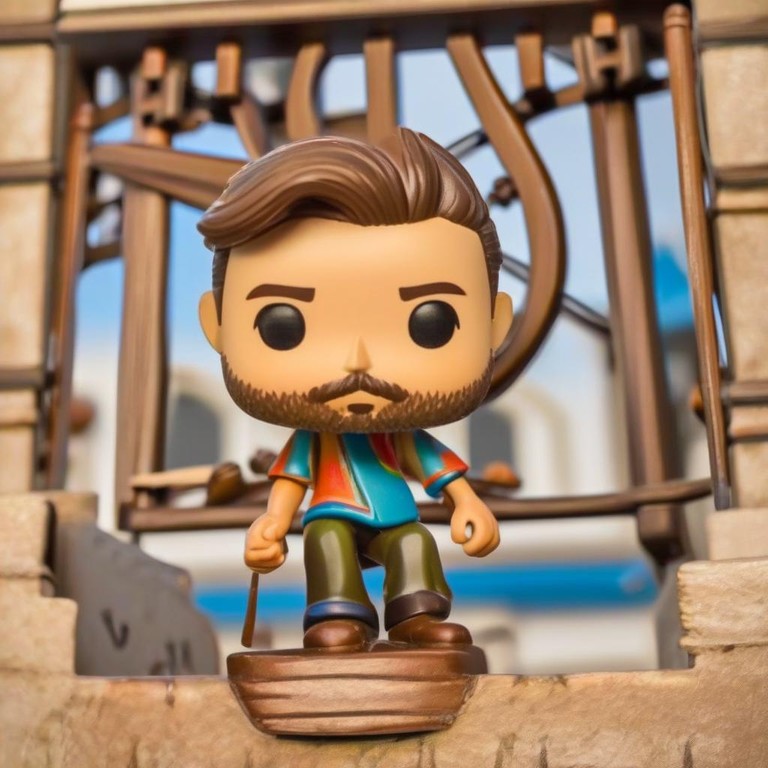} &
        \includegraphics[width=0.142\linewidth]{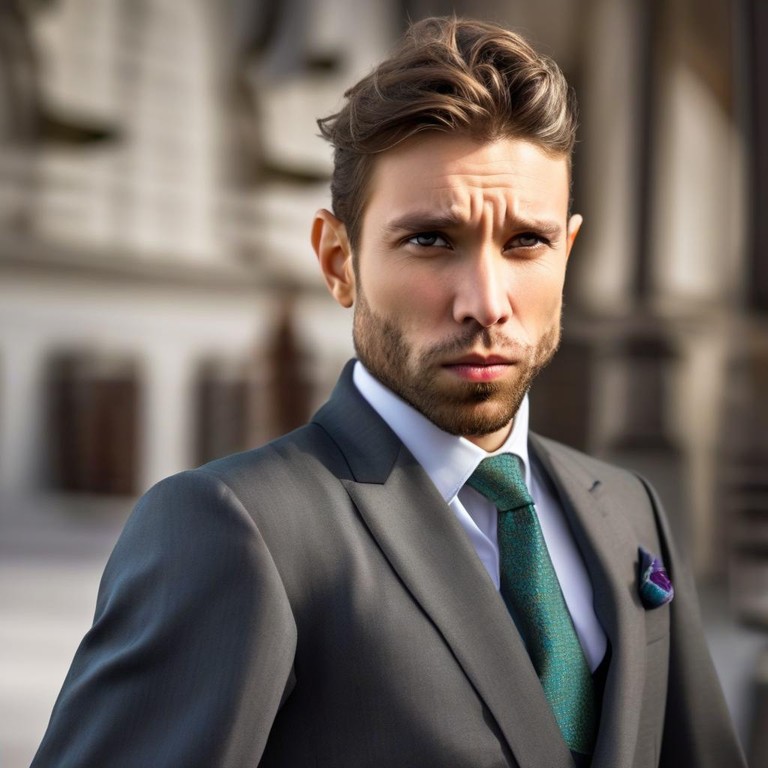} &
        \includegraphics[width=0.142\linewidth]{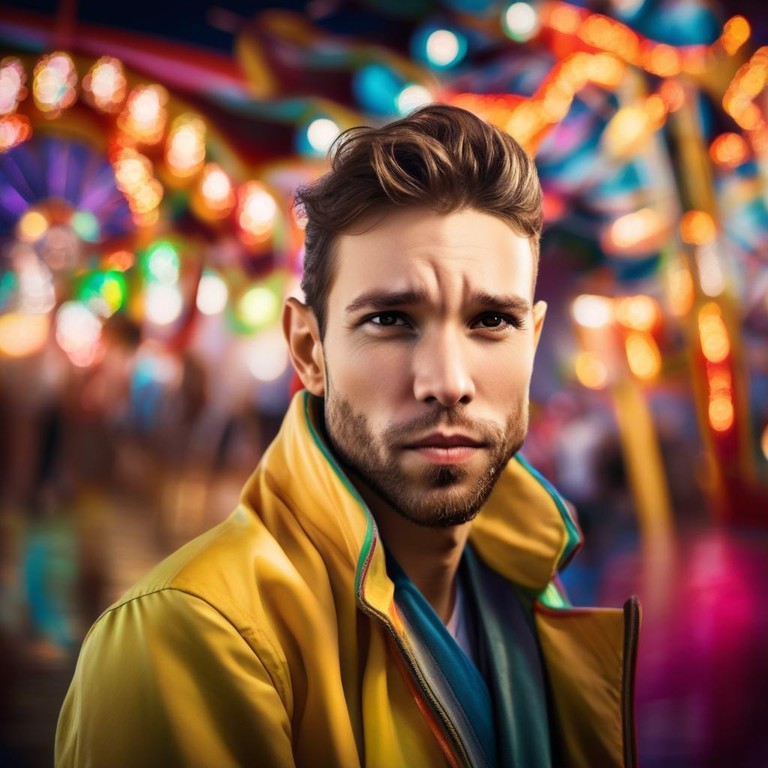} \\
        \includegraphics[width=0.142\linewidth]{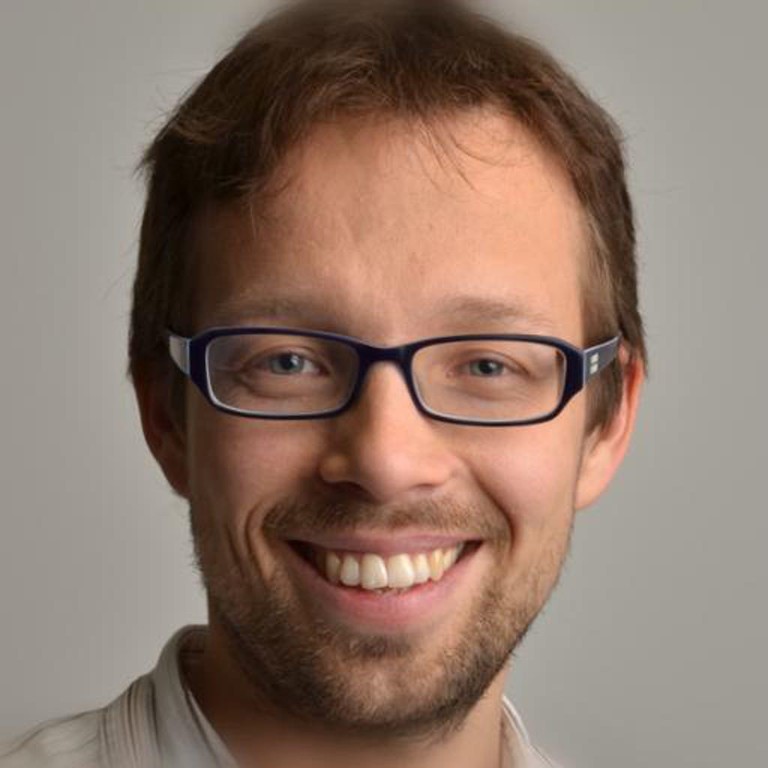} &
        \includegraphics[width=0.142\linewidth]{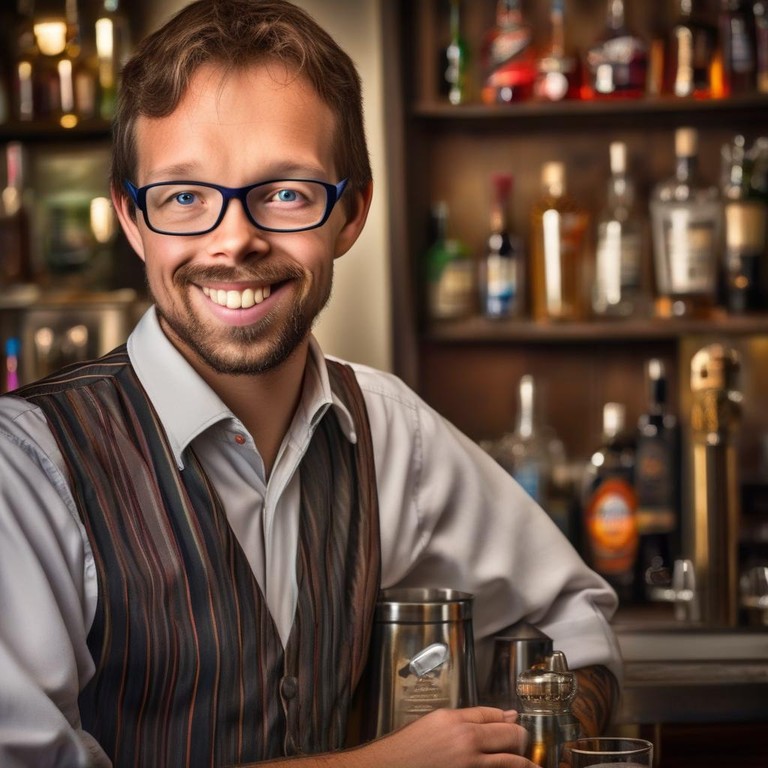} &
        \includegraphics[width=0.142\linewidth]{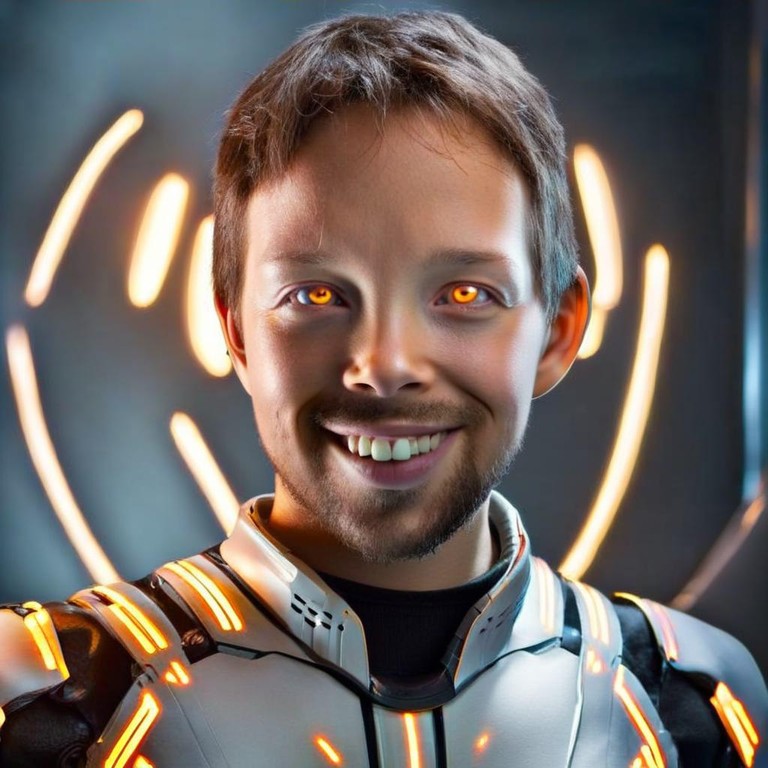} &
        \includegraphics[width=0.142\linewidth]{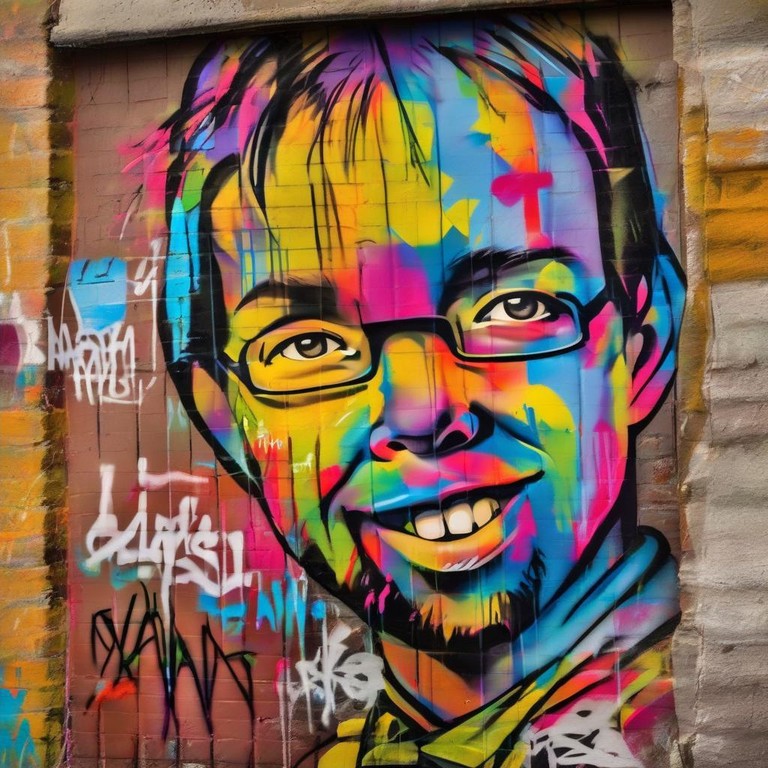} &
        \includegraphics[width=0.142\linewidth]{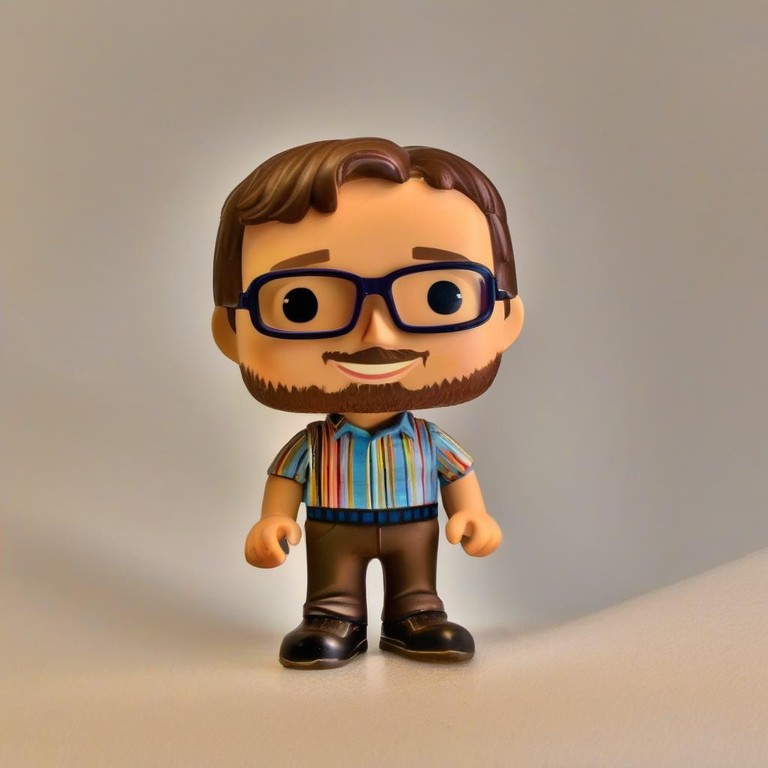} &
        \includegraphics[width=0.142\linewidth]{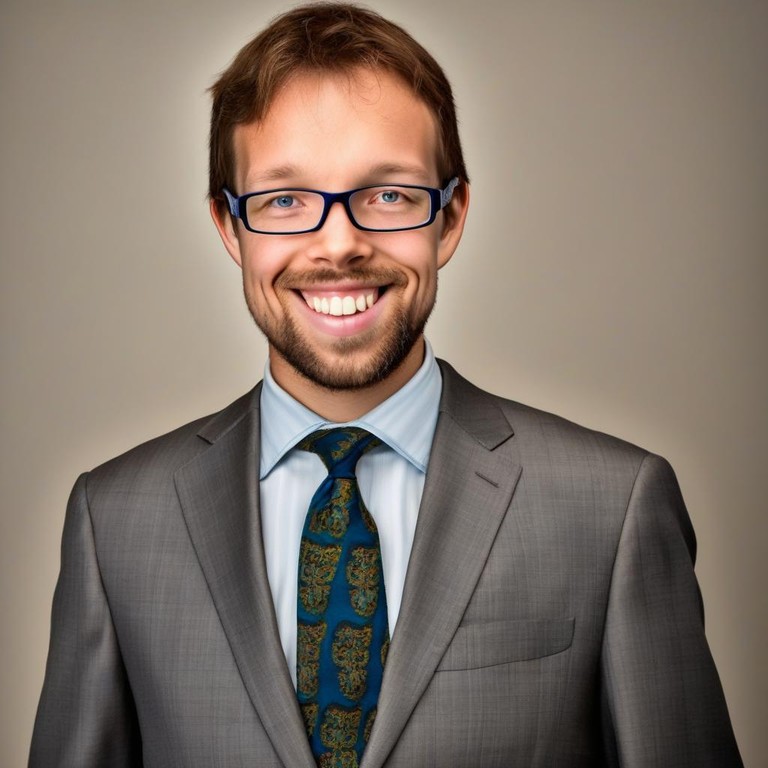} &
        \includegraphics[width=0.142\linewidth]{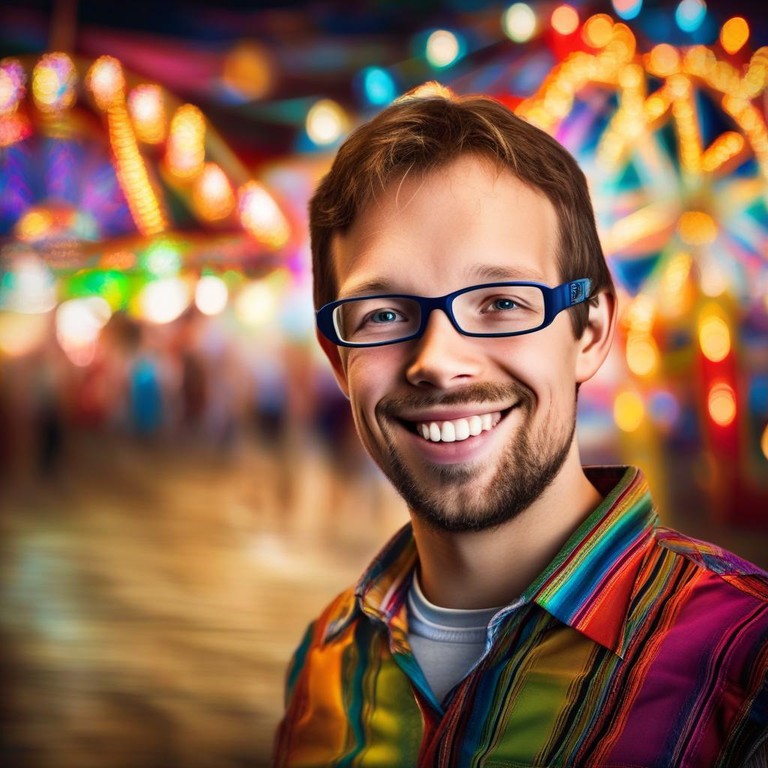} \\
        \includegraphics[width=0.142\linewidth]{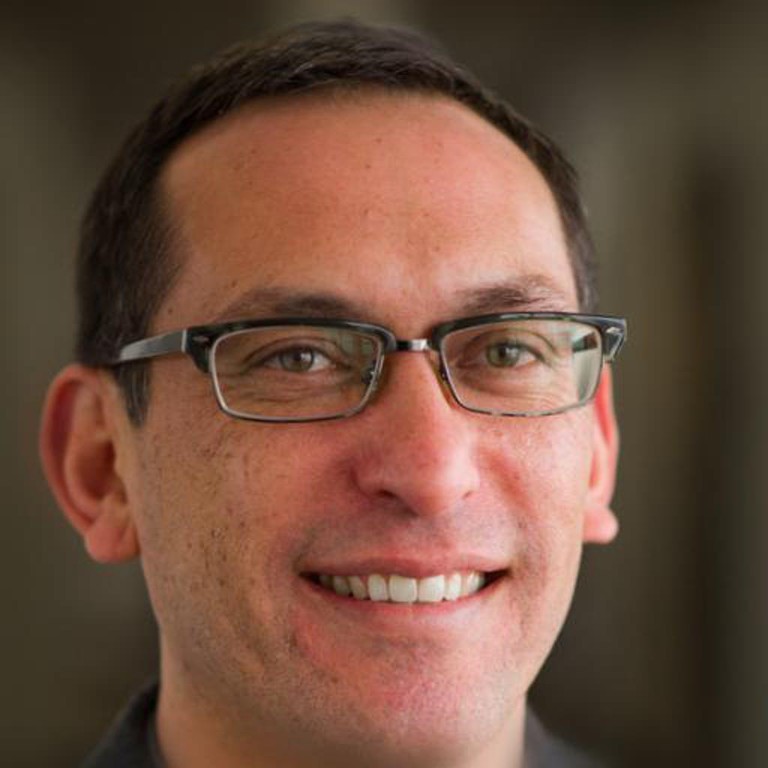} &
        \includegraphics[width=0.142\linewidth]{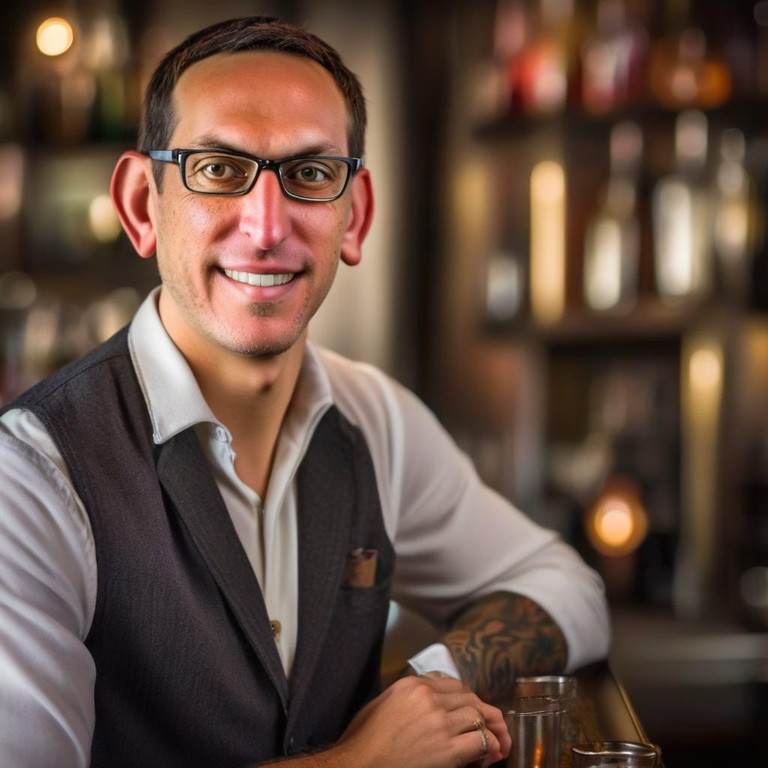} &
        \includegraphics[width=0.142\linewidth]{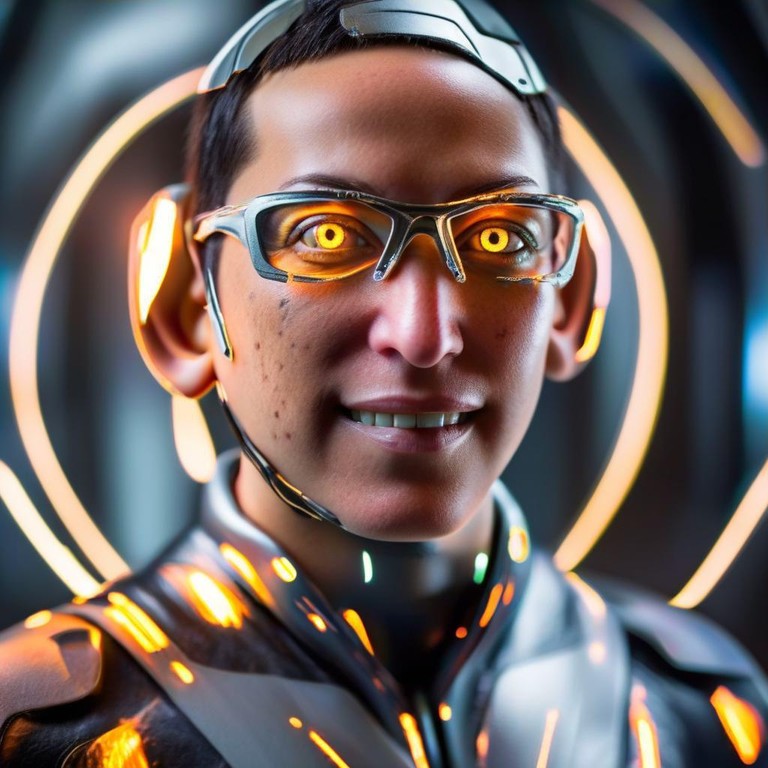} &
        \includegraphics[width=0.142\linewidth]{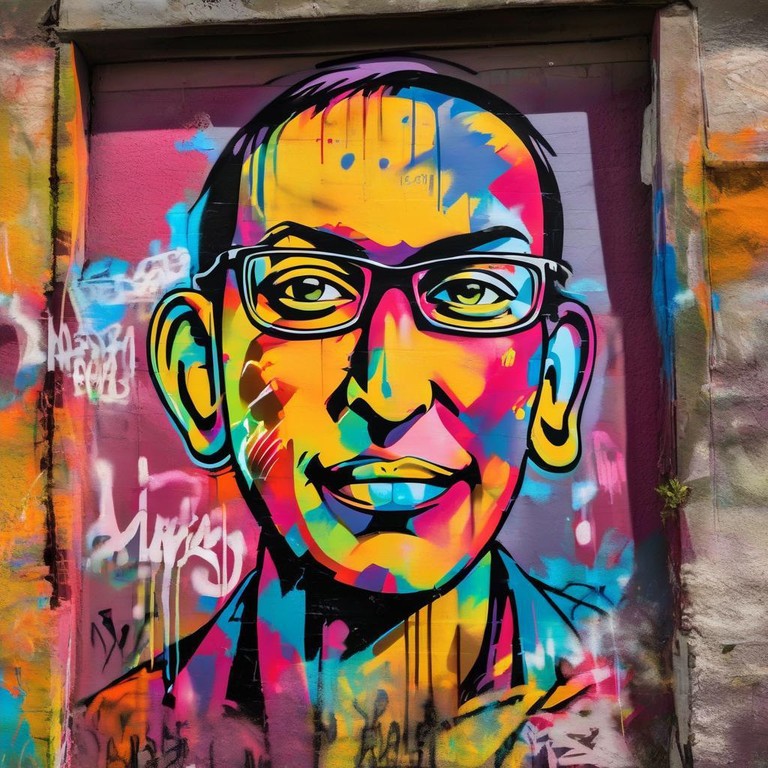} &
        \includegraphics[width=0.142\linewidth]{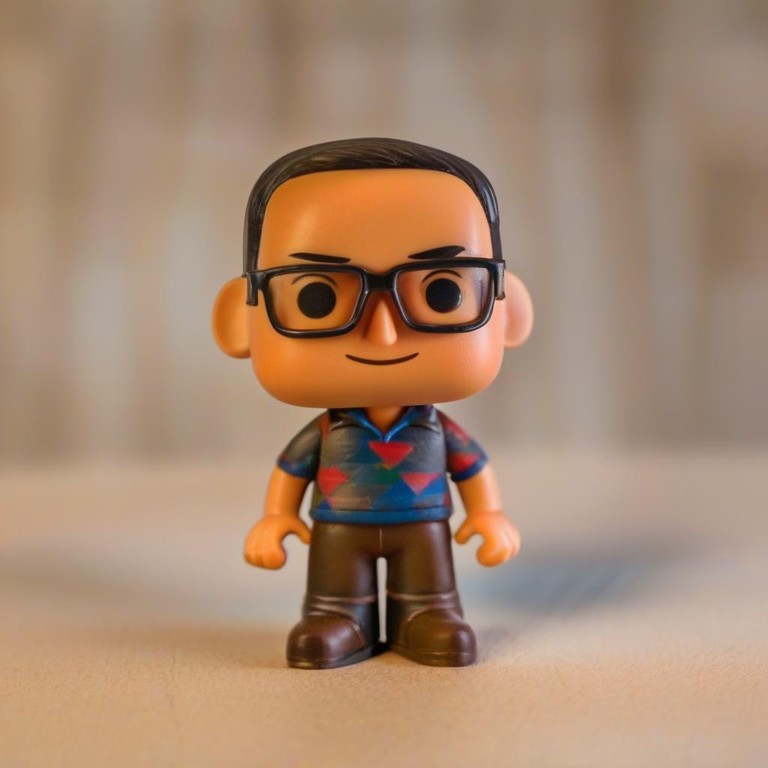} &
        \includegraphics[width=0.142\linewidth]{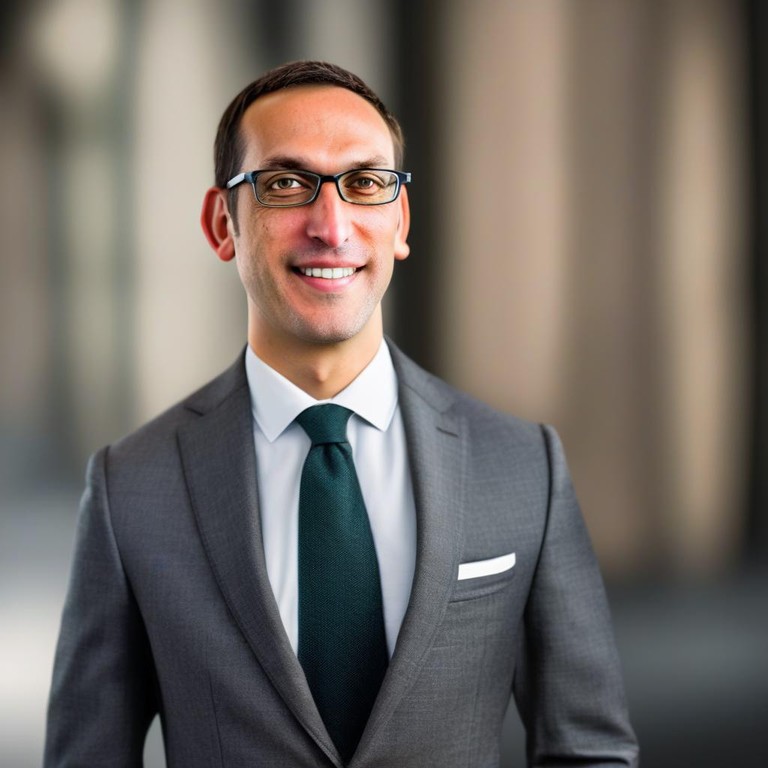} &
        \includegraphics[width=0.142\linewidth]{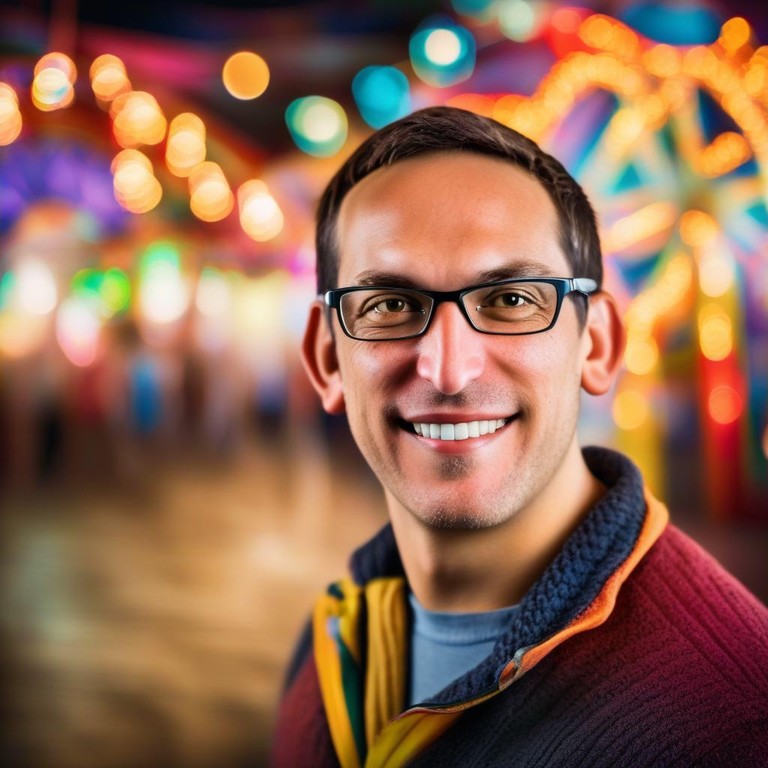} \\
        Input & ``Bar tender'' & ``Cyborg'' & ``Grafiiti'' & ``Pop figure'' & ``Wearing a suit'' & ``Carnival''
    \end{tabular}
    }
    \caption{
    Additional results on human faces. The initial noise is fixed across each column.
    }
    \label{fig:face-grid}
\end{figure*}

\begin{figure*}
    \centering
    \setlength{\tabcolsep}{1pt}
    \scriptsize{
    \begin{tabular}{ccccccc}
        \includegraphics[width=0.142\linewidth]{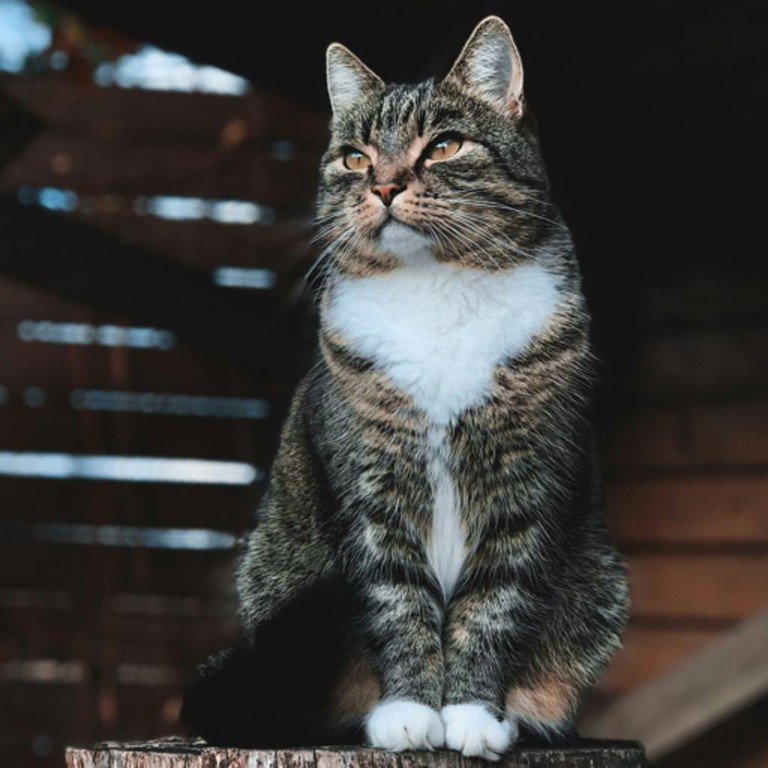} &
        \includegraphics[width=0.142\linewidth]{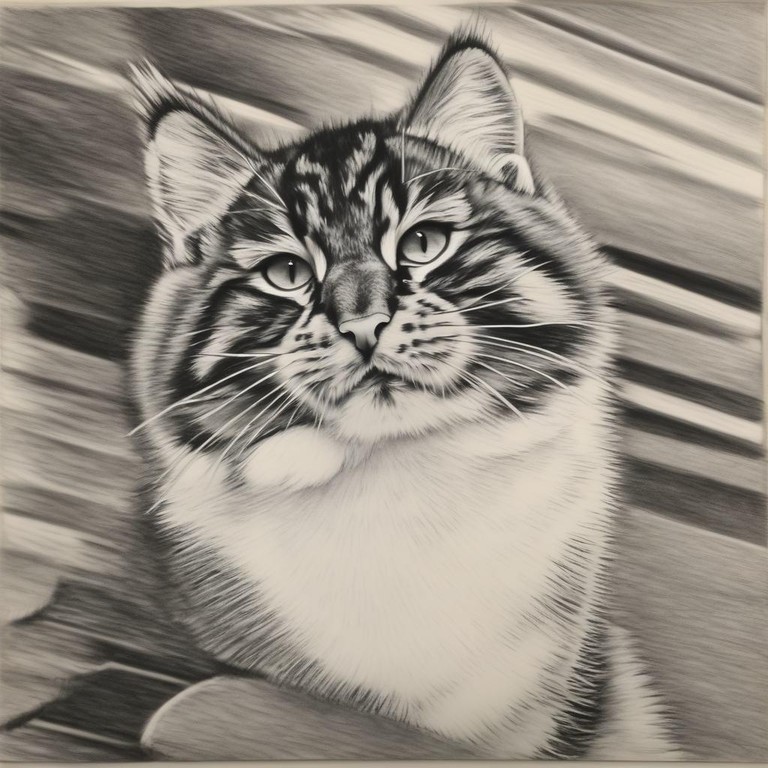} &
        \includegraphics[width=0.142\linewidth]{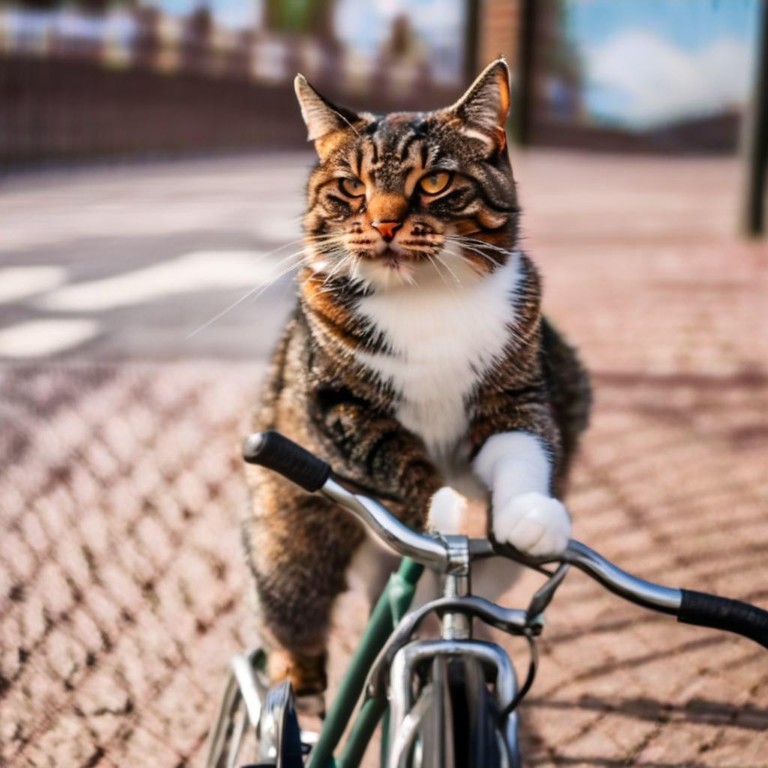} &
        \includegraphics[width=0.142\linewidth]{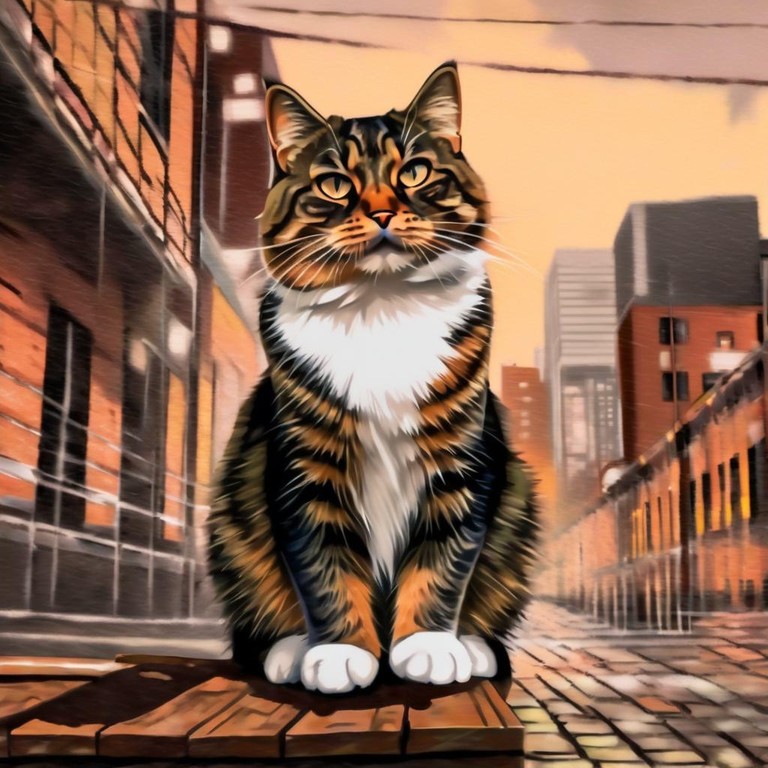} &
        \includegraphics[width=0.142\linewidth]{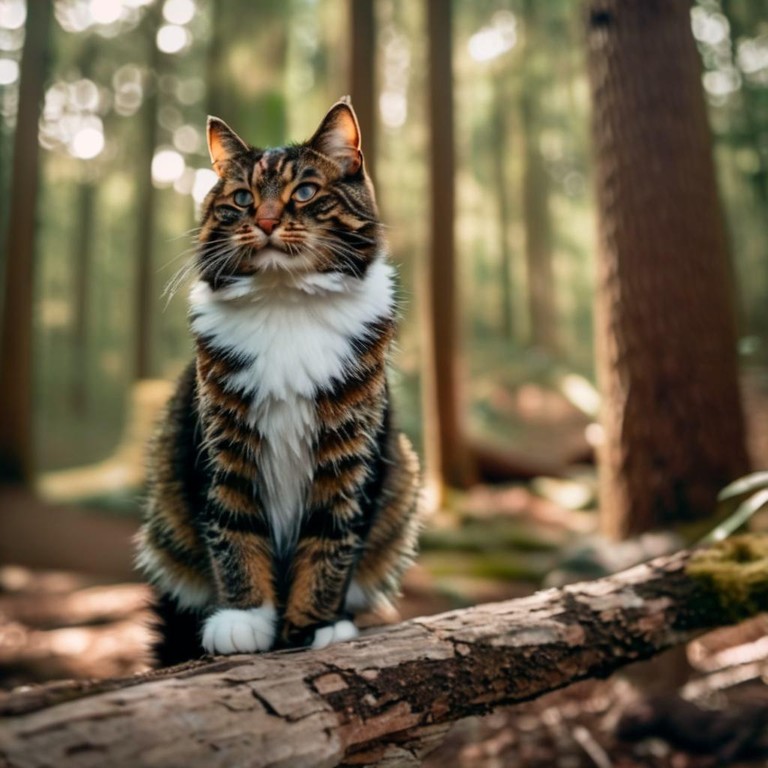} &
        \includegraphics[width=0.142\linewidth]{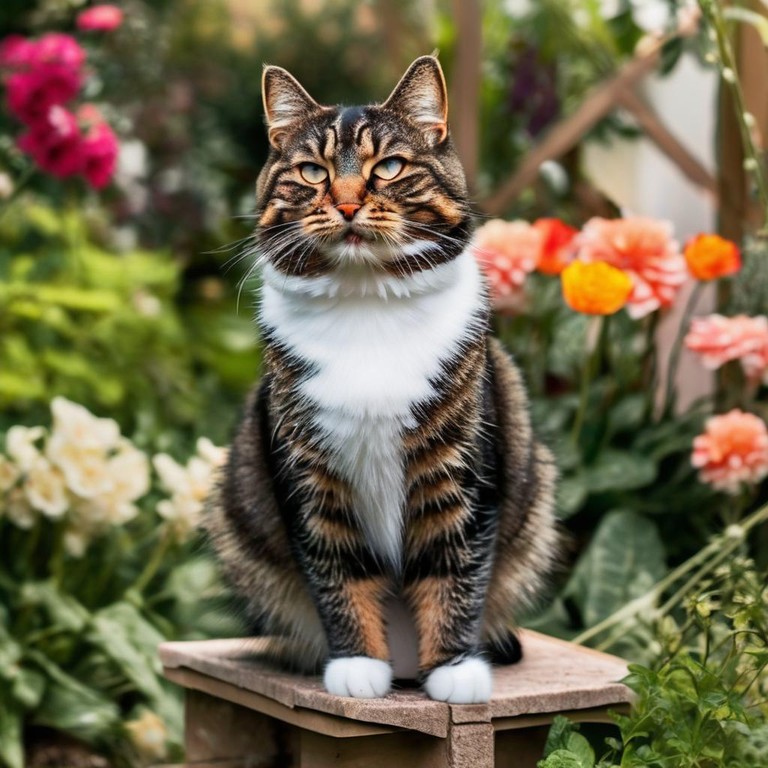} &
        \includegraphics[width=0.142\linewidth]{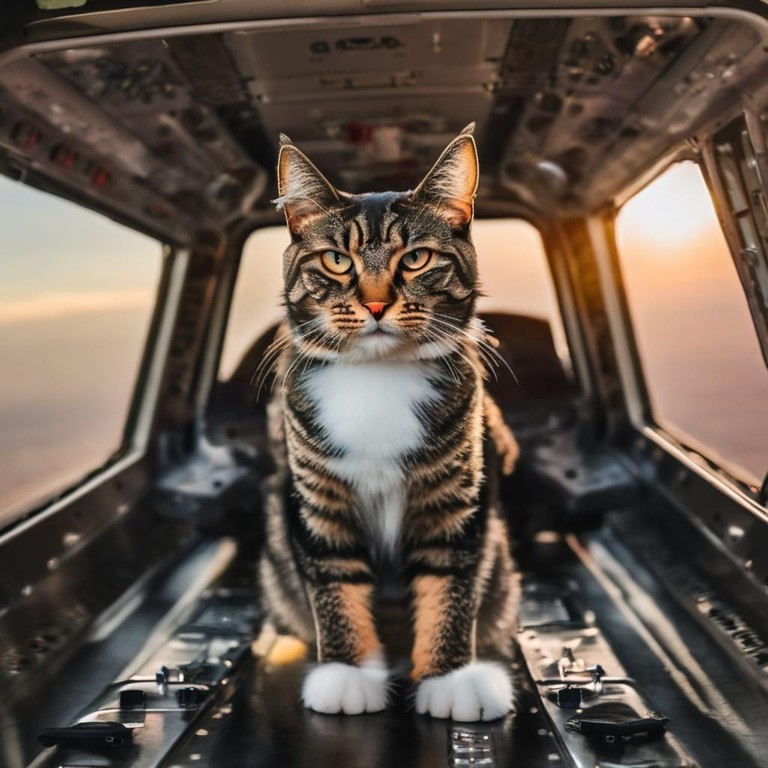} \\
        \includegraphics[width=0.142\linewidth]{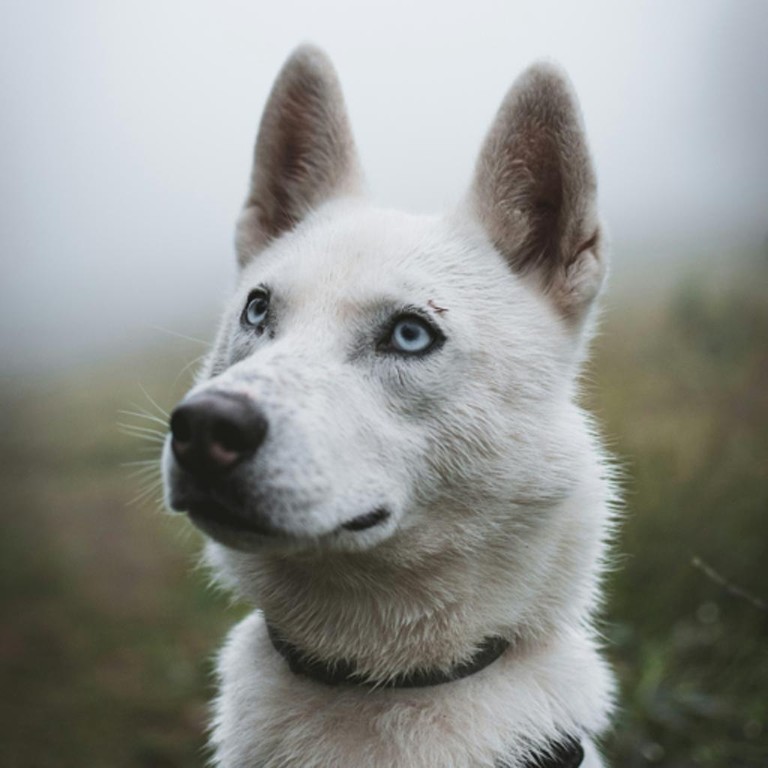} &
        \includegraphics[width=0.142\linewidth]{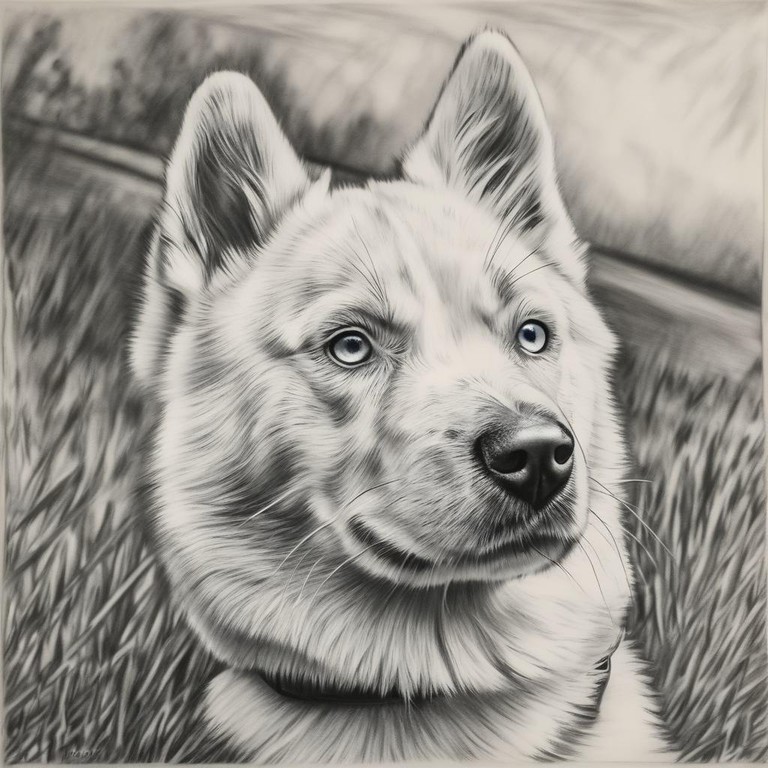} &
        \includegraphics[width=0.142\linewidth]{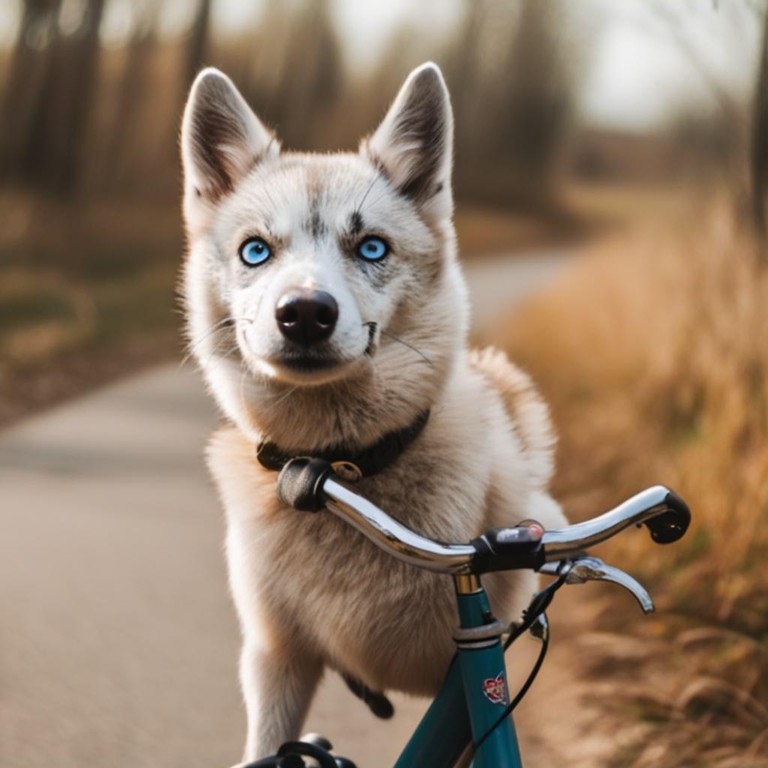} &
        \includegraphics[width=0.142\linewidth]{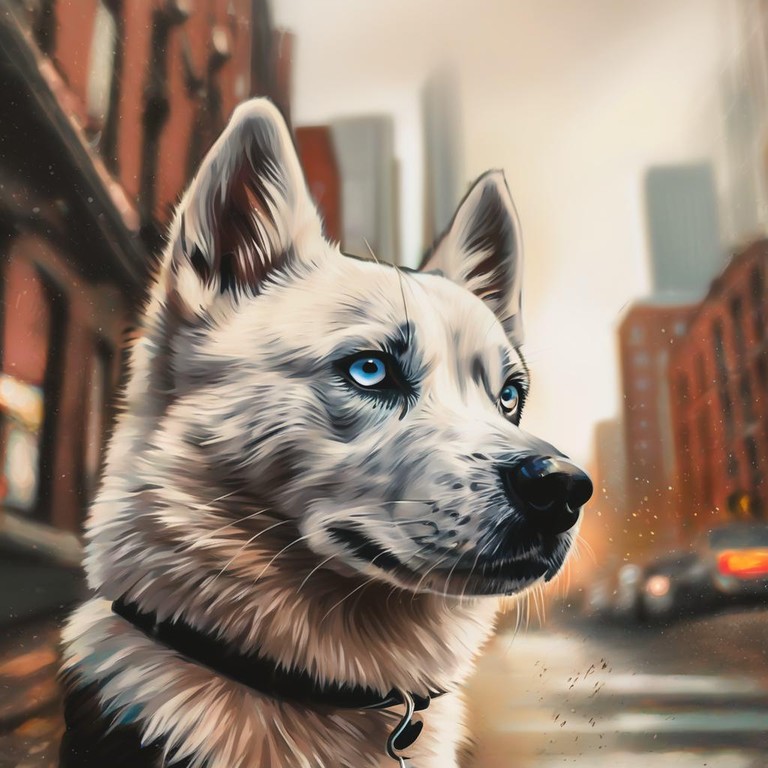} &
        \includegraphics[width=0.142\linewidth]{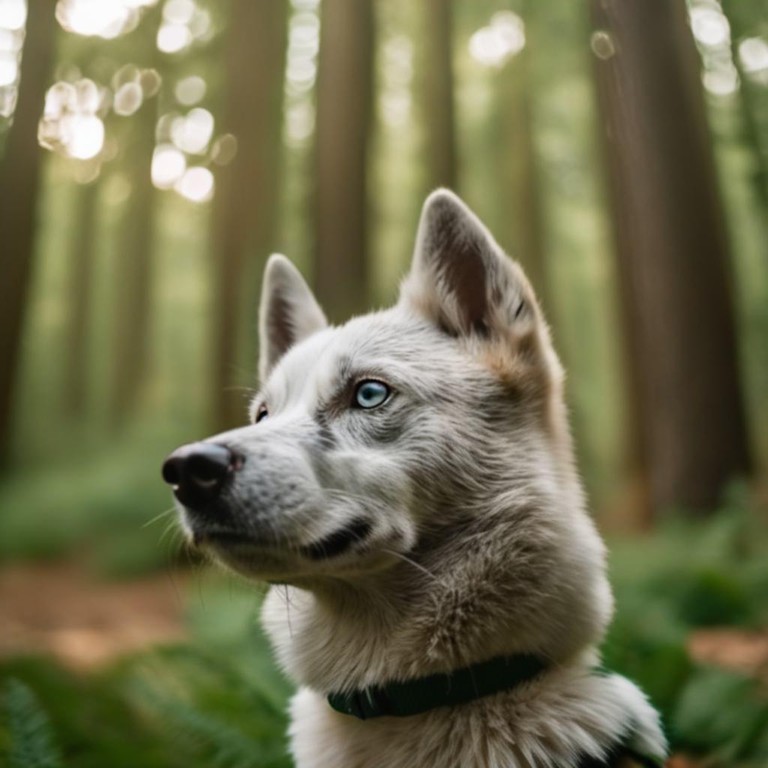} &
        \includegraphics[width=0.142\linewidth]{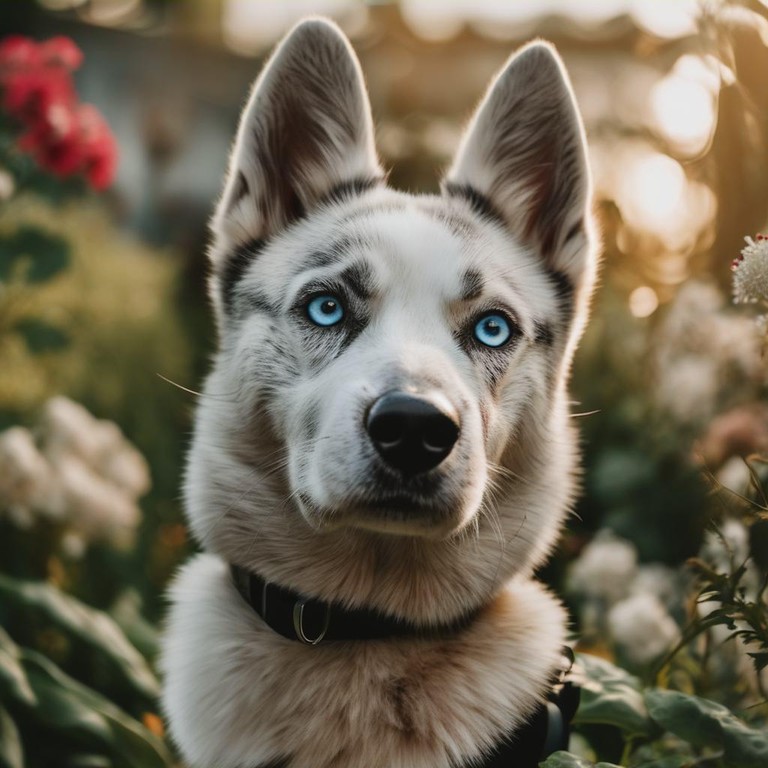} &
        \includegraphics[width=0.142\linewidth]{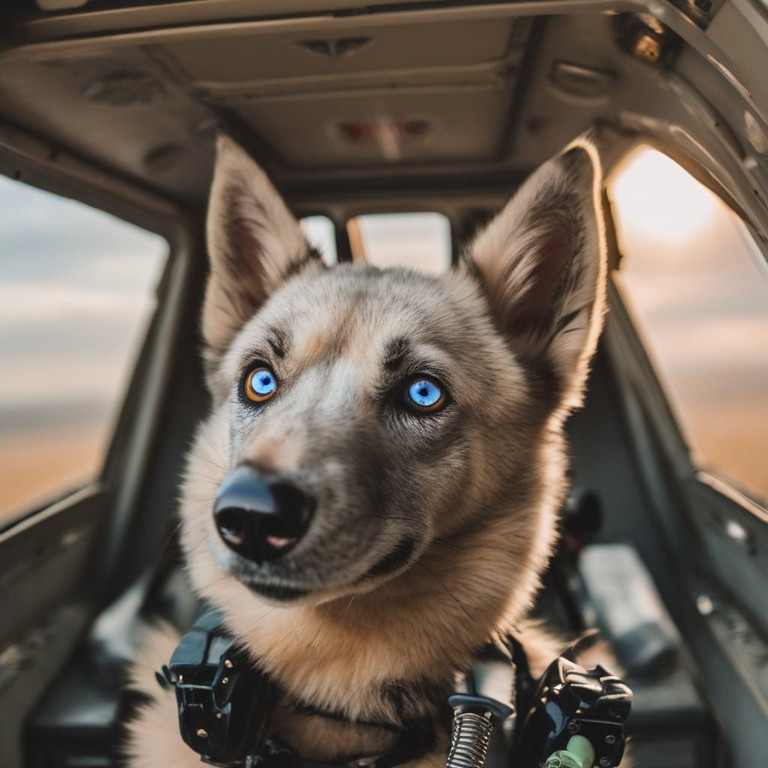} \\
        \includegraphics[width=0.142\linewidth]{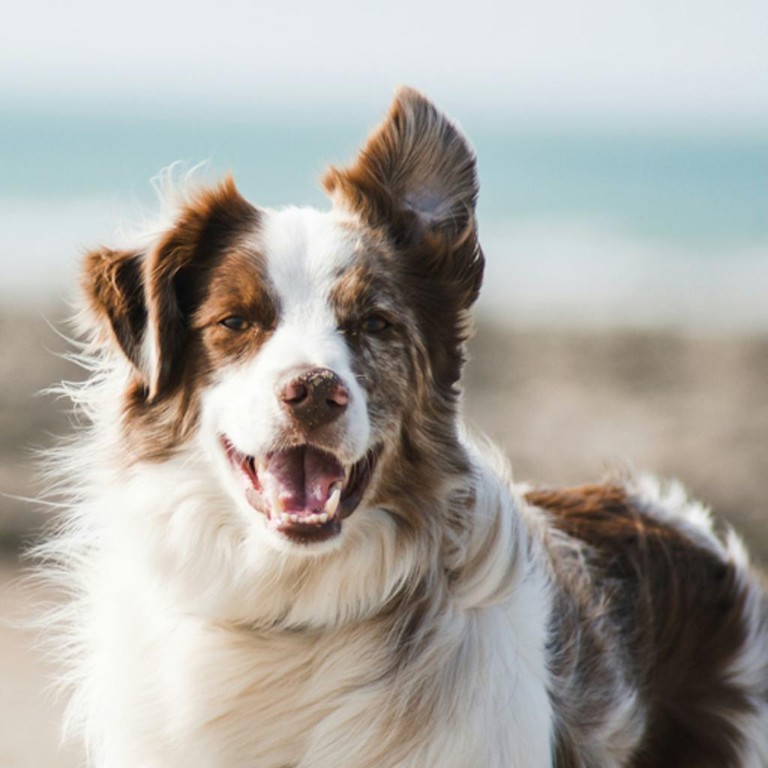} &
        \includegraphics[width=0.142\linewidth]{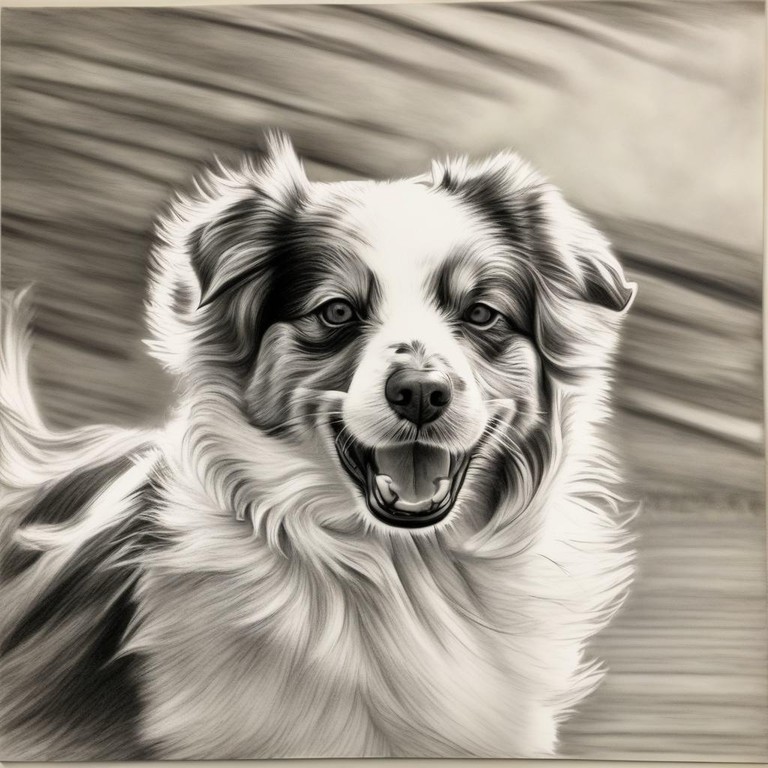} &
        \includegraphics[width=0.142\linewidth]{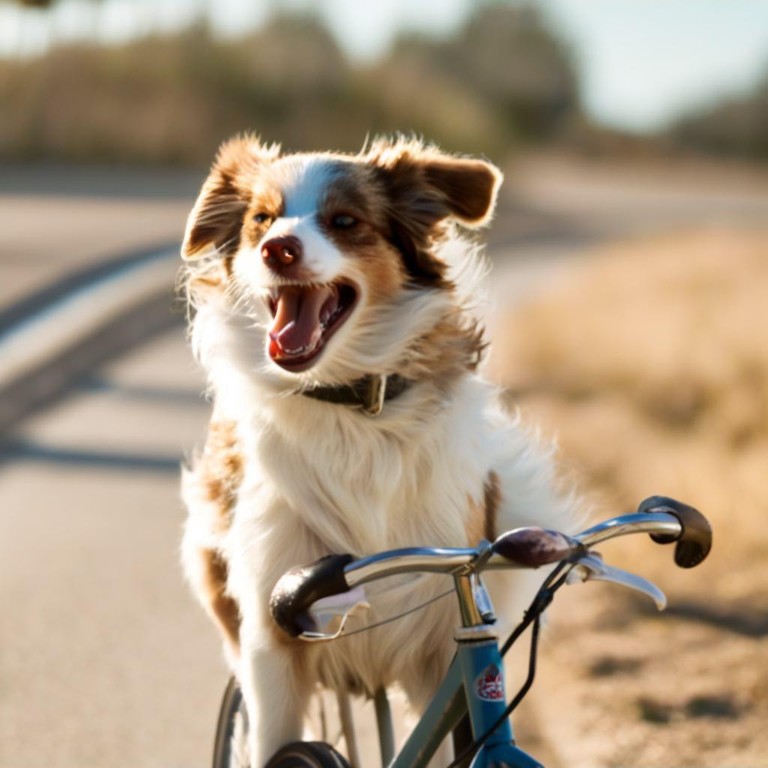} &
        \includegraphics[width=0.142\linewidth]{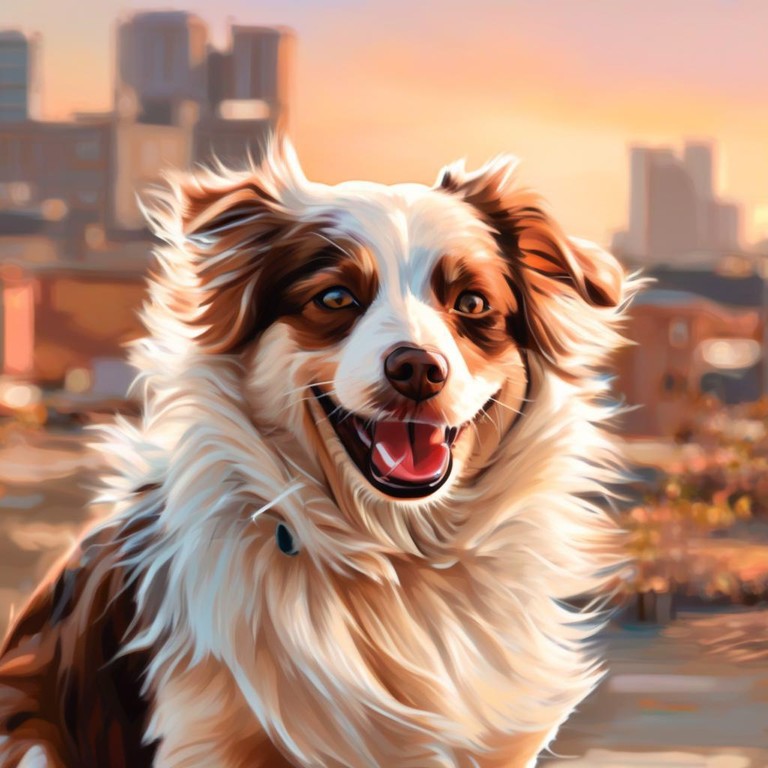} &
        \includegraphics[width=0.142\linewidth]{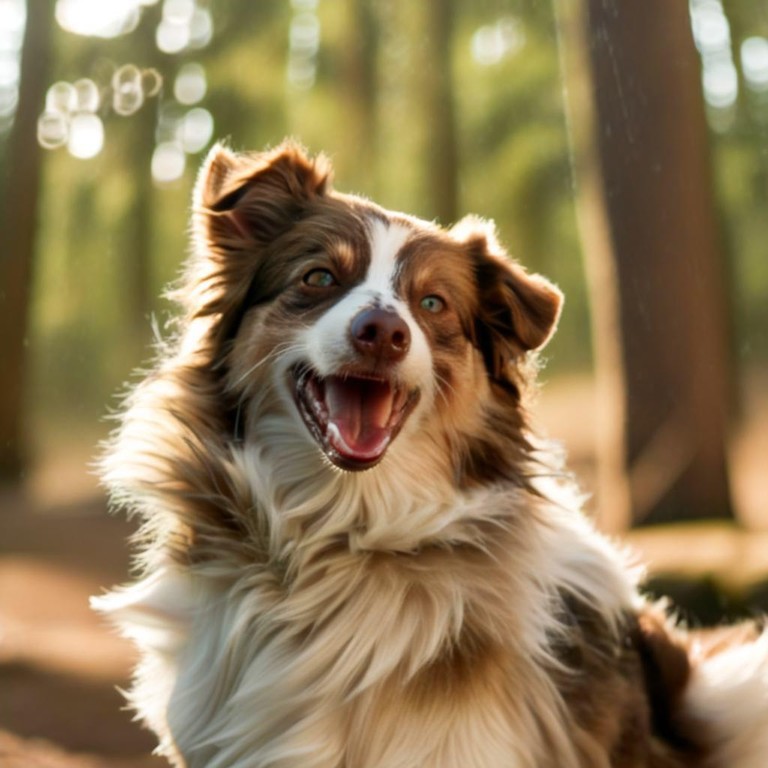} &
        \includegraphics[width=0.142\linewidth]{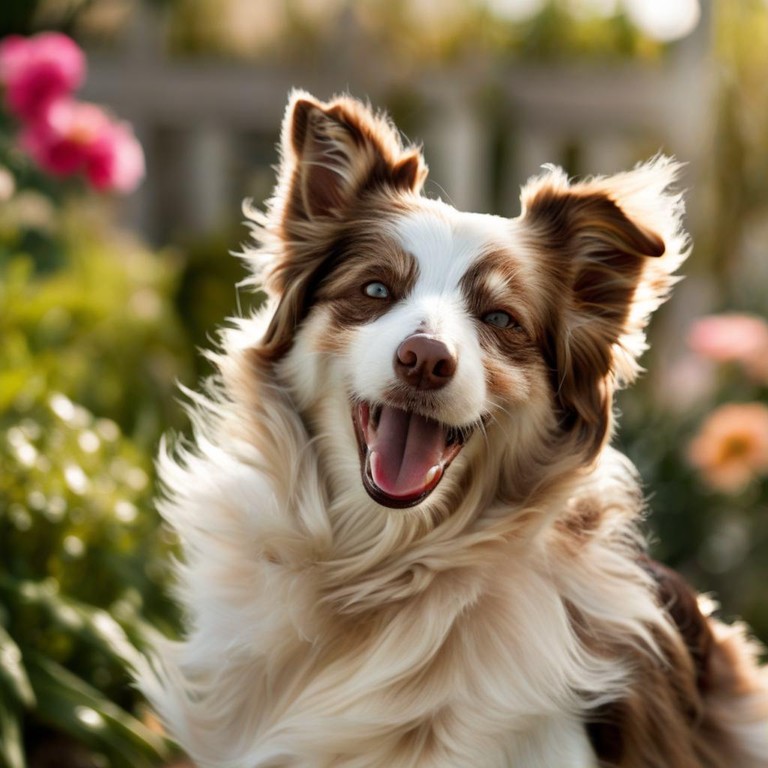} &
        \includegraphics[width=0.142\linewidth]{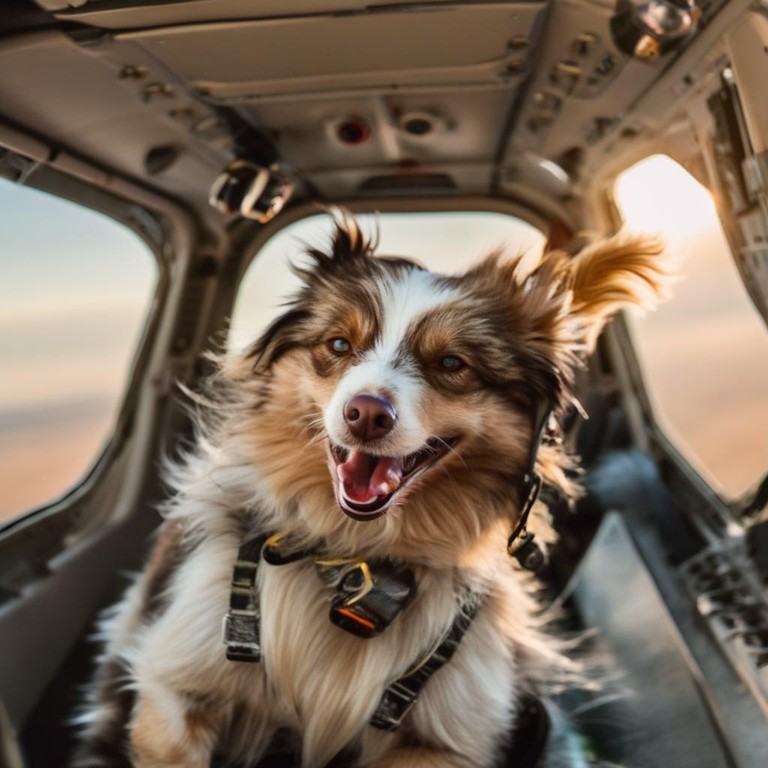} \\
        \includegraphics[width=0.142\linewidth]{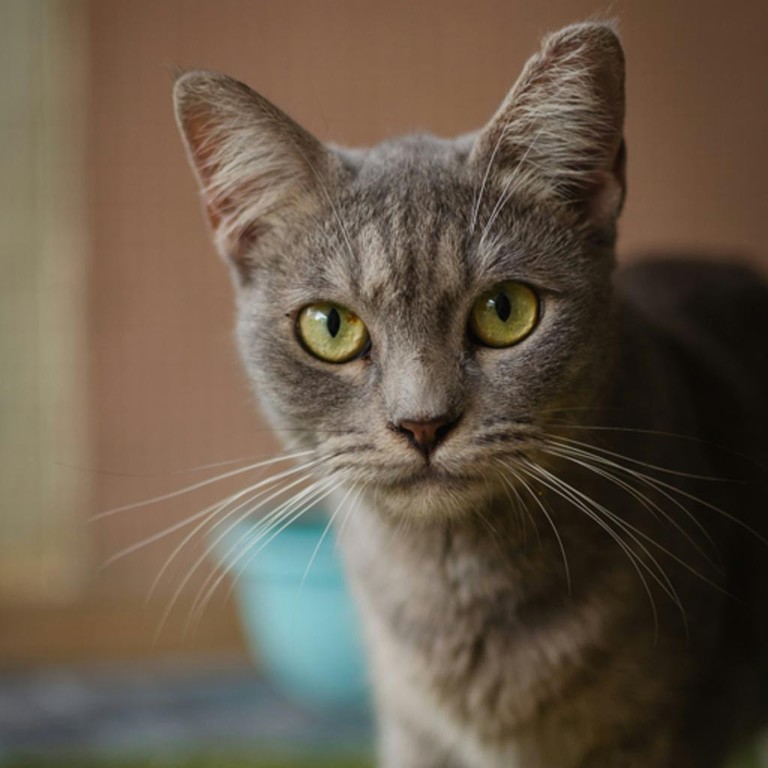} &
        \includegraphics[width=0.142\linewidth]{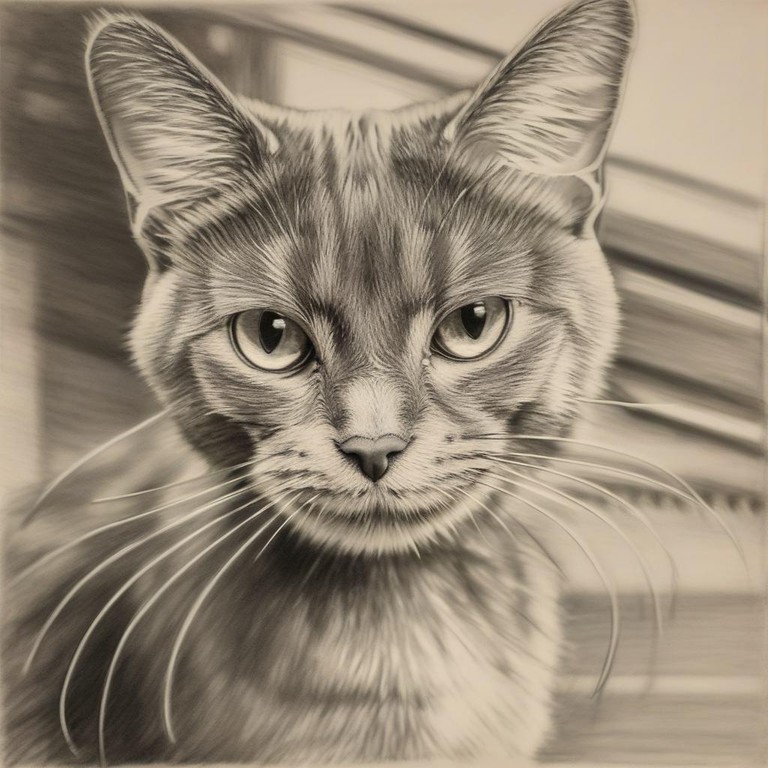} &
        \includegraphics[width=0.142\linewidth]{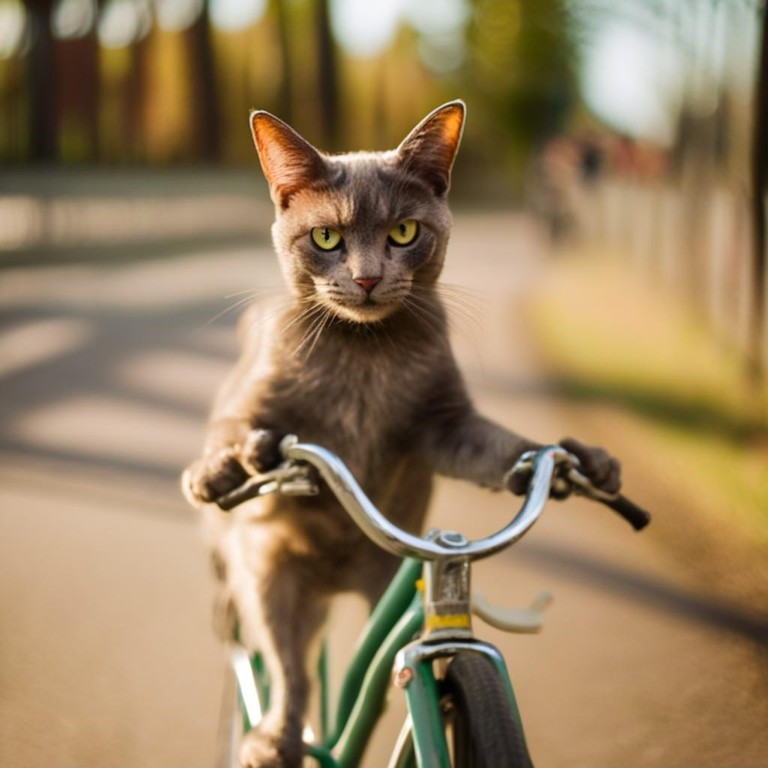} &
        \includegraphics[width=0.142\linewidth]{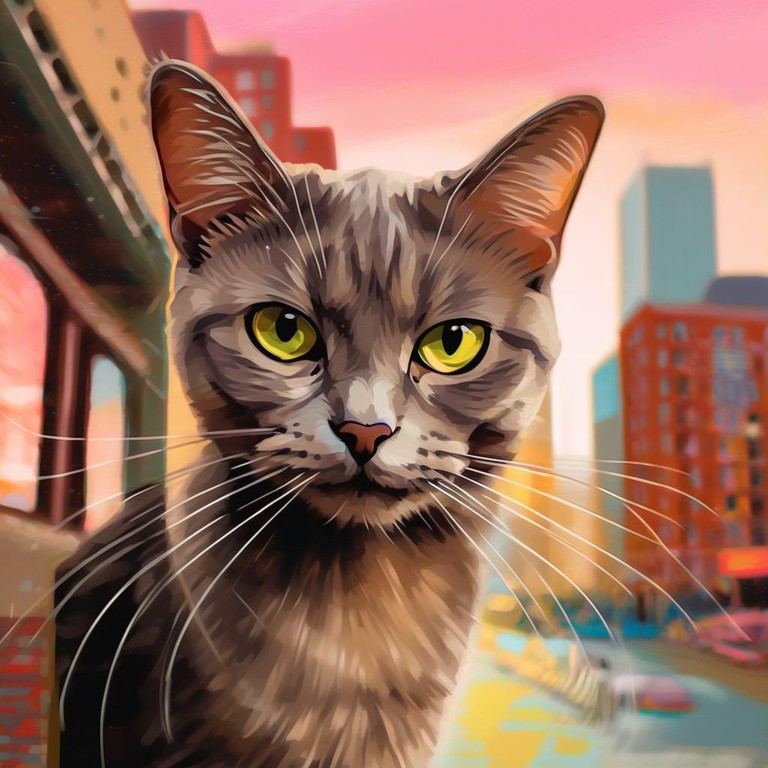} &
        \includegraphics[width=0.142\linewidth]{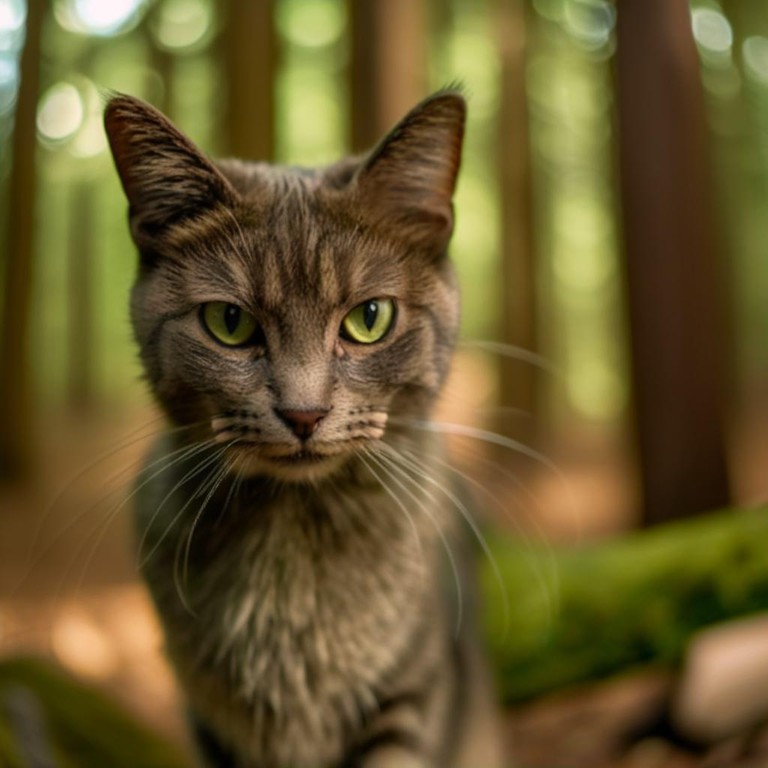} &
        \includegraphics[width=0.142\linewidth]{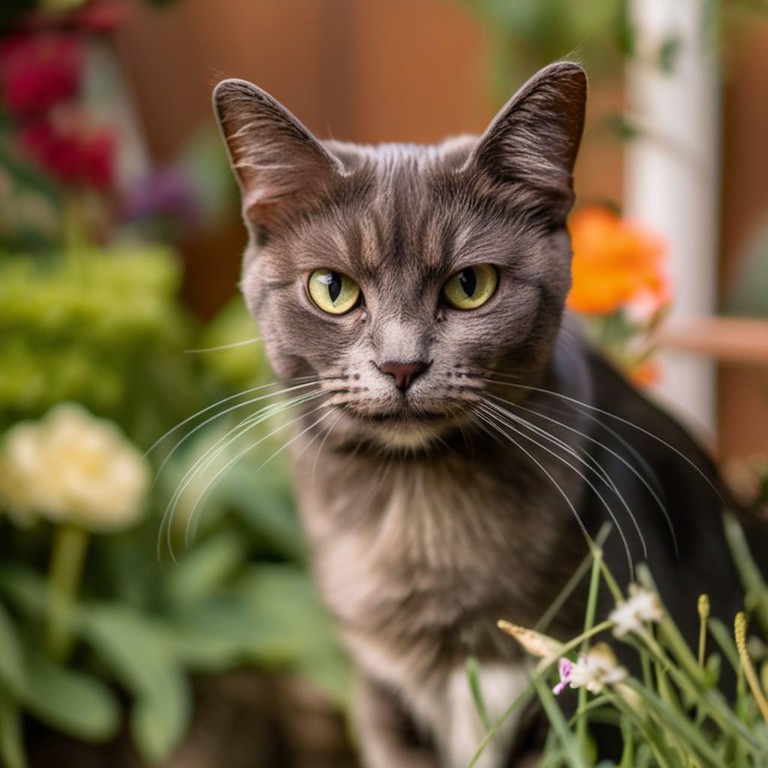} &
        \includegraphics[width=0.142\linewidth]{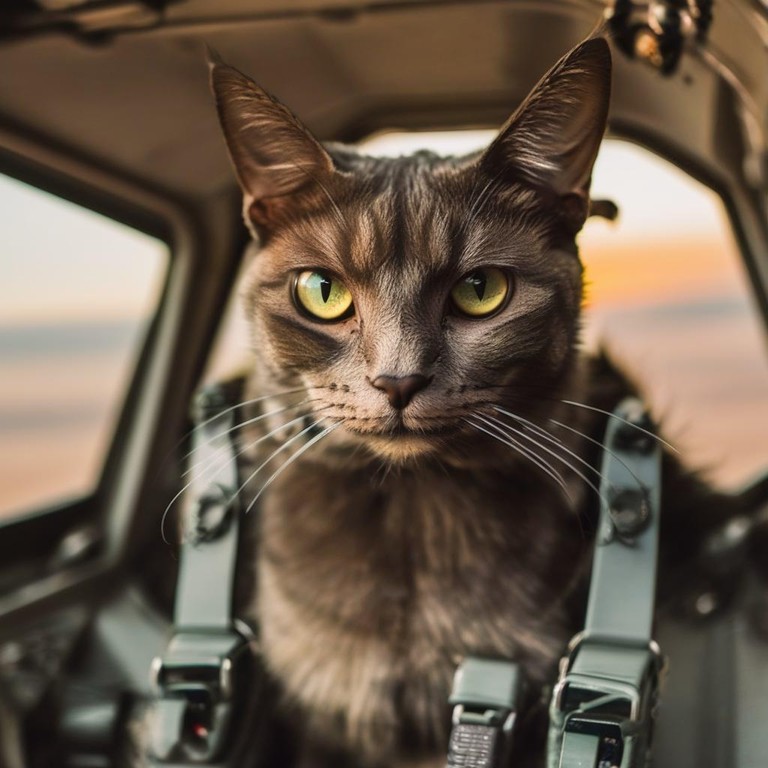} \\
        \includegraphics[width=0.142\linewidth]{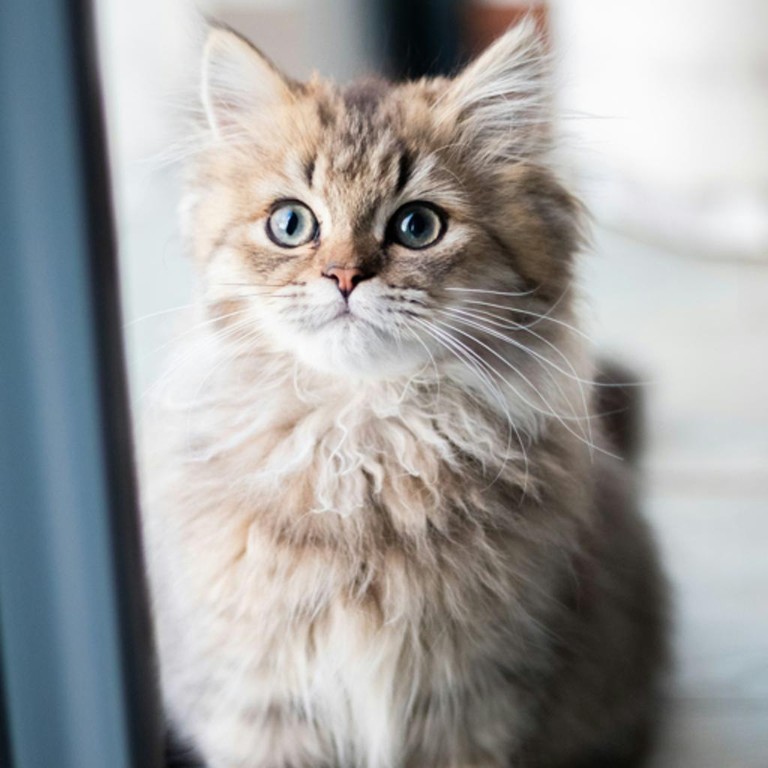} &
        \includegraphics[width=0.142\linewidth]{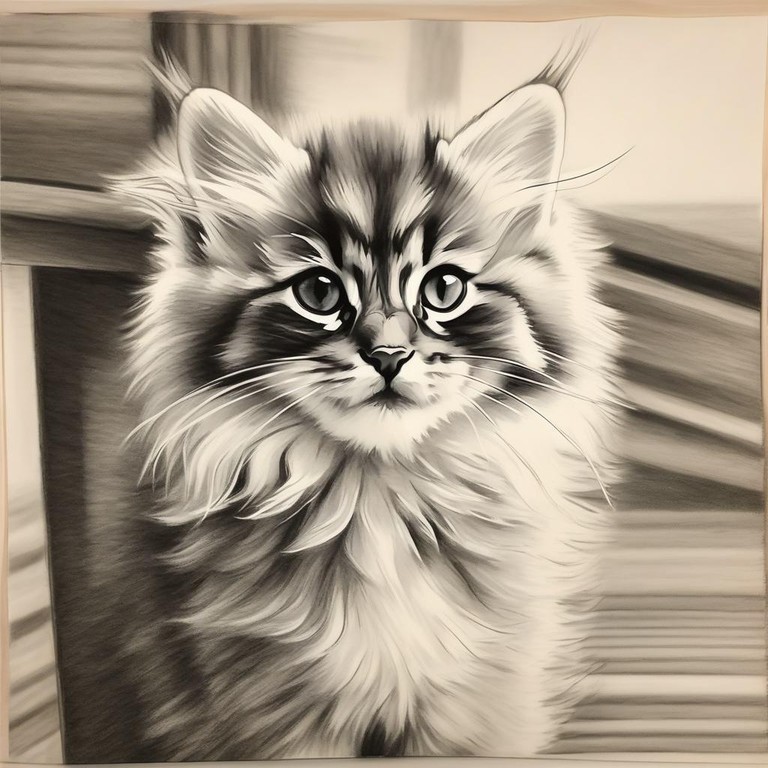} &
        \includegraphics[width=0.142\linewidth]{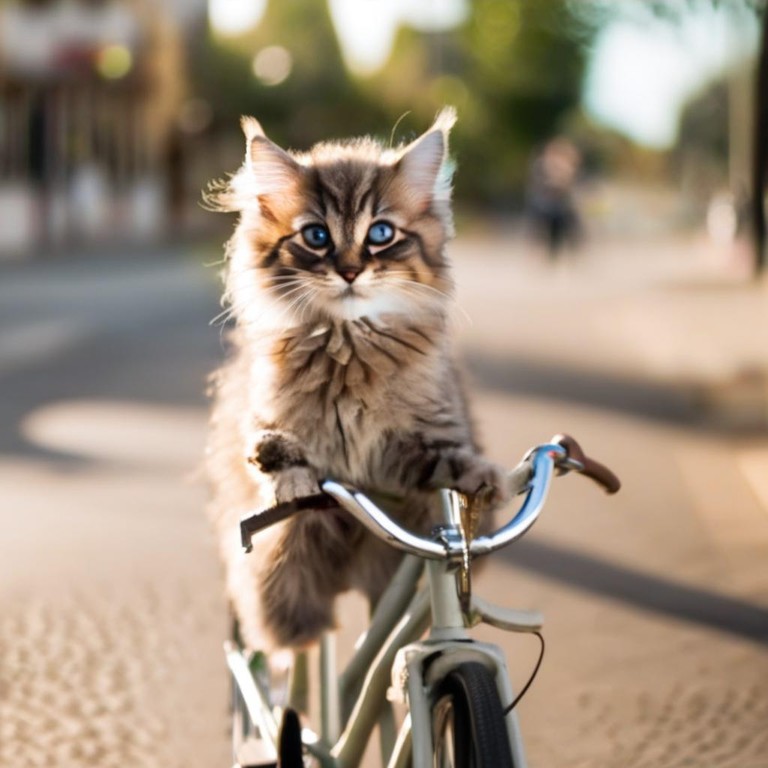} &
        \includegraphics[width=0.142\linewidth]{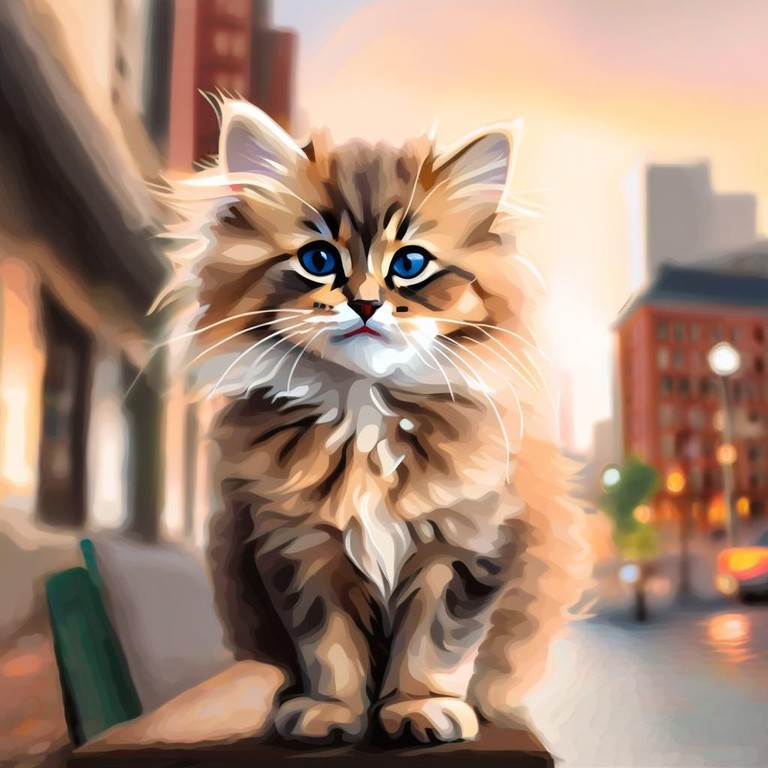} &
        \includegraphics[width=0.142\linewidth]{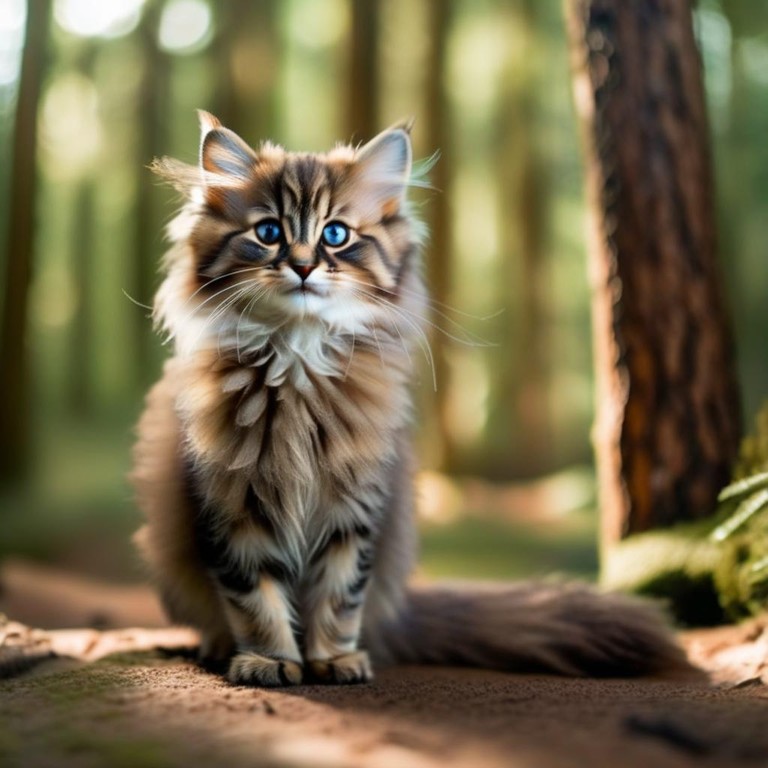} &
        \includegraphics[width=0.142\linewidth]{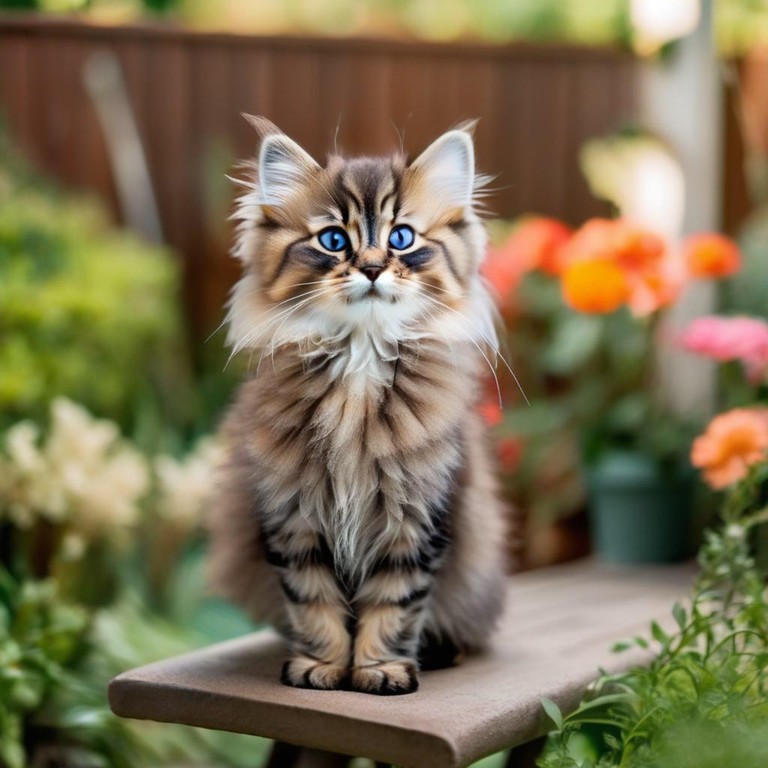} &
        \includegraphics[width=0.142\linewidth]{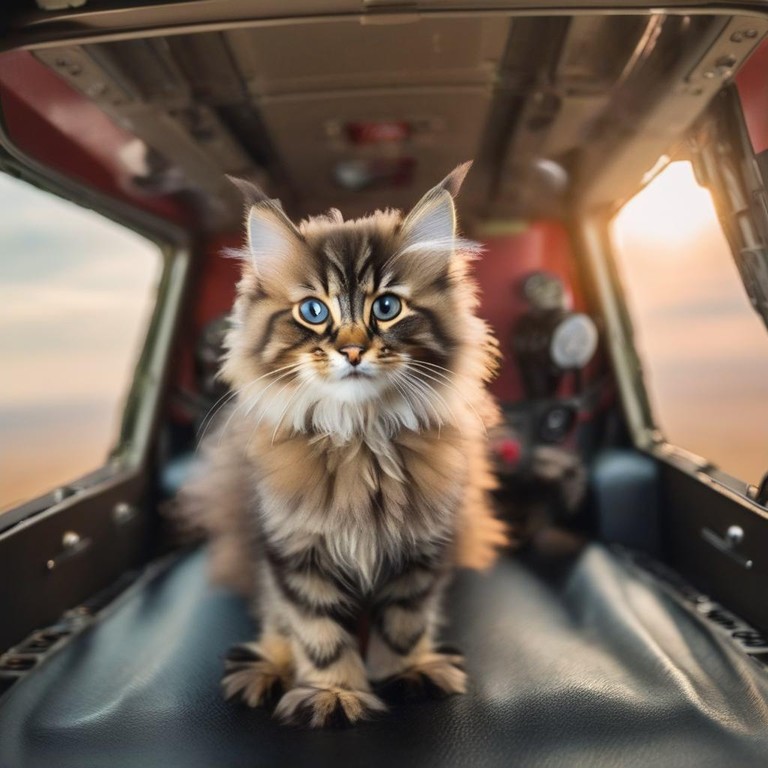} \\
        \includegraphics[width=0.142\linewidth]{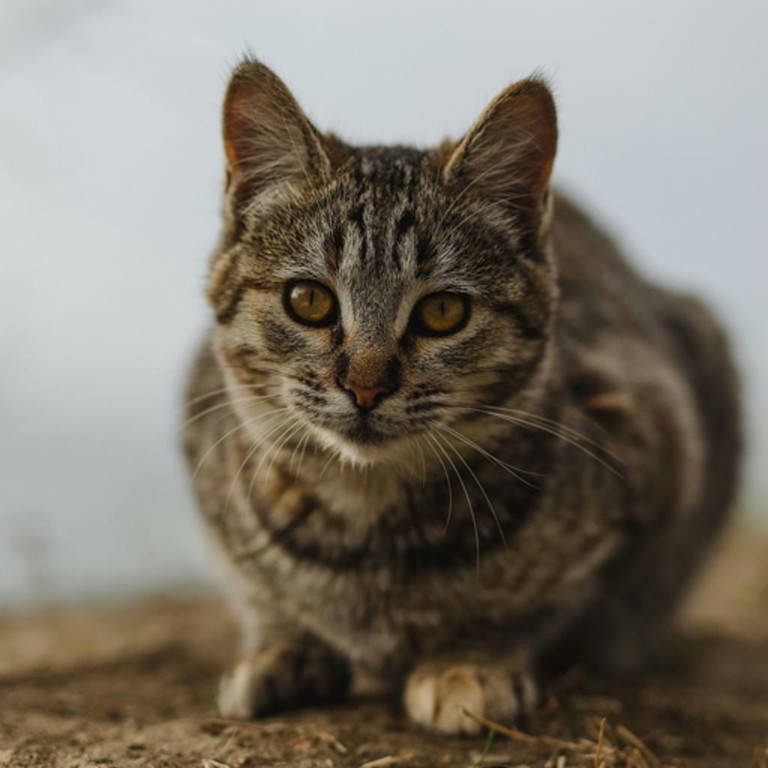} &
        \includegraphics[width=0.142\linewidth]{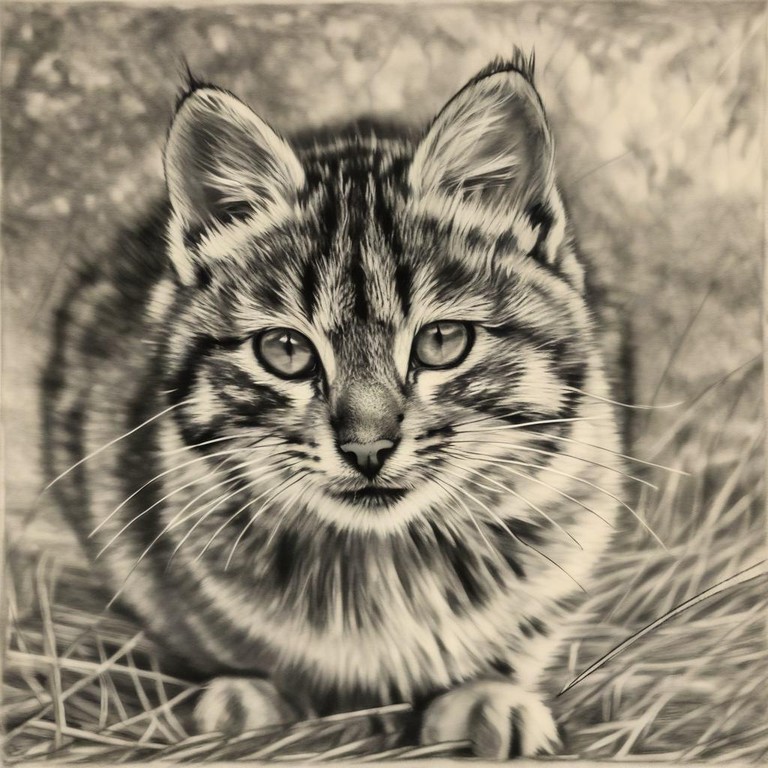} &
        \includegraphics[width=0.142\linewidth]{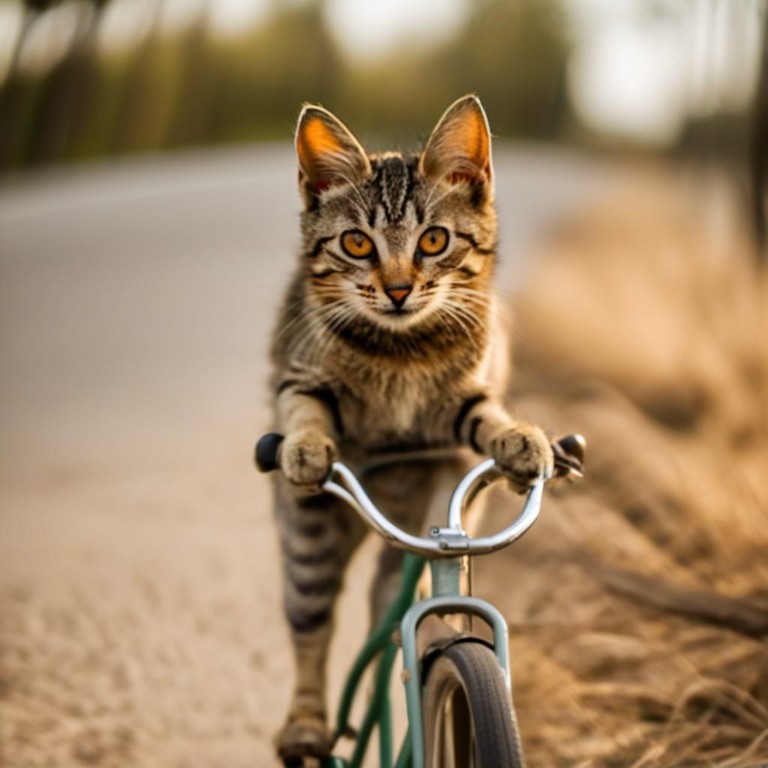} &
        \includegraphics[width=0.142\linewidth]{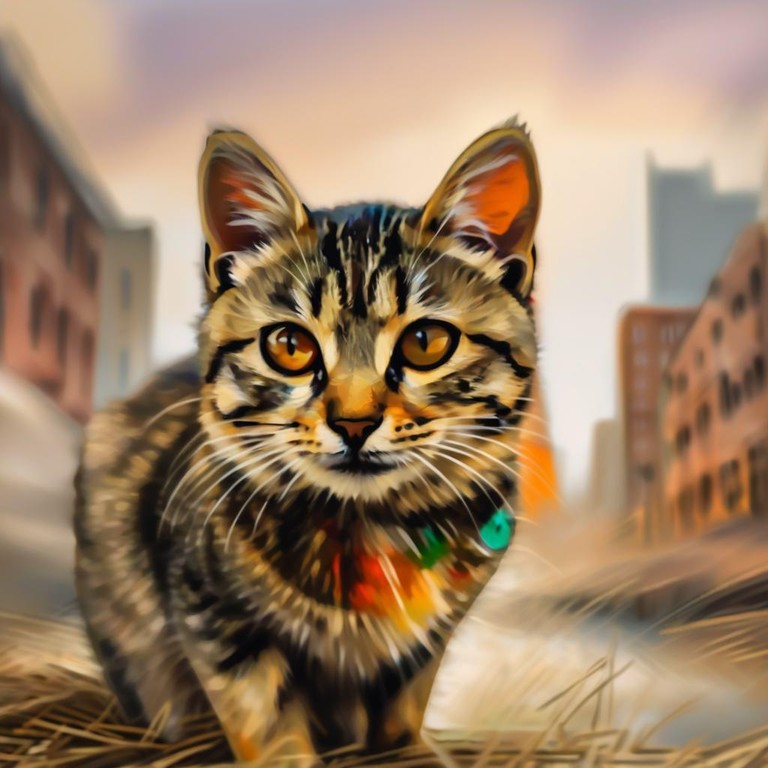} &
        \includegraphics[width=0.142\linewidth]{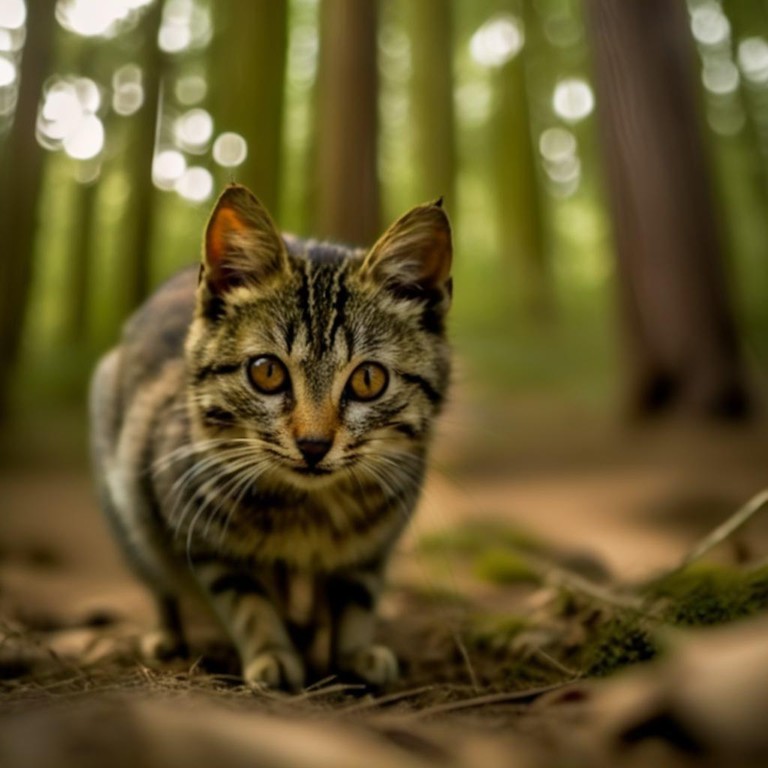} &
        \includegraphics[width=0.142\linewidth]{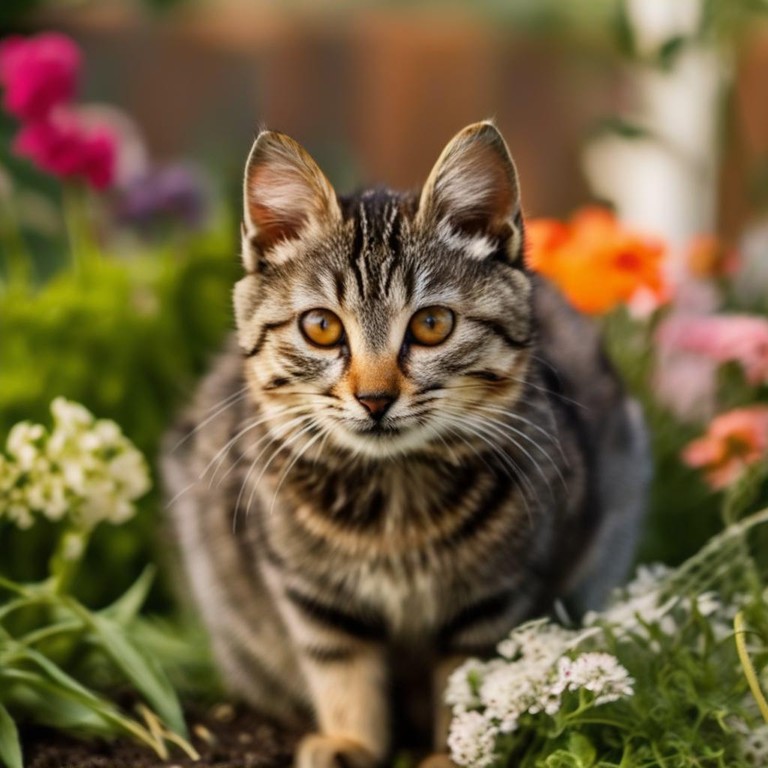} &
        \includegraphics[width=0.142\linewidth]{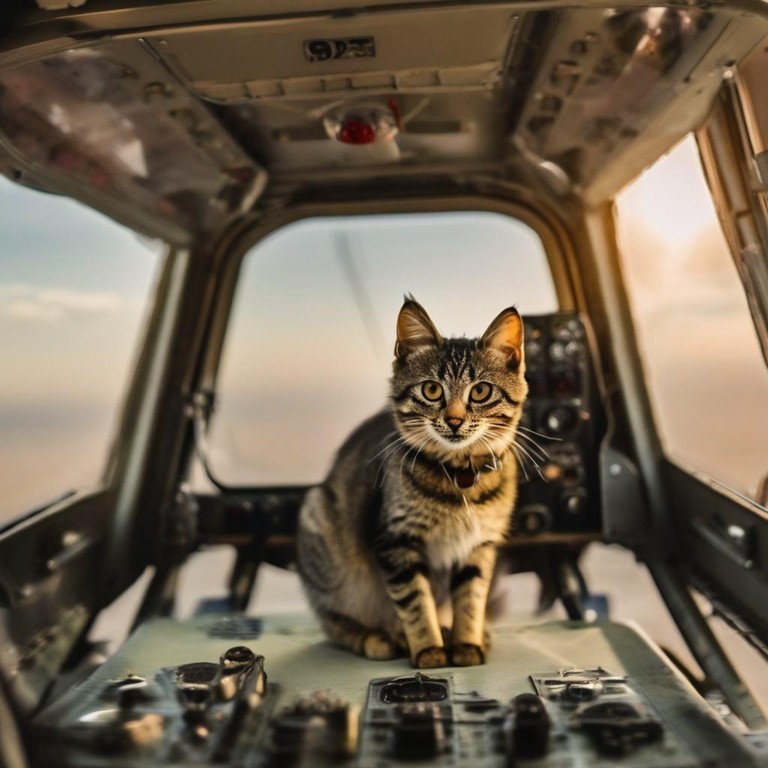} \\
        \includegraphics[width=0.142\linewidth]{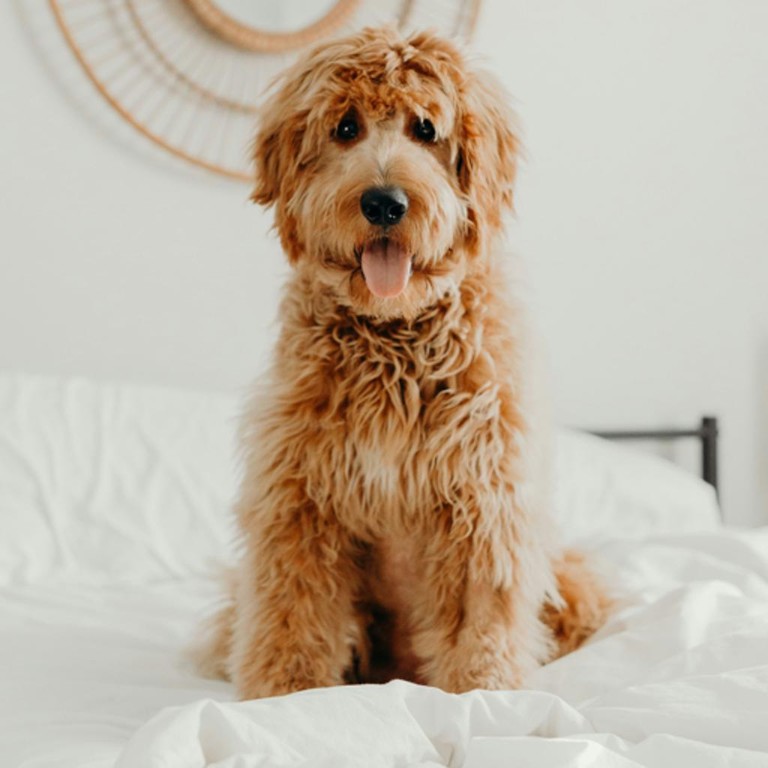} &
        \includegraphics[width=0.142\linewidth]{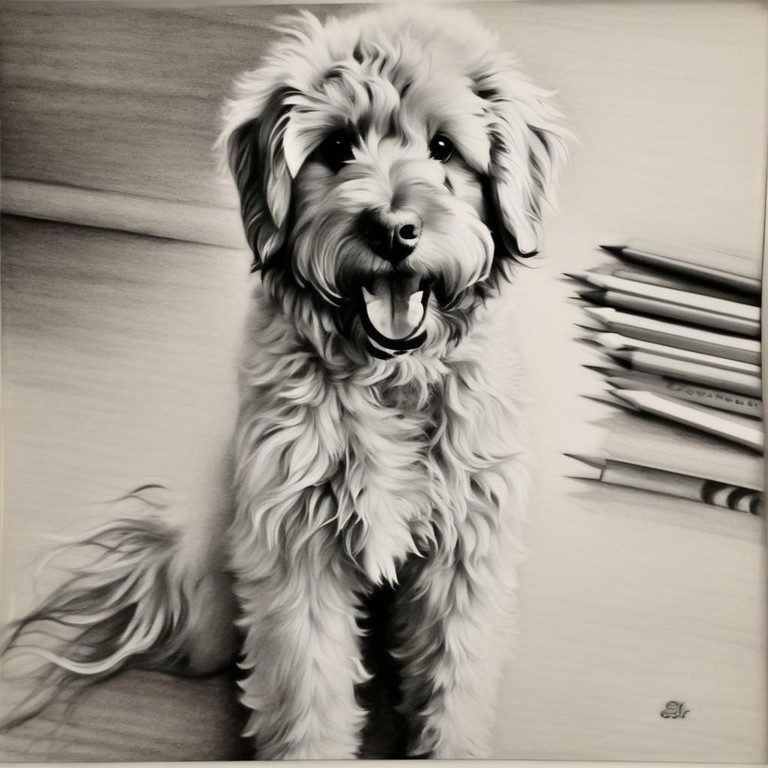} &
        \includegraphics[width=0.142\linewidth]{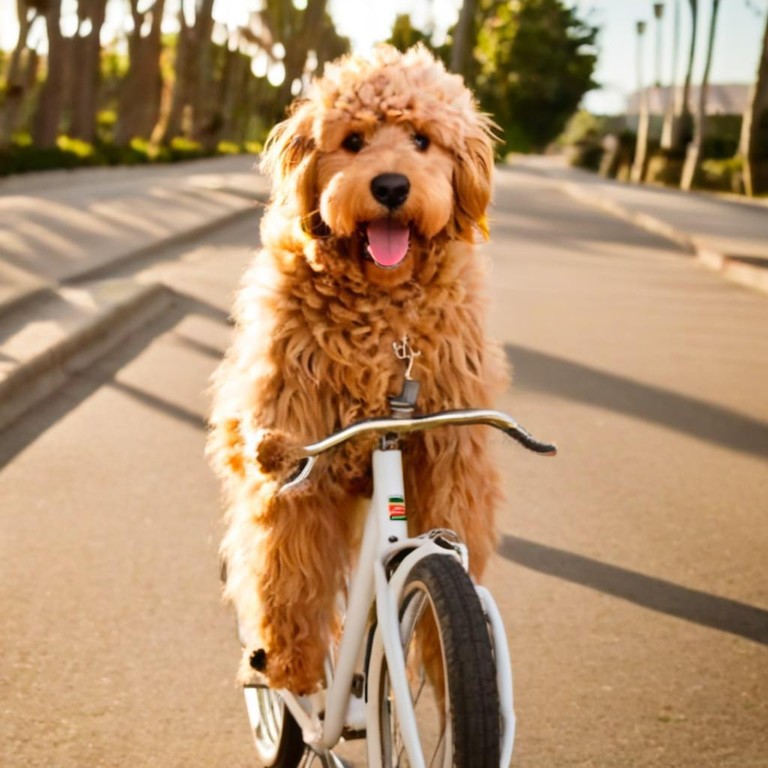} &
        \includegraphics[width=0.142\linewidth]{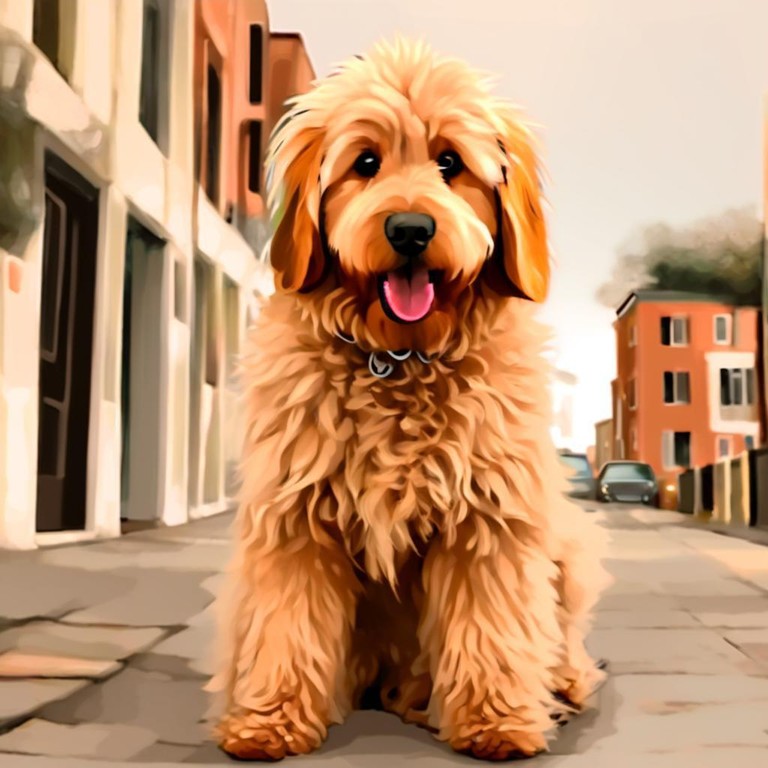} &
        \includegraphics[width=0.142\linewidth]{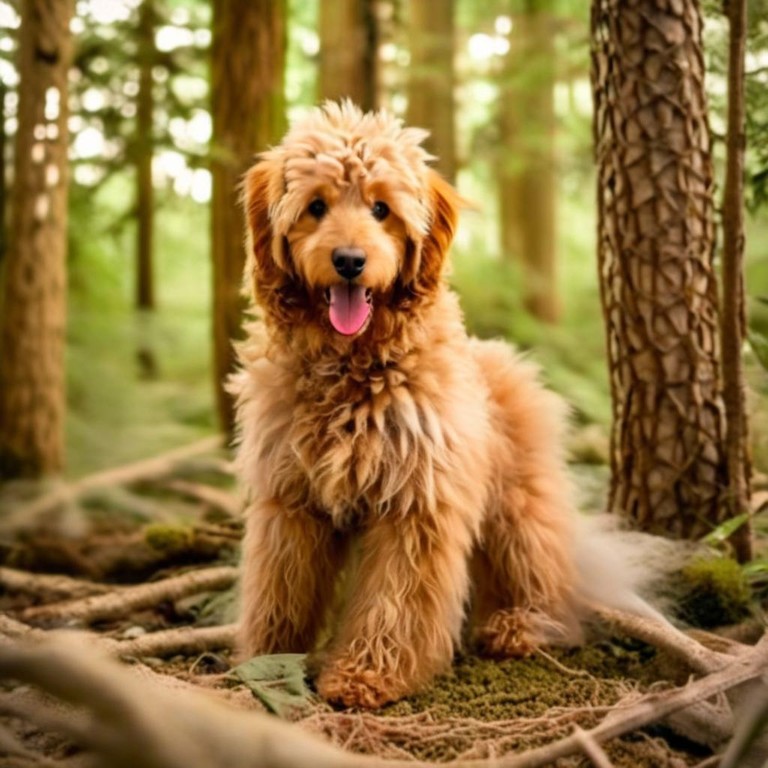} &
        \includegraphics[width=0.142\linewidth]{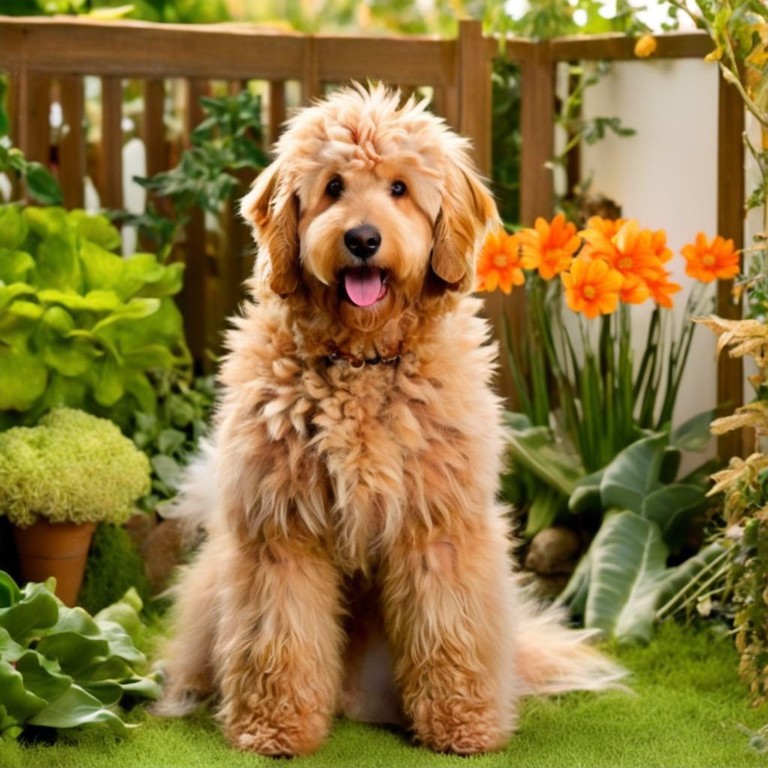} &
        \includegraphics[width=0.142\linewidth]{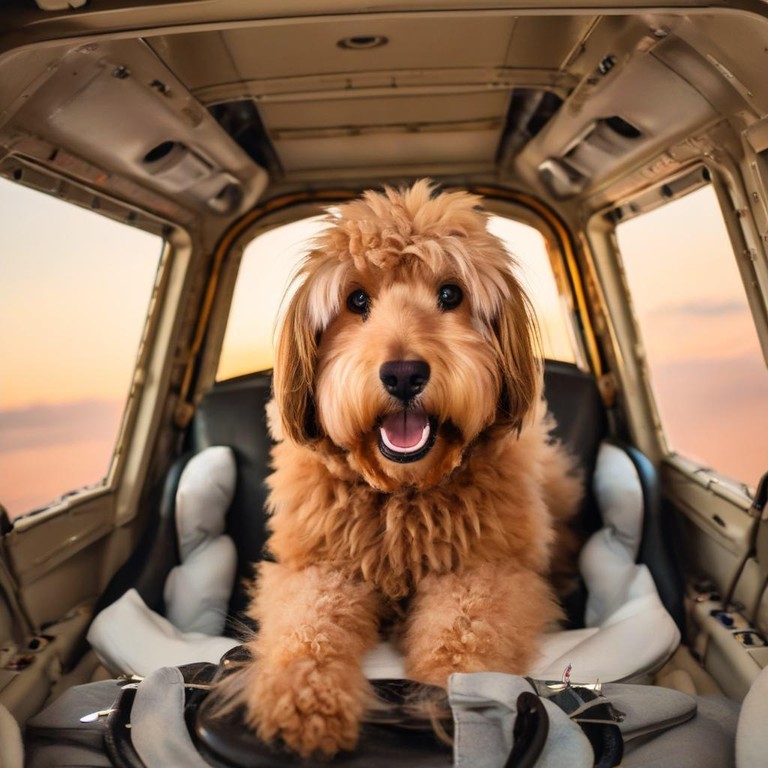} \\
        \includegraphics[width=0.142\linewidth]{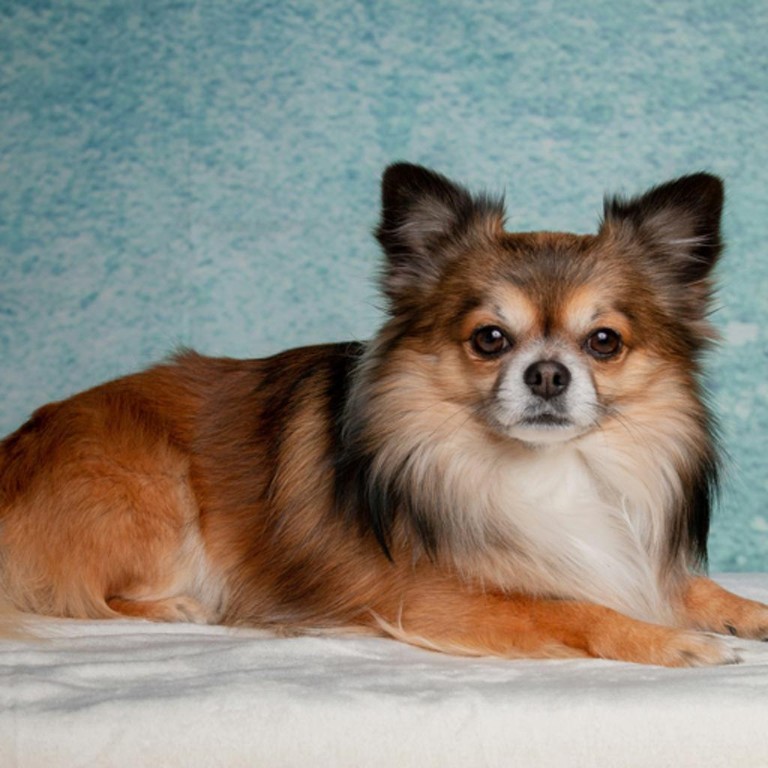} &
        \includegraphics[width=0.142\linewidth]{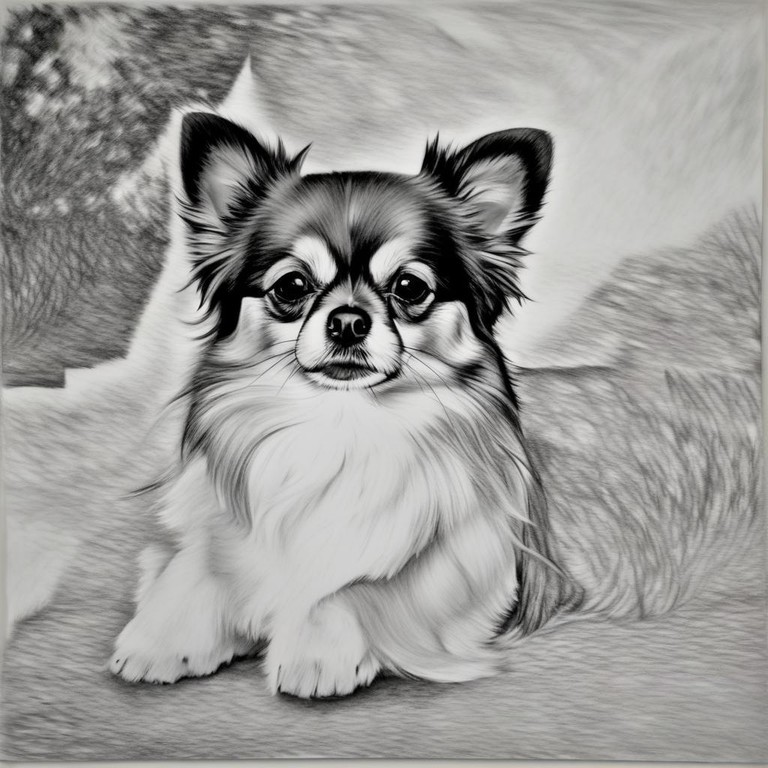} &
        \includegraphics[width=0.142\linewidth]{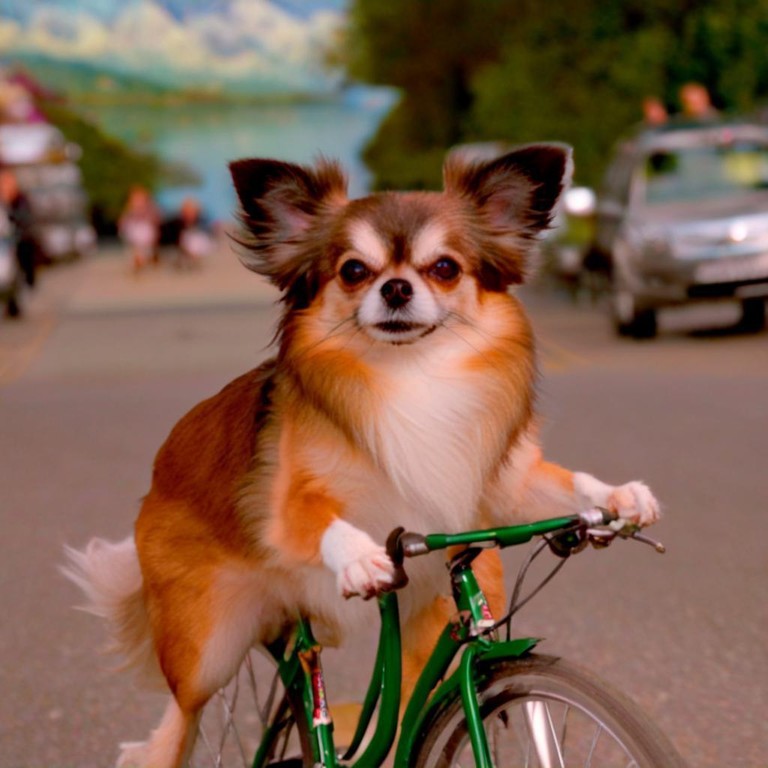} &
        \includegraphics[width=0.142\linewidth]{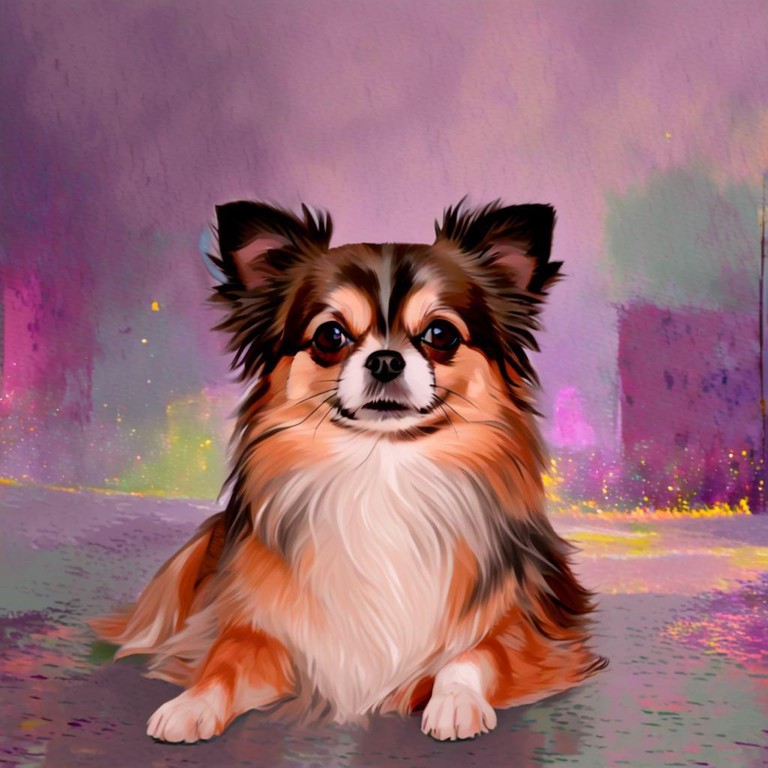} &
        \includegraphics[width=0.142\linewidth]{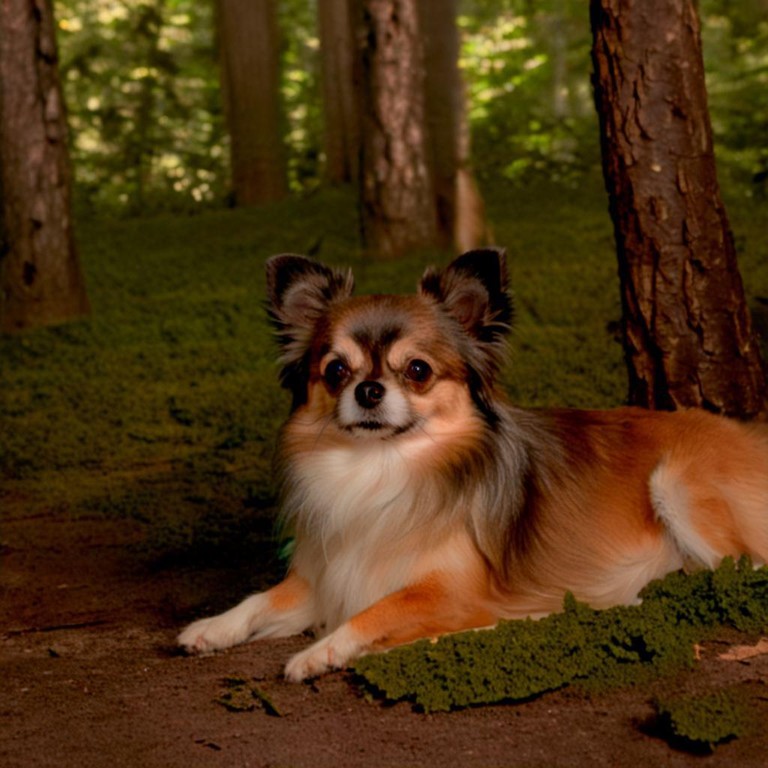} &
        \includegraphics[width=0.142\linewidth]{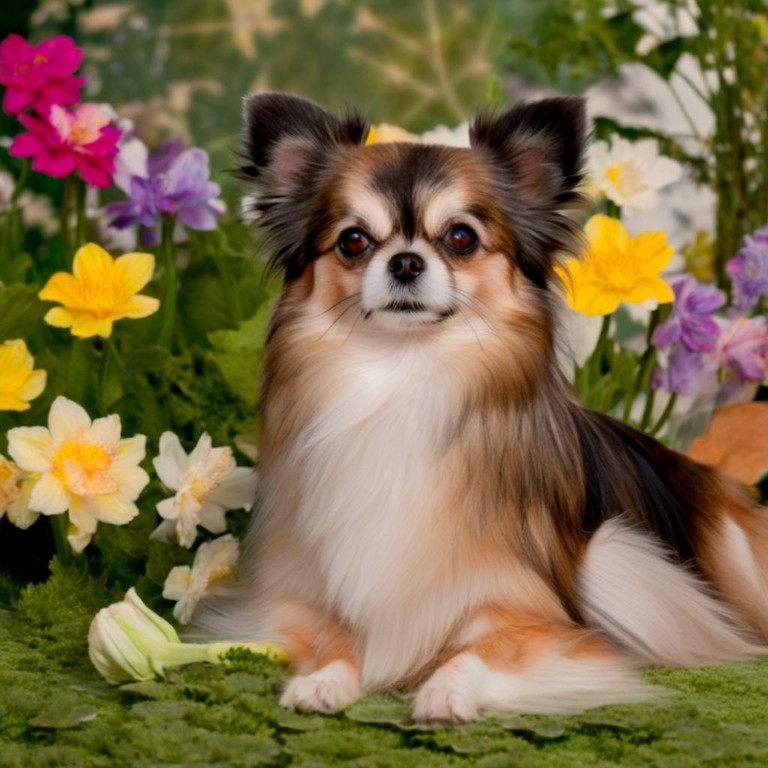} &
        \includegraphics[width=0.142\linewidth]{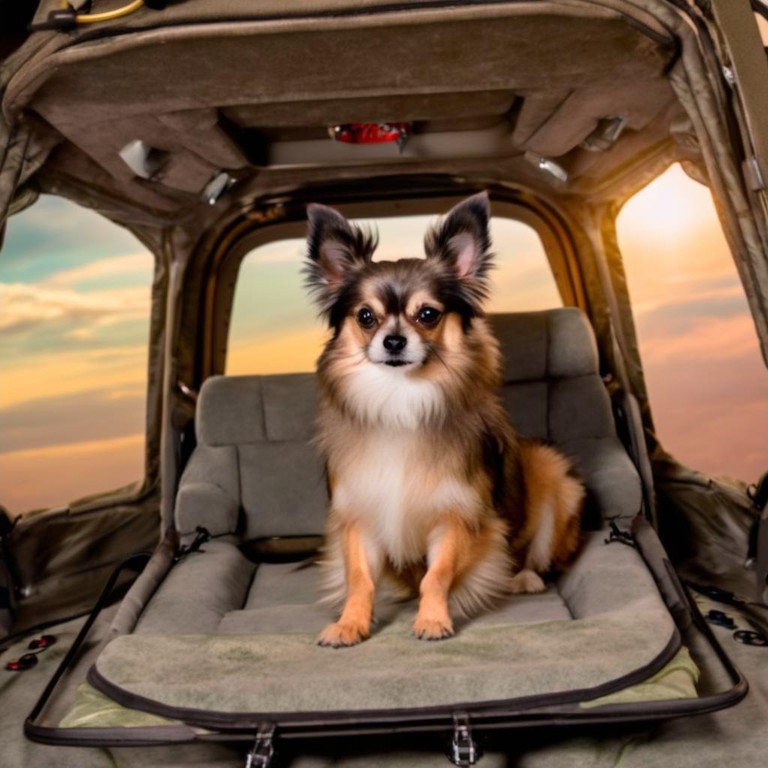} \\
        Input & ``pencil drawing'' & ``bike'' & ``digital art'' & ``forest'' & ``garden'' & ``helicopter''
    \end{tabular}
    }
    \caption{
    Additional results on pets. The initial noise is fixed across each column.
    }
    \label{fig:pet-grid}
\end{figure*}

Additional results of our method are presented in \Cref{fig:face-grid,fig:pet-grid}.

\paragraph{Identities Mixing}

\begin{figure}
    \centering
    \setlength{\tabcolsep}{1pt}
    \scriptsize{
    \begin{tabular}{cccc}
        \includegraphics[width=0.24\linewidth]{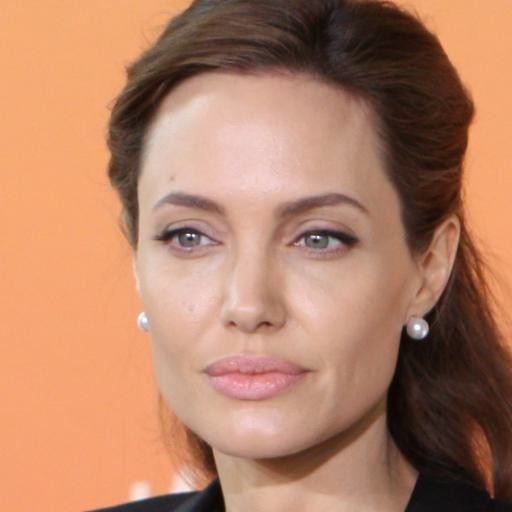} &
        \includegraphics[width=0.24\linewidth]{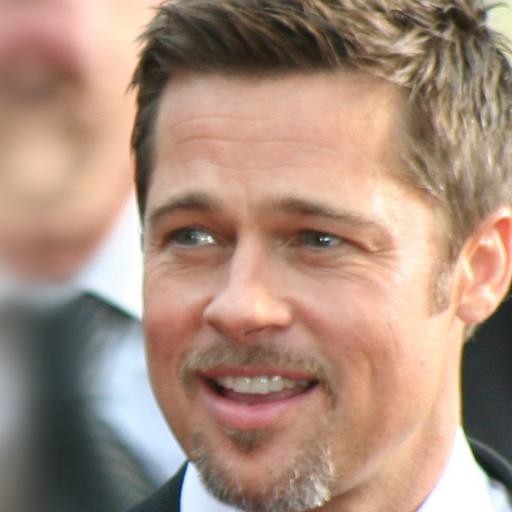} &
        \includegraphics[width=0.24\linewidth]{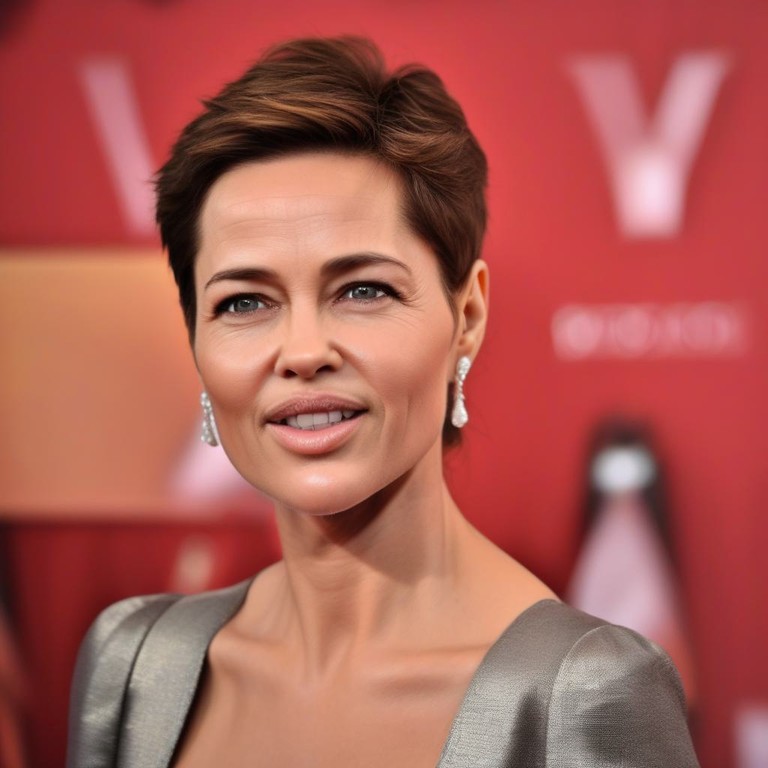} &
        \includegraphics[width=0.24\linewidth]{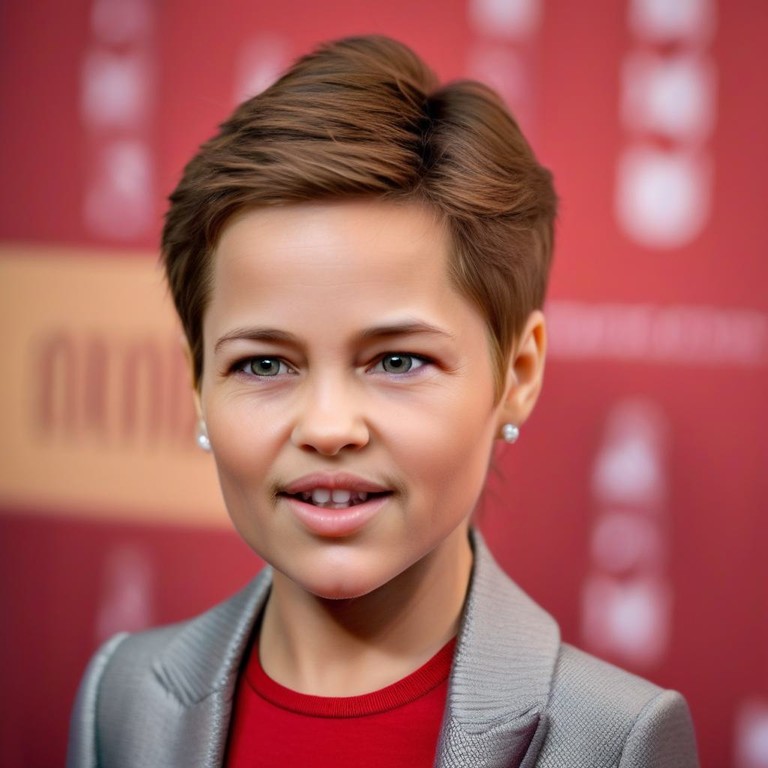} \\
        Input 1 & Input 2 & ``person'' & ``child'' \\
        \includegraphics[width=0.24\linewidth]{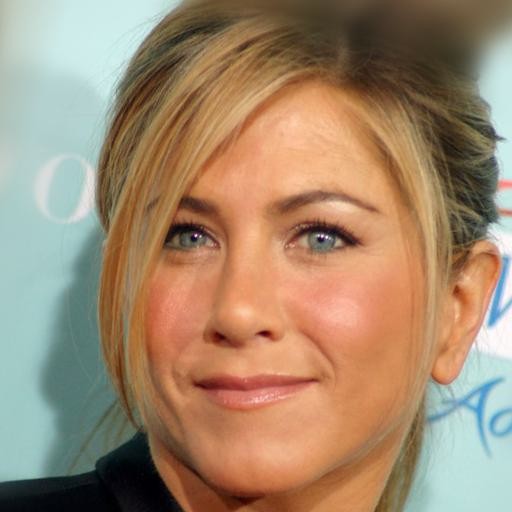} &
        \includegraphics[width=0.24\linewidth]{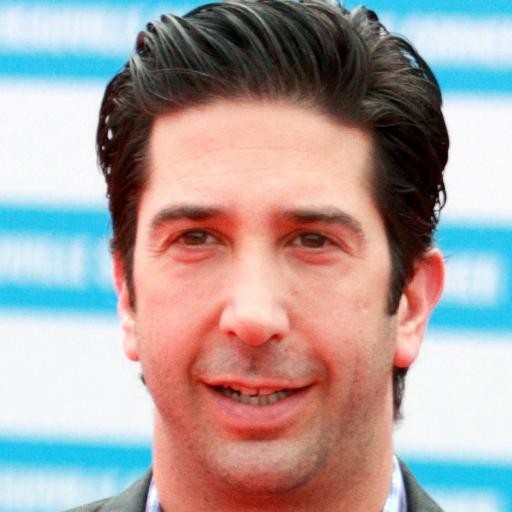} &
        \includegraphics[width=0.24\linewidth]{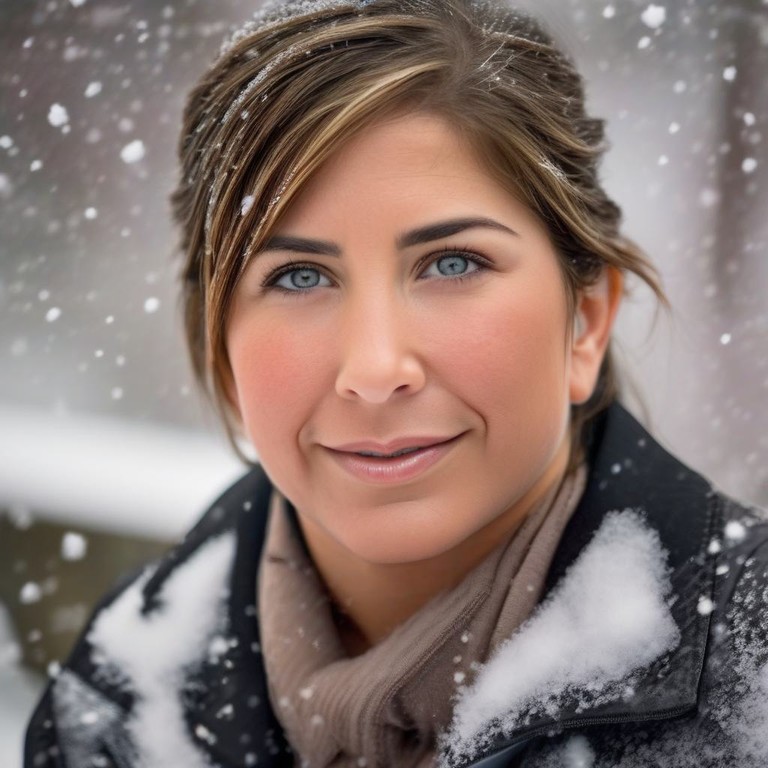} &
        \includegraphics[width=0.24\linewidth]{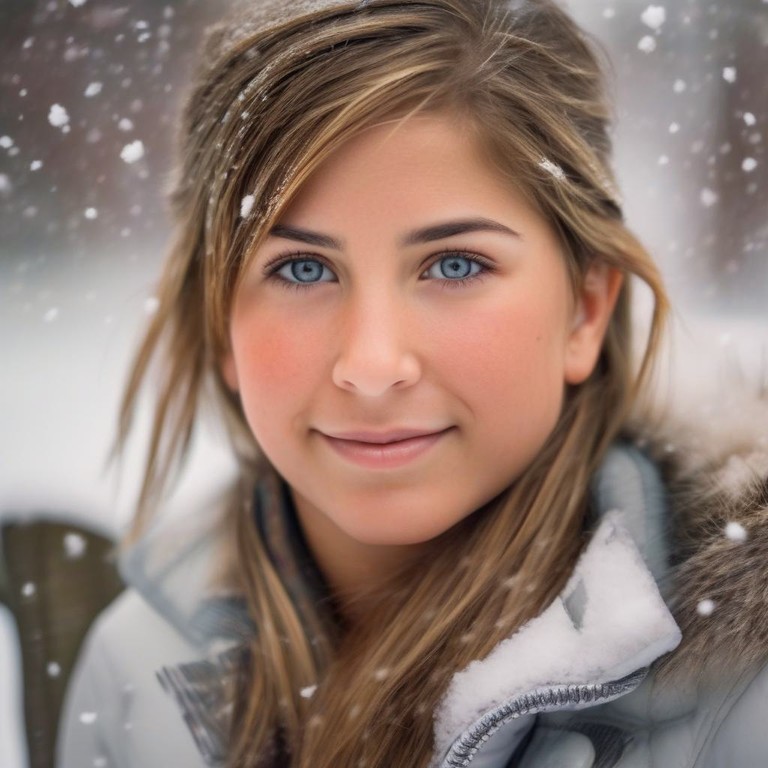} \\
        Input 1 & Input 2 & ``person'' & ``girl'' \\
    \end{tabular}
    }
    \caption{
        Our method allows mixing two identities by encoding them separately, and pass the concatenated representations to the nested attention layer.
    }
    \label{fig:mixing}
\end{figure}

Following prior works~\cite{li2023photomaker}, our method enables mixing between two identities, as illustrated in Figure~\ref{fig:mixing}. 
To achieve this, we use our trained encoder to independently encode each image and concatenate the resulting tokens.
These concatenated tokens are then fed into the nested attention layers. This approach is analogous to the technique used for handling multiple images of the same subject, except that in this case, the images represent different subjects.

\begin{figure}
    \centering
    \setlength{\tabcolsep}{1pt}
    \scriptsize{
    \begin{tabular}{cccc}
        \includegraphics[width=0.24\linewidth]{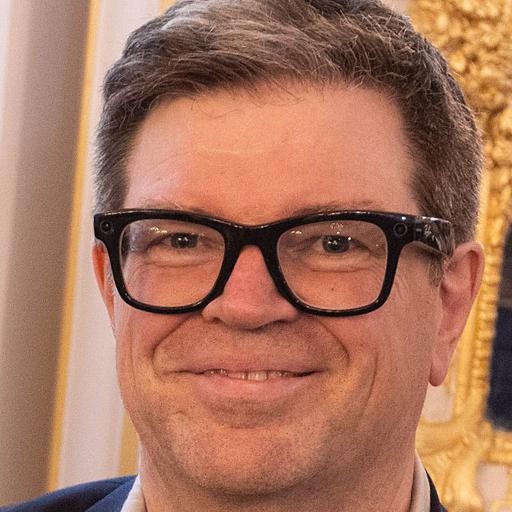} &
        \includegraphics[width=0.24\linewidth]{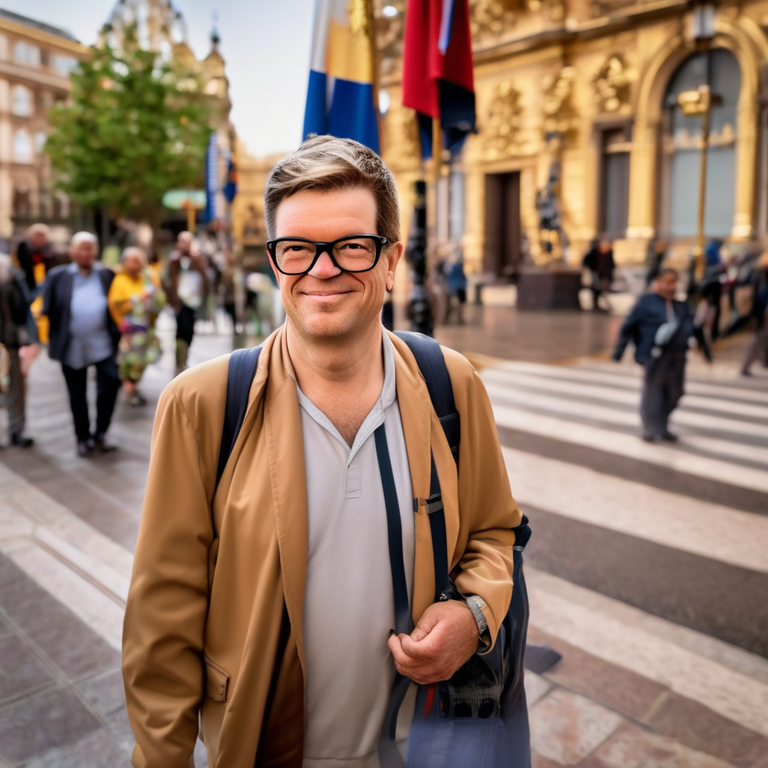} &
        \includegraphics[width=0.24\linewidth]{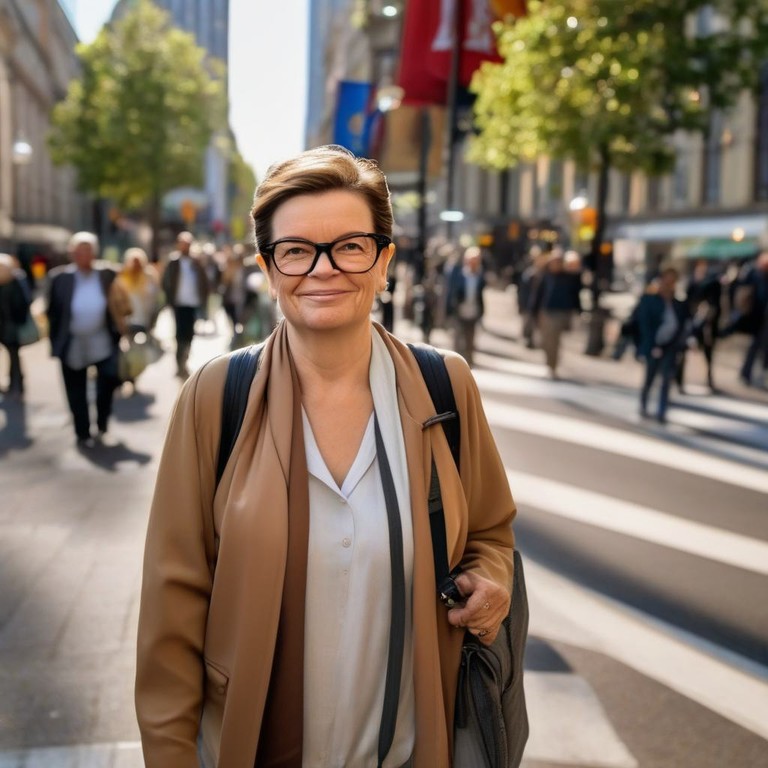} &
        \includegraphics[width=0.24\linewidth]{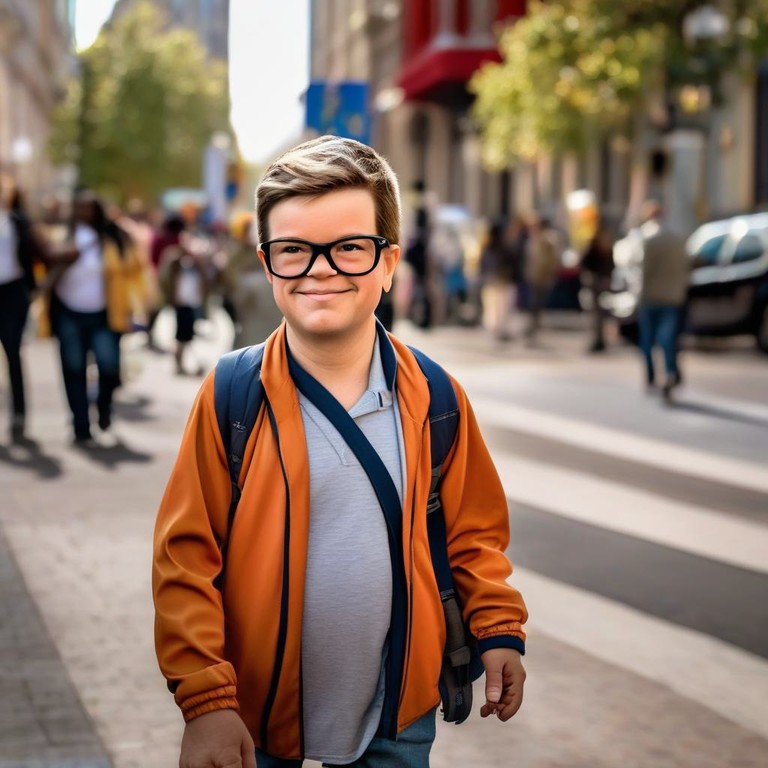} \\
        Input & ``person'' & ``woman'' & ``child'' \\
        \includegraphics[width=0.24\linewidth]{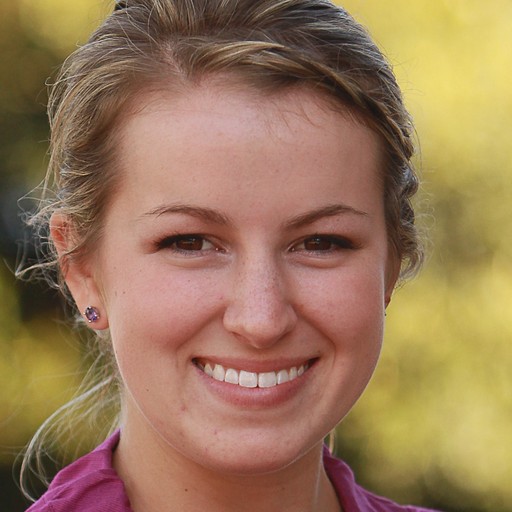} &
        \includegraphics[width=0.24\linewidth]{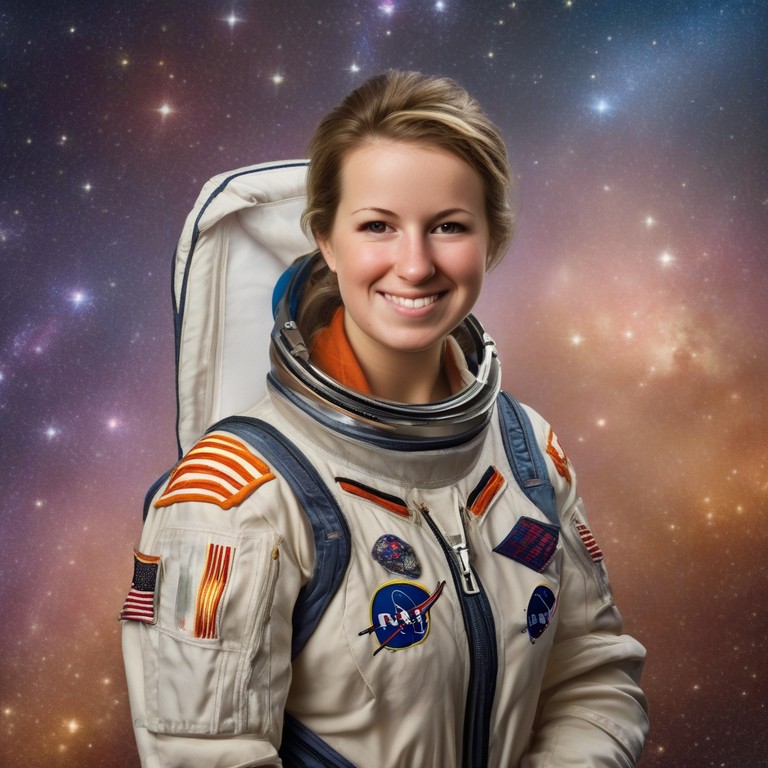} &
        \includegraphics[width=0.24\linewidth]{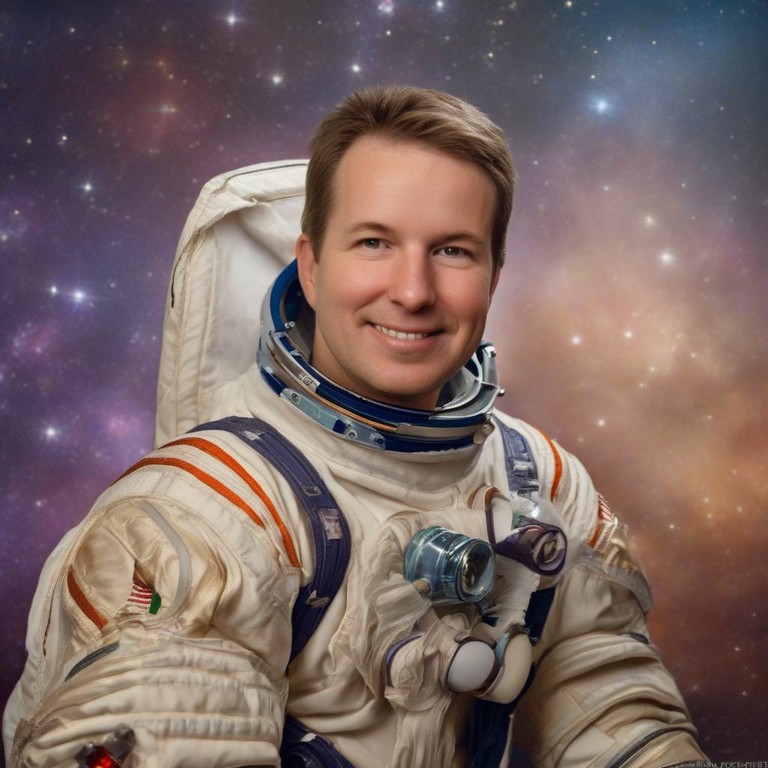} &
        \includegraphics[width=0.24\linewidth]{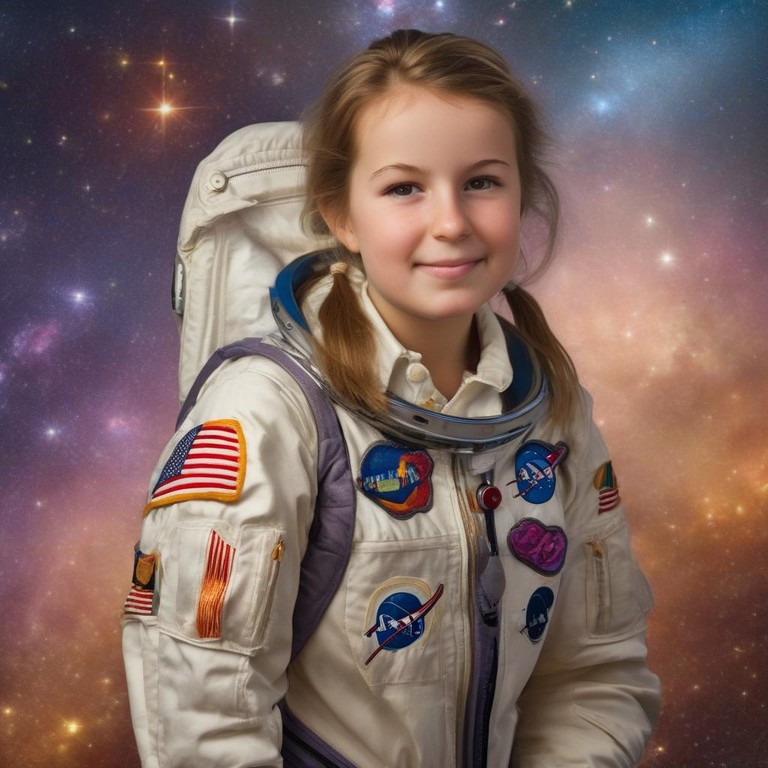} \\
        Input & ``person'' & ``man'' & ``girl''
    \end{tabular}
    }
    \caption{
            By simply changing at inference time the token to which we inject the personalized concept (\eg, ``person'' to ``woman'') we can have various semantic variations. 
    }
    \label{fig:semantic-variation}
\end{figure}

\begin{figure*}
    \centering
    \setlength{\tabcolsep}{1pt}
    \scriptsize{
    \begin{tabular}{c c ccc c ccc}
        & Input & \multicolumn{3}{c}{$\longleftarrow$ Varying $\lambda$ $\longrightarrow$} & Input & \multicolumn{3}{c}{$\longleftarrow$ Varying $\lambda$ $\longrightarrow$} \\
        \raisebox{24pt}{\rotatebox[origin=t]{90}{Simple Adapter}} &
        \includegraphics[width=0.12\linewidth]{images/mechanism_comparison/man_watercolor_input.jpg} &
        \includegraphics[width=0.12\linewidth]{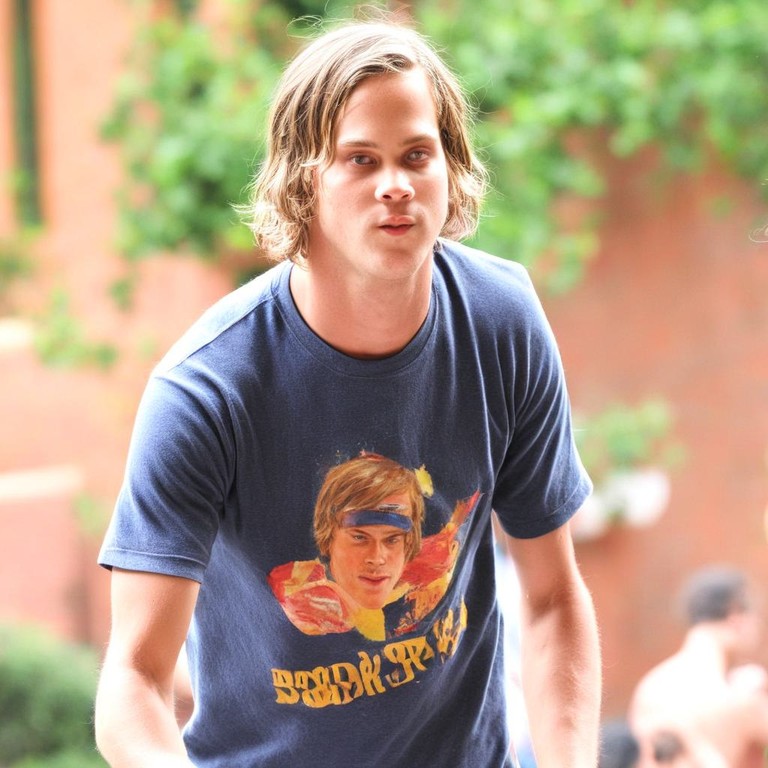} &
         \raisebox{25pt}{N/A} &
         \raisebox{25pt}{N/A} &
        \includegraphics[width=0.12\linewidth]{images/mechanism_comparison/37512.jpg} &
        \includegraphics[width=0.12\linewidth]{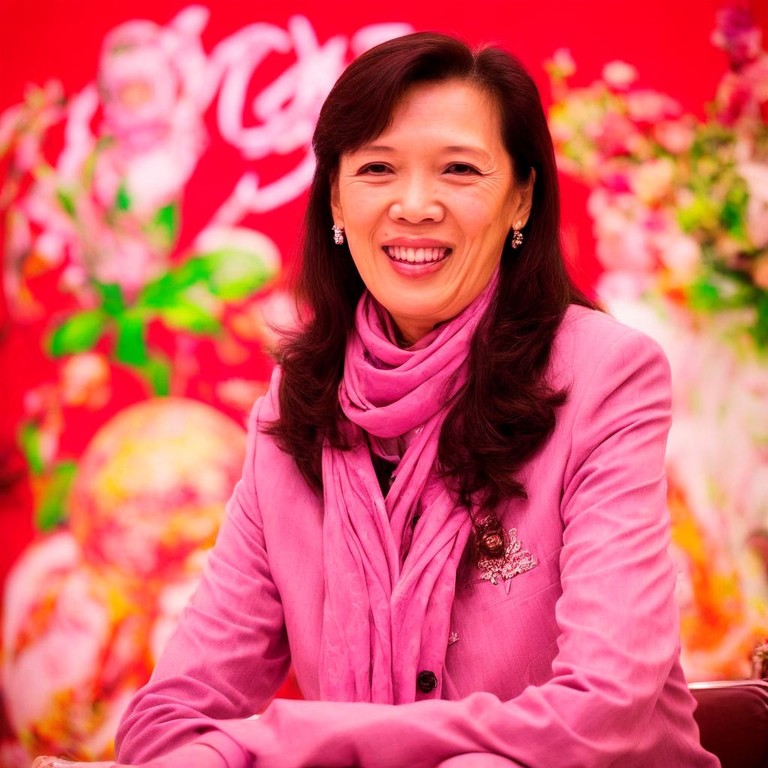} &
         \raisebox{25pt}{N/A} &
         \raisebox{25pt}{N/A}
         \\
        \raisebox{24pt}{\rotatebox[origin=t]{90}{Multiple Tokens}} &
        \includegraphics[width=0.12\linewidth]{images/mechanism_comparison/man_watercolor_input.jpg} &
        \includegraphics[width=0.12\linewidth]{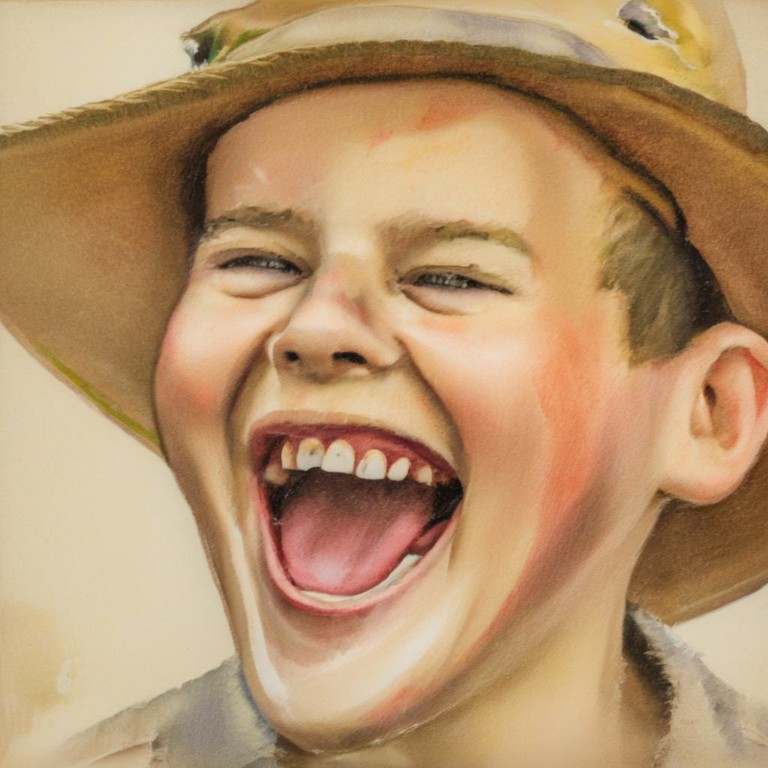} &
        \includegraphics[width=0.12\linewidth]{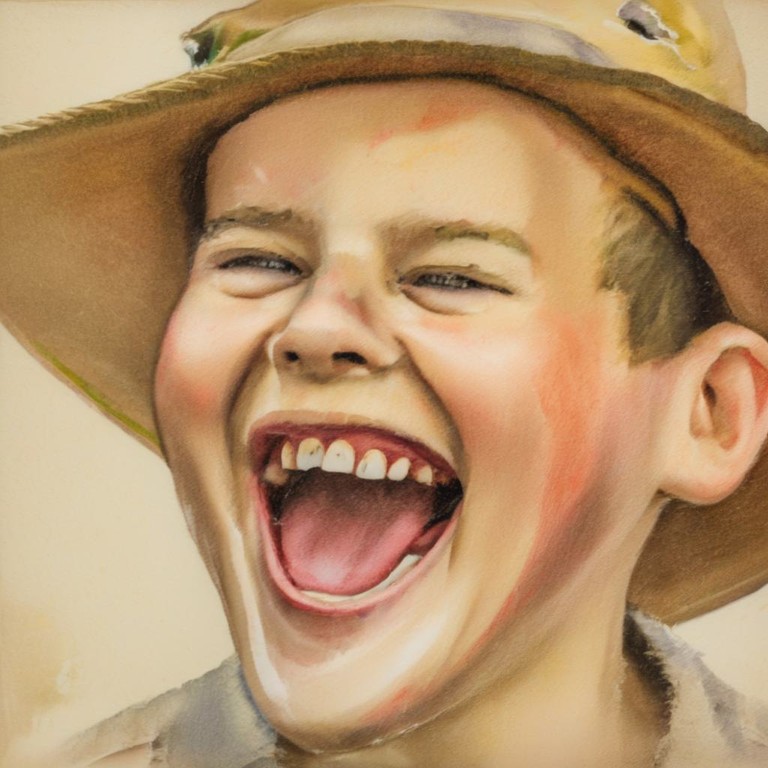} &
        \includegraphics[width=0.12\linewidth]{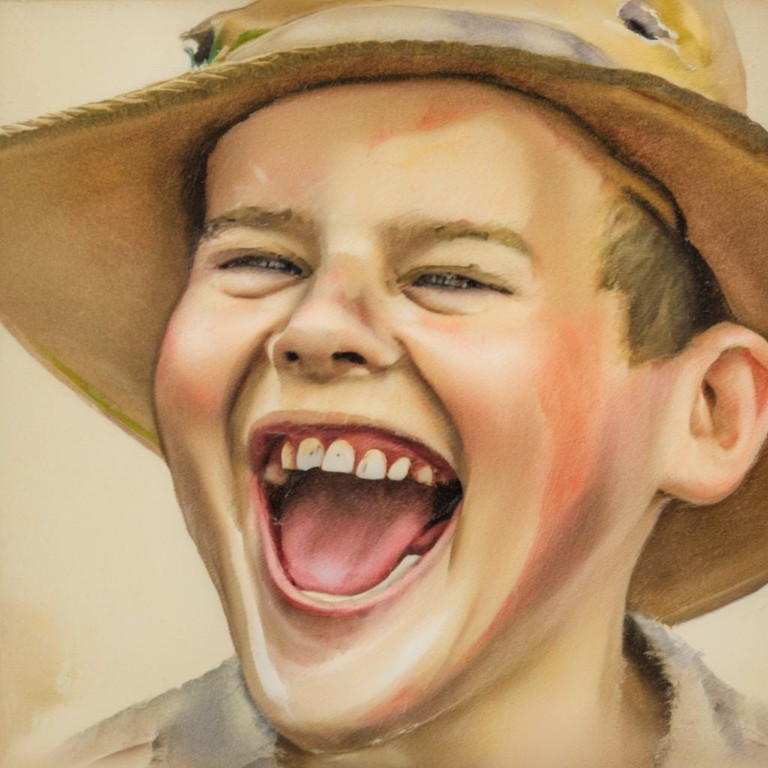} &
        \includegraphics[width=0.12\linewidth]{images/mechanism_comparison/37512.jpg} &
        \includegraphics[width=0.12\linewidth]{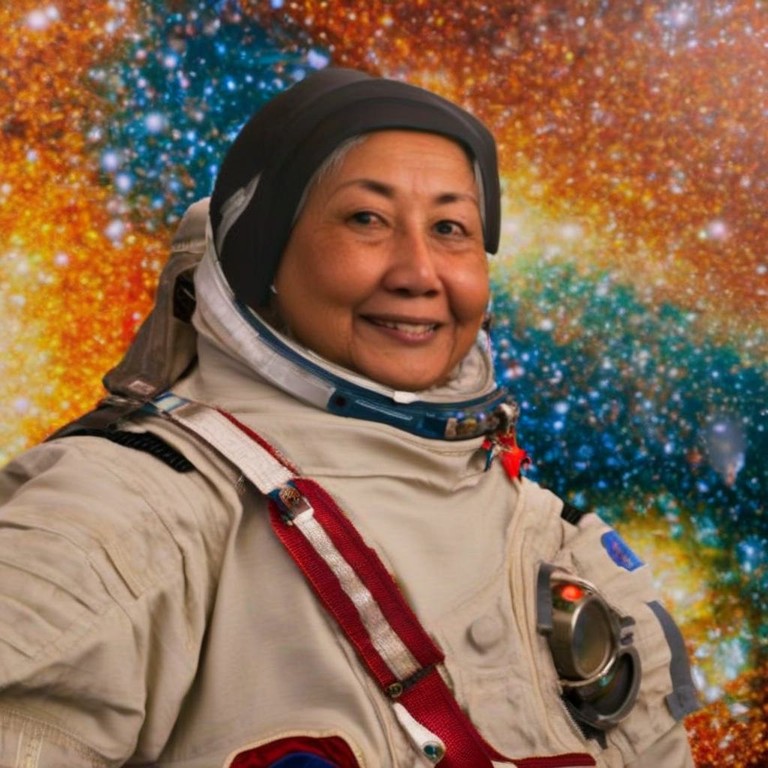} &
        \includegraphics[width=0.12\linewidth]{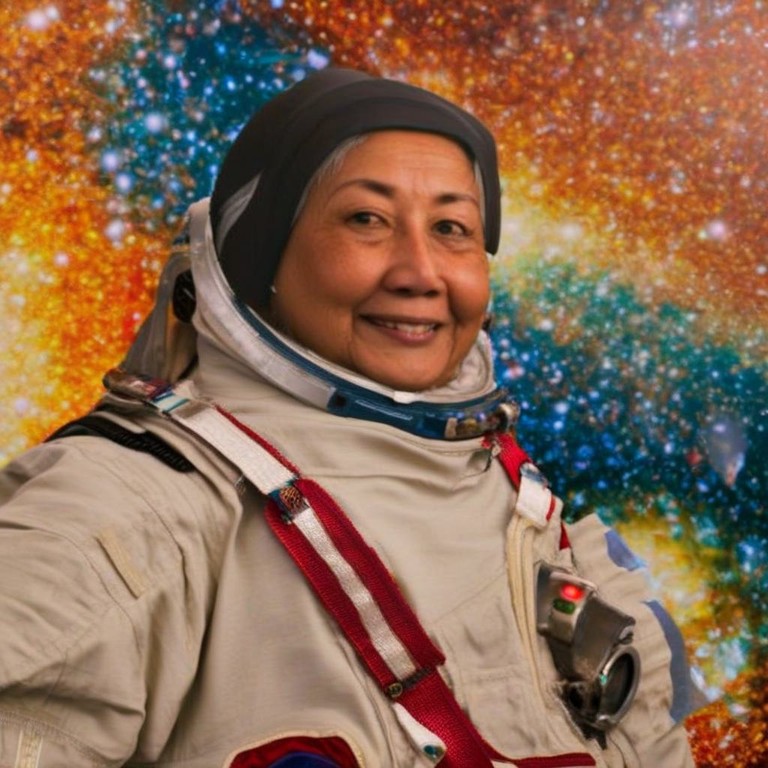} &
        \includegraphics[width=0.12\linewidth]{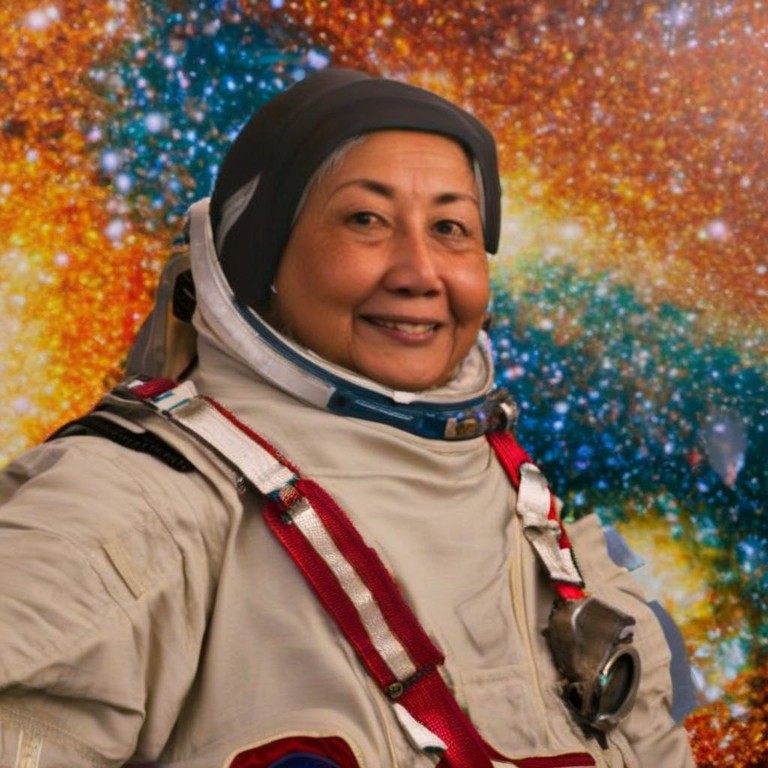} \\
        \raisebox{24pt}{\rotatebox[origin=t]{90}{Decoupled CA}} &
        \includegraphics[width=0.12\linewidth]{images/mechanism_comparison/man_watercolor_input.jpg} &
        \includegraphics[width=0.12\linewidth]{images/mechanism_comparison/man_watercolor_decoupled_0.5.jpg} &
        \includegraphics[width=0.12\linewidth]{images/mechanism_comparison/man_watercolor_decoupled_0.6.jpg} &
        \includegraphics[width=0.12\linewidth]{images/mechanism_comparison/man_watercolor_decoupled_1.0.jpg} &
        \includegraphics[width=0.12\linewidth]{images/mechanism_comparison/37512.jpg} &
        \includegraphics[width=0.12\linewidth]{images/mechanism_comparison/woman_decoupled_astronaut_0.5.jpg} &
        \includegraphics[width=0.12\linewidth]{images/mechanism_comparison/woman_decoupled_astronaut_0.6.jpg} &
        \includegraphics[width=0.12\linewidth]{images/mechanism_comparison/woman_decoupled_astronaut_1.0.jpg} \\
        \raisebox{24pt}{\rotatebox[origin=t]{90}{Global $V$}} &
        \includegraphics[width=0.12\linewidth]{images/mechanism_comparison/man_watercolor_input.jpg} &
        \includegraphics[width=0.12\linewidth]{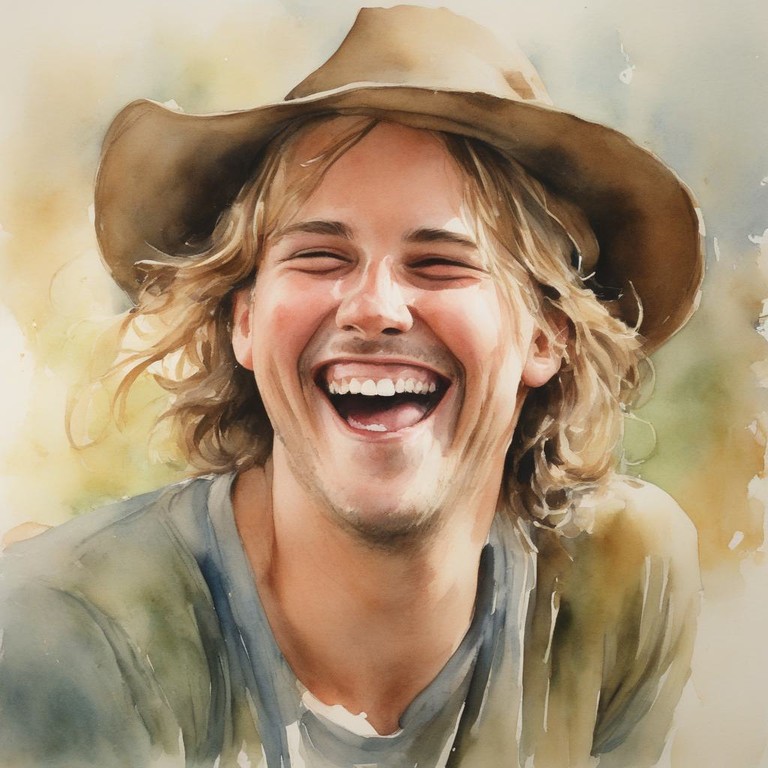} &
        \includegraphics[width=0.12\linewidth]{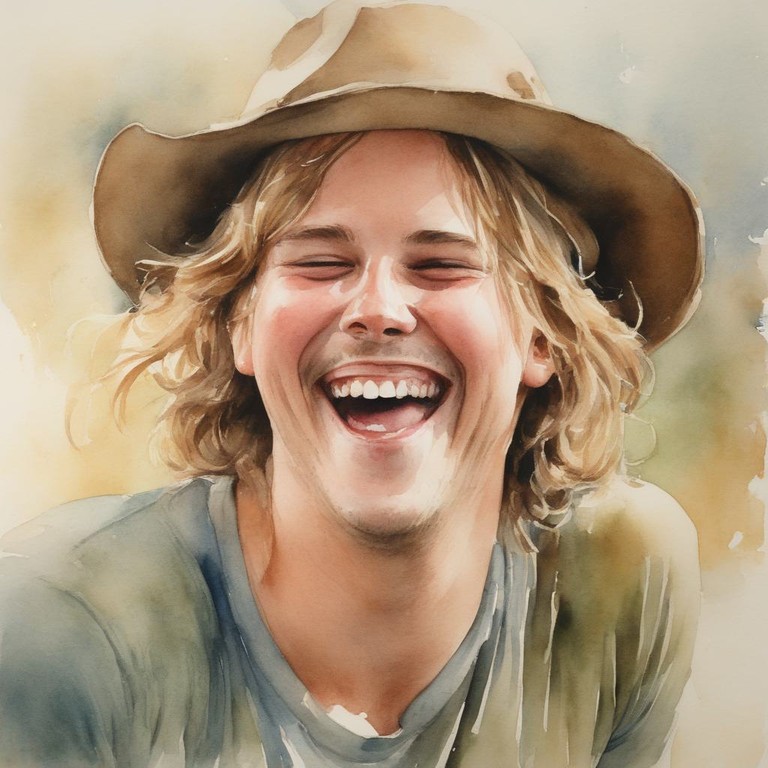} &
        \includegraphics[width=0.12\linewidth]{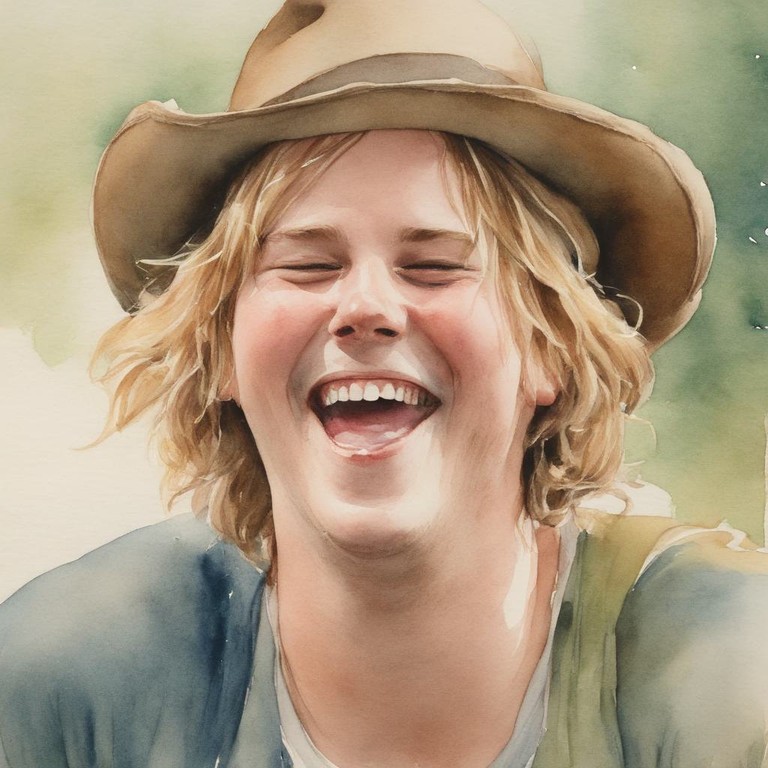} &
        \includegraphics[width=0.12\linewidth]{images/mechanism_comparison/37512.jpg} &
        \includegraphics[width=0.12\linewidth]{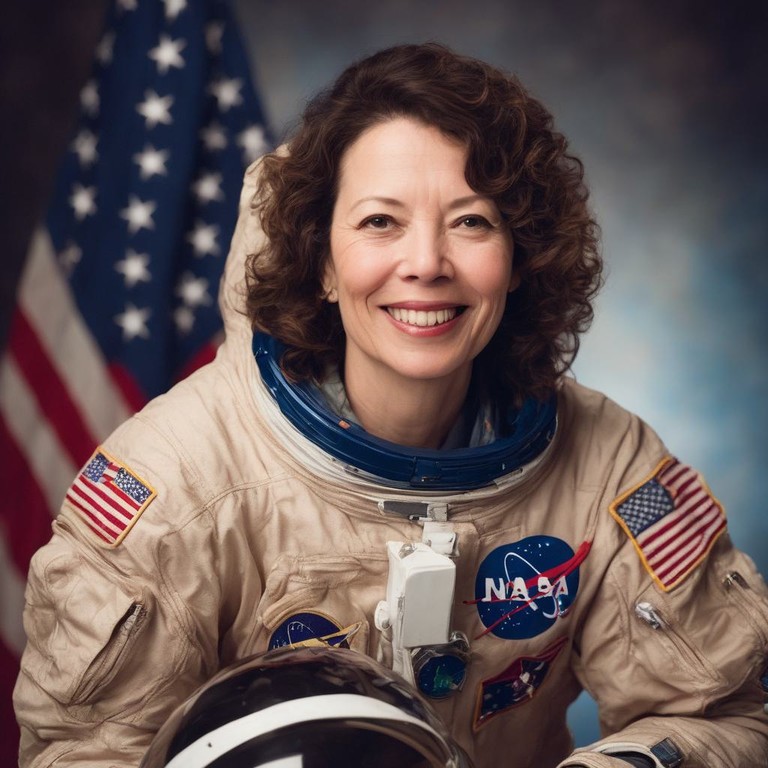} &
        \includegraphics[width=0.12\linewidth]{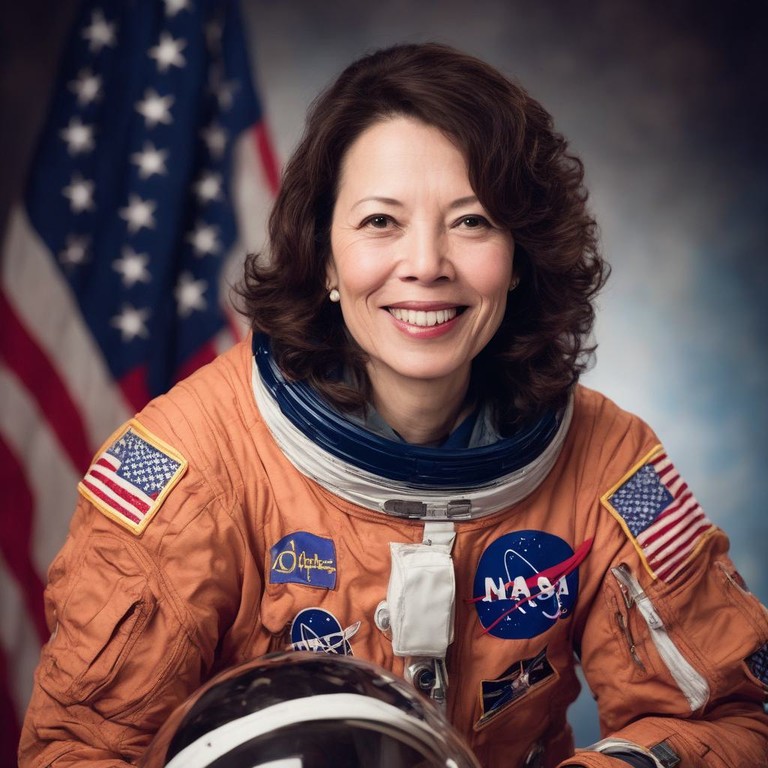} &
        \includegraphics[width=0.12\linewidth]{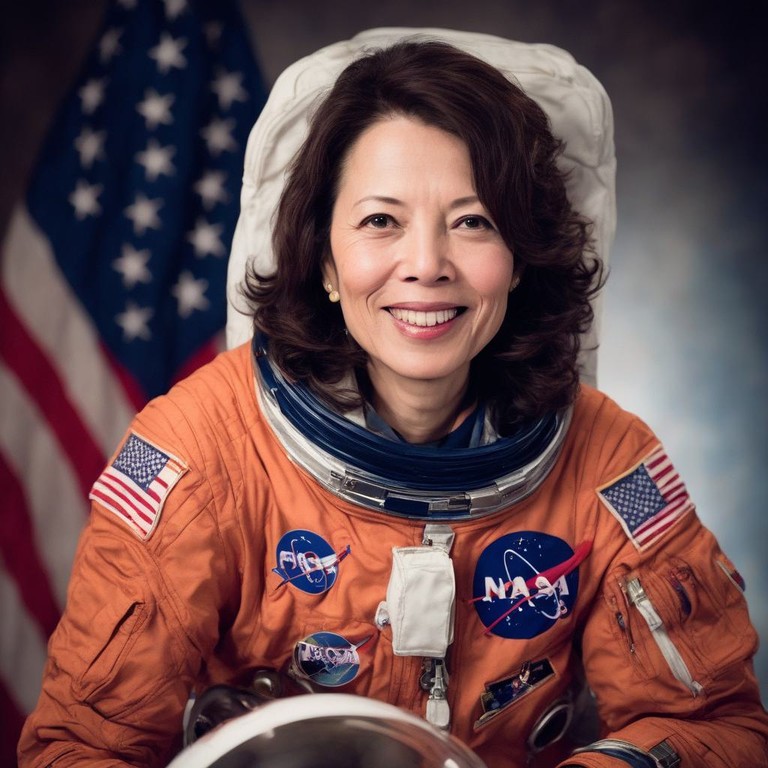} \\
        \raisebox{24pt}{\rotatebox[origin=t]{90}{Nested Attention}} &
        \includegraphics[width=0.12\linewidth]{images/mechanism_comparison/man_watercolor_input.jpg} &
        \includegraphics[width=0.12\linewidth]{images/mechanism_comparison/man_watercolor_nested_1.0.jpg} &
        \includegraphics[width=0.12\linewidth]{images/mechanism_comparison/man_watercolor_nested_2.0.jpg} &
        \includegraphics[width=0.12\linewidth]{images/mechanism_comparison/man_watercolor_nested_4.0.jpg} &
        \includegraphics[width=0.12\linewidth]{images/mechanism_comparison/37512.jpg} &
        \includegraphics[width=0.12\linewidth]{images/mechanism_comparison/woman_nested_astronaut_1.0.jpg} &
        \includegraphics[width=0.12\linewidth]{images/mechanism_comparison/woman_nested_astronaut_2.0.jpg} &
        \includegraphics[width=0.12\linewidth]{images/mechanism_comparison/woman_nested_astronaut_4.0.jpg} \\
        & \multicolumn{4}{c}{``A watercolor painting of a \emph{person} smiling, he is wearing a hat''} &
        \multicolumn{4}{c}{``A high quality photo of a \emph{person} as an astronaut''} \\
        
    \end{tabular}
    }
    \caption{
    Qualitative comparison of injection mechanism. 
    $\lambda$ balances between identity preservation and prompt alignment. 
    We use the following $\lambda$ values from left to right. Decoupled CA: 0.5, 0.6, 1.0, global $V$, multiple tokens and nested attention: 1.0, 2.0, 4.0. 
    }
    \label{fig:mechanism-comparison-supp}
\end{figure*}

\paragraph{Semantic Variations} 
When training our face model, we attach the concepts to the token \emph{person}. In \Cref{fig:semantic-variation} we show that, similarly to prior embedding-based personalization encoders~\cite{li2023photomaker}, our method supports semantic variations by changing the textual token to which we attach the subject during inference time.

\paragraph{Image injection mechanism comparison} \label{app:comparison-supp}

Qualitative results of all baseline injection methods are presented in Figure~\ref{fig:mechanism-comparison-supp}.

As can be seen in the qualitative results, the `Simple Adapter' method struggles to faithfully adhere to the input prompt, with notable deviations in style, expression, and clothing. The method's approach of concatenating a large number of image tokens with textual tokens appears to disproportionately distribute attention, prioritizing image tokens at the expense of textual tokens. This limitation is also evident by the quantitative evaluation. In IP-Adapter~\cite{ye2023ipadapter}, it has been shown that using a small amount of tokens with this approach leads to poor identity preservation.

In the `Multiple Tokens' method, the generated images adhere to the text prompt, but the identity preservation is poor. In this method, all the image tokens get the same amount of attention, and they are not query dependent. Having such a large amount of tokens that should together encode the information about the input image make the optimization process difficult. This method captures attributes such as gender and hair color, but the identity preservation is overall poor.

Decoupled Cross-Attention struggles to adhere to some text prompts, especially when they consist of non-photorealistic styles. The summation of the image-cross-attention with the text-cross-attention allows the model to generate image features that overwhelm the features coming from the text. 

In `Global $V$', averaging the tokens produced by the image encoder results in values that do not depend on the queries and hence struggle to convey all the facial details of the specific identity. The optimization in this method, however, is easier than the one in `Multiple Tokens'.

Overall, our method achieves the best tradeoff between identity preservation and prompt alignment. This is evident both in the qualitative and quantitative results.

\section{Ablation Studies}

\begin{figure}
    \centering
    \setlength{\tabcolsep}{1pt}
    \scriptsize{
    \begin{tabular}{ccccc}
        Input & 16 & 64 & 256 & 1024 \\
        \includegraphics[width=0.19\linewidth]{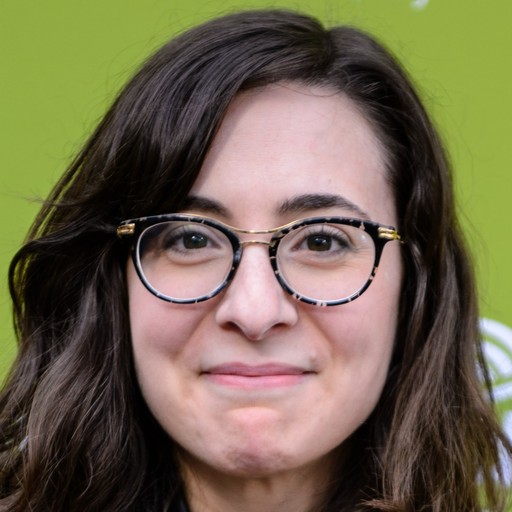} &
        \includegraphics[width=0.19\linewidth]{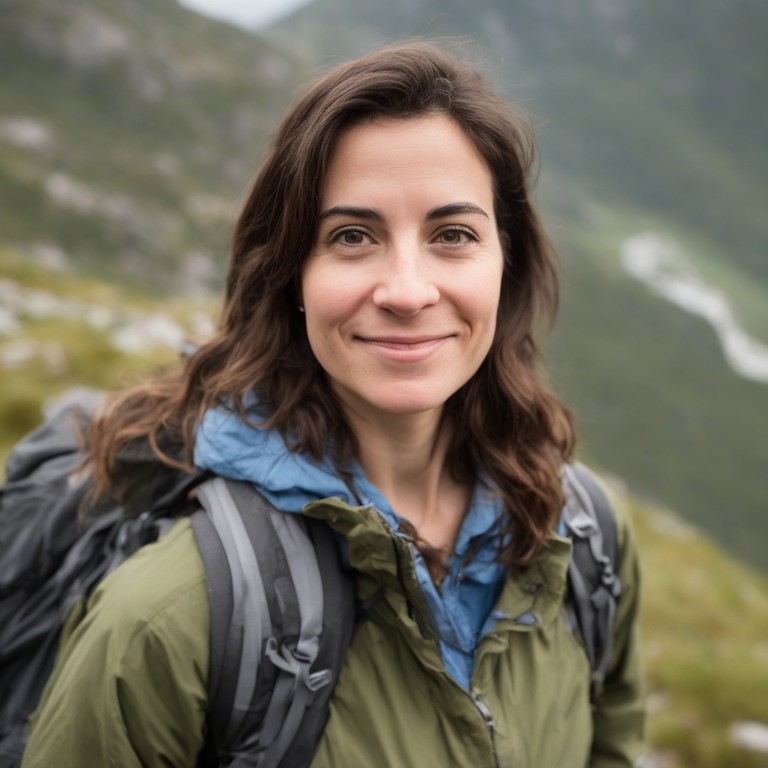} &
        \includegraphics[width=0.19\linewidth]{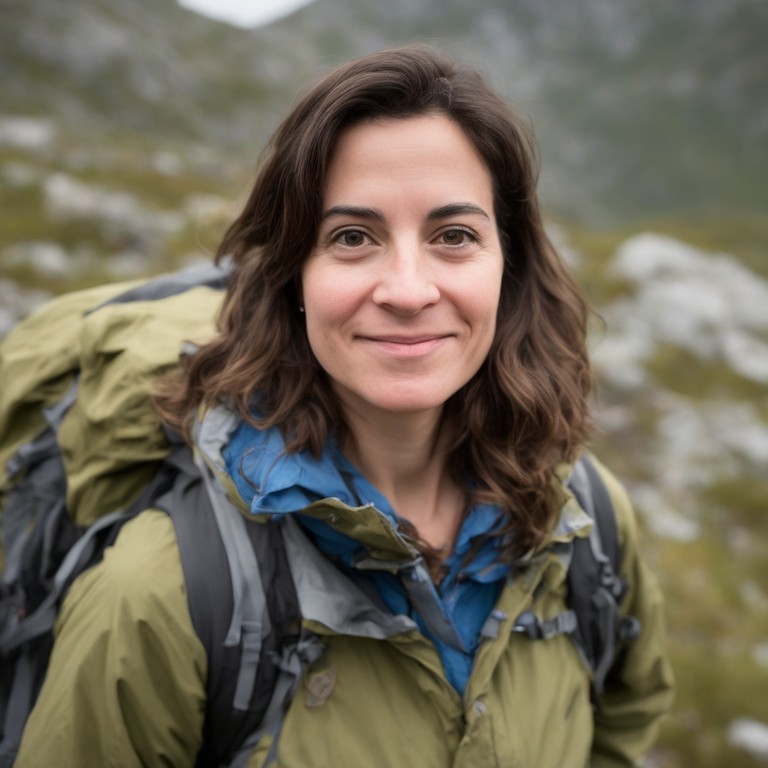} &
        \includegraphics[width=0.19\linewidth]{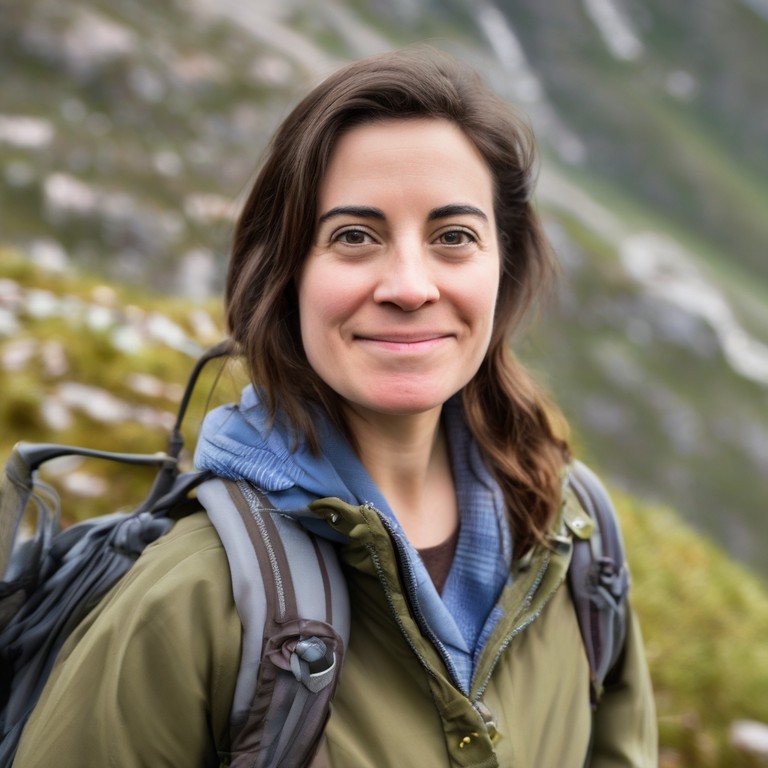} &
        \includegraphics[width=0.19\linewidth]{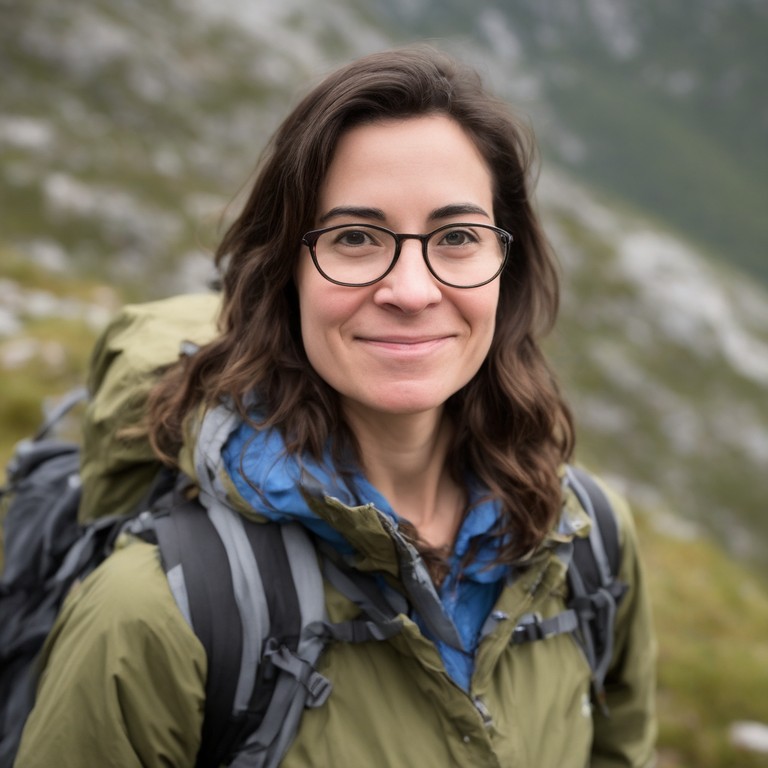} \\
        \multicolumn{5}{c}{``Hiking on a mountain''} \\
        \includegraphics[width=0.19\linewidth]{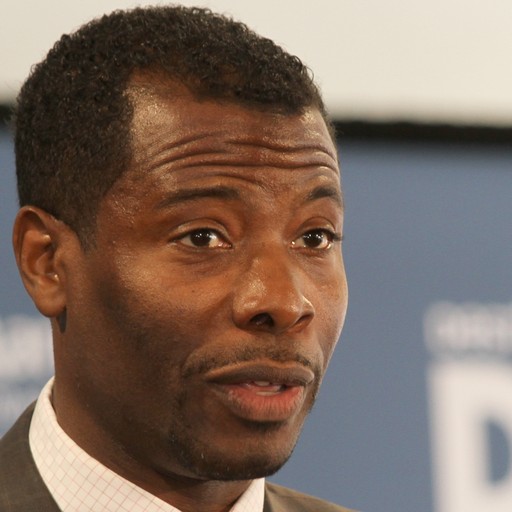} &
        \includegraphics[width=0.19\linewidth]{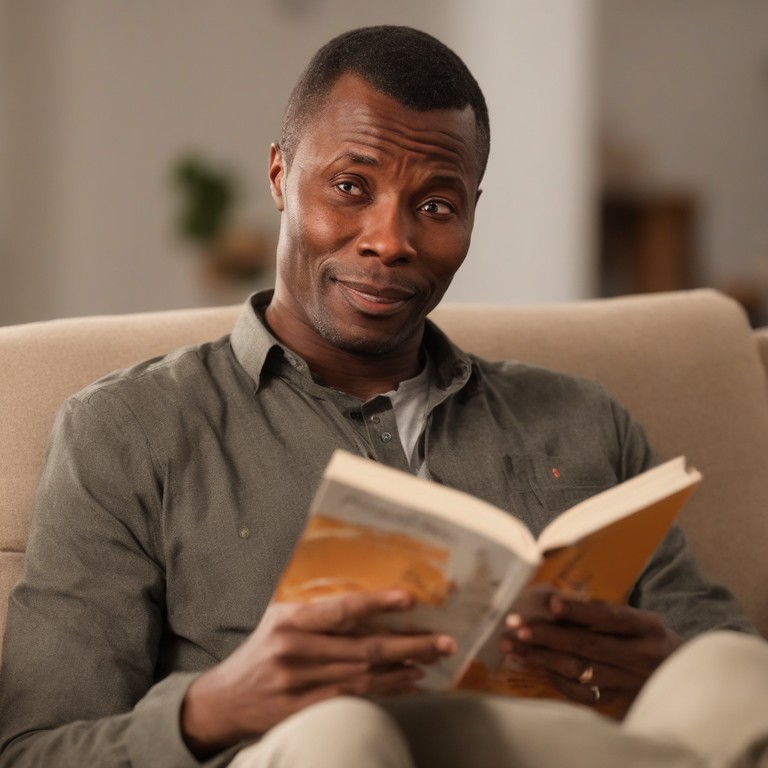} &
        \includegraphics[width=0.19\linewidth]{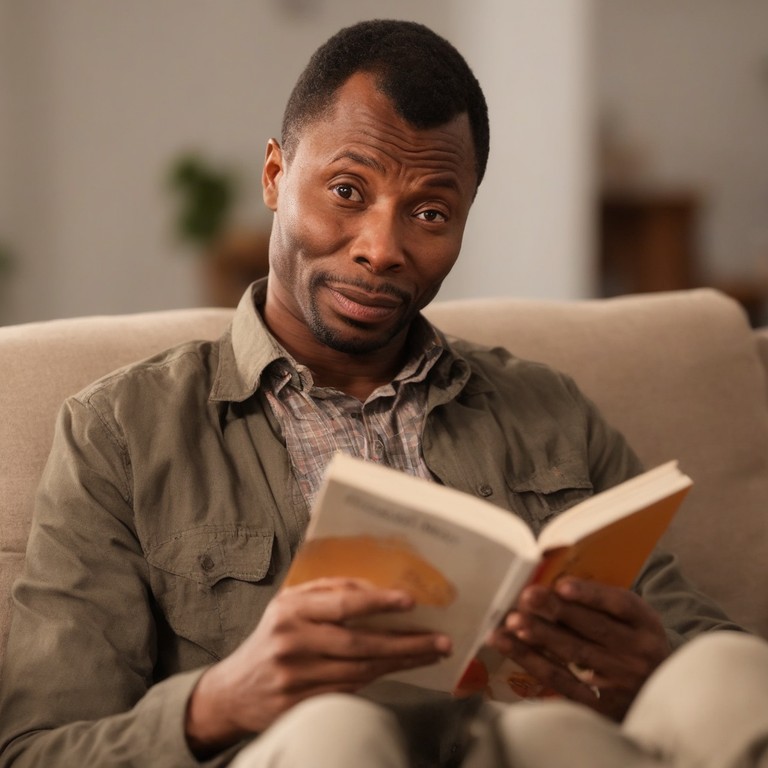} &
        \includegraphics[width=0.19\linewidth]{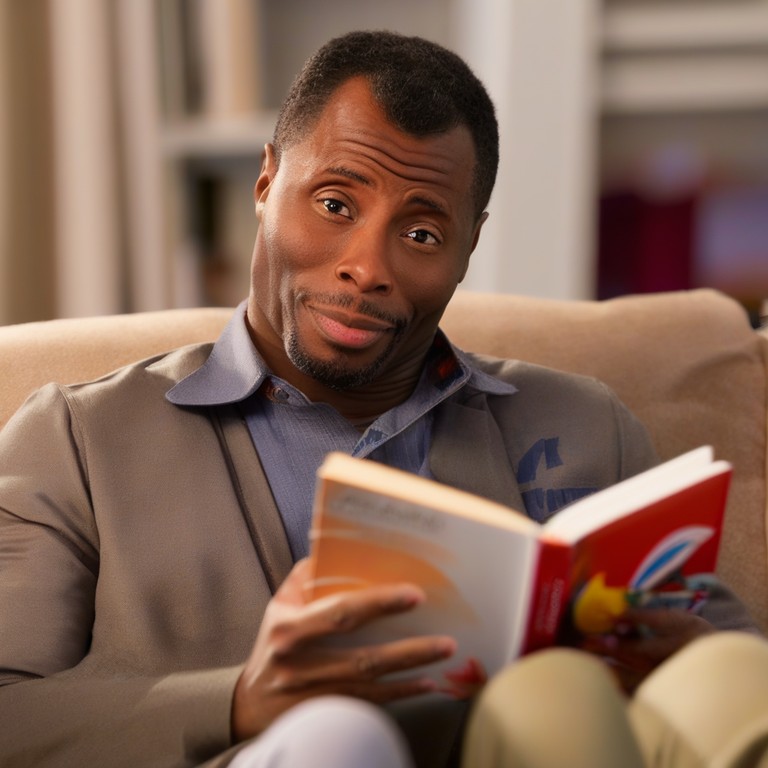} &
        \includegraphics[width=0.19\linewidth]{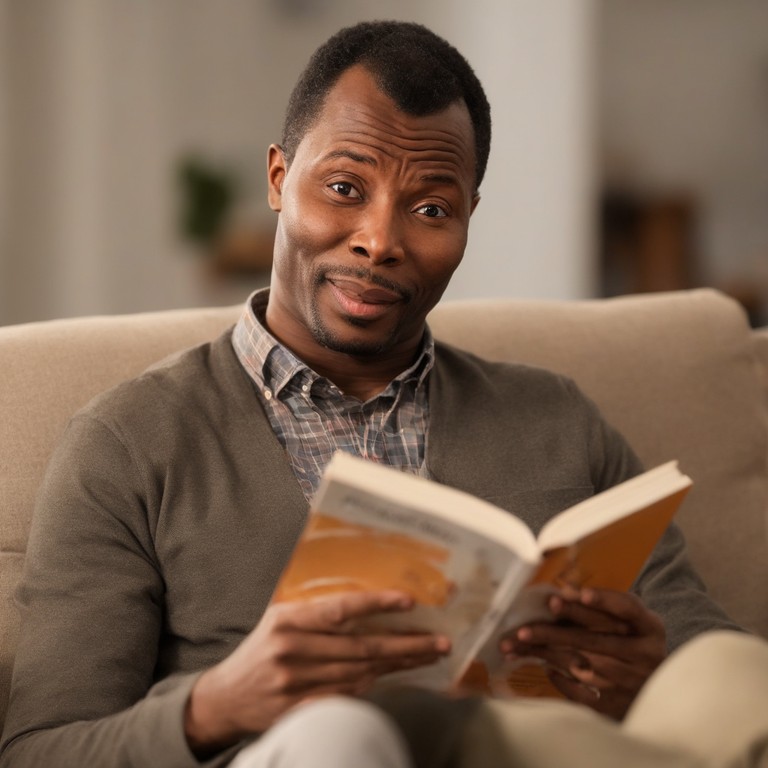} \\
        \multicolumn{5}{c}{``In a living room, reading a book''} \\[3pt]
        \textbf{ID Score} & \textbf{0.299} & \textbf{0.318} & \textbf{0.302} & \textbf{0.363}
    \end{tabular}
    }
    \caption{
    Results of models trained with varying number of learned queries. Increasing the number of learned queries improves identity preservation. All models used in this figure underwent only the first phase of training.
    }
    \label{fig:number-of-q}
\end{figure}

\paragraph{Number of learned queries}
Here, we ablate the effect of the number of learned queries in the Q-Former on model performance. Since the number of learned queries determines the number of nested keys and values, increasing them leads to a richer image representation. In Figure~\ref{fig:number-of-q}, we present results from models trained with varying numbers of learned queries. The bottom of the figure shows the average ID score computed across different prompts on our test set. All models undergo only the first training phase at a resolution of $512$. Both qualitative and quantitative results demonstrate that a higher number of learned queries enhances identity preservation and captures subtle identity features more accurately. Note that the ID scores shown here are lower than our final model's scores, as these results reflect performance after only the first training phase.

\begin{figure}
    \centering
    \setlength{\tabcolsep}{1pt}
    \scriptsize{
    \begin{tabular}{ccccc}
    Input  & w/o reg. & $\alpha=1$ & $\alpha=2$ & $\alpha=3$ \\
        \includegraphics[width=0.19\linewidth]{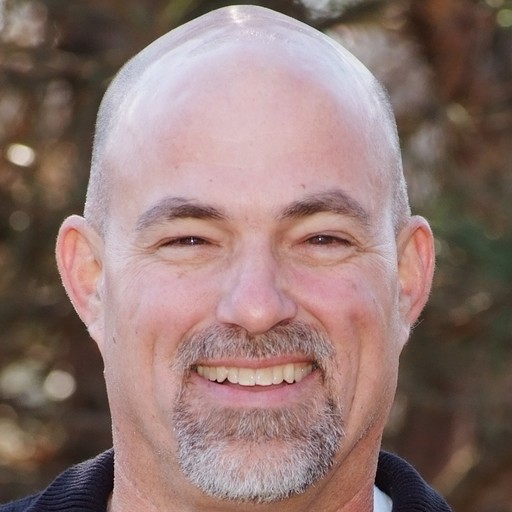} &
        \includegraphics[width=0.19\linewidth]{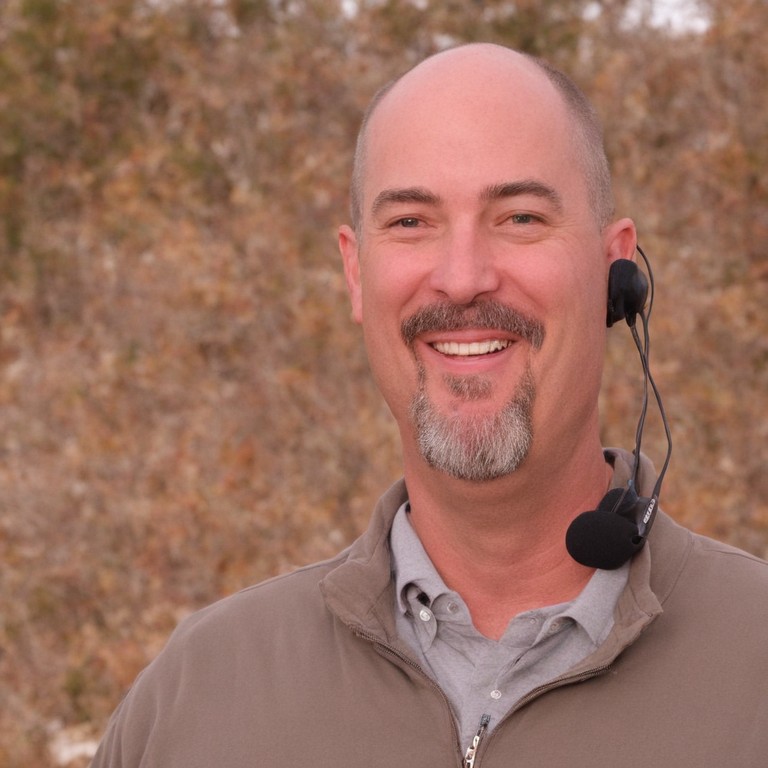} &
        \includegraphics[width=0.19\linewidth]{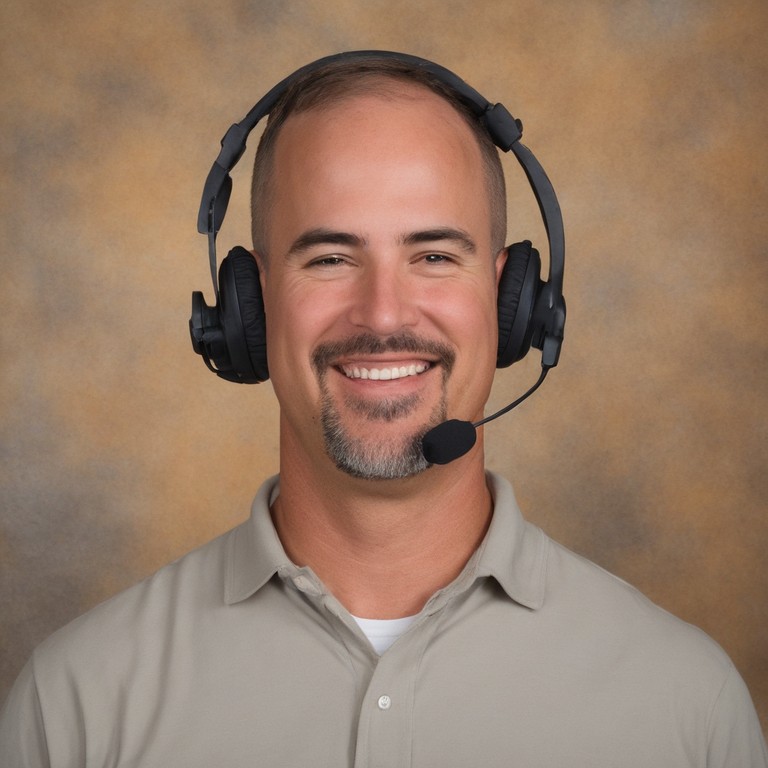} &
        \includegraphics[width=0.19\linewidth]{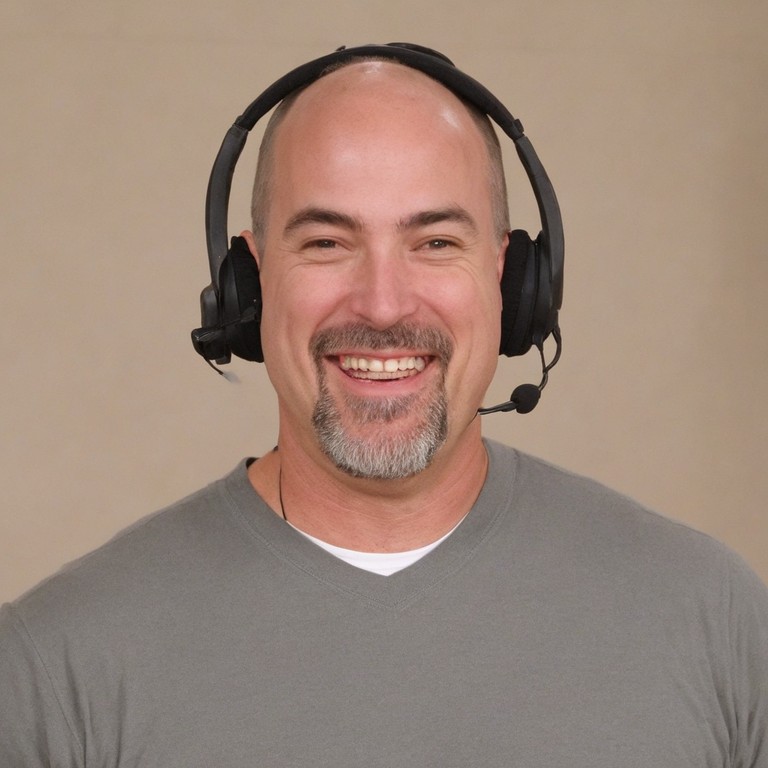} &
        \includegraphics[width=0.19\linewidth]{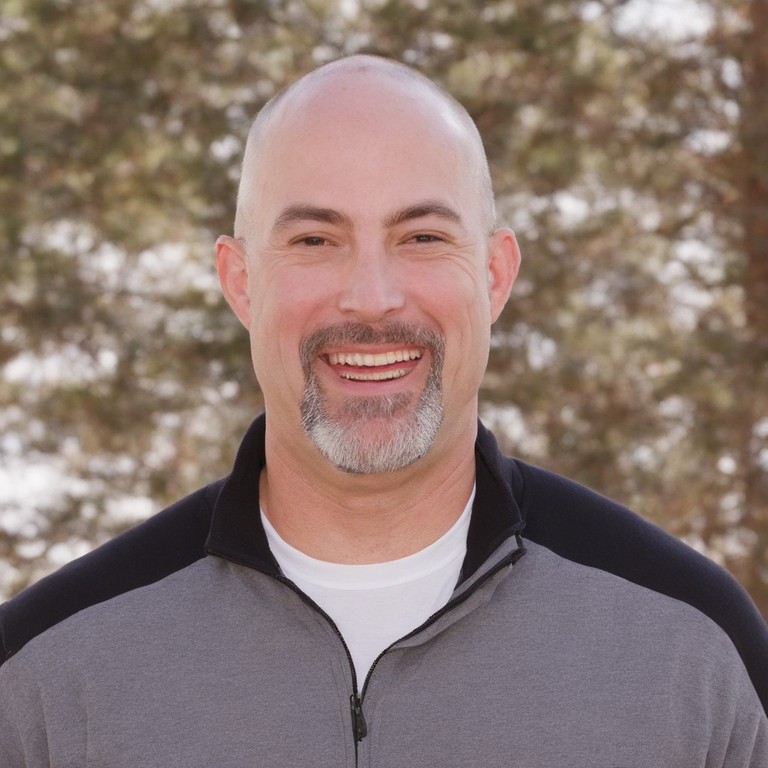} \\
        \multicolumn{5}{c}{``Wearing a headset''
} \\
        \includegraphics[width=0.19\linewidth]{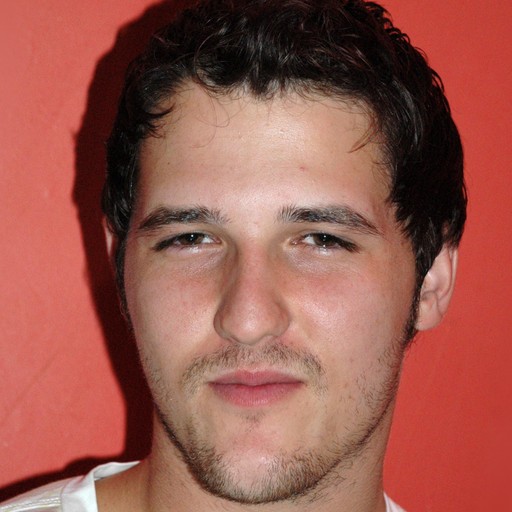} &
        \includegraphics[width=0.19\linewidth]{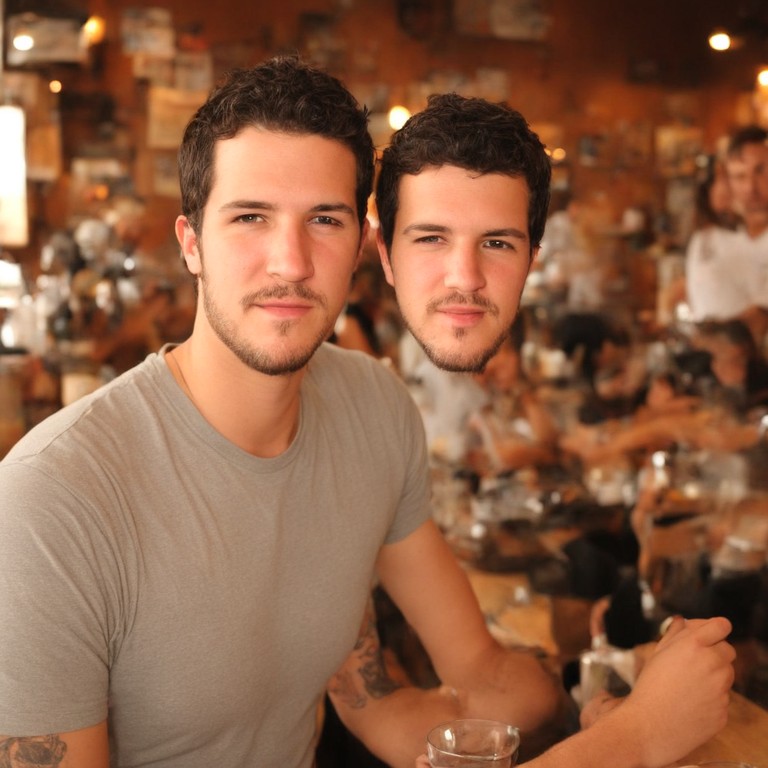} &
        \includegraphics[width=0.19\linewidth]{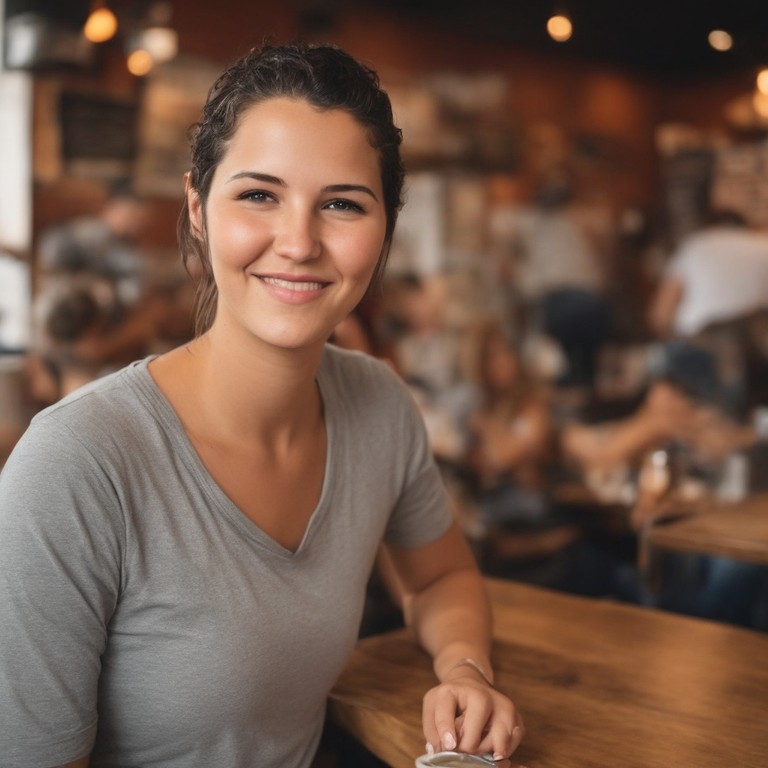} &
        \includegraphics[width=0.19\linewidth]{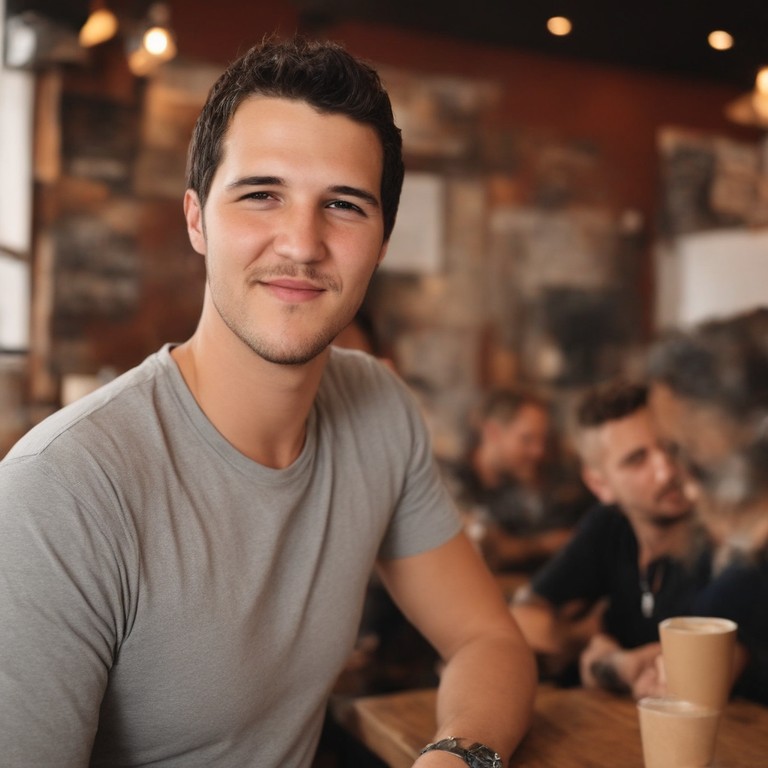} &
        \includegraphics[width=0.19\linewidth]{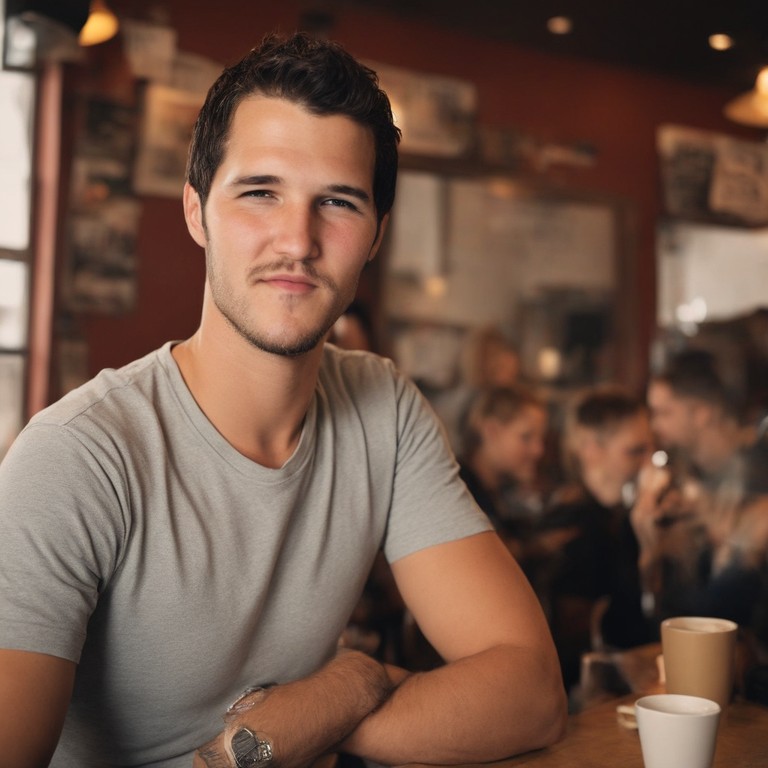} \\
        \multicolumn{5}{c}{``In a coffee shop''} \\
    \end{tabular}
    }
    \caption{
    Ablating the regularization performed on $V_q[s^*]$. Without normalization, the image looks red and contains artifacts. Setting $\alpha=2$ provides a good balance between identity preservation and prior preservation.
    }
    \label{fig:normailze-vq}
\end{figure}

\paragraph{Normalizing $\pmb{V_q[s^*]}$} \label{app:supp-ablate-normalization}
Figure~\ref{fig:normailze-vq} demonstrates the importance of regularizing $V_q[s^*]$. All models shown were trained with 256 learned queries and underwent only the first training phase (at a resolution of $512$). Without regularization, the input image dominates the output, resulting in poorer prior preservation. Additionally, unregularized models produce images with reddish tints and visible artifacts (see second column of \Cref{fig:normailze-vq}). We further ablate the choice of the regularization constant $\alpha$ (Section 3.2 in the main paper). With $\alpha=1$, artifacts are eliminated but identity preservation suffers. At $\alpha=3$, while identity preservation improves, prior preservation slightly degrades. We find that $\alpha=2$ offers a good balance between identity preservation, image quality, and prior preservation.

\end{appendices}

\end{document}